\documentclass[a4paper,12pt,times,numbered,print,oneside,custommargin]{PhDThesisPSnPDF}

\usepackage{textcomp}
\usepackage{stfloats}
\usepackage{url}
\usepackage{verbatim}
\usepackage{amssymb}  
\usepackage{color}
\usepackage{tocbibind}

\usepackage[utf8]{inputenc}

\usepackage{amsmath}
\usepackage{amssymb}
\usepackage{amsthm}
\usepackage{amsfonts}
\usepackage{cite}
\usepackage{array}
\usepackage{float} 

\usepackage{nomencl}
\makenomenclature
\renewcommand{\nomname}{List of symbols}
\renewcommand{\nompreamble}{The next list describes symbols used within this thesis.}
\usepackage{etoolbox}
\renewcommand\nomgroup[1]{%
  \item[\vspace{10pt}\bfseries
  \ifstrequal{#1}{A}{Acronyms}{%
  \ifstrequal{#1}{B}{Algebra}{%
  \ifstrequal{#1}{G}{Machine Learning Basics}{%
  \ifstrequal{#1}{F}{Operators and Functions}{%
  \ifstrequal{#1}{S}{Symbols Associated with Data}{%
  \ifstrequal{#1}{T}{Symbols Associated with Empirical Reliability}{%
  \ifstrequal{#1}{U}{Symbols Associated with Certified Reliability}{%
  \ifstrequal{#1}{O}{Operators}{%
  }}}}}}}}]%
}

\usepackage{bm}

\usepackage{booktabs}
\usepackage{bbm}

\usepackage{algorithmic}
\usepackage[ruled,linesnumbered,lined]{algorithm2e}

\newcommand{\Rmnum}[1]{\expandafter\@slowromancap\romannumeral #1@}
\usepackage{graphicx}
\usepackage{subfigure}
\usepackage{epsfig}

\usepackage{multirow}
\usepackage{makecell}
\usepackage{diagbox}

\input{Preamble/preamble}

\title{Foundations of Reliable Inference: Reliability-Efficiency Co-Design}

\author{Jiayi Huang}

\dept{The Department of Engineering}

\university{King's College London}
\crest{\includegraphics[width=0.4\textwidth]{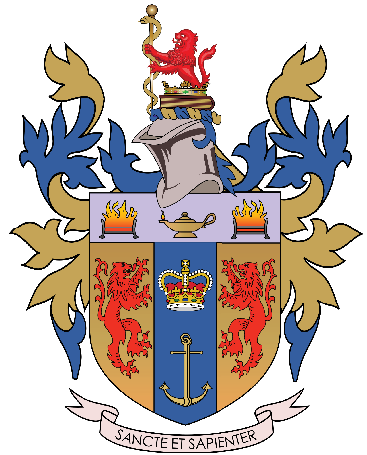}}

\supervisor{Prof. O. Simeone \newline 
Prof. F. Restuccia}

\degreetitle{Doctor of Philosophy}

\subject{LaTeX} \keywords{{LaTeX} {PhD Thesis} {Engineering} {University of
Cambridge}}

\ifdefineAbstract
\fi

\ifdefineChapter
\fi

\usepackage[acronym]{glossaries}
\makeglossaries

\begin{document}

\mainmatter

\maketitle


\begin{dedication} 

To my beloved motherland, my parents, and my brother, for their unwavering support, which has given me the freedom to chase everything I aspire to.

\end{dedication}


\begin{declaration}

I hereby declare that except where specific reference is made to the work of 
others, the contents of this dissertation are original and have not been 
submitted in whole or in part for consideration for any other degree or 
qualification in this, or any other university. This dissertation is my own 
work and contains nothing which is the outcome of work done in collaboration 
with others, except as specified in the text and Acknowledgements. This 
dissertation contains fewer than 65,000 words including appendices, 
bibliography, footnotes, tables and equations and has fewer than 150 figures.


\end{declaration}


\begin{acknowledgements}   

As time flies, my twenty-year student journey comes to an end in London, a place that I could not have imagined myself being even ten years ago. From a primary school student to a PhD, from my hometown of Guangzhou to London, the city where I have stayed the longest, this path has been unexpected. While we can not choose the starting point of our journey, we can always chase what we truly desire, even if it leads us to unfamiliar places. We are always facing uncertainty in the world and continue to experience it. As individuals with ambition, perhaps one thing we can do is to strive to be our best, continually refining our own ``models''. My research over the past three years on reliable AI has inspired me that we, as humans, are in some sense training ourselves. The way I guide and train AI models has, in turn, shaped my own thinking, thinking as a robot. Along this journey, I have experienced both small achievements and many setbacks. The completion of this thesis is not only an opportunity to summarize my work over the past three years, but also a moment to express my deepest gratitude to those who have supported me throughout. This journey would not have been possible without them.

First and foremost, I would like to express my sincere gratitude to my PhD supervisor, Prof. Osvaldo Simeone. We first met in January 2022, and I have since had the privilege of being his student. I am deeply grateful for his kind and professional guidance, both in academic research and in my career. He has always made time to meet with me, freeing me from the puzzle where I get lost, and guiding me through every aspect of research, from writing papers to presenting ideas. His ability to explain complex ideas with clarity, precision, and conciseness, along with his dedication to research and humility as a world-class scientist, has been truly inspiring and charming, which always encourages me to do the work with sincerity and respect, benefiting me throughout my life. I hope that one day I can be like him.

I would also like to extend my sincere thanks to my collaborators. To Dr. Nicola Paoletti, for his insightful comments and discussions; to Dr. Sangwoo Park, for his unwavering and generous support in paper writing, experimental implementation, and countless inspiring discussions; and to Dr. Amirmohammad Farzaneh, for his timely and encouraging advice. Their contributions have continually pushed me to challenge my limits and improve the quality of my research. It has been a great privilege to work with such outstanding researchers, whose professionalism and humility have inspired me to always strive for excellence.

In addition, this journey would not have been possible without the support of my colleagues at King’s. Alongside Sangwoo and Amir mentioned above, I would like to thank Prof. Yunchuan Zhang, Dr. Jiechen Chen, Dr. Clement Ruah, Dr. Kfir Cohen, and Dr. Zihang Song for the many engaging and interesting conversations we have shared. I am also deeply grateful to my colleagues and friends at King’s, Dr. Ivana Nikoloska, Dr. Nicolas Skachkovsky, Dr. Hari Hara Suthan Chittoor, Dr. Matteo Zecchin, Dr. Prabodh Katti, Dr. Dengyu Wu, Dr. Houssem Sifaou, Dr. Abin Varghese, Dr. Robert O'Shea, Dr. Meiyi Zhu, Dr. Qiushuo Hou, Bowen Wang, Kristian Sotirov, and Hange Lao, for the meals we have shared and the joy we have experienced together. I would also like to thank all the visiting colleagues I had the pleasure of meeting, exchanging ideas with, and sharing memorable moments with. I am truly grateful to everyone I have met over the past three years, who have made my life in London joyful.

Last but not least, my deepest gratitude goes to my family, who have always supported me in exploring the world in my own way. To my parents, who have provided me with a nice and supportive environment for both life and work, and who have stood by me, especially during the most difficult moments of my research. Although they may not fully understand my work, they have always encouraged me to enjoy life, embrace nature, and live freely. To my brother, who constantly shares new perspectives and insights from his own field, broadening my horizons.

\end{acknowledgements}

\begin{abstract}

Reliable inference requires that artificial intelligence (AI) models provide trustworthy uncertainty estimates, not merely accurate predictions. Recent advances in Bayesian learning have made significant progress toward this goal, and growing concerns about computational overhead have jointly shifted the design criterion from reliability alone to the co-design of reliability and efficiency, i.e., reducing computational overhead while preserving trustworthy uncertainty quantification. This thesis develops a unified framework from two perspectives to address the central question: \emph{can we efficiently perform reliable inference?}

From the \emph{reliability} perspective, we first propose a calibration-aware Bayesian learning framework that integrates three components: a calibration penalty to improve in-distribution (ID) calibration, out-of-distribution (OOD) confidence minimization to enhance distributional shift detection, and a selective calibration mechanism that abstains from decisions where calibration is expected to degrade, thereby reconciling conflicting ID and OOD objectives. While this framework enhances reliability through well-calibrated confidence levels, Bayesian learning remains a best-effort method offering only \emph{empirical reliability}. By contrast, statistical tools such as conformal prediction (CP) can formally guarantee the target reliability requirement by augmenting any pre-trained model with prediction sets, referred to as \emph{certified reliability}. As an instance, we provide certified reliability for optimized certainty equivalent (OCE) risk measures, controlling worst-case model behavior as required by high-stakes applications such as medical image diagnosis. However, CP tends to produce uninformative prediction sets when the underlying model is of poor quality, e.g., a small-scale model, which poses a challenge for efficient deployment. 


From the \emph{efficiency} perspective, we bridge the reliability gap between complex and compact models, balancing the trade-off between reliability and computational cost. To endow compact models with informative predictions while preserving certified reliability, we develop a conformalized calibration distillation framework that distills calibration information from a complex model, where certified reliability is preserved through prediction sets containing reference distributions with high probability. However, the inherent representational limitations of the compact model prevent it from producing reliable decisions for all inputs. To address this limitation, we further extend this framework to collaborative inference: inputs are processed locally when the compact model's decision aligns with the complex model's prediction, thereby preserving the same level of reliability as if produced by the reference complex model. This method achieves the most efficient and reliable performance at the cost of escalating only a fraction of inputs to the complex model.

Together, these contributions span from empirical to certified reliability, and from single model prediction to model collaboration, providing a principled and validated answer to the opening question of efficient, reliable inference.

\end{abstract}


\tableofcontents

\listoffigures

\listoftables


\printnomenclature

\printglossary[type=\acronymtype, title=List of Acronyms]



\newacronym{ai}{AI}{artificial intelligence}
\newacronym{id}{ID}{in-distribution}
\newacronym{ood}{OOD}{out-of-distribution}
\newacronym{bnns}{BNNs}{Bayesian neural networks}
\newacronym{fnns}{FNNs}{frequentist neural networks}
\newacronym{mmce}{MMCE}{maximum mean calibration error}
\newacronym{esd}{ESD}{expected squared difference}
\newacronym{avuc}{AvUC}{accuracy versus uncertainty calibration}
\newacronym{ocm}{OCM}{out-of-distribution confidence minimization}
\newacronym{cbnn}{CBNN}{calibration-regularized Bayesian neural network}
\newacronym{cfnn}{CFNNs}{calibration-regularized FNNs}
\newacronym{cbnn-ocm}{CBNN-OCM}{CBNN-OCM}
\newacronym{scbnn-ocm}{SCBNN-OCM}{selective CBNN-OCM}
\newacronym{sbnn-ocm}{SBNN-OCM}{selective BNN-OCM}
\newacronym{cfnns}{CFNNs}{calibration-regularized FNNs}
\newacronym{ece}{ECE}{expected calibration error}
\newacronym{kl}{KL}{Kullback-Liebler}
\newacronym{vi}{VI}{variational inference}
\newacronym{sgd}{SGD}{stochastic gradient descent}
\newacronym{tv}{TV}{total variation}
\newacronym{flop}{FLOPs}{floating-point operations}
\newacronym{cp}{CP}{conformal prediction}
\newacronym{rcps}{RCPS}{risk-controlling prediction sets}
\newacronym{fnr}{FNR}{false negative rate}
\newacronym{oce}{OCE}{optimized certainty equivlent}
\newacronym{cvar}{CVaR}{conditional value-at-risk}
\newacronym{oce-crc}{OCE-CRC}{OCE conformal risk control}
\newacronym{oce-rcps}{OCE-RCPS}{OCE risk-controlling prediction sets}
\newacronym{ucb}{UCBs}{upper confidence bounds}
\newacronym{wsr}{WSR}{Waudby-Smith and Ramdas}
\newacronym{cd-ci}{CD-CI}{conformalized distillation for credal inference}
\newacronym{ip}{IP}{imprecise probability}
\newacronym{ipml}{IPML}{Imprecise probabilistic machine learning}
\newacronym{ca}{CA}{conformal alignment}
\newacronym{lcp}{LCP}{localized CP}
\newacronym{cab}{CAb}{conformal alignment-based}
\newacronym{mht}{MHT}{multiple-hypothesis testing}
\newacronym{fdr}{FDR}{false discovery rate}
\newacronym{qa}{QA}{question-answering}
\newacronym{hms}{HMS}{highest mass set}
\newacronym{wmmce}{WMMCE}{weighted MMCE}
\newacronym{dr}{DR}{deferral rate}
\newacronym{ni}{NI}{normalized inefficiency}
\newacronym{fdp}{FDP}{false discovery proportion}
\newacronym{cbd}{CbD}{confidence-based deferral}
\newacronym{sota}{SOTA}{state-of-the-art}
\newacronym{kde}{KDE}{kernel density estimator}
\newacronym{ce}{CE}{cross-entropy}
\newacronym{llm}{LLMs}{large language models}
\newacronym{nn}{NNs}{neural networks}

\nomenclature[B]{$\mathbb{R}$}{Set of real scalars}
\nomenclature[B]{$\mathbb{R}^d$}{Set of real vectors of size $d$}

\nomenclature[G]{$\mathcal{X}$}{Input space}
\nomenclature[G]{$\mathcal{Y}$}{Output/label space}
\nomenclature[G]{$x$}{Input vector}
\nomenclature[G]{$y$}{Output vector}
\nomenclature[G]{$\mathcal{P}$}{Probability simplex}
\nomenclature[G]{$p(\cdot \mid x)$}{Pre-trained predictive distribution on input $x$}
\nomenclature[G]{$\Pr[\cdot]$}{The probability of a specific objective}
\nomenclature[G]{$(x,y) \sim p(x,y)$}{Data pair sampled from joint distribution}
\nomenclature[G]{$\hat{y}(x)$}{Hard decision}
\nomenclature[G]{$\theta$}{Model parameter vector}
\nomenclature[G]{$n_p$}{Dimension of model parameter vector}
\nomenclature[G]{$\mathcal{L}(\cdot \mid \cdot)$}{Cross-entropy loss}
\nomenclature[G]{$\ell(\cdot, \cdot)$}{Non-negative bounded loss}
\nomenclature[G]{$\mathcal{H}$}{Null hypothesis}
\nomenclature[G]{$d$}{Input dimensionality}
\nomenclature[G]{$M$}{Number of bins for expected calibration error}
\nomenclature[G]{$\alpha$}{Parameter for $\alpha$-divergence measurement}
\nomenclature[G]{$h$}{Kernel bandwidth}
\nomenclature[G]{$\varphi =[\mu,\Sigma]$}{Variational parameter vector with Gaussian mean $\mu$ and covariance $\Sigma$}
\nomenclature[G]{$\beta$}{Kullback-Leibler divergence weight in free energy}
\nomenclature[G]{$\Pr[\cdot \mid x]$}{The conditional probability of a specific objective given input $x$}
\nomenclature[G]{$p(y \mid x,\theta)$}{Parametrized model}
\nomenclature[G]{$\mathcal{N}(\cdot, \cdot)$}{Normal distribution}
\nomenclature[G]{$\delta_{\mathcal{J}}$}{Point mass at value $\mathcal{J}$}
\nomenclature[G]{$\mathcal{D}$}{Reference dataset}
\nomenclature[G]{$\mathcal{D}^{\text{tr}}$, $\mathcal{D}^{\text{val}}$, $\mathcal{D}^{\text{cal}}$, $\mathcal{D}^{\text{opt}}$, $\mathcal{D}^{\text{unl}}$, $\mathcal{D}^{\text{te}}$}{Training, validation, calibration, optimization, unlabeled, and test dataset}

\nomenclature[F]{$\text{exp}(\cdot)$}{Exponential function}
\nomenclature[F]{$\infty$}{Infinity}
\nomenclature[F]{$\lvert\cdot\rvert$}{Absolute value / Size of a set}
\nomenclature[F]{$\arg \max$}{Arguments of the maxima}
\nomenclature[F]{$\arg \min$}{Arguments of the minima}
\nomenclature[F]{$\max$}{Maximum}
\nomenclature[F]{$\mathbbm{1} (\cdot)$}{Indicator function}
\nomenclature[F]{$\log(\cdot)$}{Natural logarithm}
\nomenclature[F]{$\mathbb{E}[\cdot]$}{Expectation function}
\nomenclature[F]{$\kappa(\cdot,\cdot)$}{Kernel function}
\nomenclature[F]{$\approx$}{``approximated to''}
\nomenclature[F]{$\in$}{``element of''}
\nomenclature[F]{$\subseteq$}{``subset of''}
\nomenclature[F]{$\inf$}{Infimum}
\nomenclature[F]{$\sup$}{Supremum}
\nomenclature[F]{$\cup$}{Union}
\nomenclature[F]{$\cap$}{Intersection}
\nomenclature[F]{$\nabla_{\theta}$}{Gradient with respect to parameter $\theta$}
\nomenclature[F]{$\operatorname{KL}(\cdot,\cdot)$}{Kullback-Leibler divergence}
\nomenclature[F]{$\mathrm{D}_f (\cdot \mid \cdot)$}{$f$-divergence}
\nomenclature[F]{$\lceil \cdot \rceil$}{Ceiling function}
\nomenclature[F]{$\mathcal{O}(\cdot)$}{Computational complexity}
\nomenclature[F]{$H(\cdot)$}{Shannon entropy}
\nomenclature[F-a]{$\mathcal{F}(\cdot \mid \cdot)$}{Free energy objective}
\nomenclature[F]{$\text{Quantile}_{1-\alpha_{\text{label}}^{\text{mis}}}(\cdot)$}{Empirical $(1-\alpha_{\text{label}}^{\text{mis}})$-quantile of a vector}
\nomenclature[F]{$\lfloor \cdot \rceil$}{Rounding function}

\nomenclature[T]{$p(\theta)$}{Parameter prior}
\nomenclature[T]{$p^{\text{ID}}(r)$, $p^{\text{OOD}}(r)$}{Confidence-level distributions for in-distribution and out-of-distribution data}
\nomenclature[T]{$\phi$}{Selector parameter vector}
\nomenclature[T]{$c(\cdot)$, $r(\cdot)$, $\bar{r}(\cdot)$}{Correctness score, confidence score, and expected confidence score}
\nomenclature[T]{$z$}{Hidden-layer feature vector for outlier detection}
\nomenclature[T]{$\mathcal{D}^{\text{tr}}_{\text{id}}$, $\mathcal{D}^{\text{unl}}_{\text{ood}}$}{Training dataset for in-distribution calibration and uncertainty dataset for out-of-distribution detection}

\nomenclature[T]{$\gamma_\text{id}$, $\gamma_\text{ood}$, $\gamma_\text{sel}$}{Weights for in-distribution calibration, out-of-distribution detection, and selective calibration regularizers}

\nomenclature[T]{$\xi$}{Target in-distribution coverage rate for selective calibration}
\nomenclature[T]{$\tau_{\text{sel}}$}{Threshold for selector binarization}
\nomenclature[T]{$q(\theta \mid \varphi)$}{Variational posterior}
\nomenclature[T]{$s(\cdot \mid \theta)$}{Outlier score vector}
\nomenclature[T]{$p^{\text{OOD}}_{\text{d}}$}{Optimal out-of-distribution detection probability}

\nomenclature[U]{$\psi(u)$}{Optimized certainty equivalent cost function}
\nomenclature[U]{$\lambda$, $\hat{\lambda}$}{Risk-controlling prediction set threshold parameter and its optimal value}
\nomenclature[U]{$\Gamma_{\lambda}(\cdot)$}{Parameterized prediction set with threshold $\lambda$}
\nomenclature[U]{$t$, $t^*$}{Optimized certainty equivalent auxiliary parameter and its optimized value}
\nomenclature[U]{$\alpha_{\text{risk}}^{\text{tol}}$}{Risk tolerance level}
\nomenclature[U]{$R_{\text{OCE}}(\cdot)$}{Optimized certainty equivalent risk function}
\nomenclature[U]{$\zeta$}{Risk sensitivity parameter for conditional value-at-risk / entropoc risk}
\nomenclature[U]{$\alpha_{\text{dist}}^{\text{mis}}$, $\alpha_{\text{label}}^{\text{mis}}$}{Miscoverage rate for prediction sets of distributions and for prediction sets of labels}

\nomenclature[U]{$\mathcal{V}$}{Divergence score set}
\nomenclature[U]{$q_{L_y}$, $q_{U_y}$}{Probability lower and upper bounds on class $y$}
\nomenclature[U]{$b$}{Constant to satisfy the unit-sum constraint}
\nomenclature[U]{$\hat{q}$, $\hat{q}(\cdot)$}{Conformal prediction quantile threshold and localized conformal prediction quantile threshold}
\nomenclature[U]{$k$, $k_{\text{CA}}$}{Screening step index and screening stopping index}
\nomenclature[U]{$p^*(y|x)$, $p^e(y|x)$}{Cloud model and edge model}
\nomenclature[U]{$\Gamma^*(\cdot)$, $\Gamma^e(\cdot)$}{Cloud-generated prediction set and edge-generated prediction set}

\nomenclature[U]{$\mathcal{S}$}{Edge processed input subset}
\nomenclature[U]{$\delta$}{Tolerance level of violation}
\nomenclature[U]{$\Gamma_{\text{dist}}(\cdot)$, $\Gamma(\cdot)$}{Prediction set of predictive distributions and prediction set of labels}
\nomenclature[U]{$\mathcal{J}$}{Non-conformity score set}
\nomenclature[U]{$\tau_{\text{div}}$, $\tau_\text{def}$}{Credal set radius and self-reported confidence threshold for edge-cloud escalation}
\nomenclature[U]{$\mathcal{D}_{\text{CA}}^{\text{tr}}$, $\mathcal{D}_{\text{CA}}^{\text{val}}$}{Training and validation dataset for conformal alignment}
\nomenclature[U]{$\mathcal{D}_{(k)}^{\text{scr}}$, $\mathcal{D}_{(k)}^{\text{uns}}$}{Screened and unscreened dataset at screening step $k$}
\nomenclature[U]{$\mathcal{D}_{(k)}^{\text{uns,val}}$, $\mathcal{D}_{(k)}^{\text{uns,te}}$}{Unscreened validation and unscreened test dataset at screening step $k$}
\nomenclature[U]{$C^*(\cdot)$, $\hat{C}(\cdot)$}{Ground-truth alignment score and the corresponding predicted alignment score}

\chapter{Introduction} \label{chapter:1}  

\ifpdf
    \graphicspath{{Chapter1/Figs/}{Chapter1/Figs/PDF/}{Chapter1/Figs/}}
\else
    \graphicspath{{Chapter1/Figs/}{Chapter1/Figs/}}
\fi
\section{Motivations}
\label{sec:c1_motivation}

Modern \gls{ai} models, such as \gls{nn} and \gls{llm}, have achieved remarkable success in classification tasks and generative tasks, often surpassing human experts' performance on established benchmarks. However, deploying such \gls{ai} models in real-world systems, such as autonomous driving and medical diagnosis, faces two significant challenges: (\emph{i}) providing \emph{reliable} predictions that report what the model knows and what it does not, and (\emph{ii}) enabling \emph{efficient} inference with low computational overhead, particularly in resource-constrained scenarios. Reliability and efficiency are, however, two sides of the same coin. That said, improvements in reliability often require richer representations and more costly inference procedures, while efficiency gains tend to be obtained by reducing model complexity in ways that may compromise reliability. These two competing targets give rise to a central question: \emph{Can \gls{ai} models provide trustworthy measures of uncertainty in predictions, efficiently?} Addressing this question motivates the reliability-efficiency co-design pursued in this thesis.

\paragraph{Why reliability matters.}

The European Union's \gls{ai} Act~\citep{euaiact}, entered into force in 2024, requires \gls{ai} systems deployed in high-stakes domains, such as medical devices, critical infrastructure, and autonomous driving, to be transparent about their trustworthiness and potential errors, shifting the reliability requirement from optional to essential. More broadly, for \gls{ai} models to be trusted in safety-critical applications, they must not only be accurate about their predictions but must also reliably report \emph{what they know and what they do not know}. Existing approaches offer varying degrees of such reliability: conventional frequentist learning reports reliability with self-reported confidence scores but ignores epistemic uncertainty stemming from a lack of knowledge or incomplete data; Bayesian learning accounts for epistemic uncertainty by treating the weights of an AI model as random variables, yet it provides only empirical reliability as a best-effort approach; and statistical tools such as \gls{cp} provide finite-sample, distribution-free reliability guarantees. However, greater reliability is typically accompanied by a higher computational cost (see Sec.~\ref{sec:c1_reliability_related_work} for a detailed discussion).

\paragraph{Why efficiency matters.}

The cost of inference, performed billions of times over a model's lifetime, accounts for a growing share of global data-center energy consumption~\citep{ahmed2021review, siddik2021environmental}, and edge deployment further imposes hard constraints on memory, latency, and computational power~\citep{singh2023edge}. That said, \gls{ai} models must \emph{perform inference efficiently}. Techniques such as model compression, knowledge distillation, and collaborative inference have been developed to address this bottleneck, yet they largely ignore the requirement of reliability. Specifically, conventional model compression reduces the number of parameters available to represent the predictive distribution, which typically produces unreliable predictions, as the compressed model lacks sufficient representational capacity to preserve reliability. Knowledge distillation transfers accuracy but not necessarily reliability, as the distillation objective is designed to match the teacher model's predictions rather than to preserve its ability to capture uncertainty. Conventional model cascading relies on heuristic rules to choose the threshold for triggering the escalation, where the threshold is tuned for accuracy rather than derived from statistical guarantees, without providing formal reliability assurances (see Sec.~\ref{sec:c1_efficiency_related_work} for a detailed discussion).

\paragraph{The co-design perspective.}

Reliability and efficiency are two sides of the same coin: the most reliable methods demand the highest computational power and the greatest representational capacity to provide trustworthy decisions, while the most efficient models sacrifice reliability to operate with reduced computational budgets. Existing reliable and efficient inference methods have largely evolved in isolation. For instance, Bayesian learning frameworks aim at improving the reliability by treating model weights as random variables and are trained from scratch, regardless of their computational cost. \gls{cp} provides certified reliability, yet its prediction sets are generally uninformative when applied to a compact model, which prevents efficient inference. Knowledge distillation and collaborative inference focus on reducing computational cost while completely ignoring reliability requirements. No single approach alone answers the opening question. The challenge, therefore, lies in their \emph{co-design}: balancing the trade-off between reliability and efficiency under real-world computational constraints. This thesis develops a general framework that progressively addresses this challenge, each motivated by a concrete limitation of the preceding one.

\section{Research Questions}
\label{sec:research_questions}

This thesis addresses the following four research questions:

\begin{enumerate}[label=\textbf{RQ\arabic*:}, leftmargin=*, itemsep=6pt]
    \item \textbf{Can we enable \gls{ai} models to reliably report what they know and what they do not know, and at what cost?} We develop a unified Bayesian learning framework that enables the model to reliably report its confidence on both \gls{id} and \gls{ood} inputs via three components: a \gls{id} calibration regularization term that targets \gls{id} performance, an \gls{ood} confidence minimization objective for penalizing \gls{ood} inputs, and a selective inference mechanism that jointly guarantees \gls{id} and \gls{ood} performance. The computational cost of improving reliability arises from the need to retrain models from scratch and to maintain ensembles at inference time. (see \emph{Chapter~\ref{chapter:3}}.)

    \item \textbf{Instead of the best-effort approach, can we provide certified reliability with informative decisions?} We propose a \gls{cp}-based post-hoc inference framework that formally guarantees the reliability, e.g., ensuring that the decision risk is below a user-specified threshold with high probability. \gls{cp} augments any pre-trained model with prediction sets, without retraining, and provides certified reliability by calibrating the conformal threshold on a held-out calibration dataset. However, while certified reliability holds regardless of the pre-trained model quality, whether the resulting decisions are informative depends on the quality of the underlying model. Specifically, a powerful model achieves certified reliability with informative prediction sets, while a weak model has to preserve certified reliability by inflating the prediction sets to cover the entire output space. (see \emph{Chapter~\ref{chapter:4}}.)

    \item \textbf{Can we preserve reliability and informativeness when computational constraints force us to use a small-scale model?} We propose a conformalized credal inference framework that distills calibration information from a large-scale model to a small-scale model through offline distillation. The reliability of the small-scale model is preserved by guaranteeing that the reference predictive distribution is contained in the credal set generated by the compact model, which is a prediction set of predictive distributions, with high probability. The criterion for selecting the final predictive distribution from the conformalized credal set, e.g., maximum entropy, ensures the informativeness of the resulting decision. Therefore, the proposed certified inference framework built on the compact model provides reliable and informative decisions, without requiring the large-scale model at run time. (see \emph{Chapter~\ref{chapter:5}}.)

    \item \textbf{Can we take a reliable compact model as a gold standard for reliability-efficiency co-design, or can we push the boundary further?} Due to the limited representational capacity of the compact model, one cannot always rely on the calibration-distilled compact model to process complex inputs. This raises an operational question, i.e., should we accept the compact model's decision, or escalate it to a human expert or a more powerful model? To address this limitation, we develop a conformal alignment-based model cascading framework, which sequentially screens incoming inputs to check if each input's decision is aligned with a human expert or a more powerful model, and escalates unaligned inputs to expert processing. The proposed method provides formal conditional reliability guarantees and preserves efficiency by minimizing the escalation, surpassing what any single-model solution achieves.  (see \emph{Chapter~\ref{chapter:6}}.)
    
\end{enumerate}

\section{Overview}
\label{sec:c1_overview}

In this work, we develop methods that make \gls{ai} inference both reliable and efficient. The thesis can coarsely be divided into two parts:

\begin{enumerate}
    \item The first part focuses on \emph{reliable inference}, developing methods that improve the trustworthiness of a model's uncertainty estimates through either empirical or certified approaches:

    \begin{enumerate}
        \item \textbf{Calibrating Bayesian learning}, which synergistically integrates calibration regularization, \gls{ood} confidence minimization, and selective inference into a unified Bayesian learning framework, enabling the model to reliably report its confidence on both \gls{id} and \gls{ood} inputs. The proposed method requires training from scratch and maintaining multiple models in an ensemble at inference time. The additional computational cost is justified by the substantial improvement in reliability over standard Bayesian approaches. (see \emph{Chapter~\ref{chapter:3}}.)

        \item \textbf{Certified reliability}, which leverages \gls{cp} to statistically guarantee that a user-specified reliability criterion is satisfied, by augmenting the pre-trained model with a prediction set. Unlike the best-effort approach in (a), this \gls{cp}-based method provides a finite-sample, distribution-free guarantee on reliability, ensuring that the probability of an undesirable outcome remains below a specified level. This method requires only a limited amount of calibration data and no retraining, making it applicable to any pre-trained model regardless of the model architecture or training procedure. (see \emph{Chapter~\ref{chapter:4}}.)
    \end{enumerate}

    \item The second part focuses on \emph{reliable-efficient co-design}, combining the aforementioned reliability tools with efficient inference strategies:

    \begin{enumerate}
        \item \textbf{Conformalized calibration distillation}, which distills calibration information from a large-scale cloud model to a small-scale edge model through offline post-processing. The reliability is formally guaranteed by ensuring that the conformalized credal set generated by the compact model contains the reference predictive distribution with high probability. At run time, inference is performed entirely on the compact model, eliminating the need to query the cloud model and thereby reducing both latency and communication overhead. The proposed method is particularly suitable for deployment scenarios where computational resources are limited or network connectivity is intermittent. (see \emph{Chapter~\ref{chapter:5}}.)

        \item \textbf{Collaborative edge-cloud inference}, which introduces a conformal alignment-based model cascading mechanism that allows the edge and cloud models to collaborate at inference time. Rather than relying exclusively on the edge model, the proposed method sequentially screens incoming inputs and evaluates whether the edge model's decision can satisfy the same level of reliability as if produced by the cloud model. This method provides formal conditional reliability guarantees, while preserving efficiency by minimizing escalation. The proposed method offers a principled way to advance the frontier of efficiency and reliability, surpassing what any single-model solution achieves. (see \emph{Chapter~\ref{chapter:6}}.)
    \end{enumerate}

    We integrate certified reliability tools from the first part with efficient inference strategies, i.e., calibration distillation and collaborative inference, to build a system that produces certifiably reliable and efficient decisions. 
\end{enumerate}

\section{The Reliability-Efficiency Co-Design Perspective}
\label{sec:c1_codesign}

This section reviews the literature relevant to the opening question in Sec.~\ref{sec:c1_motivation}. Sec.~\ref{sec:c1_reliability_related_work} surveys existing approaches to reliable inference, ranging from frequentist to Bayesian learning, from selective inference to \gls{cp}. Sec.~\ref{sec:c1_efficiency_related_work} reviews techniques for efficient inference, including model compression, knowledge distillation, low-complexity Bayesian methods, and collaborative inference. Together, these two subsections provide context for the reliability-efficiency co-design methods developed in this thesis.

\subsection{Reliable Inference}
\label{sec:c1_reliability_related_work}

\paragraph{Frequentist learning.}

Standard frequentist learning optimizes a single set of model parameters by minimizing the training loss, typically the \gls{ce}, and reports its reliability via the maximum confidence score~\citep{guo2017calibration}. However, such self-reported reliability tends to be unreliable, as the single point estimate discards epistemic uncertainty entirely. The well-documented observation that modern \gls{nn} and \gls{llm} tend to produce \emph{overconfident} decisions~\citep{guo2017calibration, naeini2015obtaining}, i.e., the self-reported reliability level exceeds the true reliability, confirms this limitation. Two directions have emerged to address this miscalibration: (\emph{i}) \emph{post-hoc} calibration, including temperature scaling~\citep{guo2017calibration} and Platt scaling~\citep{platt1999probabilistic}, which learns a simple mapping from raw confidence to calibrated confidence on a held-out calibration dataset; and (\emph{ii}) \emph{trainable} calibration, including \gls{mmce}~\citep{kumar2018trainable} and focal loss~\citep{mukhoti2020calibrating}, which incorporates calibration penalties directly into the training objective. While these methods can improve calibration in practice, they provide only empirical improvements without any statistical guarantee on the achieved reliability.

\begin{figure} [t] 
    \centering
    \centerline{\includegraphics[scale=0.28]{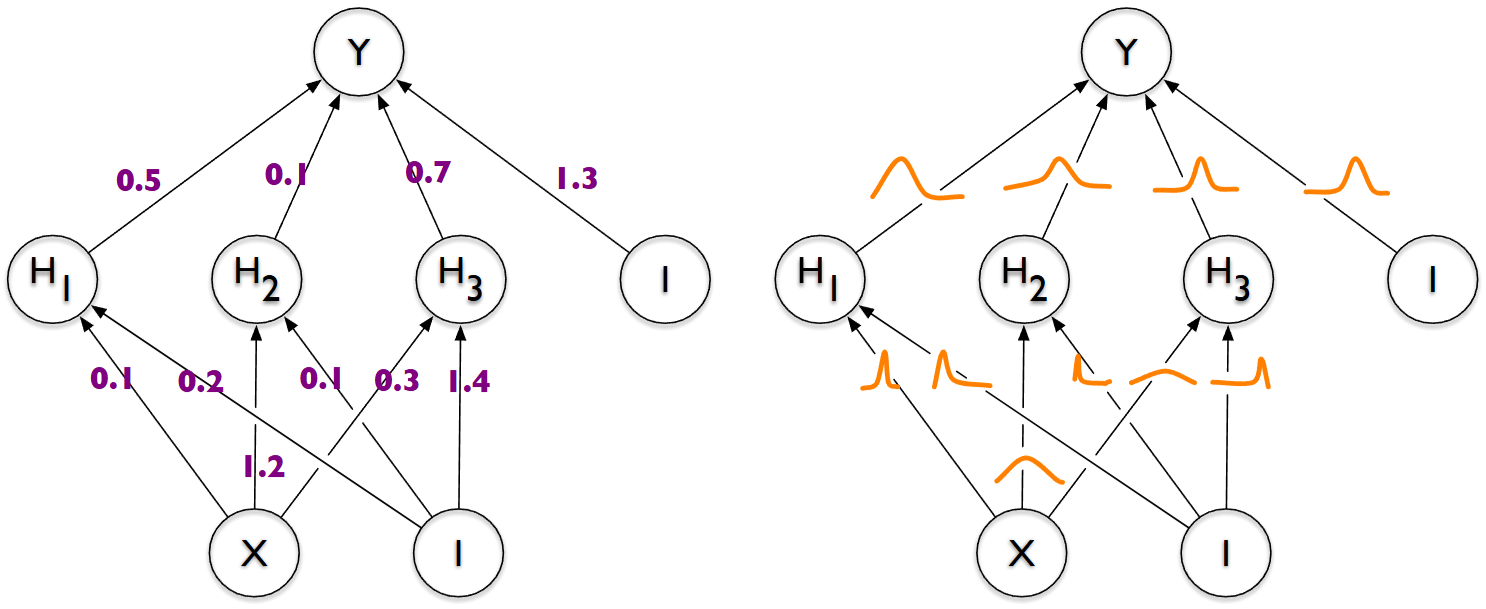}}
    \caption{A standard frequentist neural network (FNN) assigns a single deterministic value to each weight (left). A Bayesian neural network (BNN) replaces each deterministic weight with a probability distribution (right). Figure reproduced from~\citep{blundell2015weight}.}
    \label{fig:c1_fnn_bnn} 
\end{figure}

\paragraph{Bayesian learning.}

Bayesian learning~\citep{mackay2003information, simeone2022machine} addresses the epistemic limitation of frequentist learning by treating model parameters as random variables endowed with a prior distribution, and making predictions via ensembling over the posterior (see Fig.~\ref{fig:c1_fnn_bnn}). This captures epistemic uncertainty and can, in principle, produce more reliable predictions than frequentist learning. Since exact posterior inference is intractable, practical implementations rely on approximations. Specifically, \gls{vi}~\citep{blundell2015weight} approximates the true posterior with a tractable parametric distribution by minimizing \gls{kl} divergence between the two. Monte Carlo dropout~\citep{gal2016dropout} reinterprets dropout at test time as an approximate form of Bayesian inference, generating multiple stochastic forward passes to estimate the uncertainty. However, the calibration benefits of Bayesian methods are contingent on the model being well specified~\citep{masegosa2020learning, wenzel2020good}. In practice, misspecified priors, misspecified likelihoods, and the limited expressiveness of the variational family can all degrade calibration. Furthermore, Bayesian ensembling requires maintaining and running multiple models, which is computationally expensive and may exceed the resource budget of edge devices~\citep{singh2023edge}.

\paragraph{Selective inference.}

Selective inference allows a model to \emph{abstain} from predicting on inputs for which its decision is deemed unreliable, trading coverage for improved reliability on the accepted subset. Specifically, the selector filters out inputs for which the model's decisions tend to be poorly calibrated, and the remaining predictions are accepted from a region of the input space where the model's decisions are more reliable. Conventional \emph{selective classification}~\citep{geifman2019selectivenet} rejects inputs with low confidence to improve accuracy among accepted predictions. However, by favoring high-confidence inputs, it may exacerbate overconfidence on the accepted subset. \emph{Selective calibration}~\citep{fisch2022calibrated} addresses this by accepting inputs for which the gap between confidence and accuracy is expected to be small, directly targeting calibration rather than accuracy. By doing so, the selection criterion shifts from self-reported confidence to reliability, which provides a more principled basis for abstention.

\paragraph{Conformal prediction.}

A fundamentally different approach to reliable inference is provided by \gls{cp}~\citep{vovk2005algorithmic, shafer2008tutorial, angelopoulos2021gentle}, which constructs prediction sets rather than a single hard prediction using a held-out calibration dataset. \gls{cp} provides finite-sample, distribution-free coverage guarantees that hold for any pre-trained model with only a finite calibration sample. The only assumption is exchangeability between the calibration and test data. Unlike methods that calibrate a model's confidence on a point prediction, \gls{cp} shifts the reliability question from ``is the self-reported confidence well-calibrated?'' to ``does the prediction set contain the ground truth?'', statistically providing a \emph{certainty of uncertainty} without requiring retraining.

However, standard \gls{cp} provides only a \emph{marginal} coverage guarantee, while some safety-critical settings may require the stricter \emph{conditional coverage}, i.e., per-input coverage guarantees. Achieving such conditional coverage is generally impossible in a distribution-free setting~\citep{lei2014distribution, foygel2021limits}. Practical relaxations include group-conditional \gls{cp}~\citep{vovk2012conditional}, which ensures marginal coverage within each pre-defined group, and \gls{lcp}~\citep{tibshirani2019conformal, hore2025conformal}, which potentially improves conditional coverage by assigning input-dependent weights to non-conformity scores. In contrast, \emph{conformal alignment}~\citep{gui2024conformal} provides a selection-based framework that determines whether each output satisfies a desired alignment criterion, thereby approximating the conditional reliability guarantee.

The framework of \emph{\gls{ip}}~\citep{walley1991statistical} provides a complementary perspective on uncertainty representation, replacing a single predictive distribution with convex sets of distributions, known as \emph{credal sets}. Recent work has connected imprecise probability with \gls{cp} through conformalized credal set predictors~\citep{javanmardi2024conformalized, caprio2024Bayesian}, which combine the reliability guarantees of \gls{cp} with the richer uncertainty representation afforded by credal sets.

\subsection{Efficient Inference}
\label{sec:c1_efficiency_related_work}

\paragraph{Model compression.}

Model compression reduces the computational overhead of \gls{nn} and \gls{llm} to enable efficient deployment on resource-constrained devices. \emph{Quantization}~\citep{li2024evaluating, drumond2018training} reduces the numerical precision of weights and activations, e.g., from 32-bit floating point to 8-bit or lower integer representations, significantly reducing memory requirement and accelerating inference. \emph{Pruning}~\citep{lecun1989optimal} removes redundant parameters, yielding sparser models with lower computational cost, and can be applied at the level of individual weights or layers. Both techniques effectively reduce model deployment cost but inevitably introduce approximation errors that degrade not only accuracy but also, and more critically, reliability~\citep{zhu2023rethinking}. The degradation in reliability arises because the compressed model retains fewer parameters to represent the predictive distribution, which distorts the reliability estimates learned during training.

\paragraph{Knowledge distillation.}

\emph{Knowledge distillation}~\citep{hinton2015distilling} transfers the learned representations of a large \emph{teacher} model to a compact \emph{student} model by training the student to match the teacher's soft output distribution. The student learns inter-class relationships from the teacher, largely preserving predictive accuracy. Variants include attention transfer~\citep{guo2023class}, relational distillation~\citep{park2019relational}, and self-distillation~\citep{allen2020towards}. However, standard distillation objectives are designed to match the teacher's output predictions rather than to preserve the structure of its uncertainty estimates. Specifically, even when the student model achieves comparable accuracy, it does not necessarily inherit the teacher's reliability~\citep{zhu2023rethinking}, as the distillation objective does not explicitly enforce alignment between the student's and the teacher's confidence levels on individual inputs.

\paragraph{Collaborative inference.}

Rather than relying on a single model, \emph{model cascading}~\citep{marquez2018deep} allows models of different scales to collaborate at inference time. Inputs for which the lightweight model is sufficiently reliable are processed locally, while those that exceed its reliable operating range are outsourced to the more powerful model. A common strategy uses the small-scale model's maximum confidence as a proxy for escalation~\citep{fithian2014optimal, rabanser2025gatekeeper}, i.e., escalating the input to the larger model if its confidence falls below a pre-defined threshold. While such collaborative inference effectively reduces computational overhead, it lacks formal reliability guarantees and may produce unreliable deferral decisions under distribution shift. Recent work has begun integrating \gls{cp} into model cascading, e.g., using calibration data to determine the deferral threshold~\citep{yadkori2024mitigating} with statistical guarantees. However, existing methods provide only marginal coverage guarantees, meaning that reliability is ensured on average across all inputs but not for individual instances, which may be insufficient for safety-critical deployment scenarios where per-input reliability is required.

\section{Publications and Reproducibility}
\subsection{Publications}
This thesis includes the following works, listed by the order of chapter: \newline
[1] \textbf{J. Huang}, S. Park, and O. Simeone, ``Calibration-Aware Bayesian Learning,'' in \textit{IEEE 33rd International Workshop on Machine Learning for Signal Processing (MLSP 2023)}, Sept. 2023. \newline
$\Rightarrow$ Chapter \ref{chapter:3} builds upon this work, which is extended in [2], and its supplementary material is presented in Appendix \ref{app:a1}. \newline
[2] \textbf{J. Huang}, S. Park, and O. Simeone, ``Calibrating Bayesian Learning via Regularization, Confidence Minimization, and Selective Inference,'' \textit{IEEE Trans. Signal Process.}, vol. 73, pp. 4492-4505, 2025.\newline
$\Rightarrow$ This work serves as the foundation for Chapter \ref{chapter:3}, and its supplementary material is presented in Appendix \ref{app:a1}. \newline
[3] \textbf{J. Huang}, A.  Farzaneh, and O. Simeone, ``Optimized Certainty Equivalent Risk-Controlling Prediction Sets,'' \textit{arXiv}:2602.13660, 2026. (Submitted to European Signal Processing Conference) \newline
$\Rightarrow$ This work forms the basis for Chapter \ref{chapter:4}. \newline
[4] \textbf{J. Huang}, S. Park, N. Paoletti, and O. Simeone, ``Distilling Calibration via Conformalized Credal Inference,'' in \textit{International Joint Conference on Neural Networks (IJCNN 2025)}, Jul. 2025. (Oral, invited paper) \newline
$\Rightarrow$ This work forms the foundation for Chapter \ref{chapter:5}.  \newline
[5] \textbf{J. Huang}, S. Park, N. Paoletti, and O. Simeone, ``Reliable Inference in Edge-Cloud Model Cascades via Conformal Alignment,'' \textit{arXiv}:2510.17543, 2025. (Submitted to IEEE Trans. Pattern Anal. Mach. Intell.)  \newline
$\Rightarrow$ This work forms the basis for Chapter \ref{chapter:6}, and its supplementary material is presented in Appendix \ref{app:a2}.

\newpage

\subsection{Reproducibility}
For reproducibility purposes, we have made all of our code publicly available, as in Table~\ref{tab:c1_code}.
\begin{table}[ht]
\centering
\caption{Code repositories of thesis chapters.}\label{tab:c1_code}
\label{tab:comparison}
\begin{tabular}{lcc}
\toprule
\textbf{Thesis chapter} & \textbf{Code repository} \\
\midrule
Chapter~\ref{chapter:3} \citep{huang2023calibration, huang2025calibrating} & https://github.com/kclip/Calibrating-Bayesian-Learning \\
Chapter~\ref{chapter:4} \citep{huang2026optimized} & https://github.com/kclip/OCE-RCPS \\
Chapter~\ref{chapter:5} \citep{huang2025distilling} & https://github.com/kclip/Distilling-Calibration \\
Chapter~\ref{chapter:6} \citep{huang2025reliable} & https://github.com/kclip/Edge-Cloud-Conformal-Alignment \\
\bottomrule
\end{tabular}
\end{table}


\chapter{Two Sides of a Coin: Reliability and Efficiency} \label{chapter:2}

\ifpdf
    \graphicspath{{Chapter2/Figs/Raster/}{Chapter2/Figs/PDF/}{Chapter2/Figs/}}
\else
    \graphicspath{{Chapter2/Figs/Vector/}{Chapter2/Figs/}}
\fi

Reliability and efficiency are two sides of a coin, i.e, improving one often comes at the cost of the other. As motivated in Chapter~\ref{chapter:1}, a reliable model, such as a Bayesian model or a powerful cloud-based system, typically demands substantial computational resources, making deployment on resource-constrained edge devices impractical. Conversely, compact models tend to improve efficiency by sacrificing the capacity to provide trustworthy uncertainty estimates.

This chapter introduces the foundational concepts underlying both reliability and efficiency. Sec.~\ref{sec:c2_reliability} reviews two complementary notions of reliability: (\emph{i}) \emph{empirical reliability}, which improves calibration performance empirically without formal guarantees via the best-effort methods such as Bayesian learning, and (\emph{ii}) \emph{certified reliability}, which provides formal, distribution-free guarantees on prediction sets via statistical learning. Sec.~\ref{sec:c2_efficiency} introduces key strategies for efficient inference, including model compression, low-complexity Bayesian methods, knowledge distillation, and collaborative edge-cloud inference. Finally, Sec.~\ref{sec:c2_codesign_preview} connects these two perspectives and previews how the subsequent chapters progressively address the reliability-efficiency co-design challenge.

\section{Certainty of Uncertainty}
\label{sec:c2_reliability}

A reliable \gls{ai} model is one that not only makes accurate predictions, but also faithfully reports \emph{how uncertain} it is about those predictions. This section reviews two complementary approaches to achieving this goal. Sec.~\ref{sec:c2_empirical} introduces empirical reliability, which calibrates the model from scratch by improving the agreement between a model's confidence and its actual accuracy. Sec.~\ref{sec:c2_certified} introduces certified reliability, which moves beyond point predictions to \emph{set predictions} augmented with formal, distribution-free coverage guarantees, providing what we refer to as a \emph{certainty of uncertainty}.

\subsection{Empirical Reliability}
\label{sec:c2_empirical}

\paragraph{Properties of reliable probabilistic predictors.}

Given an \gls{id} training dataset $\mathcal{D}^{\text{tr}}_{\text{id}} = \{ (x_i, y_i)\}^{|\mathcal{D}^{\text{tr}}_{\text{id}}|}_{i=1}$, with $i$-th input $x_i \in \mathcal{X}$ and corresponding output $y_i \in \mathcal{Y}$ generated in an independent identically distributed (i.i.d.) manner by following an unknown, underlying joint distribution $p(x,y)$, supervised learning for classification optimizes a \emph{probabilistic predictor} $p(y|x, \mathcal{D}^{\text{tr}}_{\text{id}})$ of output $y$ given input $x$. A probabilistic predictor outputs a \emph{confidence level} $p(y|x, \mathcal{D}^{\text{tr}}_{\text{id}})$ for all possible outputs $y \in \mathcal{Y}$ given any input $x \in \mathcal{X}$ and dataset $\mathcal{D}^{\text{tr}}_{\text{id}}$. Given an input $x$, a \emph{hard} decision $\hat{y}$ can be obtained by choosing the output that has the maximum confidence level, i.e.,  
\begin{align} \label{eq:c2_hard_decision_general}
    \hat{y} (x) = \arg \max_{y \in \mathcal{Y}}  p(y|x,\mathcal{D}^{\text{tr}}_{\text{id}}).
\end{align}
Furthermore, the probabilistic predictor $p(y(x)|x, \mathcal{D}^{\text{tr}}_{\text{id}})$ provides a measure of the confidence of the prediction for a given input $x$ as
\begin{align} \label{eq:c2_basic_confidence}
    r(x) = p(\hat{y}(x)|x,\mathcal{D}^{\text{tr}}_{\text{id}}) = \max_{y \in \mathcal{Y}} p(y|x,\mathcal{D}^{\text{tr}}_{\text{id}}).
\end{align}

A probabilistic predictor is said to be \emph{perfectly \gls{id} calibrated} whenever its average prediction accuracy for any confidence level $r\in [0,1]$ equals $r$. This condition is formalized by the equality
\begin{align} \label{eq:c2_perfect_cal}
    \Pr \left[ y = \hat{y} | p(\hat{y}|x,\mathcal{D}^{\text{tr}}_{\text{id}}) = r \right] = r, \text{ for all }  r \in [0,1],
\end{align}
where the probability is taken with respect to the pairs of test data $(x,y)$. For \gls{id} calibration, one specifically assume the test pair $(x,y)$ to follow the same distribution $p(x,y)$ of the training data. Under the perfect \gls{id} calibration condition (\ref{eq:c2_perfect_cal}), among all input-output pairs $(x,y)$ that have the same confidence level $r$, the average fraction of correct decisions, with $\hat{y} (x) = y$, equals $r$. 

Importantly, perfect calibration does not imply high accuracy \citep{wang2023calibration, tao2023benchmark}. For instance, if the marginal distribution of the labels $y \in \mathcal{Y}$ is uniform, a model that disregards the input $x$, assigning a uniform confidence $p(y|x, \mathcal{D}^{\text{tr}}_{\text{id}}) = 1/|\mathcal{Y}|$,  achieves perfect calibration, while offering a generally low accuracy.

The extent to which condition (\ref{eq:c2_perfect_cal}) is satisfied is typically evaluated via the \emph{reliability diagram} \citep{degroot1983comparison} and the \emph{\gls{ece}} \citep{guo2017calibration}. Both approaches estimate the probability in (\ref{eq:c2_perfect_cal}) using a test dataset $\mathcal{D}^\text{te}=\{ (x_i^\text{te}, y_i^\text{te}) \}_{i=1}^{|\mathcal{D}^\text{te}|}$, with each pair $(x_i^\text{te}, y_i^\text{te})\sim p(x,y)$ being drawn independently of the training dataset $\mathcal{D}^{\text{tr}}_{\text{id}}$.

\begin{figure} [t] 
    \centering
    \centerline{\includegraphics[scale=0.4]{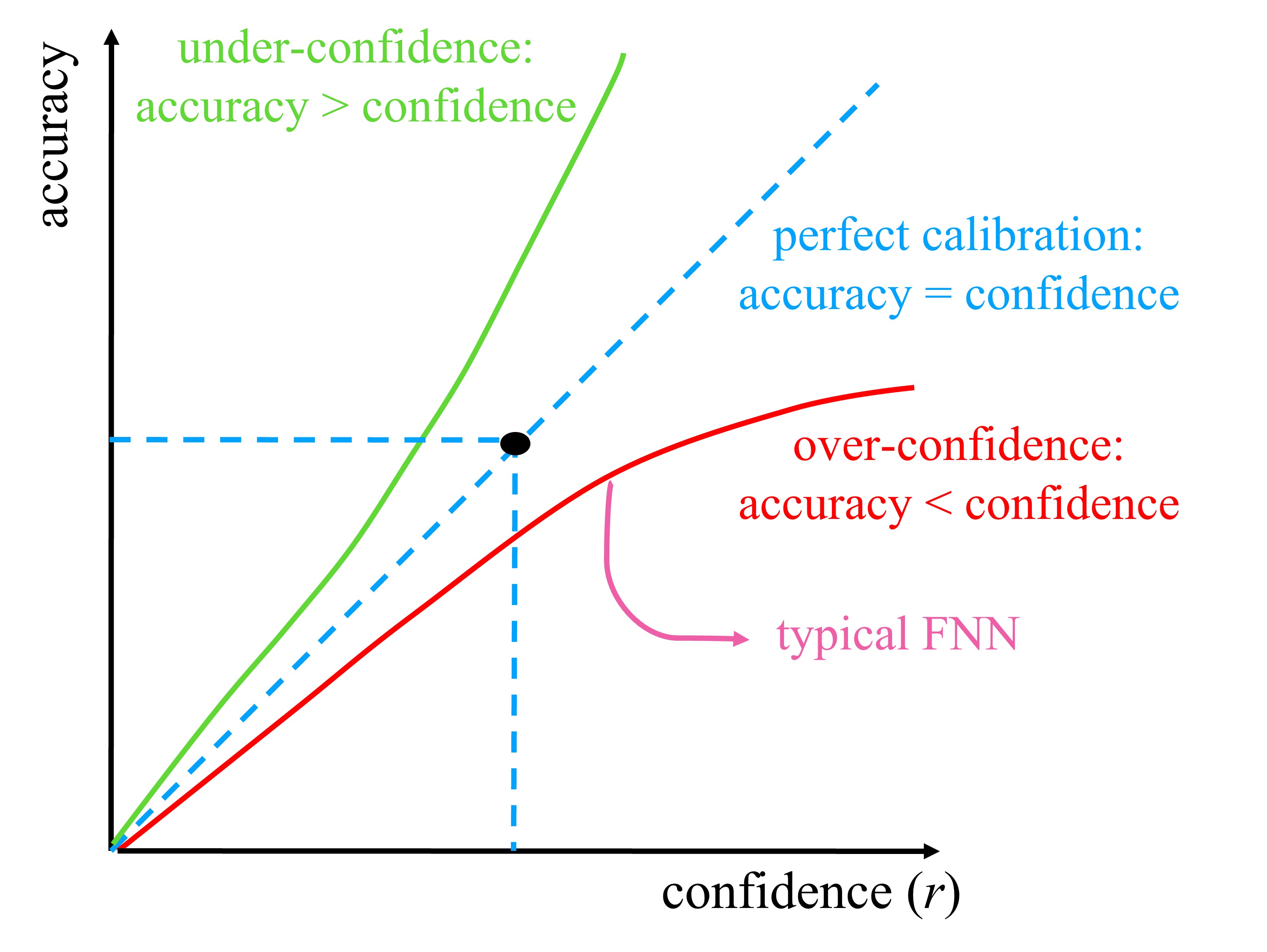}}
    \caption{Reliability diagrams visualize the calibration performance (\ref{eq:c2_perfect_cal}) of the model by evaluating the average accuracy over test examples to which the prediction has the same confidence value $r$. Typically, \gls{fnns} return over-confident decisions, for which the accuracy is lower than the confidence obtained by the model.}
    \label{fig:c2_reliability_diagram_illustration} 
\end{figure}

As illustrated in Fig.~\ref{fig:c2_reliability_diagram_illustration}, a \emph{reliability diagram} plots the average confidence produced by the model for all inputs with the same confidence level $r$ in (\ref{eq:c2_perfect_cal}). By (\ref{eq:c2_perfect_cal}), perfect calibration is obtained when the curve is aligned with the diagonal line, i.e., the dashed line in Fig.~\ref{fig:c2_reliability_diagram_illustration}. In contrast, a curve below the diagonal line, e.g., the red line in Fig.~\ref{fig:c2_reliability_diagram_illustration}, indicates over-confidence, while a curve above the diagonal line, such as the green line in Fig.~\ref{fig:c2_reliability_diagram_illustration}, implies an under-confident prediction. 

To elaborate further on calibration measure, denote the \emph{confidence score} associated with the $i$-th test input $x_i^\text{te}$ as 
\begin{align} \label{eq:c2_confidence_score}
    r_i^\text{te} = p(\hat{y}_i^\text{te} (x_i^\text{te}) | x_i^\text{te}, \mathcal{D}^{\text{tr}}_{\text{id}}),
\end{align}
given the hard decision (\ref{eq:c2_hard_decision_general}), and the corresponding \emph{accuracy score} as
\begin{align} \label{eq:c2_correctness_score}
    c_i^\text{te} = \mathbbm{1} (\hat{y}_i^\text{te} (x_i^\text{te}) = y_i^\text{te}),
\end{align}
with the indicator function $\mathbbm{1} (\cdot)$ defined as $\mathbbm{1} (\text{true}) = 1$ and $\mathbbm{1} (\text{false}) = 0$. 
To enable an estimate of the reliability diagram, we partition the test data points into \emph{bins} with approximately the same confidence level (\ref{eq:c2_confidence_score}). Specifically, the $m$-th bin $\mathcal{B}_m$ contains indices $i$ of the test inputs that have confidence scores lying in the interval $(\frac{m-1}{M}, \frac{m}{M} ]$, i.e., $\mathcal{B}_m = \{i \in \{1,...,M\}: r_i^\text{te} \in ( \frac{m-1}{M}, \frac{m}{M} ] \}$.

Each $m$-th bin is associated with a \emph{per-bin confidence} $\mathrm{conf}(\mathcal{B}_m) ={1}/{|\mathcal{B}_{m}|} \sum_{i \in \mathcal{B}_m} r_i^\text{te}$, and with a \emph{per-bin accuracy} $\mathrm{acc}(\mathcal{B}_m) = {1}/{|\mathcal{B}_{m}|} \sum_{i \in \mathcal{B}_m}  c_i^\text{te}$. An estimate of the reliability diagram is now obtained by plotting $\mathrm{acc}(\mathcal{B}_m)$ as a function of $\mathrm{conf}(\mathcal{B}_m)$ across all bins $m=1,...,M$.

A scalar calibration metric can be extracted from a reliability diagram by evaluating the average discrepancy between the per-bin confidence $\mathrm{conf}(\mathcal{B}_m)$ and the per-bin accuracy $\mathrm{acc}(\mathcal{B}_m)$. Via weighting the contribution of each bin by the corresponding fraction of samples, the \gls{ece} is defined as \citep{naeini2015obtaining}
\begin{align} \label{eq:c2_ece}
    \mathrm{ECE} = \sum^{M}_{m=1} \frac{|\mathcal{B}_{m}|}{ \sum_{m'=1}^M |\mathcal{B}_{m'}|} \left | \mathrm{acc}(\mathcal{B}_{m}) - \mathrm{conf}(\mathcal{B}_m) \right |.
\end{align}

\paragraph{Frequentist learning.}

\begin{figure} [tb] 
    \centering
    \centerline{\includegraphics[scale=0.3]{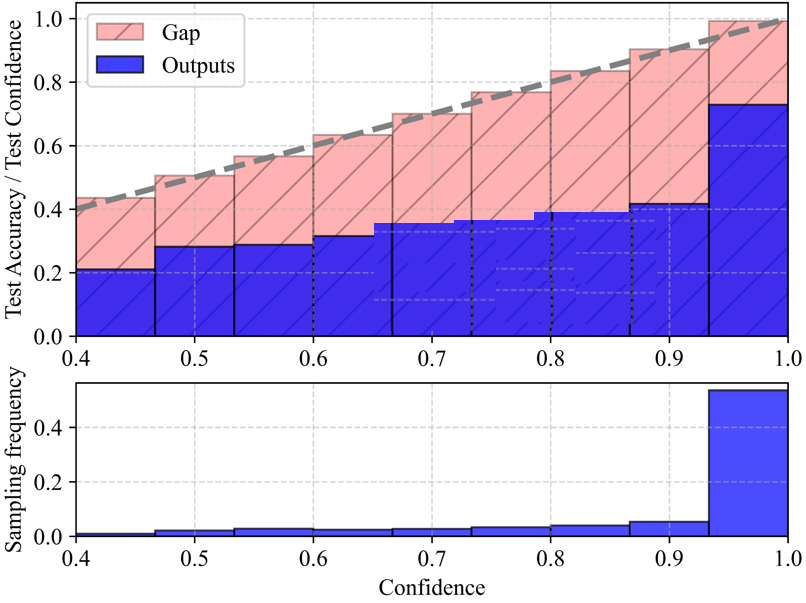}}
        \caption{Reliability diagram of FNN, illustrating the overconfidence phenomenon.}
    \label{fig:c2_fnn} 
\end{figure}

Given a parameterized vector of probabilistic predictor $p(y|x,\theta)$, standard FNN training aims at finding a parameter vector $\theta^\text{FNN}$ that minimizes the training loss $\mathcal{L}(\theta|\mathcal{D}^{\text{tr}}_{\text{id}})$. The loss function is typically defined as the negative log-likelihood, or \gls{ce}, yielding the solution    
\begin{align} \label{eq:c2_FNN}
    \theta^\text{FNN} = \arg\min_{\theta} \bigg\{ \mathcal{L}(\theta|\mathcal{D}^{\text{tr}}_{\text{id}}) = -\sum_{(x,y)\in\mathcal{D}^{\text{tr}}_{\text{id}} } \log p(y|x,\theta) \bigg\}.
\end{align} 
Accordingly, the trained probabilistic predictor is obtained as
\begin{align}
    p(y|x,\mathcal{D}^{\text{tr}}_{\text{id}}) = p(y|x, \theta^{\text{FNN}}).
\end{align}
Then, given an input $x$, the trained model produces a \emph{point prediction}
\begin{equation}
    \hat{y}(x) = \arg\max_{y \in \mathcal{Y}} p(y|x, \theta^{\text{FNN}}),
    \label{eq:c2_hard_decision_fnn}
\end{equation}
with associated confidence level and correctness score
\begin{equation}
    r(x) = \max_{y \in \mathcal{Y}} p(y|x, \theta^{\text{FNN}}), \quad c(x) = \mathbbm{1} (\hat{y}(x) = y).
    \label{eq:c2_confidence}
\end{equation}
As shown in Fig.~\ref{fig:c2_fnn}, the confidence $r(x)$ is often miscalibrated, systematically exceeding the true probability of correctness, i.e., a phenomenon known as \emph{overconfidence} \citep{guo2017calibration}, which makes the model's probabilistic outputs unreliable for downstream decision-making, particularly in safety-critical applications.

However, this approach is statistically valid only when the empirical loss $\mathcal{L}(\theta|\mathcal{D}^{\text{tr}}_{\text{id}})$ approximates the population loss well, i.e.,
\begin{equation}
    \mathcal{L}(\theta|\mathcal{D}^{\text{tr}}_{\text{id}}) \approx - \sum_{(x,y) \sim p(x,y)} \log p(y|x, \theta).
    \label{eq:c2_population_loss}
\end{equation}

An ideal learner would find a parameter vector $\theta^{\text{FNN}}$ that yields a \emph{well-calibrated} predictor matching the ground-truth conditional distribution, i.e.,
\begin{equation}
    p(y|x, \theta^{\text{FNN}}) \approx p(y|x, \theta^{\text{gt}}),
    \label{eq:c2_ideal_calibration}
\end{equation}
where $\theta^\text{gt}$ denotes the ground-truth parameter vector. For this ideal outcome (\ref{eq:c2_ideal_calibration}) to be realized, two assumptions are required. The first, known as \emph{well-specification} \citep{simeone2022machine}, requires that the model family $\{p(y|x, \theta) : \theta \in \Theta\}$ is expressive enough to represent the ground truth conditional distribution, i.e., there exists some $\theta \in \Theta$ such that
\begin{equation}
    p(y|x, \theta^\text{gt}) = p(y|x, \theta), \quad \text{ for some } \theta \in \Theta.
    \label{eq:c2_well_specification}
\end{equation}
The second assumption requires sufficient training data $\mathcal{D}^{\text{tr}}_{\text{id}}$ so that minimizing the empirical risk $\mathcal{L}(\theta|\mathcal{D}^{\text{tr}}_{\text{id}})$ yields a parameter vector close to the ideal one, i.e., $\theta^\text{FNN} \approx \theta^\text{gt}$.

In practice, both assumptions are inevitably violated. Finite training data causes $\theta^{\text{FNN}}$ to deviate from $\theta^{\text{gt}}$, introducing \emph{epistemic uncertainty} \citep{simeone2022machine, mackay2003information}. Frequentist learning, by committing to a single point estimate $\theta^{\text{FNN}}$, discards this uncertainty entirely.

\paragraph{Bayesian learning}

The overconfidence of frequentist learning stems partly from its reliance on a single parameter vector $\theta^{\text{FNN}}$, discarding the \emph{epistemic uncertainty}. Bayesian learning addresses this by treating $\theta$ as a random variable with a prior distribution $p(\theta)$~\citep{mackay2003information, simeone2022machine}.

Specifically, \gls{bnns} training leverages \emph{prior} knowledge on the \emph{distribution} $p(\theta)$ of the model parameter vector $\theta$, and it aims at finding a \emph{distribution} $q(\theta)$ over the parameter vector $\theta$ that represents the learner's uncertainty in the model parameter space. The learning objective $\mathcal{F}(q|\mathcal{D}^{\text{tr}}_{\text{id}})$, known as \emph{free energy}, accounts for the average loss $\mathbb{E}_{\theta \sim q(\theta) }[\mathcal{L}(\theta|\mathcal{D}^{\text{tr}}_{\text{id}})]$, as well for discrepancy between the distribution $q(\theta)$ and prior knowledge $p(\theta)$ as per the sum
\begin{align} \label{eq:c2_free_energy}
    \mathcal{F}(q|\mathcal{D}^{\text{tr}}_{\text{id}}) = \mathbb{E}_{\theta \sim q(\theta) }[\mathcal{L}(\theta|\mathcal{D}^{\text{tr}}_{\text{id}})] + \beta \cdot \operatorname{KL}(q(\theta) || p(\theta)).
\end{align}
In (\ref{eq:c2_free_energy}), the discrepancy between distribution $q(\theta)$ and prior $p(\theta)$ is captured by the \gls{kl} divergence
\begin{align} \label{eq:c2_kl_term}
    \operatorname{KL} (q(\theta) \| p(\theta)) = \mathbb{E}_{q(\theta)} \left[\log \left(\frac{q(\theta)}{p(\theta)}\right) \right],
\end{align}
where we have introduced a hyperparameter $\beta > 0$. The \gls{kl} term within free energy (\ref{eq:c2_free_energy}) reduces the calibration error by enforcing the adherence of the model parameter distribution $q(\theta)$ to the prior distribution $p(\theta)$.

Accordingly, BNN learning yields the optimized distribution $q^\text{BNN}(\theta)$ by addressing the minimization of the free energy as per
\begin{align} \label{eq:c2_BNN}
    q^\text{BNN}(\theta) = \arg\min_{q(\theta)} \mathcal{F}(q|\mathcal{D}^{\text{tr}}_{\text{id}}),
\end{align}
\begin{figure} [tb] 
    \centering
    \centerline{\includegraphics[scale=0.26]{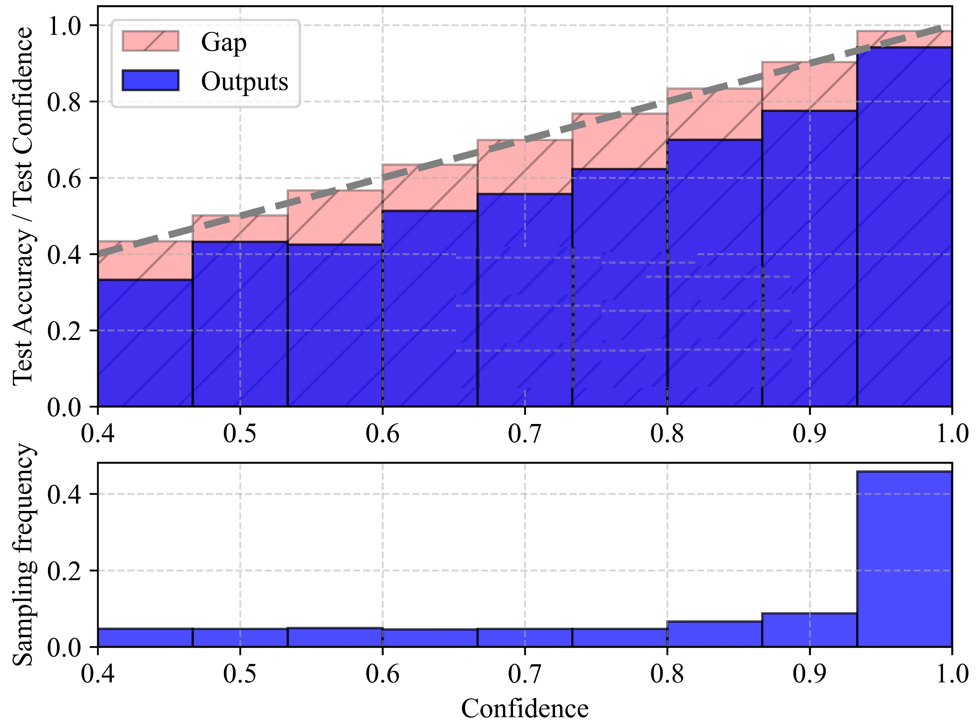}}
    \caption{Reliability diagram for BNN, showing improved calibration over FNN.}
    \label{fig:c2_bnn} 
\end{figure}and the corresponding trained probabilistic predictor is obtained as the \emph{ensemble} \citep{simeone2022machine}
\begin{align} \label{eq:c2_ensmble}
    p(y|x,\mathcal{D}^{\text{tr}}_{\text{id}}) = \mathbb{E}_{ \theta \sim q^\text{BNN}(\theta)} \big[ p(y|x, \theta)\big]. 
\end{align}
Accordingly, the predictive probability distribution (\ref{eq:c2_ensmble}) requires averaging the output distribution $p(y|x,\theta)$ over the model parameter vector $\theta$ sampled from the distribution $q^\text{BNN}(\theta)$, which accounts for model uncertainty and can produce better-calibrated predictions than a frequentist model (\ref{eq:c2_FNN}).

In practice, the optimization (\ref{eq:c2_BNN}) over distribution $q(\theta)$ is often carried out over a parameterized class of distribution via \emph{\gls{vi}} \citep{blundell2015weight, gal2016dropout} (see \citep[Ch. 12]{simeone2022machine} for an overview). Specifically, denoting as $q(\theta|\varphi)$ a distribution over parameter vector $\theta$ parameterized by a vector $\varphi$, the problem (\ref{eq:c2_BNN}) is simplified as
\begin{align} \label{eq:c2_simple_BNN}
    \varphi^{\text{BNN}} = \arg \min_{q(\theta|\varphi)} \mathcal{F}(q|\mathcal{D}^{\text{tr}}_{\text{id}}),
\end{align}
yielding the optimized variational distribution $q^{\text{BNN}}(\theta) = q(\theta|\varphi^{\text{BNN}})$. The parameter vector $\theta$ is typically assumed to follow a Gaussian distribution, with mean $\mu$ and covariance $\Sigma$ included in the variational parameters as $\varphi = [\mu, \Sigma]$. Furthermore, the covariance $\Sigma$ is often constrained to be diagonal with each $i$-th diagonal term modelled as $\Sigma_{ii} = \exp(\rho_i)$ for a learnable vector $\rho = [\rho_1,...,\rho_{n_p}]$, where $n_p$ is the size of the model parameter vector $\theta$. For a Gaussian variational distribution, problem (\ref{eq:c2_simple_BNN}) can be addressed via the reparametrization trick \citep{mohamed2020monte}. 

However, the calibration benefits of Bayesian learning are contingent on the model being well specified \citep{masegosa2020learning, wenzel2020good}. In practice, misspecified priors, misspecified likelihoods, and the limited expressiveness of the variational family can all degrade calibration, yielding predictions no better than frequentist models, which motivates the calibration-aware training methods developed in Chapter~\ref{chapter:3}.

\paragraph{Selective inference.}

A complementary approach to improving reliability is to allow the model to \emph{abstain} from predicting on inputs for which the target criterion cannot be satisfied. This idea, known as \emph{selective inference}, augments the model with a \emph{selector} $g(x|\phi) \in \{0, 1\}$ parameterized by a vector $\phi$. An input is accepted if the selector outputs $g(x|\phi) =1$, and is rejected if $g(x|\phi)=0$, i.e.,
\begin{align} \label{eq:c2_selector}
    g(x|\phi) = \begin{cases}
        1, & \text{input } x \text{ is accepted,} \\
        0, & \text{input } x \text{ is abstained.}
    \end{cases}
\end{align}

As an example, given a \emph{pre-trained} model $ p(y|x, \mathcal{D}^{\text{tr}}_{\text{id}})$, conventional selective classification targets accuracy among accepted predictions. Specifically, to design the corresponding parametrized selector $g(x|\phi)$, assuming that we have access to a held-out \emph{validation dataset} $\mathcal{D}^\text{val}=\{ (x_i, y_i) \}_{i=1}^{|\mathcal{D}^\text{val}|}$, the \emph{selector validation loss} is the loss evaluated only on the accepted samples of the dataset $\mathcal{D}^{\text{val}}$, i.e.,
\begin{align} \label{eq:c2_sel-cla-loss}
    \mathcal{R}(\phi|\mathcal{D}^{\text{val}}) = - \frac{\sum_{(x,y)\in \mathcal{D}^{\text{val}}} g(x|\phi) \cdot \log p(y|x, \mathcal{D}^{\text{tr}}_{\text{id}})}{\sum_{(x,y)\in \mathcal{D}^{\text{val}}} g(x|\phi)},
\end{align}
where $\sum_{x \in \mathcal{D}^\text{val}} g(x|\phi) $ is the number of accepted validation examples.

The optimization of the selective classification is formulated in \citep{geifman2019selectivenet} as the minimization of the selective validation loss (\ref{eq:c2_sel-cla-loss}) subject to a target coverage rate $\xi$, with $0 \leq \xi \leq 1$, i.e.,
\begin{align} \label{eq:c2_sel-cla}
    \phi^{\text{S-Cla}} = \arg \min_{\phi} \mathcal{R}(\phi|\mathcal{D}^{\text{val}}) \quad \text{s.t.} \quad \frac{1}{|\mathcal{D}^{\text{val}}|} \sum_{x \in \mathcal{D}^{\text{val}}} g(x|\phi) \geq \xi.
\end{align}

By adopting the design criterion (\ref{eq:c2_sel-cla-loss}), selective classification aims at choosing only the inputs $x$ for pairs $(x,y)$ with large confidence levels $p(y|x, \mathcal{D}^{\text{tr}}_{\text{id}})$. As anticipated, this objective is sensible if the underlying model $p(y|x, \mathcal{D}^{\text{tr}}_{\text{id}})$ is sufficiently well calibrated. Reference \citep{geifman2019selectivenet} addressed the constrained optimization in (\ref{eq:c2_sel-cla}) by converting it into an unconstrained problem via a quadratic penalty function (see \citep[Eq. (2)]{geifman2019selectivenet}). 

However, selective classification targets accuracy rather than calibration, and by favoring high-confidence inputs, it may exacerbate overconfidence on the accepted subset, which motivates the investigation of \emph{selective calibration}, detailed in Chapter \ref{chapter:3}.

\subsection{Certified Reliability}
\label{sec:c2_certified}

The empirical reliability tools reviewed in Sec.~\ref{sec:c2_empirical} only claim to improve the calibration performance empirically, without providing any formal reliability guarantee. Their performance remains vulnerable to model misspecification and limited data, and cannot certify the extent to which a model's decisions are reliable. \emph{Certified reliability} addresses these limitations by moving beyond point predictions to \emph{set predictions} with formal, distribution-free coverage guarantees.

\paragraph{Properties of set predictions.}

The point prediction $\hat{y}(x)$ in (\ref{eq:c2_hard_decision_general}) returns a single label whose reliability depends on the model's confidence estimates, which, as shown in Fig.~\ref{fig:c2_fnn}, is often miscalibrated. An alternative is to produce a \emph{prediction set} $\Gamma(x) \subseteq \mathcal{Y}$, i.e., a subset of labels. This shifts the question from ``is the self-reported confidence reliable?'' to ``does the prediction set contain the truth?''

Given a function $J(x, y)$ measuring the discrepancy between the prediction of a model $p(y|x, \mathcal{D}^{\text{tr}}_{\text{id}})$ and the label $y$, such as the negative log-loss $J(x, y) = -\log p(y|x, \mathcal{D}^{\text{tr}}_{\text{id}})$. Then, given an input $x$, one can construct a prediction set by including all labels with discrepancy below a threshold $q$, i.e.,
\begin{equation}
    \Gamma(x) = \{y \in \mathcal{Y} : J(x,y) \leq q\}.
    \label{eq:c2_general_set}
\end{equation}
More generally, the prediction set can be a subset of the label space as in (\ref{eq:c2_general_set}) or a subset of the probability simplex $\mathcal{P}$ over $\mathcal{Y}$, as detailed in Chapter \ref{chapter:5}.

The quality of a prediction set is characterized by two properties: (\emph{i}) \emph{reliability}, known as \emph{coverage}, is measured by the probability that the prediction set contains the ground truth label, i.e.,
\begin{align} \label{eq:c2_coverage_def}
    \Pr\left[y \in \Gamma(x)\right],
\end{align}
where the probability is taken with respect to the test pair $(x, y)$, and (\emph{ii}) \emph{informativeness}, known as \emph{inefficiency}, is measured by the expected normalized size, i.e.,
\begin{align} \label{eq:c2_inefficiency}
    \mathbb{E}\left[\frac{|\Gamma(x)|}{|\mathcal{Y}|}\right],
\end{align}
where the expectation is taken with respect to the test input $x$.

The threshold $q$ (\ref{eq:c2_general_set}) governs the coverage-inefficiency trade-off. A trivially set $\Gamma(x) = \mathcal{Y}$ maximizes coverage but conveys no information, and a singleton set $\Gamma(x) = \{\hat{y}(x)\}$ is maximally informative but may fail to cover the ground truth label. A well-designed set predictor achieves target coverage with minimal inefficiency.

A fundamental question remains: \emph{how should the threshold $q$ be chosen?} If the model were well-calibrated, the threshold $q$ could be set from the model's own confidence. However, as established in Sec.\ref{sec:c2_empirical}, model self-reported confidence is often unreliable. This motivates a \emph{data-driven} approach to threshold selection (\ref{eq:c2_general_set}) with formal guarantees, which is precisely what \gls{cp} provides.

\paragraph{Conformal prediction.}

\gls{cp}~\citep{vovk2005algorithmic, shafer2008tutorial} provides a principled method for selecting threshold $q$ (\ref{eq:c2_general_set}) so that the prediction set $\Gamma(x)$ satisfies a formal coverage guarantee. The key idea is to determine the threshold empirically from a held-out calibration dataset, leveraging exchangeability to translate empirical quantiles into population-level guarantees.

The standard split \gls{cp} pipeline operates as follows. Given a pre-trained model and a calibration dataset $\mathcal{D}^{\text{cal}} = \{(x_i, y_i)\}_{i=1}^{|\mathcal{D}^{\text{cal}}|}$ drawn exchangeably with the test data, one computes the nonconformity score $J(x,y)$ on the calibration dataset as
\begin{align} \label{eq:c2_cal_scores}
    \mathcal{J} = \{J(x_i, y_i)\}_{i=1}^{|\mathcal{D}^{\text{cal}}|}.
\end{align}
Then, the threshold $q$ is selected as the $\lceil(1 - \alpha_{\text{label}}^{\text{mis}})(1 + |\mathcal{D}^{\text{cal}}|)\rceil$-th smallest element of $\mathcal{J} \cup \{\infty\}$, i.e.,
\begin{align} \label{eq:c2_cp_quantile}
    q = \text{Quantile}_{1-\alpha_{\text{label}}^{\text{mis}}}\left(\frac{1}{1+|\mathcal{D}^{\text{cal}}|} \sum_{i=1}^{|\mathcal{D}^{\text{cal}}|} \delta_{J(x_i,y_i)} + \frac{1}{1+|\mathcal{D}^{\text{cal}}|} \delta_{\infty}\right),
\end{align}
where $\delta_J$ is a point mass at $J$ and the function $\text{Quantile}_{1-\alpha_{\text{label}}^{\text{mis}}}(\cdot)$ returns the smallest value with cumulative weight at least $1 - \alpha_{\text{label}}^{\text{mis}}$. The point mass at $\infty$ ensures validity in the finite-sample regime. 

Augmented with the threshold $q$ (\ref{eq:c2_cp_quantile}), \gls{cp} provides marginal validity guarantees \citep[Eq.~(1)]{angelopoulos2021gentle}, that is, the prediction set $\Gamma(x)$ \label{eq:c2_general_set} satisfies the condition
\begin{align} \label{eq:c2_cp_guarantee}
    \Pr[y \in \Gamma(x)] \geq 1-\alpha_{\text{label}}^{\text{mis}},
\end{align}
where the probability is taken with respect to both the test pair $(x, y)$ and calibration data $\mathcal{D}^{\text{cal}}$. The marginal guarantee (\ref{eq:c2_cp_guarantee}) is \emph{distribution-free}, which holds regardless of the data distribution, model architecture, or model quality, referred to as a \emph{certainty of uncertainty}.

However, while coverage is guaranteed regardless of model quality, \emph{informativeness} is not. For instance, a poorly calibrated model produces large confidence scores, resulting in high threshold $q$ and uninformative sets, i.e., formally valid but practically useless. This observation motivates the framework developed in Chapters \ref{chapter:5} and Chapter \ref{chapter:6}, i.e., when the base model is weak and poorly calibrated, certified methods alone are insufficient, and one must either improve the model via distillation, see Chapter~\ref{chapter:5}, or selectively leverage a stronger model, see Chapter~\ref{chapter:6}.

\paragraph{Marginal versus conditional coverage.}
The marginal coverage guarantee (\ref{eq:c2_cp_guarantee}) averages over the input population, i.e.,
\begin{align} \label{eq:c2_marginal_decomposition}
    \Pr\left[y \in \Gamma(x)\right] = \mathbb{E}_{x \sim p(x)}\left[\Pr\left[y \in \Gamma(x) \,\middle|\, x\right]\right] \geq 1 - \alpha_{\text{label}}^{\text{mis}},
\end{align}
which does not prevent coverage from being poor on specific inputs. For instance, a model achieving $99\%$ coverage on easy inputs but $50\%$ on hard inputs may satisfy (\ref{eq:c2_marginal_decomposition}) while being unreliable where it matters most.

Unlike marginal coverage guarantee (\ref{eq:c2_cp_guarantee}), \emph{conditional coverage guarantee} requires strict per-input coverage guarantee, i.e.,
\begin{align} \label{eq:c2_conditional_coverage}
    \Pr\left[y \in \Gamma(x) \,\middle|\, x\right] \geq 1 - \alpha_{\text{label}}^{\text{mis}}, \quad \text{for all } x \in \mathcal{X}.
\end{align}
As observed, any set predictor $\Gamma(x)$ satisfying (\ref{eq:c2_conditional_coverage}) also satisfies (\ref{eq:c2_cp_guarantee}), but not conversely.

Exact conditional coverage guarantee (\ref{eq:c2_conditional_coverage}) is generally impossible in a distribution-free setting \citep{lei2014distribution, hore2025conformal}, since it requires localizing coverage guarantees to individual test inputs without knowledge of the ground truth conditional distribution. This has motivated relaxations including group-conditional \gls{cp} \citep{vovk2012conditional},  \gls{lcp} \citep{tibshirani2019conformal, hore2025conformal}, and the conformal alignment framework in Chapter \ref{chapter:6}.

\section{Strategies for Effective Inference}
\label{sec:c2_efficiency}

As discussed in Sec.~\ref{sec:c2_reliability}, reliable models tend to be complex and demand substantial computational resources, making efficient deployment impractical \citep{singh2023edge}. This section reviews key strategies for efficient inference, including model compression in Sec.~\ref{sec:c2_compression}, low-complexity Bayesian methods in Sec.~\ref{sec:c2_laplace}, knowledge distillation in Sec.~\ref{sec:c2_distillation}, and collaborative inference in Sec.~\ref{sec:c2_cascading}.

\subsection{Model Compression}
\label{sec:c2_compression}

\emph{Model quantization}~\citep{li2024evaluating} reduces the computational overhead of a model by representing weights and activations with reduced precision. Given a full-precision weight $w_{full} \in \mathbb{R}$, \emph{uniform quantization} maps it to
\begin{align} \label{eq:c2_quantization}
    \hat{w}_{quant} = \tau_{quant} \cdot \min \left ( \max \left(\left\lfloor \frac{w_{full}}{\tau_{quant}} \right\rceil, \, -2^{n_b-1}, \right) \, 2^{n_b-1} - 1\right),
\end{align}
where $n_b$ is the number of bits used for quantization, $\tau_{quant} > 0$ is a scaling factor, and $\lfloor \cdot \rceil$ operation denotes rounding. For a model with $n_p$ parameters, quantization reduces memory to $n_b \times n_p$ bits.

While effective at reducing memory and accelerating inference, the quantization strategy may degrade accuracy and calibration performance as the approximation errors introduced by the rounding operation in (\ref{eq:c2_quantization}). This degradation motivates the calibration distillation method in Chapter \ref{chapter:5}, which demonstrates that the proposed calibration distillation method can compensate for the reliability loss induced by small model.

\subsection{Bayesian Inference via Laplace Approximation}
\label{sec:c2_laplace}
The Bayesian learning framework in Sec.~\ref{sec:c2_empirical} provides a principled route to calibration via ensembling, but the variational inference procedure (\ref{eq:c2_simple_BNN}) can be computationally expensive, particularly for large models. The \emph{Laplace approximation} \citep{daxberger2021laplace, simeone2022machine} offers a low-complexity alternative by constructing a Gaussian approximation to the posterior centered at a pre-trained point estimate.

To elaborate, given a small model $p^e(y|x) = p^e(y|x, \hat{\theta})$ with pre-trained parameter vector $\hat{\theta}$, the Laplace approximation approximates the posterior distribution of model parameter $\theta$ given a training dataset $\mathcal{D}^{\text{tr}}_{\text{id}}$ as
\begin{align} \label{eq:c2_laplace_posterior}
    p(\theta|\mathcal{D}^{\text{tr}}_{\text{id}}) \approx \mathcal{N}(\hat{\theta}, \Sigma), \quad \text{with} \quad \Sigma := \left( \nabla^2_{\theta} \mathcal{L}(\theta|\mathcal{D}^{\text{tr}}_{\text{id}}) \big|_{\theta = \hat{\theta}} \right)^{-1},
\end{align}
where $\mathcal{L}(\theta|\mathcal{D}^{\text{tr}}_{\text{id}})$ is the \gls{ce} training loss \citep{daxberger2021laplace}. In practice, the covariance matrix $\Sigma$ can be approximated in several ways, e.g., via the Gauss-Newton method \citep{lecun1989optimal}. 

Given the approximate posterior (\ref{eq:c2_laplace_posterior}), a predictive distribution can be made via Bayesian model ensembling as
\begin{align} \label{eq:c2_laplace_prob}
    q^{\text{La}}(y|x) = \mathbb{E}_{\theta \sim p(\theta|\mathcal{D}^{\text{tr}}_{\text{id}})} \left[ p(y|x,\theta) \right].
\end{align}
The average in (\ref{eq:c2_laplace_prob}) can be estimated via Monte Carlo sampling from the distribution (\ref{eq:c2_laplace_posterior}). The Laplace approximation is attractive as a \emph{post-hoc} method requiring no retraining, but its quality is limited by the Gaussian assumption. In Chapter \ref{chapter:5}, it serves as a baseline, and the proposed conformalized calibration distillation method is shown to outperform it.

\subsection{Knowledge Distillation}
\label{sec:c2_distillation}
\begin{figure} [tb] 
    \centering
    \centerline{\includegraphics[scale=0.28]{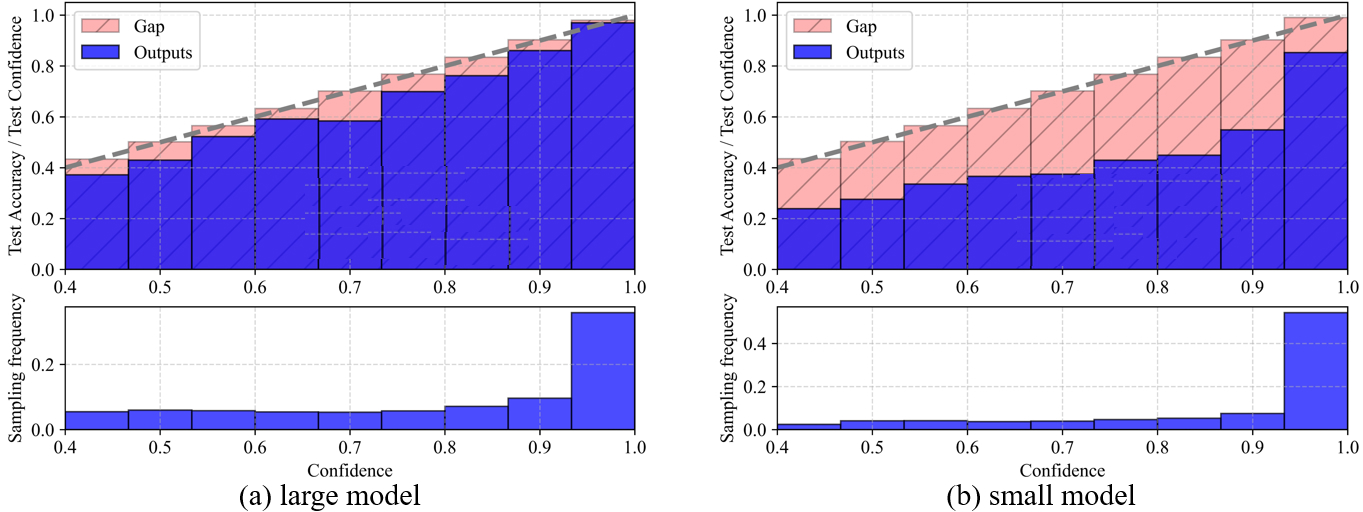}}
    \caption{Reliability diagrams for the large teacher model (left) and the small student model (right), showing that knowledge distillation does not necessarily transfer calibration.}
    \label{fig:c2_distill} 
\end{figure}

\emph{Knowledge distillation}~\citep{hinton2015distilling} transfers information from a large \emph{teacher} model $p^*(y|x)$ to a smaller \emph{student} model $p^e(y|x)$ by training the student to match the teacher's predictive distribution. Given a training dataset $\mathcal{D}^{\text{tr}}_{\text{id}} = \{x_i\}_{i=1}^{|\mathcal{D}^{\text{tr}}_{\text{id}}|}$, the standard knowledge distillation objective minimizes the \gls{kl} divergence between the teacher and student predictive distributions, i.e., yielding the solution 
\begin{align} \label{eq:c2_kd}
    \theta^e = \arg \min_{\theta} \sum_{x \in \mathcal{D}^{\text{tr}}_{\text{id}}} \operatorname{KL}\left(p^*(\cdot|x) || p^e(\cdot|x, \theta)\right).
\end{align}
Then, the distilled student model can be obtained as
\begin{align}
    p^e(y|x) = p^e(y|x, \theta^e).
\end{align}
By training on the teacher's soft output distribution rather than hard one-hot labels, the student learns inter-class relationships captured by the teacher, often improving generalization \citep{hinton2015distilling}.

However, the training objective (\ref{eq:c2_kd}) transfers \emph{accuracy} but not necessarily \emph{calibration}. A student matching the teacher's rankings may still be poorly calibrated, as reported in Fig.~\ref{fig:c2_distill}. This motivates \emph{calibration distillation}, where the goal is to transfer the teacher's reliability. As discussed in Sec.~\ref{sec:c2_certified}, applying post-hoc certified methods to a weak student yields valid but uninformative prediction sets. Calibration distillation bridges this gap, as developed in Chapter \ref{chapter:5}.

\subsection{Collaborative Inference via Model Cascading}
\label{sec:c2_cascading}

Rather than relying entirely on a compressed or distilled edge model, a \emph{model cascade} \citep{marquez2018deep} allows the edge model $p^e(y|x)$ and cloud model $p^*(y|x)$ to collaborate at inference time. The edge model processes each input first; if its prediction is deemed sufficiently reliable, the edge prediction set is returned, otherwise the input is \emph{deferred} to the cloud. Formally, the cascade produces
\begin{align} \label{eq:c2_cascade}
    \Gamma(x) = \begin{cases}
        \Gamma^e(x), & \text{if } x \text{ is processed at the edge,} \\
        \Gamma^*(x), & \text{if } x \text{ is deferred to the cloud,}
    \end{cases}
\end{align}
where $\Gamma^e(x)$ and $\Gamma^*(x)$ are the edge and cloud prediction sets, respectively.

The simplest deferral strategy uses the edge model's maximum confidence as a proxy for reliability~\citep{fithian2014optimal, rabanser2025gatekeeper}:
\begin{align} \label{eq:c2_confidence_cascade}
    \Gamma(x) = \begin{cases}
        \Gamma^*(x), & \text{if } \max_{y \in \mathcal{Y}} p^e(y|x) < \tau_{\text{def}}, \\
        \Gamma^e(x), & \text{if } \max_{y \in \mathcal{Y}} p^e(y|x) \geq \tau_{\text{def}},
    \end{cases}
\end{align}
where $\tau_{\text{def}} \in [0, 1]$ is a pre-determined threshold. However, this heuristic rule lacks formal reliability guarantees and may produce unreliable deferral decisions. Chapter \ref{chapter:6} develops a principled alternative based on conformal alignment that provides formal reliability guarantees.

\section{Towards Reliability-Efficiency Co-Design}
\label{sec:c2_codesign_preview}

As described, Sec.~\ref{sec:c2_reliability} and Sec.~\ref{sec:c2_efficiency} have introduced the reliability and efficiency toolboxes separately. However, achieving reliable and efficient inference requires combining tools from both. The subsequent chapters progressively integrate these tools, with each combination addressing a limitation of the previous one:

\begin{itemize}
    \item \textbf{Empirical reliability (Chapter~\ref{chapter:3}).} Bayesian learning can produce well-calibrated models, but requires training from scratch and maintaining ensembles at inference time, limiting efficiency.

    \item \textbf{Certified reliability (Chapter~\ref{chapter:4}).} \gls{cp} is inherently post-hoc and lightweight, making it computationally efficient. However, the informativeness of the resulting prediction sets is fundamentally limited by the quality of the underlying model. That said, applying \gls{cp} to a weak model yields formally valid but uninformatively prediction sets.

    \item \textbf{Certified reliability + calibration distillation (Chapter~\ref{chapter:5}).} To resolve this quality bottleneck, calibration distillation transfers reliability from a powerful cloud model to a compact edge model, so that certified methods produce informative prediction sets without requiring the cloud at run time. Yet the distilled edge model is not guaranteed to be reliable for all inputs.

    \item \textbf{Certified reliability + collaborative inference (Chapter~\ref{chapter:6}).} Model cascading allows the edge and cloud to collaborate online, but conventional cascading lacks formal reliability guarantees. Chapter~\ref{chapter:6} bridges this gap by augmenting cascading with a certified conditional reliability guarantee, ensuring cloud-level reliability while minimizing reliance on the cloud.
\end{itemize}

\chapter{Calibrating Bayesian Learning via Regularization, Confidence Minimization, and Selective Inference} \label{chapter:3}

\ifpdf
    \graphicspath{{Chapter3/Figs/}{Chapter3/Figs/PDF/}{Chapter3/Figs/}}
\else
    \graphicspath{{Chapter3/Figs/}{Chapter3/Figs/}}
\fi

\section{Overview}


As discussed in Chapter~\ref{chapter:2}, the application of \gls{ai} models in safety-critical fields is limited by the known difficulty of quantifying the reliability of an \gls{ai}'s decision. Bayesian learning offers a principled route to calibration by capturing epistemic uncertainty through ensembling over a distribution of model parameters. However, in practice, model misspecification and approximate variational inference often degrade calibration, while the capacity to detect \gls{ood} inputs remains a separate and often conflicting objective. A well-calibrated \gls{ai} model must correctly report its accuracy on \gls{id} inputs, while also enabling the detection of \gls{ood} inputs. Conventional solutions are effective at addressing only one of these two conflicting requirements. 

To address this challenge, this chapter proposes a novel solution that endows Bayesian learning with enhanced \gls{id} calibration and \gls{ood} detection capabilities by integrating calibration regularization for improved \gls{id} performance, confidence minimization for \gls{ood} detection, and selective calibration to ensure a synergistic use of calibration regularization and confidence minimization. Selective calibration rejects inputs for which the calibration performance is expected to be insufficient, supporting effective \gls{ood} detection, while also ensuring \gls{id} calibration. Prior art had only considered these ideas in isolation and for frequentist learning. Numerical results illustrate the trade-offs between \gls{id} accuracy, \gls{id} calibration, and \gls{ood} calibration, showing that the proposed novel Bayesian approach achieves the best \gls{id} and \gls{ood} performance compared to existing \gls{sota} approaches, at the cost of rejecting a fraction of the inputs.

\section{Introduction} \label{sec:c3_intro}
\subsection{Context and Motivation}


Modern \gls{ai} models, including deep \gls{nn} and \gls{llm}, have achieved great success in many domains, even surpassing human experts in specific tasks \citep{herbold2023large}.  However, it is common to hear concerns voiced by experts on the application of \gls{ai} for safety-critical fields such as engineering  \citep{ovadia2019can} or health care \citep{esteva2017dermatologist}. These concerns hinge on the known difficulty in quantifying the reliability of an \gls{ai}'s decision, as in the well-reported phenomenon of the ``hallucinations'' of \gls{llm} \citep{detommaso2024multicalibration, kumar2023conformal, quach2023conformal}. Indeed, in safety-critical applications, \emph{calibration} -- i.e., the property of a model to know when it does not know -- is arguably just as important as accuracy \citep{zecchin2024forking, lindemann2023safe, ren2023robots}.
\begin{table}[t]
\centering
\caption{Comparison with \gls{sota}.}
\begin{tabular}{@{}cccccc@{}}
\toprule
    \textbf{Reference} & $\textbf{Learning}^*$ & \thead{\textbf{Calibration} \\ \textbf{regularization}} & \thead{\textbf{OOD} \\ \textbf{confidence}\\ \textbf{minimization}} & \thead{\textbf{Selective} \\ \textbf{inference}}  \\ \midrule
 \citep{kumar2018trainable, bohdal2021meta, yoon2023esd, mukhoti2020calibrating, karandikar2021soft} & F & \checkmark &  & \\

 \citep{krishnan2020improving}, \citep{huang2023calibration}  & B & \checkmark &  &  \\

 \citep{choi2023conservative} & F &  & \checkmark & \\

 \citep{fisch2022calibrated} & F &  &  & \checkmark\\
Ours  & F, B & \checkmark & \checkmark & \checkmark \\ \bottomrule
\end{tabular}
\flushleft{\footnotesize{* ``F'' $=$ frequentist learning; ``B'' $=$ Bayesian learning.}}
\label{tab:c3_my_label}
\end{table}


A well-calibrated \gls{ai} model must correctly report its accuracy on \emph{\gls{id}} inputs, i.e., on inputs following the same statistics as training data,  while also enabling the detection of \emph{\gls{ood}} inputs, i.e., of inputs that are not covered by the training data distribution. 

A conventional approach to improve \gls{id} calibration is the application of Bayesian \emph{ensembling}. \emph{\gls{bnns}} capture epistemic uncertainty via a distribution over the weights of the model, making it possible to evaluate reliability via the level of consensus between models drawn from the model distribution \citep{ovadia2019can, krishnan2020improving,simeone2022machine}. However, owing to computational limitations and model misspecification, practical ensembling strategies do not necessarily enhance \gls{id} calibration \citep{masegosa2020learning, knoblauch2019generalized, wenzel2020good}. Furthermore, improvements in \gls{id} calibration often degrade \gls{ood} detection performance \citep{ovadia2019can, wald2021calibration, henning2021bayesian}.

Focusing on conventional \emph{\gls{fnns}}, prior art has introduced a number of notable methods to separately enhance \gls{id} calibration and \gls{ood} detection. In particular, first, \emph{calibration regularization} improves \gls{id} performance by penalizing excessively confident prediction \citep{kumar2018trainable}. Second, \emph{confidence minimization} enhances \gls{ood} detection by exposing the model to \gls{ood} samples during training \citep{choi2023conservative}. Finally, \emph{selective calibration} adaptively decides to reject inputs for which the calibration error is estimated to be excessively large \citep{fisch2022calibrated}. This work proposes a novel training strategy that endows \gls{bnns} with enhanced \gls{id} calibration and \gls{ood} detection capabilities by integrating calibration regularization for improved \gls{id} performance, confidence minimization for \gls{ood} detection, and selective calibration to ensure a synergistic use of calibration regularization and confidence minimization. A table summarizing the comparison of our paper to existing studies can be found in Table~\ref{tab:c3_my_label}.

\begin{figure} [tb] 
    \centering
    \centerline{\includegraphics[width=\textwidth]{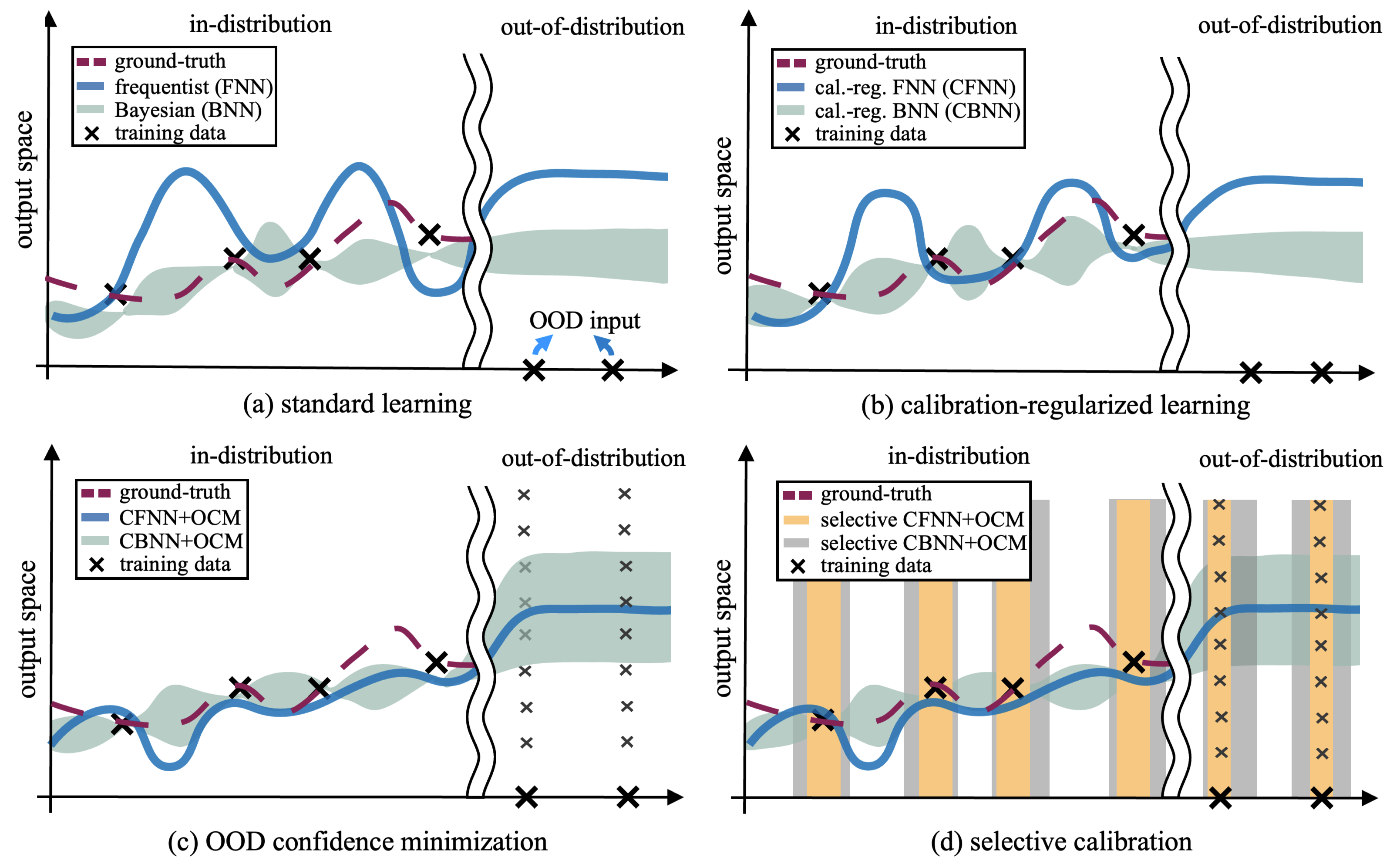}}
        \caption{(a) Standard \gls{fnns} generally fail to provide well-calibrated decisions, and improved \gls{id} calibration can be achieved via \gls{bnns} \citep{huang2023calibration}. (b) \emph{Calibration regularization} improves \gls{id} calibration via a regularizer that penalizes calibration errors \citep{kumar2018trainable}. (c) \emph{\gls{ocm}} injects \gls{ood} examples during training to improve \gls{ood} detection \citep{choi2023conservative}. (d) \emph{Selective calibration} further improves both \gls{id} and \gls{ood} calibration by only producing decisions for inputs at which uncertainty quantification is deemed to be sufficiently reliable. Prior works \citep{kumar2018trainable, choi2023conservative, fisch2022calibrated} introduced calibration regularization, \gls{ocm}, and selective calibration as separate methods for \gls{fnns}. In contrast, this work presents an integrated training method for \gls{bnns} that integrates calibration regularization for improved \gls{id} performance, confidence minimization for \gls{ood} detection, and selective calibration to ensure a synergistic use of calibration regularization and confidence minimization.}
    \label{fig:c3_6} 
\end{figure}

\subsection{Related Work}

As shown in Fig.~\ref{fig:c3_6}(a), conventional \gls{fnns} do not capture epistemic uncertainty, i.e., the uncertainty that arises due to limited access to training data. As a result, they often fail to provide a reliable quantification of uncertainty for both \gls{id} data -- i.e., in between data points -- and \gls{ood} data -- i.e., away from the domain spanned by training data \citep{guo2017calibration, tao2023benchmark}. \gls{ood} calibration is measured in terms of the capacity of the model to detect inputs that are too different from available training data, which requires the capacity to measure epistemic uncertainty \citep{henning2021bayesian,tran2022plex}.

In order to enhance the \gls{id} calibration of \gls{fnns}, \emph{calibration regularization} modifies the training loss by adding measures of \gls{id} calibration, such as the \emph{\gls{mmce}} \citep{kumar2018trainable}, the \emph{\gls{esd}} \citep{yoon2023esd}, the \emph{focal loss} \citep{mukhoti2020calibrating}, and the \emph{soft \gls{avuc}} \citep{karandikar2021soft}. As shown in Fig.~\ref{fig:c3_6}(b), \gls{fnns} with calibration regularization can improve \gls{id} calibration, but it does not resolve the issue that \gls{fnns} cannot model epistemic uncertainty, as they treat model weights as deterministic quantities, rather than as random variables in \gls{bnns}.

As mentioned, improving \gls{id} calibration often degrades \gls{ood} detection performance, and tailored methods are required to enhance \gls{ood} performance  \citep{choi2023conservative, hendrycks2018deep,ovadia2019can, lakshminarayanan2017simple, krishnan2020improving,  tran2022plex}. As illustrated in Fig.~\ref{fig:c3_6}(c), \emph{\gls{ocm}} augments the dataset with \gls{ood} inputs, which are used to specify a regularizer that favors high uncertainty outside the training set  \citep{choi2023conservative}. However, in so doing, \gls{ocm} can hurt \gls{id} calibration, making the model excessively conservative.

Overall, there is generally a \emph{trade-off between \gls{ood} and \gls{id} calibration}, and schemes designed to optimize either criterion may hurt the other \citep{choi2023conservative}.

\emph{Selective inference} techniques typically focus on enhancing the accuracy of the model by selecting inputs with the largest model confidence among the selected inputs \citep{geifman2019selectivenet, huang2020self, pugnana2024deep, fisch2022calibrated}. As a result, these techniques may end up exacerbating issues with \gls{id} calibration by favoring over-confidence. In contrast, as shown in Fig.~\ref{fig:c3_6}(d),  \emph{selective calibration} learns how to select inputs that are expected to have a low gap between accuracy and confidence, thus boosting \gls{id} calibration \citep{fisch2022calibrated}.

All the papers presented so far have focused on improving either \gls{id} or \gls{ood} calibration. Furthermore, by focusing on \gls{fnns}, these methods are limited in the capacity to provide reliable decisions that fully account for epistemic uncertainty.

\subsection{Main Contributions}

This chapter proposes a novel methodology for BNN training that integrates calibration regularization for improved \gls{id} performance, \gls{ocm} for \gls{ood} detection, and selective calibration to ensure a synergistic use of calibration regularization and \gls{ocm}. As illustrated in Fig.~\ref{fig:c3_6}, this chapter constructs this scheme successively by first introducing \gls{cbnn}; then incorporating \gls{ocm} to yield CBNN-OCM; and finally integrating also selective calibration to produce the proposed \gls{scbnn-ocm}. 

Overall, the main contributions are as follows.
\begin{itemize}
    \item We propose \gls{cbnn}, a novel BNN training scheme that improves the \gls{id} calibration performance in the presence of computational complexity constraints and model misspecification by adding a calibration-aware regularizer (see Fig.~\ref{fig:c3_6}(b)). 
    \item In order to improve \gls{ood} detection, \gls{cbnn} is extended by incorporating an additional regularizer that penalizes confidence on \gls{ood} data via \gls{ocm}. The resulting scheme is referred to as CBNN-OCM (see Fig.~\ref{fig:c3_6}(c)).  
    \item Since CBNN-OCM can enhance \gls{ood} detection performance at the cost of \gls{id} performance, we finally propose to further generalize CBNN-OCM via selective calibration (see Fig.~\ref{fig:c3_6}(d)). The proposed scheme, \gls{scbnn-ocm}, selects inputs that are likely to be well calibrated, avoiding inputs whose \gls{id} calibration may have been damaged by \gls{ocm}.
    \item Extensive experimental results on real-world image classification task, including CIFAR-100 dataset \citep{krizhevsky2010cifar} and TinyImageNet dataset \citep{liang2017principled}, illustrate the trade-offs between \gls{id} accuracy, \gls{id} calibration, and \gls{ood} calibration attained by both FNN and BNN. Among the main conclusions, \gls{scbnn-ocm} is seen to achieve best \gls{id} and \gls{ood} performance as compared to existing \gls{sota} approaches as long as a fraction of inputs is rejected.
\end{itemize}
Versions of \gls{cbnn} have appeared in \citep{huang2023calibration, krishnan2020improving}, with \citep{huang2023calibration} being an earlier conference version of this chapter. The authors in \citep{krishnan2020improving} proposed a new calibration-aware regularization term for Bayesian learning, \gls{avuc}, whose limitations as a calibration-aware regularizer were demonstrated in \citep{karandikar2021soft}. Reference \citep{huang2023calibration} studied a more general \gls{cbnn} framework compared to \citep{krishnan2020improving}, while focusing solely on \gls{id} calibration performance.

The remainder of this chapter is organized as follows. Sec.~\ref{sec:c3_ Calibration-Regularized}  summarizes necessary background on \gls{id} calibration, calibration-aware training, and Bayesian learning, and it presents \gls{cbnn}. \gls{ood} detection is discussed in Sec.~\ref{sec:c3_OOD}, which introduces CBNN-OCM. Sec.~\ref{sec:c3_general}  describes the selective calibration, and proposes \gls{scbnn-ocm}. Sec.~\ref{sec:c3_results} illustrates experimental setting and results. Finally, Sec.~\ref{sec:c3_conclusion} concludes the paper.

\section{Calibration-Regularized Bayesian Learning} \label{sec:c3_ Calibration-Regularized}

As discussed in Sec~\ref{sec:c2_empirical}, standard Bayesian learning can improve calibration over frequentist learning by capturing epistemic uncertainty, but model misspecification and approximate inference often limit this benefit in practice. In this section, to seek a well-calibrated model satisfying condition (\ref{eq:c2_perfect_cal}), we aim to improve the calibration performance of \gls{bnns} by introducing a data-dependent regularizer that directly penalizes calibration errors (see Fig.~3.1(b)). The approach extends the \emph{calibration-regularized \gls{fnns} (CFNNs)} of~\citep{kumar2018trainable} to CBNNs.

\subsection{State-of-the-Art: Calibration-Regularized Frequentist Learning} \label{subsection:cfl}


Frequentist learning often yields over-confident outputs that fall far short of satisfying the calibration condition (\ref{eq:c2_perfect_cal})  \citep{guo2017calibration, blundell2015weight, lakshminarayanan2017simple}. To address this problem, reference \citep{kumar2018trainable} proposed to modify the training loss by adding a data-dependent \emph{calibration-based} regularizer, namely \gls{mmce}, which is defined as 
\begin{align} \label{eq:c3_mmce}
 \mathcal{E}(\theta|\mathcal{D}^{\text{tr}}_{\text{id}}) = \Bigg( \sum_{i=1}^{|\mathcal{D}^{\text{tr}}_{\text{id}}|} \sum_{j=1}^{|\mathcal{D}^{\text{tr}}_{\text{id}}|} \frac{(c_i - {r}_i) (c_j - {r}_j) \kappa({r}_i , {r}_j)}{|\mathcal{D}^{\text{tr}}_{\text{id}}|^2} \Bigg)^{\frac{1}{2}},
\end{align}
where $\kappa(\cdot,\cdot)$ is a kernel function. The \gls{mmce} (\ref{eq:c3_mmce}) provides a differentiable measure of the extent to which the calibration condition (\ref{eq:c2_perfect_cal}) is satisfied, and the confidence score $r_i$ and correctness score $c_i$ are defined as in (\ref{eq:c2_confidence_score}) and (\ref{eq:c2_correctness_score}), respectively, with the only difference that they are evaluated based on $i$-th training example $(x_i,y_i)$. Other possible choices for the regulairzer $\mathcal{E}(\theta|\mathcal{D}^{\text{tr}}_{\text{id}})$ include the \emph{\gls{wmmce}} \citep{kumar2018trainable}, the \emph{differentiable \gls{ece}} \citep{bohdal2021meta}, the \emph{\gls{esd}} \citep{yoon2023esd}, the \emph{focal loss} \citep{mukhoti2020calibrating}, and the \emph{soft \gls{avuc}} \citep{karandikar2021soft}.

Equipped with a data-dependent calibration-aware regularizer $\mathcal{E}(\theta|\mathcal{D}^{\text{tr}}_{\text{id}})$, \gls{cfnn} aim at finding a deterministic parameter vector $\theta^\text{CFNN}$ that minimizes the regularized \gls{ce} loss as per the optimization \citep{kumar2018trainable}
\begin{align} \label{eq:c3_ca_fnn}
    \theta^\text{CFNN} = \arg\min_{\theta} \big\{\mathcal{L}(\theta|\mathcal{D}^{\text{tr}}_{\text{id}}) + \gamma_{\text{id}} \cdot \mathcal{E}(\theta|\mathcal{D}^{\text{tr}}_{\text{id}})\big\}
\end{align}
given a hyperparameter $\gamma_{\text{id}}>0$. Accordingly, the trained probabilistic predictor is obtained as
\begin{align}
    p(y|x,\mathcal{D}^{\text{tr}}_{\text{id}}) = p(y|x, \theta^{\text{CFNN}}).
\end{align}
Note that in (\ref{eq:c3_ca_fnn}) the regularizer $\mathcal{E}(\theta|\mathcal{D}^{\text{tr}}_{\text{id}})$ depends on the same training data $\mathcal{D}^{\text{tr}}_{\text{id}}$ used in the \gls{ce} loss $\mathcal{L}(\theta|\mathcal{D}^{\text{tr}}_{\text{id}})$. An alternative, investigated in \citep{yoon2023esd}, is to split the training data into two disjoint parts $\mathcal{D}^{\text{tr}}_{\text{id},1}$ and $\mathcal{D}^{\text{tr}}_{\text{id},2}$ with $\mathcal{D}^{\text{tr}}_{\text{id},1} \cup \mathcal{D}^{\text{tr}}_{\text{id},2} = \mathcal{D}^{\text{tr}}_{\text{id}}$ and evaluate the regularizer using $\mathcal{D}^{\text{tr}}_{\text{id},2}$, yielding the problem $\min_{\theta \in \Theta} \{\mathcal{L}(\theta|\mathcal{D}^{\text{tr}}_{\text{id},1}) + \gamma_{\text{id}} \cdot \mathcal{E}(\theta|\mathcal{D}^{\text{tr}}_{\text{id},2})\}$, where the \gls{ce} loss is estimated using dataset $\mathcal{D}^{\text{tr}}_{\text{id},1}$.

\subsection{Proposed Method: Calibration-Regularized Bayesian Learning} \label{subsec:c3_CB}

In the previous subsections, we have summarized existing ways to improve the \gls{id} calibration performance either via a calibration-based, data-dependent, regularizer (Sec.~\ref{subsection:cfl}) or via a prior-based, data-independent, regularizer (Sec.~\ref{sec:c2_empirical}). In this subsection, we first describe some limitations of the existing approaches as solutions to improve \gls{id} calibration, and then we propose the \gls{cbnn} framework.

As described in Sec.~\ref{subsection:cfl}, CFNN training disregards epistemic uncertainty, i.e., uncertainty in the model parameter space, by optimizing over a single parameter vector $\theta^\text{CFNN}$ that minimizes a regularized training loss. This may cause the resulting CFNN predictor to remain over-confident on \gls{id} test inputs.

By capturing the epistemic uncertainty of the \gls{nn} via ensembling as in (\ref{eq:c2_ensmble}), BNN learning has the potential to yield a better \gls{id}-calibrated probabilistic predictor that accounts for model-level {uncertainty}. However, in practice, \gls{bnns} may still yield poorly calibrated predictors in the presence of \emph{model misspecification} \citep{masegosa2020learning, morningstar2022pacm, zecchin2023robust}, as well as due to \emph{approximations} such as the parametrization assumed by \gls{vi} motivated by computational limitations. Model misspecification may refer to settings with a misspecified prior and/or to a misspecified likelihood $p(y|x,\theta)$, which do not reflect well the true data-generating distribution.

In order to handle the limitations of both CFNN and BNN, we propose \gls{cbnn}. To this end, we first extend the calibration-based regularizer $\mathcal{E} (\theta | \mathcal{D}^{\text{tr}}_{\text{id}})$, defined in (\ref{eq:c3_mmce}) for a deterministic model parameter vector $\theta$, to a random vector $\theta \sim q(\theta)$ as 
\begin{align} \label{eq:c3_cal_reg_q}
    \mathcal{E} (q | \mathcal{D}^{\text{tr}}_{\text{id}}) = \mathbb{E}_{\theta \sim q(\theta)}\big[\mathcal{E} (\theta | \mathcal{D}^{\text{tr}}_{\text{id}})\big].
\end{align}
The key idea underlying \gls{cbnn} is to leverage the calibration regularizer (\ref{eq:c3_cal_reg_q}) as a way to correct for the \gls{id} calibration errors arising from model misspecification and approximate Bayesian inference. Accordingly, adopting \gls{vi}, \gls{cbnn} learning aims at finding a variational parameter $\varphi$ by addressing the minimization
    \begin{align} \label{eq:c3_CA-BNN_gen_recipe}
        \varphi^\text{CBNN} = \arg\min_{q(\theta|\varphi)} \big\{\mathcal{F}(q|\mathcal{D}^{\text{tr}}_{\text{id}}) + \gamma_{\text{id}}\cdot \mathcal{E}(q|\mathcal{D}^{\text{tr}}_{\text{id}})\big\},
    \end{align} 
for some hyperparameter $\gamma_{\text{id}} > 0$. In practice, the optimization in (\ref{eq:c3_CA-BNN_gen_recipe}) can be carried out via \gls{sgd} on the variational parameter $\varphi$ via the reparameterization trick, in a manner similar to BNN learning (see Sec.~\ref{sec:c2_empirical}).

Finally, the trained probabilistic predictor is obtained as the ensemble
\begin{align}
    p(y|x, \mathcal{D}^{\text{tr}}_{\text{id}}) = \mathbb{E}_{\theta \sim q(\theta|\varphi^\text{CBNN})}[p(y|x,\theta)].
\end{align}

\begin{figure} [tb] 
    \centering
    \centerline{\includegraphics[width=\textwidth]{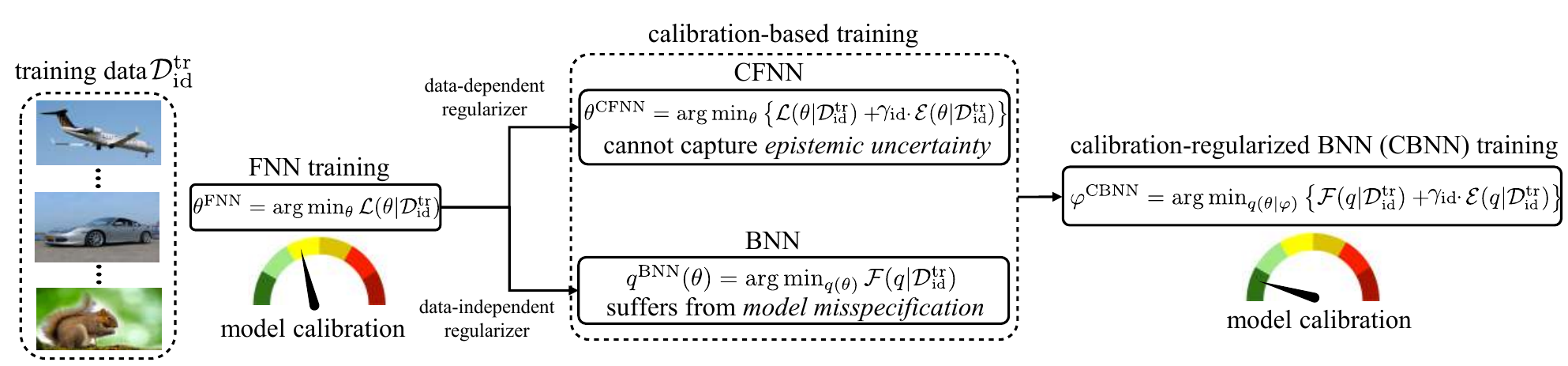}}
    \caption{Standard FNN training \citep{simeone2022machine} minimizes the \gls{ce} loss $\mathcal{L}(\theta|\mathcal{D}^{\text{tr}}_{\text{id}})$; CFNN training \citep{kumar2018trainable} minimizes the regularized \gls{ce} loss (\ref{eq:c3_ca_fnn}); and BNN learning optimizes the free-energy loss in (\ref{eq:c2_free_energy}) \citep{simeone2022machine}. The proposed \gls{cbnn} optimizes a regularized free-energy loss with the aim of capturing epistemic uncertainty, like \gls{bnns}, while also accounting directly for calibration performance as \gls{cfnn}. }
    \label{fig:c3_training_based_cal} 
\end{figure}

As anticipated, as compared to existing schemes, \gls{cbnn}s have two potential advantages. On the one hand, they can improve the calibration performance of \gls{cfnn} by taking epistemic uncertainty into account via the use of a variational distribution $q(\theta|\varphi)$ on the model parameter space. On the other hand, they can improve the calibration performance of BNN in the presence of model misspecification and/or approximation errors by modifying the free-energy loss function (\ref{eq:c2_simple_BNN}) through a calibration-driven regularizer. A summary of \gls{cbnn} with a comparison to FNN, CFNN, and BNN is illustrated in Fig.~\ref{fig:c3_training_based_cal}.

\section{Bayesian Learning with \gls{ood} Confidence Minimization} \label{sec:c3_OOD}

In the previous section, we have designed the \gls{cbnn} training methodology with the aim at improving \gls{id} calibration performance. As discussed in Sec.~\ref{sec:c3_intro}, another important requirement of reliable models is \emph{\gls{ood} detection} \citep{tran2022plex, geng2020recent}, i.e., the ability to distinguish \gls{ood} data from \gls{id} data. In this section, we describe an extension of \gls{cbnn} that aims at improving the \gls{ood} detection performance via the introduction of \gls{ocm} \citep{choi2023conservative}. To this end, we first describe the \gls{ood} detection problem; and then, we review the \gls{ocm} regularizer presented in \citep{choi2023conservative} to improve \gls{ood} detection for \gls{fnns}. Finally, we propose the inclusion of an \gls{ocm} regularizer within the training process of \gls{cbnn}s, defining the proposed CBNN-OCM scheme (see Fig.~\ref{fig:c3_6}(c)).

\subsection{Problem Definition: \gls{ood} Detection}
An important facet of calibration is the capacity of a model to detect \gls{ood} inputs, that is, inputs that suffer from a distributional shift with respect to training conditions \citep{tran2022plex}. \gls{ood} detection aims at distinguishing \gls{ood} inputs from \gls{id} inputs by relying on the observation of the confidence levels $p(y|x, \mathcal{D}^{\text{tr}}_{\text{id}})$ produced by the model  \citep{ren2019likelihood,hendrycks2016baseline,  choi2023conservative}. The underlying principle is that, if the predictor $p(y|x, \mathcal{D}^{\text{tr}}_{\text{id}})$ is well calibrated, \gls{ood} inputs should be characterized by a high level of predictive uncertainty, i.e., by a higher-entropy conditional distribution $p(y|x, \mathcal{D}^{\text{tr}}_{\text{id}})$, as compared to \gls{id} inputs. 

To quantity the effectiveness of \gls{ood} detection, one can thus use the discrepancy between the typical confidence levels $r(x) = p(\hat{y}(x)|x, \mathcal{D}^{\text{tr}}_{\text{id}})$ in (\ref{eq:c2_basic_confidence}) assigned to the hard decision $\hat{y} (x)$ in (\ref{eq:c2_hard_decision_general}) for \gls{id} inputs and the typical confidence levels $r(x) = p(\hat{y}(x)|x, \mathcal{D}^{\text{tr}}_{\text{id}})$ for \gls{ood} inputs $x$. Indeed, if the confidence level $r(x)$ tends to be different for \gls{id} and \gls{ood} inputs, then \gls{ood} detection is likely to be successful.

To elaborate, we define $p^{\text{ID}}(r)$ as the distribution of the confidence levels $r(x)$ when $x$ is an \gls{id} input, and $p^{\text{OOD}}(r)$ as the distribution of the confidence levels $r(x)$ when $x$ is an \gls{ood} input. Then, using standard results on the optimal probability of error for binary detection (see, e.g.,  \citep{Polyanskiy_Wu_2024}), one can use the \emph{\gls{tv}} distance
\begin{align} \label{eq:c3_tv_distance}
    \text{TV} = \frac{1}{2} \int_{0}^{1} \Big|p^{\text{ID}}(r) - p^{\text{OOD}}(r) \Big| \mathrm{d}r
\end{align}
as a performance measure for \gls{ood} detection. In fact, when \gls{id} and \gls{ood} data are equally likely, the \emph{optimal \gls{ood} detection probability} is given by
\begin{align} \label{eq:c3_OOD-detection-probability}
    p_{\text{d}}^{\text{OOD}} = \frac{1}{2}(1+\text{TV}).
\end{align}

\subsection{State-of-the-Art: \gls{ood} Confidence Minimization} \label{sub:ocm_fre}
In this subsection, we summarize \gls{ocm} \citep{choi2023conservative}, a \gls{sota} frequentist learning strategy, which improves the \gls{ood} detection performance of \gls{fnns} by leveraging \emph{unlabeled} \gls{ood} inputs $\mathcal{D}^\text{unl}_{\text{ood}}=\{x^\text{u}_i\}_{i=1}^{|\mathcal{D}^\text{unl}_{\text{ood}}|}$. We emphasize that reference \citep{choi2023conservative} concluded that \gls{ocm} outperforms the other benchmarks on \gls{ood} detection, such as MaxLogit \citep{hendrycks2019scaling} and Energy Score \citep{liu2020energy}. The dataset $\mathcal{D}^\text{unl}_{\text{ood}}$ contains $|\mathcal{D}^\text{unl}_{\text{ood}}|$ inputs $x^\text{u}_i$ that are generated from the marginal \gls{ood} input distribution $p_{\text{OOD}}(x)$ \citep{hendrycks2016baseline,  lakshminarayanan2017simple}. Following \citep{choi2023conservative}, the unlabeled dataset $\mathcal{D}^\text{unl}_{\text{ood}}$ is referred to as the \emph{uncertainty dataset}.

Equipped with the uncertainty dataset $\mathcal{D}^\text{unl}_{\text{ood}}=\{x^\text{u}_i\}_{i=1}^{|\mathcal{D}^\text{unl}_{\text{ood}}|}$, \gls{ocm} aims at \emph{maximizing} the uncertainty of the predictor when applied to inputs in $\mathcal{D}^\text{unl}_{\text{ood}}$. This is done by considering a fictitious labeled dataset in which all inputs $x^\text{u}_i$ appear with all possible labels in set $\mathcal{Y}$. Accordingly, FNN training with \gls{ocm}, referred to as FNN-OCM, addresses the problem
\begin{align} \label{eq:c3_CM}
    \theta^{\text{FNN-OCM}} = \arg \min_{\theta} \big\{\mathcal{L}(\theta|\mathcal{D}^{\text{tr}}_{\text{id}}) + \gamma_{\text{ood}} \cdot \mathcal{C} (\theta | \mathcal{D}^\text{unl}_{\text{ood}})\big\},
\end{align}
where the \emph{\gls{ocm} regularization} term is defined as the log-likelihood of the fictitious labeled dataset, i.e.,
\begin{align} \label{eq:c3_freq_OCM}
    \mathcal{C}(\theta|\mathcal{D}^\text{unl}_{\text{ood}}) = - \sum_{i=1}^{|\mathcal{D}^\text{unl}_{\text{ood}}|} \sum_{y\in \mathcal{Y}} \log p(y|x^\text{u}_i,\theta),
\end{align}
given some hyperparameter $\gamma_{\text{ood}} > 0$. It was shown in \citep{choi2023conservative} that fine-tuning the pre-trained frequentist model $p(y|x,\theta^\text{FNN})$, with $\theta^\text{FNN}$ in (\ref{eq:c2_FNN}), by solving problem (\ref{eq:c3_CM}) can significantly aid the \gls{ood} detection task. With the obtained solution, the trained probabilistic predictor for \gls{fnns} with \gls{ocm} is obtained as
\begin{align}
    p(y|x,\mathcal{D}^{\text{tr}}_{\text{id}}) = p(y|x, \theta^{\text{FNN-OCM}}).
\end{align}

\begin{figure} [tb] 
    \centering
    \centerline{\includegraphics[width=\textwidth]{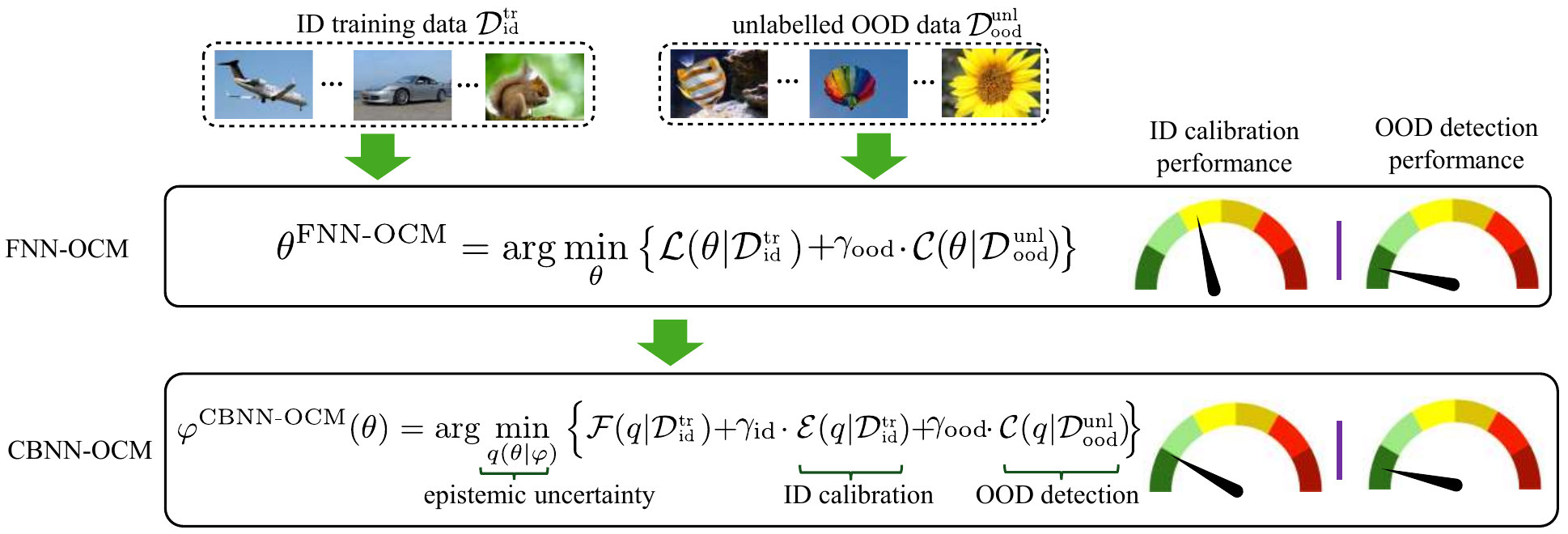}}
    \caption{Unlike standard \gls{fnns} that aim at maximizing the accuracy on the \gls{id} training dataset, FNN-OCM caters to the \gls{ood} detection task by maximizing the uncertainty for an unlabeled dataset that contains \gls{ood} inputs. As a result, the model tends to assign different confidence levels $r(x) = p(\hat{y}(x)|x, \mathcal{D}^{\text{tr}}_{\text{id}})$ to \gls{id} and \gls{ood} inputs. The proposed CBNN-OCM accounts for both \gls{id} calibration and \gls{ood} detection by capturing epistemic uncertainty as well as \gls{ood} uncertainty.}
    \label{fig:c3_OCM} 
\end{figure}

\subsection{Proposed Method: Calibration-Regularized Bayesian Learning with \gls{ood} Confidence Minimization} \label{subsec:c3_CB-OCM}

Even though FNN-OCM can help separating \gls{id} and \gls{ood} inputs, it inherits the poor \gls{id} calibration performance of \gls{fnns}, as it is unable to capture epistemic uncertainty. In contrast, \gls{bnns} and \gls{cbnn}s can enhance \gls{id} calibration, while often failing at the \gls{ood} detection task due to computational limitations and model misspecification \citep{ovadia2019can, wald2021calibration, henning2021bayesian}. In order to enhance \gls{ood} detection, while still benefiting from the advantages of \gls{cbnn} in \gls{id} calibration, in this subsection, we introduce CBNN-OCM by integrating \gls{cbnn} and \gls{ocm}. 

To this end, we first generalize the \gls{ocm} regularizer (\ref{eq:c3_freq_OCM}) so that it can be applied to a random model parameter vector $\theta \sim q(\theta)$ as the average 
\begin{align} \label{eq:c3_infer_CM}
\mathcal{C} (q | \mathcal{D}^\text{unl}_{\text{ood}}) =  \mathbb{E}_{\theta \sim q(\theta)}\big[\mathcal{C} (\theta | \mathcal{D}^\text{unl}_{\text{ood}})\big] = - \mathbb{E}_{\theta \sim q(\theta)} \Bigg[\sum_{i=1 }^{|\mathcal{D}^\text{unl}_{\text{ood}}|}  \sum_{y \in \mathcal{Y}} \log p(y|x^\text{u}_i,\theta)\Bigg].
\end{align}

Then, with (\ref{eq:c3_infer_CM}), we propose to modify the objective of \gls{cbnn} in (\ref{eq:c3_CA-BNN_gen_recipe}) by adding the \gls{ocm} regularizer as
\begin{align} \label{eq:c3_general_CM_gen_recipe}
\varphi^\text{CBNN-OCM}(\theta) = \arg\min_{q(\theta|\varphi)} \Big \{ \mathcal{F}(q|\mathcal{D}^{\text{tr}}_{\text{id}})  + \gamma_{\text{id}} \cdot \mathcal{E}(q|\mathcal{D}^{\text{tr}}_{\text{id}}) + \gamma_{\text{ood}} \cdot \mathcal{C}(q|\mathcal{D}^\text{unl}_{\text{ood}}) \Big\}. 
\end{align}

Finally, the trained probabilistic predictor is implemented as the ensemble
\begin{align} \label{eq:c3_CBNN-OCM-predictor}
    p(y|x, \mathcal{D}^{\text{tr}}_{\text{id}}) = \mathbb{E}_{\theta \sim q(\theta|\varphi^\text{CBNN-OCM})}[p(y|x,\theta)].
\end{align}

With the introduction of both data-dependent and data-independent regularizers in (\ref{eq:c3_general_CM_gen_recipe}), CBNN-OCM has the potential to provide reliable \gls{id} predictions, while also facilitating the detection of \gls{ood} inputs. A summary of the relation between FNN-OCM and CBNN-OCM is illustrated in Fig.~\ref{fig:c3_OCM}.

\section{Bayesian learning with Selective Calibration} 
\label{sec:c3_general}

While CBNN-OCM is beneficial for \gls{ood} detection, it may potentially deteriorate the \gls{id} calibration performance of \gls{cbnn} by modifying the operation of the model on examples that
are hard to identify as \gls{id} or \gls{ood}. In this section, we introduce \emph{selective calibration} as a way to correct this potential deterioration in \gls{id} performance. Selective calibration endows the model with the capacity to reject examples for which the gap between confidence and accuracy is expected to be excessive \citep{fisch2022calibrated}.

Like conventional selective classification \citep{geifman2019selectivenet}, selective calibration adds the option for a model to refuse to produce a decision for some inputs. However, conventional selective classification aims at rejecting examples with expected low accuracy. Accordingly, in order to ensure the successful selection of low-accuracy examples, selective classification requires the underlying model to be well calibrated, though possibly inaccurate. In contrast, selective calibration accepts examples with both high and low confidence levels, as long as the selector deems the confidence level to be close to the true accuracy. As a result, selective calibration does not require the underlying model to be well calibrated. Rather, it aims at enhancing calibration on the selected examples by making decisions only on inputs for which the model is expected to be well calibrated.

In this section, we first give a brief introduction to selective classification; then, we describe the selective calibration framework proposed in \citep{fisch2022calibrated} for \gls{fnns}; and finally we present the \gls{scbnn-ocm} framework that combines selective calibration and CBNN-OCM (see Fig.~\ref{fig:c3_6}(d)).

\subsection{State-of-the-Art: Selective Calibration} \label{sec:c3_background_sel_ca}

Like selective classification, given a pre-trained FNN model $p(y|x, \theta^{\text{FNN}})$, \emph{selective calibration} introduces a selector network $g(x|\phi)$ that reject or accept an input $x$ by setting $g(x|\phi) =0$ or $g(x|\phi) =1$, respectively. However, rather than targeting high-accuracy examples, selective calibration aims at accepting inputs $x$ for which the confidence level $r(x) = p(\hat{y}(x)|x,\theta^\text{FNN})$ is expected to match the true accuracy of the hard decision $\hat{y}(x)$. Accordingly, the goal of the selector is to ensure the calibration condition (\ref{eq:c2_perfect_cal}) when conditioning on the accepted examples. The resulting \emph{perfect selective \gls{id} calibration} condition can be formalized as 
\begin{align} \label{eq:c3_perfect_cal_sel}
    \Pr \Big[ y = \hat{y} | p(\hat{y}(x)|x,\theta^\text{FNN} ) = r,  g(x|\phi) = 1 \Big] = r, \text{ for all }  r \in [0,1].
\end{align}
The condition (\ref{eq:c3_perfect_cal_sel}) is less strict than the perfect \gls{id} calibration condition (\ref{eq:c2_perfect_cal}), since it is limited to accepted inputs.


\emph{1) Training the Selector: }Generalizing the \gls{mmce} in (\ref{eq:c3_mmce}), the \gls{ece} corresponding to the calibration criterion (\ref{eq:c3_perfect_cal_sel}) can be estimated by using the accepted examples within the validation dataset.

This yields the \emph{selective \gls{mmce}}
\begin{align}  \label{eq:c3_sel-cal-loss}
    &\mathcal{E}^{\text{S-Cal}}(\phi|\mathcal{D}^{\text{val}}) = \left( \frac{\sum_{i=1}^{|\mathcal{D}^\text{val}|} \sum_{j=1}^{|\mathcal{D}^\text{val}|}(c_i - {r}_i) (c_j - {r}_j) g(x_i^{\text{val}}|\phi)g(x_j^{\text{val}}|\phi) \kappa({r}_i , {r}_j)}{ \sum_{i=1}^{|\mathcal{D}^\text{val}|} \sum_{j=1}^{|\mathcal{D}^\text{val}|} g(x_i^{\text{val}}|\phi)g(x_j^{\text{val}}|\phi)}   \right)^{\frac{1}{2}},
\end{align}
with confidence scores $r_i = p(\hat{y}_i (x_i) | x_i, \theta^\text{FNN})$ and correctness scores $c_i = \mathbbm{1} (\hat{y}_i (x_i) = y_i)$. Note that the selective \gls{mmce} recovers to the \gls{mmce} (\ref{eq:c3_mmce}) when the selector accepts all examples, i.e., when $g(x|\phi)=1$ for all inputs $x$. The parameter vector $\phi$ of the selector is obtained by optimizing the selective \gls{mmce} (\ref{eq:c3_sel-cal-loss}) by considering the targeted coverage rate constraint $\xi$, with $0 \leq \xi \leq 1$, as
\begin{align} \label{eq:c3_sel-cal}
    \phi^{\text{S-Cal}} = \arg \min_{\phi}  \mathcal{E}^{\text{S-Cal}}(\phi|\mathcal{D}^{\text{val}}) \quad \text{s.t. } \frac{1}{|\mathcal{D}^{\text{val}}|} \sum_{x \in \mathcal{D}^{\text{val}}} g(x|\phi) \geq \xi.
\end{align}

Reference \citep{fisch2022calibrated} addressed the constrained optimization in (\ref{eq:c3_sel-cal}) by turning it into an unconstrained problem via the introduction of regularizer that accounts for the fraction of selected examples. To this end, the discrete-output function $g(x|\phi)$ is relaxed via a continuous-output function $\tilde{g}(r(x|\theta^\text{FNN}), s(x|\theta^\text{FNN})| \phi)$ that takes values within the interval $[0,1]$. The function $\tilde{g}(r(x|\theta^\text{FNN}), s(x|\theta^\text{FNN})| \phi)$ takes the confidence score
\begin{align}
    r(x|\theta^\text{FNN}) = p(\hat{y}(x)|x,\theta^\text{FNN})
\end{align}
as input, as well as the \emph{non-parametric outlier score vector} $s(x|\theta^\text{FNN})$ which will be specified shortly. All in all, problem (\ref{eq:c3_sel-cal}) is approximately addressed by solving the problem
\begin{align} \label{eq:c3_calibration_selector}
    \phi^{\text{S-Cal}} = \arg \min_{\phi}  \Bigg\{\tilde{\mathcal{E}}^{\text{S-Cal}} (\phi|\mathcal{D}^{\text{val}})  - \gamma_{\text{sel}} \cdot \sum_{x\in\mathcal{D}^\text{val}} \log \left( \tilde{g}(r(x|\theta^\text{FNN}), s(x|\theta^\text{FNN})| \phi) \right)   \Bigg\},
\end{align}
where $\tilde{\mathcal{E}}^{\text{S-Cal}} (\phi|\mathcal{D}^{\text{val}})$, defined as 
\begin{align} \label{eq:c3_tidle_E}
    \tilde{\mathcal{E}}^{\text{S-Cal}}(\phi|\mathcal{D}^{\text{val}})  = \left( \sum_{i=1}^{|\mathcal{D}^\text{val}|} \sum_{j=1}^{|\mathcal{D}^\text{val}|}(c_i - {r}_i) (c_j - {r}_j) \tilde{g}(r_i, s_i| \phi) \tilde{g}(r_j, s_j| \phi) \kappa({r}_i , {r}_j)\right)^{\frac{1}{2}}, 
\end{align}
uses confidence score $r_i = r(x_i|\theta^\text{FNN})$ and non-parametric outlier score vector $s_i = s(x_i|\theta^\text{FNN})$, with a coefficient $\gamma_{\text{sel}} \geq 0$  that balances selective calibration performance and coverage. Note that function (\ref{eq:c3_tidle_E}) is the numerator of (\ref{eq:c3_sel-cal-loss}) with the relaxed selector $\tilde{g}(r(x|\theta^\text{FNN}), s(x|\theta^\text{FNN})| \phi)$ in lieu of $g(x|\phi)$, and that $-\log(\cdot)$ plays the role of a barrier function to enforce the coverage rate constraint \citep{boyd2004convex}.

\emph{2) Non-Parametric Outlier Score Vector: }The non-parametric outlier score vector $s(x|\theta^\text{FNN})$ of an input $x$ has the goal of quantifying the extent to which input $x$ conforms with the input in the training dataset $\mathcal{X}^\text{tr} = \{ x_i  \}_{i=1}^{|\mathcal{D}^{\text{tr}}_{\text{id}}|}$. The rationale for adding $s(x|\theta^\text{FNN})$ as an input to the selector is that inputs $x$ that are too ``far'' from the training set may yield more poorly calibrated decisions by the model $p(y|x, \theta^\text{FNN})$. The authors of \citep{fisch2022calibrated} included four different features in vector $s(x|\theta^\text{FNN})$, which are obtained from a kernel density estimator, an isolation forest, a one-class support vector machine, and a $k$-nearest neighbor distance, constructing a $4\times1$ score vector $s(x|\theta^\text{FNN})$. In particular, the first element of vector $s(x|\theta^\text{FNN})$ is obtained by the \gls{kde} \citep{scott2015multivariate}
\begin{align} \label{eq:c3_KDE}
    s_1 (x|\theta^\text{FNN}) = \frac{1}{|\mathcal{X}^{\text{tr}}|} \sum_{i=1}^{|\mathcal{X}^{\text{tr}}|} \kappa \left( || z-z^{\text{tr}}_{i} || \right),
\end{align}
where $\kappa(\cdot)$ is a kernel function; $z$ is the output of the last hidden layer of the parametrized network $p(y|x, \theta^\text{FNN})$; and, similarly, $z_i^{\text{tr}}$ is the corresponding feature vector for $p(y|x_i, \theta^\text{FNN})$. Details regarding the other elements of the score vector $s(x|\theta^\text{FNN})$ can be found in Appendix~\ref{sec:a1_outlier_score}.

\emph{3) Inference with the Selector: }During testing, the binary selector $g(x|\phi^\text{S-Cal})$ is obtained via thresholding as
\begin{align} \label{eq:c3_cont_to_discrete_sel_cal}
    g(x|\phi^{\text{S-Cal}}) = \mathbbm{1} \left( \tilde{g}(r(x|\theta^\text{FNN}), s(x|\theta^\text{FNN})| \phi^{\text{S-Cal}}) \geq \tau_{\text{sel}} \right),
\end{align}
where $\tau_{\text{sel}}$ is a hyperparameter controlling the fraction of selected examples. In this regard, we note that the targeted coverage rate $\xi$ can be in principle met by changing only the hyperparameter $\gamma_{\text{sel}}$ in (\ref{eq:c3_calibration_selector}) without requiring the choice of the threshold $\tau_{\text{sel}}$ in (\ref{eq:c3_cont_to_discrete_sel_cal}). However, this would require retraining of the selector any time the desired coverage level $\xi$ is changed. Hence, reference \citep{fisch2022calibrated} introduced the additional hyperparameter $\tau_{\text{sel}}$ that can adjust the coverage given a fixed, trained, selector $\phi^\text{S-Cal}$.

\subsection{Proposed Method: Bayesian Learning with Selective Calibration}
\begin{figure} [tb] 
    \centering
    \centerline{\includegraphics[width=\textwidth]{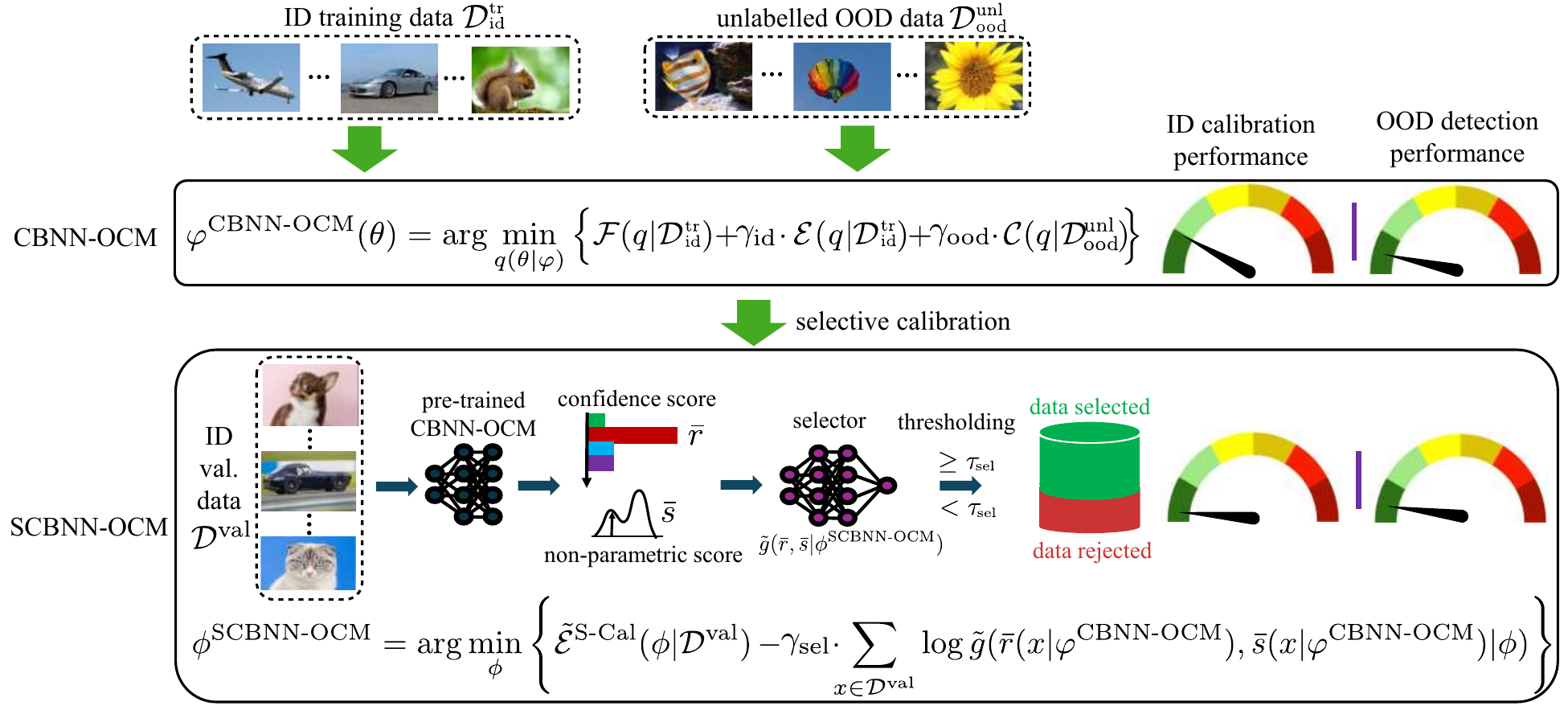}}
    \caption{Given a fixed, pre-trained model parameter vector $\theta \sim q(\theta|\varphi^\text{CBNN-OCM})$, selective calibration aims at achieving well-calibrated decisions (\ref{eq:c2_perfect_cal}) on the selected inputs (hence aiming at (\ref{eq:c3_perfect_cal_sel}) by rejecting inputs on which the discrepancy between confidence and accuracy is expected to be large.} 
    \label{fig:c3_CBS_CM} 
\end{figure}

In order to extend selective calibration to CBNN-OCM, we propose to apply the selective \gls{mmce} criterion in (\ref{eq:c3_sel-cal-loss}) with confidence score and outlier score vector averaged over the model distribution $q(\theta| \varphi^\text{CBNN-OCM})$ in (\ref{eq:c3_general_CM_gen_recipe}). Specifically, the average confidence score $\Bar{r}(x|\varphi^\text{CBNN-OCM})$ is defined as 
\begin{align} \label{eq:c3_r-bar}
    \Bar{r}(x|\varphi^\text{CBNN-OCM}) = \max_{y \in\mathcal{Y}} \mathbb{E}_{\theta \sim q(\theta|\varphi^\text{CBNN-OCM})} \left[ p(y|x,\theta) \right],
\end{align}
and the average non-parametric outlier score vector is given by
\begin{align} \label{eq:c3_s-bar}
    \Bar{s}(x|\varphi^\text{CBNN-OCM}) =  \mathbb{E}_{\theta \sim q(\theta|\varphi^\text{CBNN-OCM})} \left[ s(x|\theta) \right],
\end{align}
with vector $s(x|\theta)$ defined as in the previous subsection.

With these definitions, the selector parameter vector $\phi^\text{SCBNN-OCM}$ is obtained by addressing the problem 
\begin{align} \label{eq:c3_selector_gen_gen_recipe}
    \phi^{\text{SCBNN-OCM}}  = \arg \min_{\phi}  \Bigg\{  \tilde{\mathcal{E}}^{\text{S-Cal}} (\phi|\mathcal{D}^{\text{val}})  - \gamma_{\text{sel}} \cdot \sum_{x\in\mathcal{D}^\text{val}} \log \tilde{g}(\Bar{r}(x|\varphi^\text{CBNN-OCM}), \Bar{s}(x|\varphi^\text{CBNN-OCM})| \phi)  \Bigg\},
\end{align}
where $\tilde{\mathcal{E}}^{\text{S-Cal}} (\phi|\mathcal{D}^{\text{val}}) $, defined as 
\begin{align} \label{eq:c3_sel-cal-loss-bayes}
    \tilde{\mathcal{E}}^{\text{S-Cal}} (\phi|\mathcal{D}^{\text{val}}) = \left( \sum_{i=1}^{|\mathcal{D}^\text{val}|} \sum_{j=1}^{|\mathcal{D}^\text{val}|}(c_i - {\Bar{r}}_i) (c_j - {\Bar{r}}_j) \tilde{g}(\Bar{r}_i, \Bar{s}_i| \phi) \tilde{g}(\Bar{r}_j, \Bar{s}_j| \phi) \kappa({\Bar{r}}_i , {\Bar{r}}_j)    \right)^{\frac{1}{2}},
\end{align}
uses average confidence score $\Bar{r}_i = \Bar{r}(x_i|\varphi^\text{CBNN-OCM})$ and average non-parametric outlier score vector $\Bar{s}_i = \Bar{s}(x_i|\varphi^\text{CBNN-OCM})$.

Finally, during inference, the binary selector $g(x|\phi^{\text{SCBNN-OCM}})$ is obtained via the thresholding in a manner similar to (\ref{eq:c3_cont_to_discrete_sel_cal}), i.e., 
\begin{align} \label{eq:c3_scbnn-ocm}
    g(x|\phi^{\text{SCBNN-OCM}})  = \mathbbm{1} \left( \tilde{g}(\Bar{r}(x|\varphi^\text{CBNN-OCM}), \Bar{s}(x|\varphi^\text{CBNN-OCM})| \phi^{\text{SCBNN-OCM}}) \geq \tau_{\text{sel}} \right), 
\end{align}
for some threshold $\tau_{\text{sel}}$. A summary of \gls{scbnn-ocm} is illustrated in Fig.~\ref{fig:c3_CBS_CM}.

\section{Experiments} \label{sec:c3_results}

In this section, we report on the effectiveness of calibration-based training, \gls{ocm}, and selective calibration for frequentist and Bayesian learning in terms of \gls{id} and \gls{ood} calibration performance. Additionally, the discussion on the differentiable calibration regularizer can be seen in Appendix~\ref{sec:a1_differentiable}. The ablation study and backbone study can be checked in Appendix~\ref{sec:a1_ablation} and in Appendix~\ref{sec:a1_backbone}, respectively.

\subsection{Setting and Metrics} \label{subsec:c3_setting_and_metrics}
All experiments use the CIFAR-100 dataset \citep{krizhevsky2010cifar} to generate \gls{id} samples and the TinyImageNet (resized) dataset \citep{liang2017principled} to generate \gls{ood} samples \citep{choi2023conservative}. The parameterized predictor $p(y|x,\theta)$ adopts the WideResNet-40-2 architecture \citep{zagoruyko2016wide}, and the variational distribution $q(\theta|\varphi)$ is Gaussian with mean and diagonal covariance being the variational parameter $\varphi$. For the selective inference task, we use a $3$-layer ReLU activated feed-forward neural network $g(x|\phi)$ with $64$ dimensional hidden layer \citep{fisch2022calibrated}.

We consider the following evaluation criteria: (\emph{i}) the \emph{reliability diagram}, which plots accuracy against confidence as defined in Sec.~\ref{sec:c2_empirical} with number of bins $M=15$ (we only 
\begin{figure} [tb] 
    \centering
    \centerline{\includegraphics[scale=0.3]{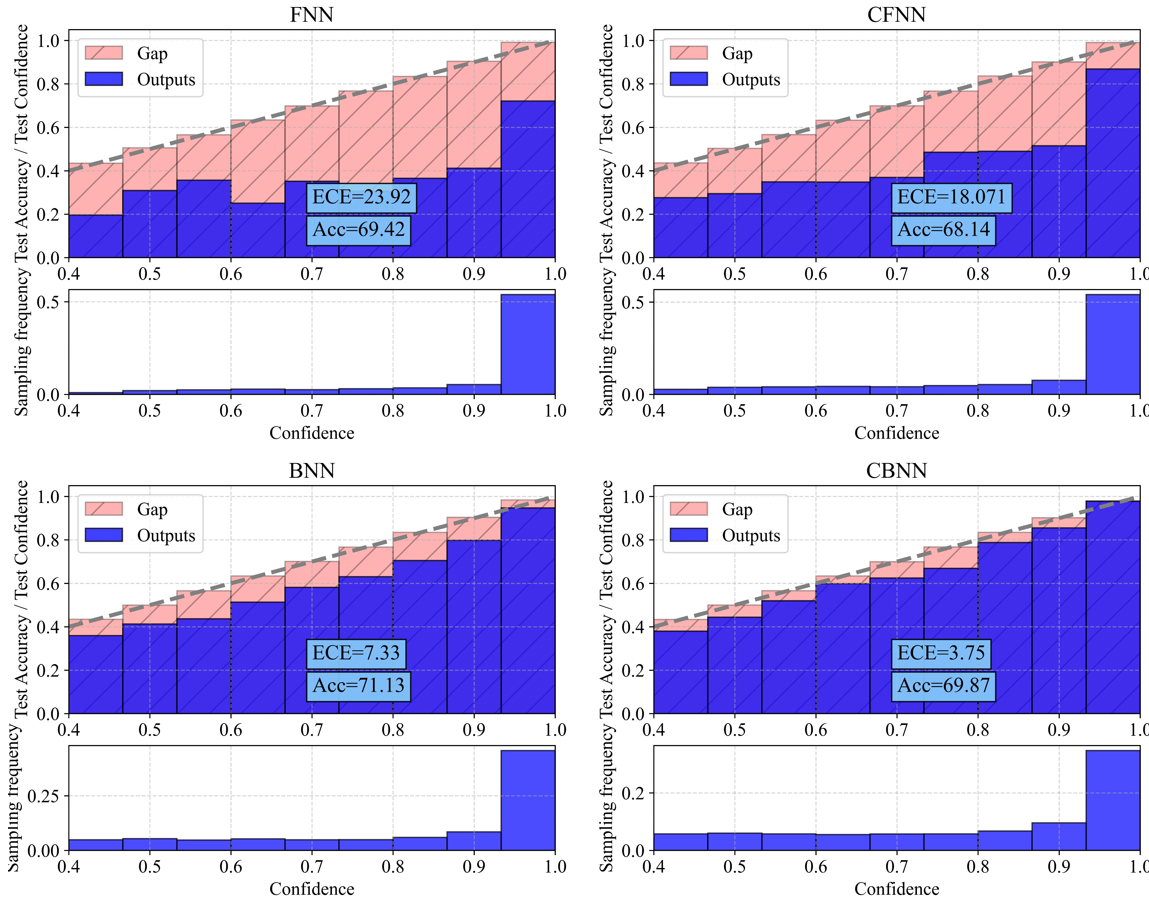}}
    \caption{Reliability diagrams for the CIFAR-100 classification task given the predictor trained using (\emph{i}) FNN (top left); (\emph{ii}) CFNN (top right, benchmark); (\emph{iii}) BNN (bottom left); and (\emph{iv}) \gls{cbnn} (bottom right, ours).}
    \label{fig:c3_CA-RD} 
\end{figure}visualize the bins that have number of samples no smaller than $100$ to avoid undesired statistical noise \citep{raviv2023modular}); (\emph{ii}) the \emph{\gls{ece}}, which evaluates the average discrepancy between per-bin accuracy and per-bin confidence as defined in (\ref{eq:c2_ece}); (\emph{iii}) \emph{\gls{id} accuracy}, marked as ``Acc'', which is the proportion of the \gls{id} data for which a correct prediction is made; (\emph{iv}) \emph{\gls{ood} detection probability}, which is the optimal probability of successfully detecting \gls{ood} data, as defined in (\ref{eq:c3_OOD-detection-probability}); and (\emph{v}) \emph{\gls{id} coverage rate}, which is the fraction of selected data points on the \gls{id} test dataset for selective calibration. We refer to Appendix~\ref{sec:a1_experiments} for all the experimental details\footnotemark[1].
\footnotetext[1]{Code can be found at \url{https://github.com/kclip/Calibrating-Bayesian-Learning}.}

\subsection{Can Calibration-Regularization Enhance \gls{id} Calibration for Bayesian Learning?}

We first evaluate the \gls{id} calibration performance for (\emph{i}) FNN, defined as in (\ref{eq:c2_FNN}); (\emph{ii}) CFNN with calibration-based regularization \citep{kumar2018trainable}, as defined in (\ref{eq:c3_ca_fnn}); (\emph{iii}) BNN as per (\ref{eq:c2_BNN}); and (\emph{iv}) the proposed \gls{cbnn} as per (\ref{eq:c3_CA-BNN_gen_recipe}). To this end, we present reliability diagrams, as described in Sec.~\ref{sec:c3_ Calibration-Regularized}, as well as the trade-off between accuracy and calibration obtained by varying the hyperparameter $\gamma_{\text{id}}$ in the designed objectives (\ref{eq:c3_ca_fnn}) and (\ref{eq:c3_CA-BNN_gen_recipe}).

To start, in Fig.~\ref{fig:c3_CA-RD}, we show the reliability diagrams for all the four schemes, which are evaluated on the CIFAR-100 dataset. Note that the figure also plots test accuracy, reported as ``Acc'', and \gls{ece}. BNN is observed to yield better calibrated predictions than FNN with a slight improvement in accuracy, while calibration-based regularization improves the calibration performance for both FNN and BNN. In particular, the \gls{ece} is decreased by more than $20\%$ for frequentist learning and by $50\%$ for Bayesian learning.
\begin{figure} [tb] 
    \centering
    \centerline{\includegraphics[scale=0.25]{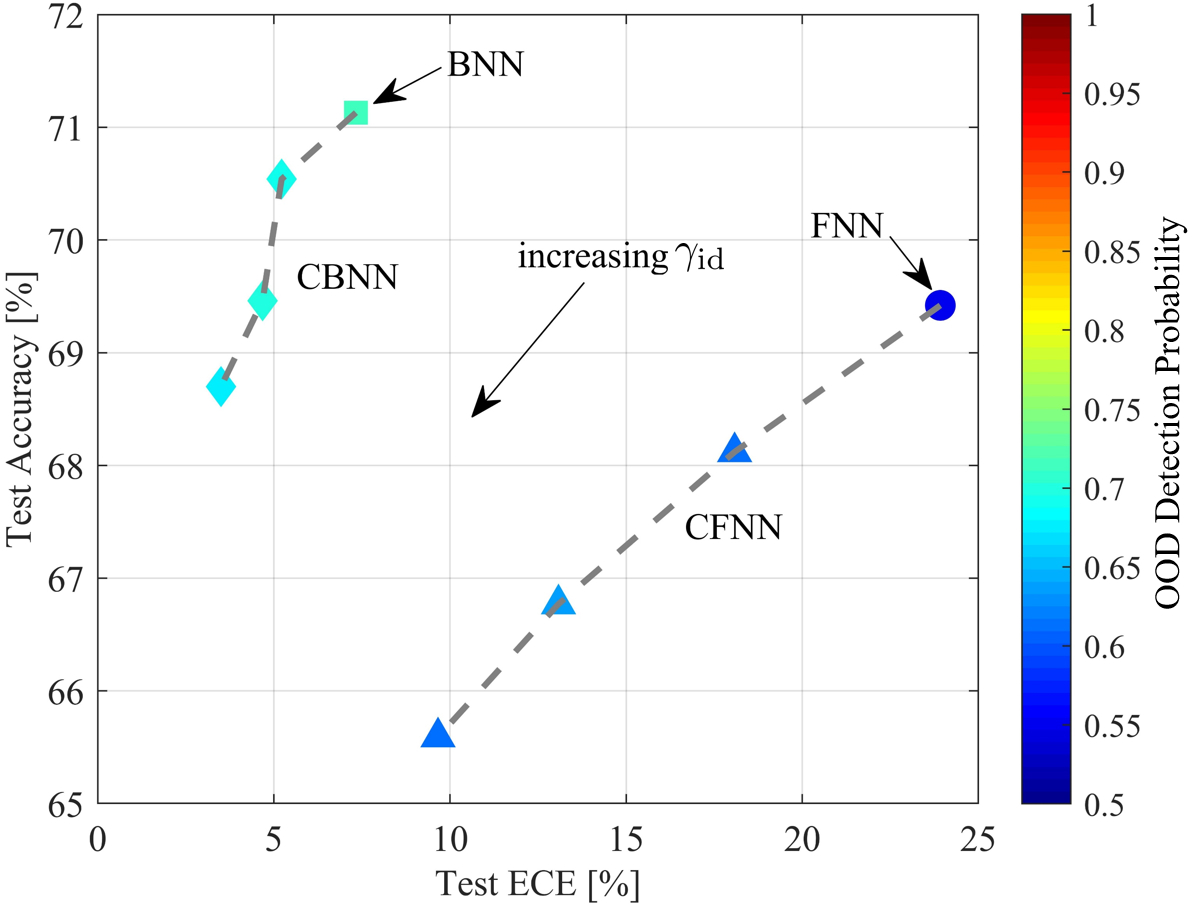}}
    \caption{Accuracy versus \gls{ece} obtained by changing the hyperparameter $\gamma_{\text{id}}$ on CIFAR-100 dataset for FNN, CFNN (benchmark), BNN, and \gls{cbnn} (ours), with \gls{ood} detection probability indicated by the marker's color. }
    \label{fig:c3_CA-pareto} 
\end{figure} 

Fig.~\ref{fig:c3_CA-pareto} demonstrates the trade-off between accuracy and \gls{ece} obtained by varying the hyperparameter $\gamma_{\text{id}}$ in (\ref{eq:c3_ca_fnn}) and (\ref{eq:c3_CA-BNN_gen_recipe}). Note that with $\gamma_{\text{id}}=0$, one recovers FNN and BNN from CFNN and \gls{cbnn}, respectively. Increasing the value of $\gamma_{\text{id}}$ is seen to decrease the \gls{ece} for both CFNN and \gls{cbnn} and to gradually affect the accuracy, tracing a trade-off between \gls{id} calibration and accuracy. This figure also report on the \gls{ood} detection probability, which will be discussed later in this section.

\subsection{Can Confidence Minimization Improve \gls{ood} Detection without Affecting \gls{id} Performance?}

\begin{figure} [tb] 
    \centering
    \centerline{\includegraphics[scale=0.5]{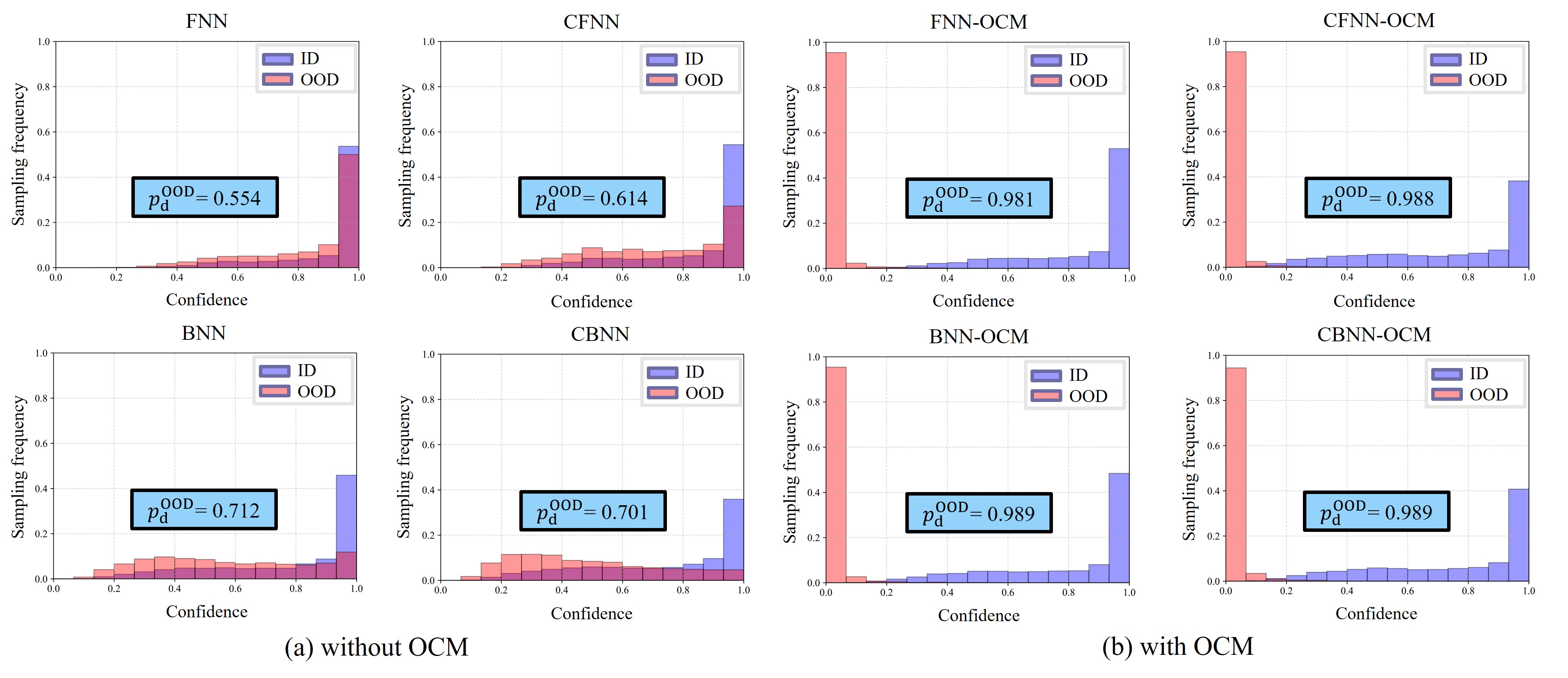}}
    \caption{Confidence histograms for \gls{id} and \gls{ood} data drawn from CIFAR-100 and TinyImageNet datasets, respectively, for FNN, CFNN, FNN-OCM (benchmark), CFNN-OCM, BNN, \gls{cbnn}, BNN-OCM, and CBNN-OCM (ours), with hyperparameter $\gamma_{\text{ood}} = 0.5$ \citep{choi2023conservative}.}
    \label{fig:c3_OCM} 
\end{figure}

To evaluate the performance in terms of \gls{ood} detection, we select \gls{id} data from the CIFAR-100 dataset, which is also used for training, while \gls{ood} is selected from the TinyImageNet dataset. As a reference, the color map in Fig.~\ref{fig:c3_CA-pareto} shows that the \gls{ood} detection probability (\ref{eq:c3_OOD-detection-probability}) does not necessarily increase as the \gls{id} calibration performance improved, which is most notable for BNN. This highlights the importance of introducing mechanism tailored to enhancing \gls{ood} detection performance, such as \gls{ocm}.

To evaluate the benefits of \gls{ocm}, Fig.~\ref{fig:c3_OCM} plots the histograms of the confidence levels produced by different models for \gls{id} and \gls{ood} data. The more distinct the two distributions are, the larger the \gls{ood} detection probability (\ref{eq:c3_OOD-detection-probability}) is \citep{choi2023conservative, Polyanskiy_Wu_2024}. First, it is observed that BNN can improve \gls{ood} detection as compared to FNN, but the \gls{ood} confidence levels tend to be approximately uniformly distributed. Second, calibration-regularization, while improving \gls{id} calibration, does not help with \gls{ood} detection, as it focuses solely on \gls{id} performance. Finally, \gls{ocm} can drastically enhance \gls{ood} detection performance for both FNN and BNN, with and without calibration-regularizers. In particular, thanks to \gls{ocm}, the model produces low confidence levels on \gls{ood} inputs, enhancing the \gls{ood} detection probability (\ref{eq:c3_OOD-detection-probability}).

\begin{figure} [tb] 
    \centering
    \centerline{\includegraphics[scale=0.25]{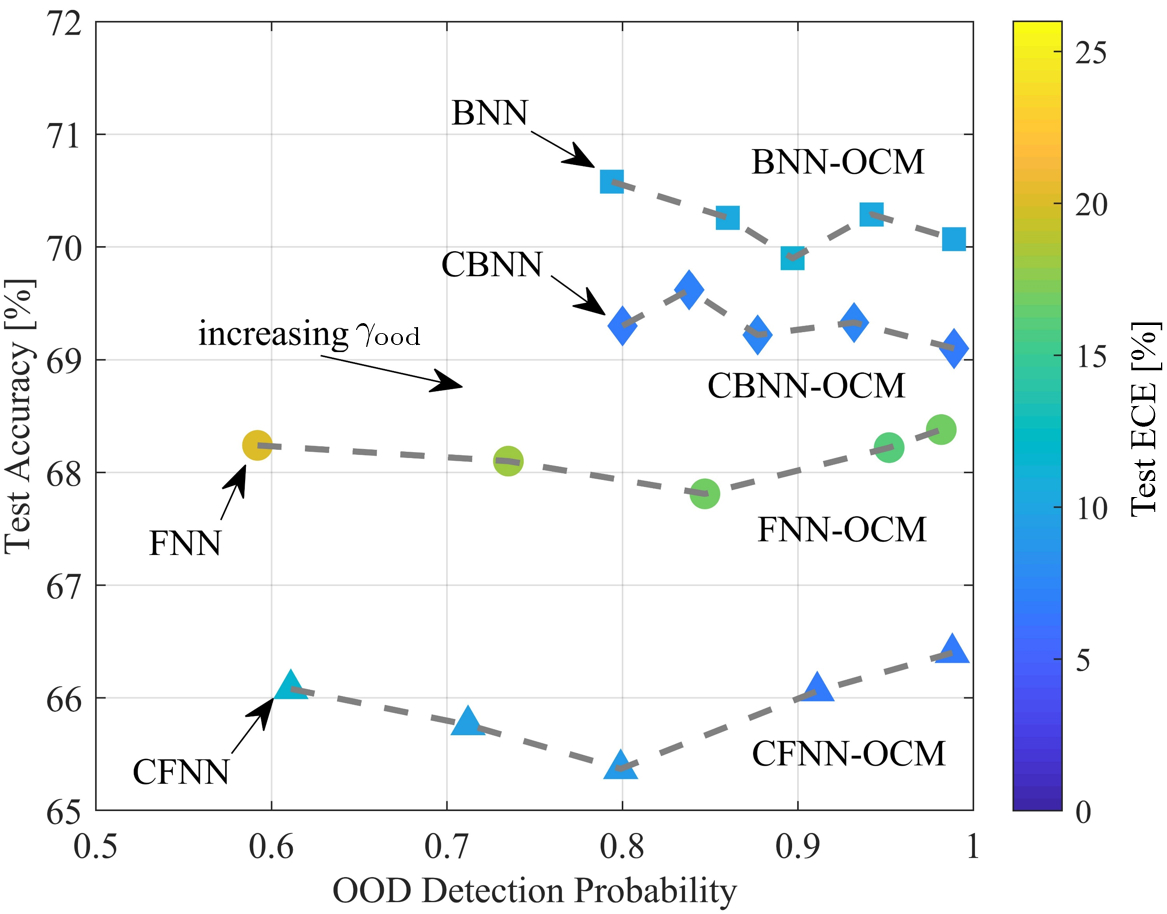}}
    \caption{Test accuracy versus \gls{ood} detection probability for FNN-OCM (benchmark), CFNN-OCM, BNN-OCM, and CBNN-OCM (ours), with test \gls{ece} as the marker's color. Note that hyperparameter $\gamma_{\text{ood}} = 0$ recovers the original schemes without \gls{ocm} regularizer.}
    \label{fig:c3_OCM-trade-off} 
\end{figure}

To further investigate the interplay between \gls{id} and \gls{ood} performance, Fig.~\ref{fig:c3_OCM-trade-off} plots the \gls{id} test accuracy versus the \gls{ood} detection probability by varying the hyperparameter $\gamma_{\text{ood}}$ in (\ref{eq:c3_CM}) and (\ref{eq:c3_general_CM_gen_recipe}). BNN yields better test accuracy for any given \gls{ood} detection probability level, and adding calibration-regularizer improves \gls{id} calibration performance, but at the cost of a reduced \gls{id} accuracy for a given \gls{ood} detection probability level.

\subsection{Can Selective Calibration Compensate for the \gls{id} Performance Loss of \gls{ocm}?}

\begin{figure} [tb] 
    \centering
    \centerline{\includegraphics[width=\textwidth]{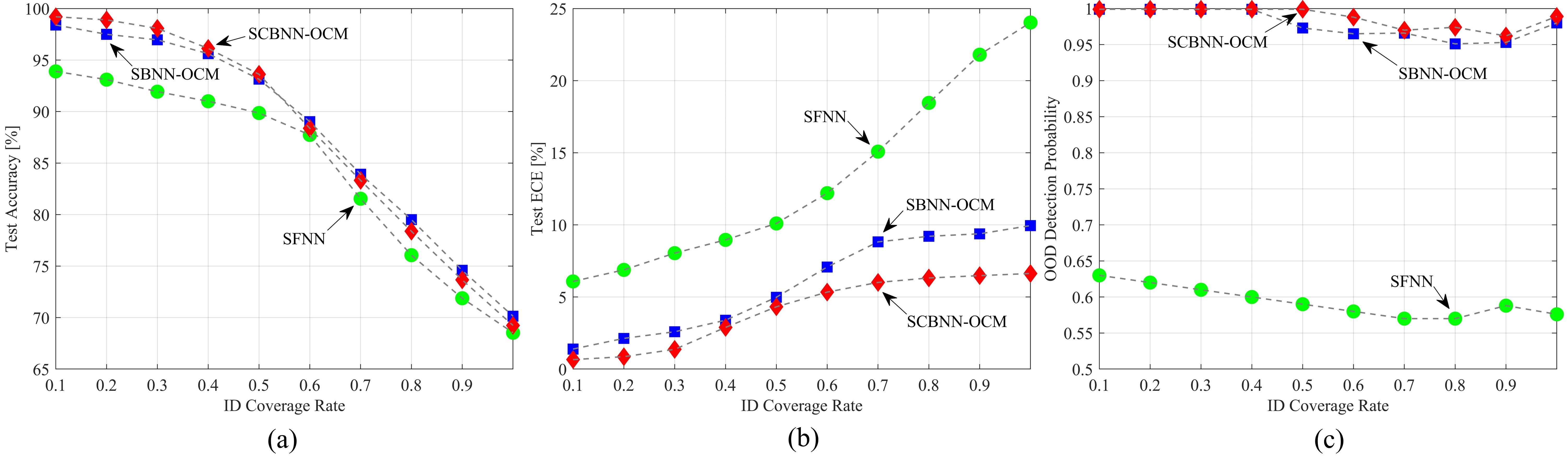}}
    \caption{Test accuracy, \gls{ece}, and \gls{ood} detection probability versus \gls{id} coverage rate for SFNN (benchmark), \gls{sbnn-ocm} and \gls{scbnn-ocm} (ours).}
    \label{fig:c3_SEL-cal} 
\end{figure}

So far, we have seen that there is generally a trade-off between \gls{id} and \gls{ood} performance. In particular, from Fig.~\ref{fig:c3_OCM-trade-off}, it was concluded that, for a given fixed level of \gls{ood} detection probability, calibration-regularized learning improved \gls{id} calibration at the cost of the \gls{id} accuracy for BNN-OCM. In this subsection, we ask whether selective calibration can ensure a synergistic use of \gls{ocm} and calibration-aware regularization, guaranteeing that CBNN-OCM achieves the best \gls{id} and \gls{ood} performance levels. To address this question, we vary the \gls{id} coverage rate, i.e., the fraction of accepted \gls{id} samples, by changing the hyperparameter $\tau_{\text{sel}}$ in (\ref{eq:c3_scbnn-ocm}). The resulting \gls{id} and \gls{ood} performance is shown in Fig.~\ref{fig:c3_SEL-cal} for \gls{sbnn-ocm} and \gls{scbnn-ocm}.

The figure demonstrates that, thanks to selective calibration, \gls{scbnn-ocm} outperforms \gls{sbnn-ocm} in terms of all metrics, namely \gls{id} accuracy, \gls{id} calibration, and \gls{ood} detection probability, for a sufficiently low \gls{id} coverage rate, here smaller than $50\%$. We thus conclude that, by selecting examples with sufficiently good \gls{id} calibration performance, \gls{scbnn-ocm} can fully benefit from calibration-aware regularization, enhancing \gls{id} performance, as well as from \gls{ocm}, improving \gls{ood} detection performance. That said, the need to accept a sufficiently small number of inputs to ensure this result highlights the challenges of guaranteeing both \gls{id} and \gls{ood} performance levels.

\subsection{Memory Requirements and Computational Cost} Finally, we analyze the memory requirements and computational cost of the different schemes studied in this paper. To this end, we focus on the setting adopted for the experiments reported in Fig.~\ref{fig:c3_CA-RD}, Fig.~\ref{fig:c3_OCM}, and Fig.~\ref{fig:c3_SEL-cal}. We evaluate the number of model parameters and the number of \gls{flop} during inference. The number of \gls{flop} is reported normalized by the size of the ensemble, and is evaluated using a standard profiler \citep{chitale2023task}. The results, reported in Table~\ref{tab:c3_overhead}, can be used to determine the trade-off between the performance improvements observed in Fig.~\ref{fig:c3_CA-RD}, Fig.~\ref{fig:c3_OCM}, and Fig.~\ref{fig:c3_SEL-cal} and the memory and computational costs. The overall number of \gls{flop} can be obtained from Table~\ref{tab:c3_overhead} by multiplying by the ensemble size, which is set to $20$ for inference and to $1$ for training. Furthermore, the number of training \gls{flop} per iteration must account for the cost of gradient evaluation, which is typically approximated as $2-5$ times the inference cost \citep{goodfellow2016deep, kaplan2020scaling}.

\begin{table}[ht]
  \centering
  \caption{Number of parameters and number of inference \gls{flop} per ensemble size (\gls{flop} are calculated with image size $32 \times 32$).}
  \label{tab:c3_overhead}
  \begin{tabular}{@{} l l r r r @{}}
    \toprule
    \textbf{Model}                         & \textbf{Parameters}  &   \textbf{\gls{flop}/ensemble size}\\
    \midrule
    FNN   &  2.25M        &  0.3G      \\ 
    CFNN   & 2.25M         & 0.3G         \\
    BNN               & 4.5M         & 0.34G           \\
    \gls{cbnn}            &4.5M         & 0.34G    \\
    \midrule
    FNN-OCM    &  2.25M         & 0.3G        \\
    CFNN-OCM   &  2.25M         & 0.3G       \\
    BNN-OCM                & 4.5M         &0.34G       \\
    CBNN-OCM             & 4.5M        & 0.34G       \\
    \midrule
    Selector   &  0.005M          &  0.001G       \\ 
    \bottomrule
  \end{tabular}
\end{table}

\section{Conclusion} \label{sec:c3_conclusion}
In this chapter, we have proposed \gls{scbnn-ocm}, a general framework that enhances variational inference-based Bayesian learning to target \emph{both} \gls{id} and \gls{ood} calibration. To improve \gls{id} calibration, we have introduced a regularizer based on the calibration error, while \gls{ood} calibration is enhanced by means of data augmentation based on confidence minimization. In order to facilitate the synergistic use of calibration-aware regularization and \gls{ocm} regularization, we have finally introduced a selective calibration strategy that rejects examples that are likely to have poor \gls{id} calibration performance. Numerical results have illustrated the challenges in ensuring both \gls{id} and \gls{ood} performance, as schemes designed for \gls{id} calibration may end up hurting \gls{ood} calibration, and vice versa. That said, thanks to selective calibration, \gls{scbnn-ocm} was observed to achieve the best \gls{id} and \gls{ood} performance as compared to the other benchmarks as long as one allows for a sufficiently large number of rejected samples. For instance, as compared to standard FNN, \gls{scbnn-ocm} achieves a $25\%$ improvement in accuracy, a $20\%$ drop in \gls{ece}, and a $50\%$ improvement in \gls{ood} detection probability, at the cost of reduced \gls{id} coverage rate of around $50\%$. Interesting directions for future work include extending the proposed framework to likelihood-free, simulation-based inference \citep{falkiewicz2024calibrating}, to continual learning \citep{li2024calibration}, and to large language models \citep{detommaso2024multicalibration}. 

\chapter{Optimized Certainty Equivalent Risk-Controlling Prediction Sets} \label{chapter:4}

\ifpdf
    \graphicspath{{Chapter4/Chapter4/Figs/}{Chapter4/Chapter4/Figs/PDF/}{Chapter4/Chapter4/Figs/}}
\else
    \graphicspath{{Chapter4/Chapter4/Figs/}{Chapter4/Chapter4/Figs/}}
\fi

\section{Overview} \label{sec:c4_overview}

The training-time calibration methods developed in Chapter~\ref{chapter:3} achieve strong empirical reliability by integrating \gls{id} calibration regularization, \gls{ocm}, and selective inference within a Bayesian framework. However, these methods require training the model from scratch and maintaining Bayesian ensembles at inference time, incurring significant computational cost. A natural alternative is post-hoc calibration, which operates on any pre-trained model using only a held-out calibration dataset, thereby eliminating the need for retraining. In particular, in safety-critical applications such as medical image segmentation, prediction systems must provide reliability guarantees that extend beyond conventional expected loss control. While \gls{rcps} offer probabilistic guarantees on the expected risk, they fail to capture tail behavior and worst-case scenarios that are crucial in high-stakes settings.

To address this, this chapter introduces \gls{oce-rcps}, a novel post-hoc framework that provides high-probability guarantees on general \gls{oce} risk measures, including \gls{cvar} and entropic risk. \gls{oce-rcps} leverages \gls{ucb} to identify prediction set parameters that satisfy user-specified risk tolerance levels with provable reliability. We establish theoretical guarantees showing that \gls{oce-rcps} satisfies the desired probabilistic constraint for loss functions such as miscoverage and \gls{fnr}. Experiments on image segmentation demonstrate that \gls{oce-rcps} consistently meets target satisfaction rates across various risk measures and reliability configurations, while \gls{oce-crc} fails to provide probabilistic guarantees.

\begin{figure}[t] 
    \centering
    \centerline{\includegraphics[width=\textwidth]{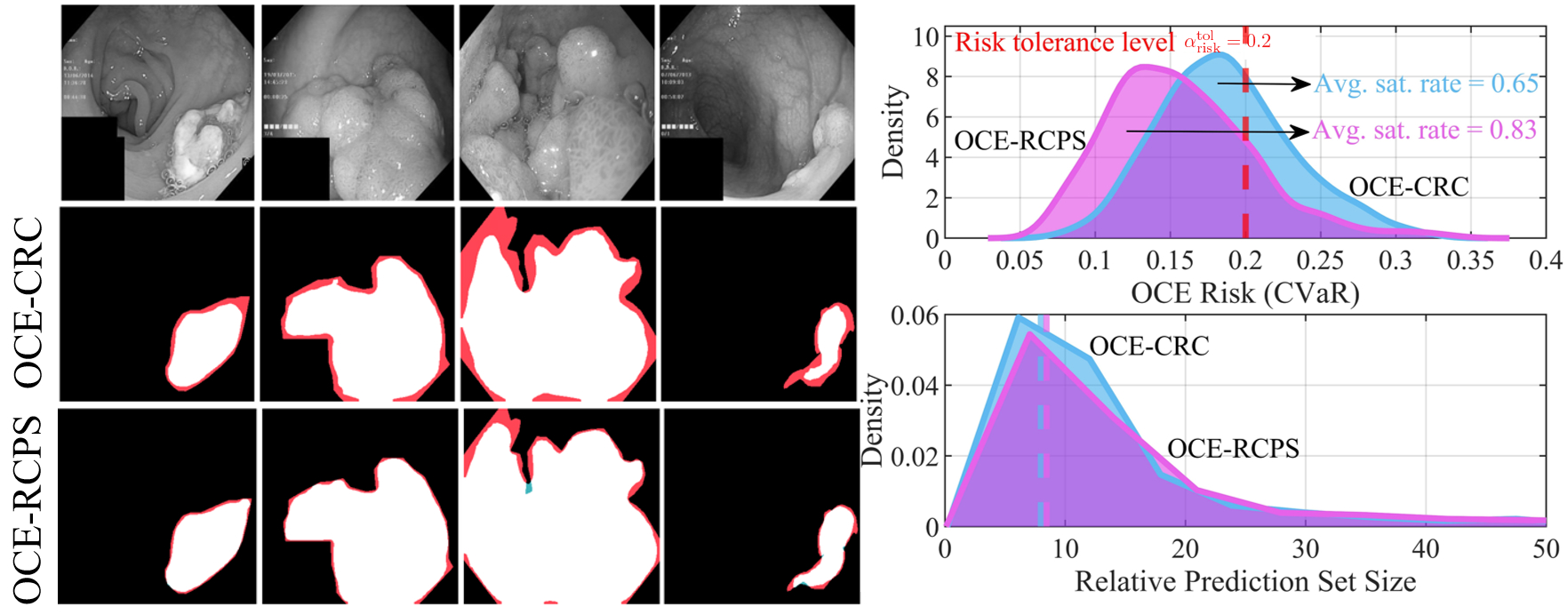}}
    \caption{Given an input $x$, e.g., an image, and a pre-trained model $p(y|x)$, \gls{oce-crc} \citep{yeh2025conformal} and the proposed \gls{oce-rcps} find a hyperparameter configuration $\hat{\lambda}$, e.g., an inclusion threshold, such that the reliability constraint $R_{\text{OCE}}(\hat{\lambda}) \leq \alpha_{\text{risk}}^{\text{tol}}$ is satisfied for any given \gls{oce} risk measure $R_{\text{OCE}}(\cdot)$, e.g., the \gls{cvar}. \gls{oce-crc} \citep{yeh2025conformal} ensures this condition only on average with respect to the calibration data used to optimize the hyperparameter $\lambda$. In contrast, \gls{oce-rcps} ensures that the condition $R_{\text{OCE}}(\hat{\lambda}) \leq \alpha_{\text{risk}}^{\text{tol}}$ holds with probability no smaller than a target level $1-\delta$. The left panel of the figure shows representative results for the \gls{cvar} of the \gls{fnr} loss in a segmentation task \citep{pogorelov2017kvasir}, with white, red, and green pixels denoting correct predictions, false negatives, and false positives, respectively. The right panel of the figure compares the empirical distribution of the \gls{cvar} and of the relative prediction set size with targets $\alpha_{\text{risk}}^{\text{tol}}=0.2$ and $1-\delta=0.8$. While \gls{oce-crc} yields hyperparameters $\hat{\lambda}$ violating the requirement $R_{\text{OCE}}(\hat{\lambda}) \leq \alpha_{\text{risk}}^{\text{tol}}$ with probability larger than the requirement $\delta=0.2$, the proposed \gls{oce-rcps} meets the target satisfaction rate $1-\delta=0.8$.}
    \label{fig:c4_overview}
\end{figure}

\section{Introduction} \label{sec:c4_intro}

\subsection{Motivation} \label{sec:c4_motivation}

In safety-critical applications such as medical diagnosis, autonomous driving, and financial decision-making, predictive models must provide not only accurate predictions but also rigorous reliability guarantees. Traditional point predictions, even when accompanied by confidence scores from probabilistic models, often fail to quantify uncertainty reliably due to miscalibration \citep{guo2017calibration, huang2025calibrating}. This limitation can lead to catastrophic failures, such as missed cancer diagnoses in medical imaging \citep{abdar2021uncertainty} or false negatives in anomaly detection systems \citep{zhu2025conformal}. 

To address these challenges, recent advances in \emph{\gls{cp}} have introduced prediction sets that provide distribution-free coverage guarantees \citep{shafer2008tutorial}. Among these methods, \emph{\gls{rcps}} \citep{bates2021distribution} offer a principled approach to control the expected loss of the prediction set with high probability. Specifically, the loss, averaged over the test data point, is kept below a user-defined threshold with high probability with respect to the realization of the held-out dataset used for calibration.

However, \gls{rcps} focuses exclusively on the average risk, which fails to capture critical aspects of risk-sensitive decision-making, particularly the tail behavior of the loss distribution. As an example, consider the medical image segmentation task illustrated in Fig.~\ref{fig:c4_overview}. Given an input image and a pre-trained segmentation model $p(y|x)$, the objective is to identify a set of pixels that constitute a reliable prediction of the object of interest. In this setting, controlling the average \gls{fnr}, i.e., the average fraction of missed object pixels, may be insufficient, as medical practitioners often require stronger guarantees that account for worst-case test scenarios. A system that performs well on average, but fails catastrophically on a non-negligible fraction of test data points may be unacceptable in clinical settings.

This motivates the need for \gls{oce} risk measures \citep{ben2007old}, which provide a flexible framework for controlling risk aversion through measures such as \gls{cvar} and entropic risk. The recently introduced \emph{\gls{oce-crc}} \citep{yeh2025conformal} guarantees that the \gls{oce} risk remains below a target level $\alpha_{\text{risk}}^{\text{tol}}$. However, this guarantee holds only on average over the possible held-out datasets used for calibration. As illustrated in Fig.~\ref{fig:c4_overview}, this implies that \gls{oce-crc} can fail to produce prediction sets that satisfy the risk constraint for a high fraction of calibration datasets, leading to unreliable decisions in practice.

\begin{table}[t]
\centering
\caption{Comparison of prediction set calibration schemes: The table lists the types of guarantees provided with respect to (w.r.t.) test and calibration data.}\label{tab:c4_summarize}
\begin{tabular}{lcc}
\toprule
\textbf{Scheme} & \textbf{w.r.t. test data} & \textbf{w.r.t. calibration data} \\
\midrule
\gls{rcps} \citep{bates2021distribution} & Average & High probability \\
\gls{oce-crc} \citep{yeh2025conformal} & \gls{oce} & Average \\
\gls{oce-rcps} [ours] & \gls{oce} & High probability \\
\bottomrule
\end{tabular}
\end{table}




\subsection{Main Contributions} \label{sec:c4_contributions}

In this work, we introduce \gls{oce-rcps}, which extends \gls{rcps} to provide high-probability guarantees on general \gls{oce} risk measures (see Table~\ref{tab:c4_summarize}). By constructing \gls{ucb} on the \gls{oce} risk, \gls{oce-rcps} ensures that the risk constraint holds with high probability. This probabilistic guarantee enables practitioners to control the tail risk of their prediction systems with respect to both test data and calibration data, making \gls{oce-rcps} particularly suitable for high-stakes applications. 


    

The remainder of this chapter is organized as follows. Sec.~\ref{sec:c4_problem_definition} formalizes the problem setting and introduces \gls{oce} risk measures. Sec.~\ref{sec:c4_oce_crc} reviews \gls{oce-crc} as background. Sec.~\ref{sec:c4_oce_rcps} presents our proposed \gls{oce-rcps} method along with theoretical guarantees. Sec.~\ref{sec:c4_experiments} provides experimental results on medical image segmentation. Finally, Sec.~\ref{sec:c4_conclusion} concludes the paper.

\section{Problem Definition} \label{sec:c4_problem_definition}

Consider an inference setting, in which a pre-designed probabilistic model $p(y|x)$ is deployed to predict the target variable $y \in \mathcal{Y}$ given the input covariates $x \in \mathcal{X}$, where $p(y|x)$ is a conditional distribution in the target space $\mathcal{Y}$ given input $x$. In safety-critical applications, such as medical image segmentation, a conventional point prediction such as $\hat{y}(x) = \arg \max_{y \in \mathcal{Y}} p(y|x)$, can lead to missed diagnoses. Furthermore, the probabilistic distribution $p(y|x)$ is typically miscalibrated, and thus it cannot be directly used to quantify predictive uncertainty \citep{guo2017calibration, huang2025calibrating}. Calibration methods like \gls{cp} address these limitations by leveraging held-out data with the goal of identifying a subset $\Gamma_{\lambda}(x)$ of the target domain $\mathcal{Y}$ that satisfies a user-specified reliability requirement.

The prediction set $\Gamma_{\lambda}(x)$ is constructed to include all output values $y \in \mathcal{Y}$ whose score $\pi(y|x)$ exceeds a threshold $1-\lambda$, i.e.,
\begin{align} \label{eq:c4_set_construction}
    \Gamma_{\lambda}(x) = \{ y \in \mathcal{Y}: \pi(y|x) \geq 1 - \lambda \}.
\end{align}
The score $\pi(y|x)$ is a function of the predictive distribution $p(\cdot|x)$, with a typical example being the confidence level $\pi(y|x) = p(y|x)$ \citep{shafer2008tutorial}.

For example, as illustrated in Fig.~\ref{fig:c4_overview}, in medical image segmentation, the set $\mathcal{Y}$ contains all pixels in an image, and the prediction set $\Gamma_{\lambda}(x)$ corresponds to the collection of pixels whose score, representing the predicted probability of being positive exceeds, a threshold $1-\lambda$. Note that the prediction set (\ref{eq:c4_set_construction}) satisfies the monotonicity condition
\begin{align} \label{eq:c4_set_monotonicity}
    \Gamma_{\lambda_1}(x) \subseteq \Gamma_{\lambda_2}(x), \qquad \text{if $\lambda_1 \leq \lambda_2$},
\end{align}
with respect to the parameter $\lambda \in \mathbb{R}$.

The hyperparameter $\lambda$ controls the trade-off between the prediction size and reliability, and it is selected based on calibration data to meet target reliability requirements. In order to define the reliability requirements, let $\ell(y, \Gamma_{\lambda}(x))$ denote a non-negative bounded loss function that measures the discrepancy between the prediction set $\Gamma_{\lambda}(x)$ and output $y \in \mathcal{Y}$. We require that the loss function satisfies the monotonicity property
\begin{align} \label{eq:c4_nesting_property}
    \ell(y, \Gamma_{\lambda_1}(x)) \geq \ell(y, \Gamma_{\lambda_2}(x)), \qquad \text{if $ \Gamma_{\lambda_1}(x) \subseteq \Gamma_{\lambda_2}(x)$},
\end{align}
so that larger prediction sets correspond to lower losses. A canonical example is the miscoverage loss $\ell(y, \Gamma_{\lambda}(x)) = \mathbbm{1}(y \notin \Gamma_{\lambda}(x))$, where $\mathbbm{1}(\cdot)$ is the indicator function defined as $\mathbbm{1}(\text{true})=1$ and $\mathbbm{1}(\text{false})=0$. Other examples include the \gls{fnr} in multi-label classification and image segmentation \citep{bates2021distribution}. 

In this work, we consider the general class of \gls{oce} risks, which include standard measures such as the expected loss, the entropic risk, and the \gls{cvar} (see Table~\ref{tab:c4_oce_risks}). This family provides flexible and mathematically convenient metrics to control the level of risk aversion in the evaluation and optimization of decision policies \citep{ben2007old}. Formally, the class of \gls{oce} risks include metrics of the form
\begin{align} \label{eq:c4_oce_risk}
    R_{\text{OCE}}(\lambda) = \inf_{t \in \mathbb{R}} \left\{ R(\lambda, t) =  t + \mathbb{E}[\psi \left(\ell(y, \Gamma_{\lambda}(x)) - t \right)] \right\},
\end{align}
where $\psi(\cdot)$ is a nondecreasing, closed, and convex cost function, and the inner expectation is taken with respect to the joint distribution $p(x,y)$ of the pair $(x,y)$. As summarized in Table~\ref{tab:c4_oce_risks}, the choice of cost function $\psi(u)$ determines the specific risk measure induced by the \gls{oce} framework.

\begin{table}[t]
\centering
\caption{Examples of \gls{oce} risk measures.}\label{tab:2}
\label{tab:c4_oce_risks}
\begin{tabular}{lc}
\toprule
\textbf{\gls{oce} risk measure} & \textbf{Cost function $\psi(u)$ in (\ref{eq:c4_oce_risk})}\\
\midrule
Average risk & $\psi(u) = u$ \\
Entropic risk & $\psi(u) = \dfrac{1}{\zeta} \left( e^{\zeta u} - 1\right),  \zeta > 0$ \\
\gls{cvar}  & $\psi(u) = \dfrac{1}{1 - \zeta} \max(u, 0), \zeta \in [0, 1)$ \\
\bottomrule
\end{tabular}
\end{table}

Given a calibration dataset $\mathcal{D}^{\text{cal}} = \{(x_i, y_i)\}_{i=1}^{|\mathcal{D}^{\text{cal}}|}$ with i.i.d. samples $(x,y) \sim p(x,y)$, we wish to identify a parameter $\hat{\lambda}$ such that the \gls{oce} risk (\ref{eq:c4_oce_risk}) satisfies the reliability condition $R_{\text{OCE}}(\hat{\lambda}) \leq \alpha_{\text{risk}}^{\text{tol}}$ for some tolerance level $\alpha_{\text{risk}}^{\text{tol}}$ with probability no smaller than a user-defined level $1-\delta \in [0,1]$. This requirement is formally expressed as the inequality
\begin{align} \label{eq:c4_oce_rcps_requirement}
    \Pr \big[ R_{\text{OCE}}(\hat{\lambda}) \leq \alpha_{\text{risk}}^{\text{tol}}  \big] \geq 1-\delta,
\end{align}
where the probability is taken over the distribution of the parameter $\hat{\lambda}$. When the \gls{oce} risk coincides with the average risk, the requirement (\ref{eq:c4_oce_rcps_requirement}) coincides with the objective of \gls{rcps} \citep{yeh2025conformal}. Our goal is thus to generalize \gls{rcps} to any \gls{oce} risk.

\section{\gls{oce} Conformal Risk Control} \label{sec:c4_oce_crc}
In this section, we briefly review \gls{oce-crc} \citep{yeh2025conformal}, which provides guarantees on the average \gls{oce} risk. Specifically, given a calibration dataset $\mathcal{D}^{\text{cal}}$, \gls{oce-crc} seeks a risk-controlled parameter $\hat{\lambda}$ such that the \gls{oce} risk $R_{\text{OCE}}(\hat{\lambda})$ does not exceed the target $\alpha_{\text{risk}}^{\text{tol}}$ on average with respect to the realizations of the calibration data $\mathcal{D}^{\text{cal}}$, i.e.,
\begin{align}\label{eq:c4_oce-crc}
    \mathbb{E}\big[ R_{\text{OCE}}(\hat{\lambda}) \big] \leq \alpha_{\text{risk}}^{\text{tol}}.
\end{align}

To this end, for a fixed $t \in \mathbb{R}$, \gls{oce-crc} selects the smallest value of the hyperparameter $\lambda$ such that an upper bound on the population \gls{oce} risk does not exceed a target risk tolerance level $\alpha_{\text{risk}}^{\text{tol}}$, i.e.,
\begin{align} \label{eq:c4_optimal_lambda_crc}
    \hat{\lambda} = \inf \left\{ \lambda \in \mathbb{R}: \frac{|\mathcal{D}^{\text{cal}}|}{|\mathcal{D}^{\text{cal}}| + 1} \hat{R}^{\text{cal}}(\lambda,t) + \frac{B(\lambda, t)}{|\mathcal{D}^{\text{cal}}| + 1} \leq \alpha_{\text{risk}}^{\text{tol}} \right\},
\end{align}
where $\hat{R}^{\text{cal}}(\lambda,t)$ is the empirical estimate of $R(\lambda,t)$ in (\ref{eq:c4_oce_risk}) evaluated on the calibration dataset $\mathcal{D}^{\text{cal}}$; $B(\lambda, t) = t + \psi (\ell_{\max}(\lambda) - t)$ is an upper bound on the function $R(\lambda,t)$ in (\ref{eq:c4_oce_risk}), with $\ell_{\max}(\lambda) = \sup_{(x,y)} \ell(y, \Gamma_{\lambda}(x))$. As an example, for a bounded loss $\ell(y, \Gamma_{\lambda}(x)) \leq 1$, such as the \gls{fnr}, we have $\ell_{\max}(\lambda) = 1$ for all values of $\lambda$.

While the guarantee (\ref{eq:c4_oce-crc}) with the selected hyperparameter $\hat{\lambda}$ in (\ref{eq:c4_optimal_lambda_crc}) holds for any fixed $t$, a poorly chosen parameter $t$ results in an unnecessarily larger prediction set $\Gamma_{\hat{\lambda}}(x)$. To improve the efficiency of the prediction set $\Gamma_{\hat{\lambda}}(x)$, \gls{oce-crc} optimizes the parameter $t$ on a separate held-out optimization dataset $\mathcal{D}^{\text{opt}} = \{(x_i, y_i)\}_{i=1}^{|\mathcal{D}^{\text{opt}}|}$ with i.i.d. samples $(x,y) \sim p(x,y)$ by addressing the convex problem
\begin{align} \label{eq:c4_optimal_t}
    t^* = \arg \min_{t \in \mathbb{R}} \hat{R}^{\text{opt}}(\lambda, t),
\end{align}
where $\hat{R}^{\text{opt}}(\lambda, t)$ is the empirical estimate of $R(\lambda,t)$ in (\ref{eq:c4_oce_risk}) evaluated on the optimization dataset $\mathcal{D}^{\text{opt}}$. The use of a separate dataset $\mathcal{D}^{\text{opt}}$ ensures that the value $t^*$ in (\ref{eq:c4_optimal_t}) is independent of the calibration data $\mathcal{D}^{\text{cal}}$, which is necessary to preserve the validity of the \gls{oce} risk guarantee in (\ref{eq:c4_oce-crc}).

\section{\gls{oce} Risk-Controlling Prediction Set} \label{sec:c4_oce_rcps}

\gls{oce-crc} guarantees the \gls{oce} risk requirement $R_{\text{OCE}}(\hat{\lambda}) \leq \alpha_{\text{risk}}^{\text{tol}}$ only on average with respect to calibration data as in (\ref{eq:c4_oce-crc}), providing no control over the probability that the \gls{oce} risk constraint (\ref{eq:c4_oce-crc}) is violated on any given trial. As discussed in Sec.~\ref{sec:c4_intro}, in safety-critical applications such as medical image analysis, practitioners require assurance that the risk constraint is satisfied not merely on average, but with high probability with respect to the available calibration data (see Fig.~\ref{fig:c4_overview}). To describe the proposed solution, in this section, we first review \gls{rcps} \citep{bates2021distribution}, and then define \gls{oce-rcps}.

\subsection{Risk-Controlling Prediction Set}
\gls{rcps} \citep{bates2021distribution} controls the expected loss $R_{\text{avg}}(\lambda) = \mathbb{E} \left[ \ell(y, \Gamma_{\lambda}(x)) \right]$, which corresponds to a special case of the \gls{oce} risk (see Table~\ref{tab:c4_oce_risks}). Specifically, \gls{rcps} identifies a hyperparameter $\hat{\lambda}$ such that condition (\ref{eq:c4_oce_rcps_requirement}) is satisfied for the expected loss $R_{\text{avg}}(\hat{\lambda})$. To this end, using the calibration dataset $\mathcal{D}^{\text{cal}}$, \gls{rcps} constructs an UCB  $\hat{R}^+_{\text{avg}}(\lambda, \delta)$ on the expected loss $R_{\text{avg}}(\lambda)$, satisfying the inequality $R_{\text{avg}}(\lambda) \leq \hat{R}^+_{\text{avg}}(\lambda, \delta)$ with probability at least $1-\delta$, i.e.,
\begin{align} \label{eq:c4_rcps_inequality}
    \Pr\big[ R_{\text{avg}}(\lambda) \leq \hat{R}^+_{\text{avg}}(\lambda, \delta) \big] \geq 1 - \delta.
\end{align}

Then, \gls{rcps} selects the smallest value of hyperparameter $\lambda$ such that the UCB does not exceed the target value $\alpha_{\text{risk}}^{\text{tol}}$ for all $\lambda' \geq \lambda$, i.e.,
\begin{align} \label{eq:c4_optimal_lambda_rcps}
    \hat{\lambda} = \inf \left\{ \lambda \in \mathbb{R}: \hat{R}_{\text{avg}}^+(\lambda', \delta) \leq \alpha_{\text{risk}}^{\text{tol}} \text{ for all } \lambda' \geq \lambda   \right\}. 
\end{align}

\subsection{\gls{oce} Risk-Controlling Prediction Set}
\gls{oce-rcps} extends \gls{rcps} to offer control over any \gls{oce} risk. For any fixed value of parameter $t \in \mathbb{R}$, the definition of the \gls{oce} (\ref{eq:c4_oce_risk}) implies the inequality 
\begin{align} \label{eq:c4_inequality_t}
    R_{\text{OCE}}(\lambda) \leq R(\lambda, t).
\end{align}
The quantity $R(\lambda, t)$ in (\ref{eq:c4_oce_risk}) can be estimated using the calibration dataset $\mathcal{D}^{\text{cal}}$. Specifically, in a manner similar to \gls{rcps}, we use the calibration dataset $\mathcal{D}^{\text{cal}}$ to construct an UCB on the function $R(\lambda, t)$. Here we adopt the \gls{sota} \gls{wsr} UCB \citep{bates2021distribution}.

Given the calibration dataset $\mathcal{D}^{\text{cal}}$ and a target probability $\delta$, the \gls{wsr} UCB on the risk $R(\lambda,t)$ is given by
\begin{align} \label{eq:c4_wsr_bound}
    \hat{R}^{+}(\lambda, t, \delta) =\inf\left\{
    R\ge 0:\;
    \max_{i \in \mathcal{D}^{\text{cal}}}\mathcal{K}_i(R, \lambda)> \frac{1}{\delta},
    \right\}
\end{align}
where function $\mathcal{K}_i(R,\lambda)$ is defined as
\begin{align}
    \mathcal{K}_i(R,\lambda) =\prod_{j=1}^{i} \left( 1+\eta_j\Bigl(t + \psi\bigl(\ell(y_j,\Gamma_{\lambda}(x_j))-t\bigr) - R\Bigr) \right),
\end{align}
with $\eta_j > 0$. The constant $\eta_j$ can be potentially optimized as a function of the previous $j-1$ observations $(x_i,y_i)_{i=1}^{j-1}$ in the calibration dataset (see \citep[Prop.~5]{bates2021distribution}). For our example, we adopt the online Newton step scheme \citep{bates2021distribution}. The \gls{wsr} UCB in (\ref{eq:c4_wsr_bound}) satisfies the inequality
\begin{align} \label{eq:c4_single_ucb}
    \Pr \big[ R(\lambda, t) \leq \hat{R}^+(\lambda, t, \delta) \big] \geq 1 - \delta,
\end{align}
where the probability is taken over the calibration data.

The following lemma establishes that the UCB $\hat{R}^+(\lambda, t, \delta)$ serves as a valid UCB on the \gls{oce} risk $R_{\text{OCE}}(\lambda)$ for any $t\in\mathbb{R}$.

\textbf{Lemma 4.1:} \textit{For any fixed value of parameter $t \in \mathbb{R}$, the UCB $\hat{R}^+(\lambda, t, \delta)$ satisfies an inequality
\begin{align} \label{eq:c4_lemma}
    \Pr \big[ R_{\text{OCE}}(\lambda) \leq \hat{R}^+(\lambda, \delta, t) \big] \geq 1 - \delta.
\end{align}
Proof:} \text{Use (\ref{eq:c4_inequality_t}) in (\ref{eq:c4_single_ucb})}. \hfill $\blacksquare$

Finally, \gls{oce-rcps} chooses the smallest value of hyperparameter $\lambda$ such that the UCB in (\ref{eq:c4_wsr_bound}) does not exceed the target $\alpha_{\text{risk}}^{\text{tol}}$ for all $\lambda' \geq \lambda$, i.e.,
\begin{align} \label{eq:c4_optimal_lambda_oce_rcps}
    \hat{\lambda} = \inf \left\{ \lambda \in \mathbb{R}: \hat{R}^+(\lambda', \delta, t) \leq \alpha_{\text{risk}}^{\text{tol}} \text{ for all } \lambda' \geq \lambda   \right\}.
\end{align}
As for \gls{oce-crc}, the tightness of the resulting set $\Gamma_{\hat{\lambda}}(x)$ depends on the choice of hyperparameter $t$. As in \gls{oce-crc}, we propose to choose an optimized value of parameter $t$ as in (\ref{eq:c4_optimal_t}) based on held-out data $\mathcal{D}^{\text{opt}}$.

\subsection{Theoretical Guarantees}
\gls{oce-rcps} satisfies the desired reliability requirement (\ref{eq:c4_oce_rcps_requirement}).

\textbf{Theorem 4.1:} \textit{Let $\hat{\lambda}$ be selected as in (\ref{eq:c4_optimal_lambda_oce_rcps}), with parameter $t^*$ in (\ref{eq:c4_optimal_t}). Under the monotonicity condition (\ref{eq:c4_set_monotonicity}), \gls{oce-rcps} satisfies the requirement
\begin{align}
    \Pr \big[ R_{\text{OCE}}(\hat{\lambda}) \leq \alpha_{\text{risk}}^{\text{tol}}  \big] \geq 1-\delta.
\end{align}}

\section{Experiments} \label{sec:c4_experiments}
In this section, to validate the proposed \gls{oce-rcps} scheme, we report empirical results for the tumor segmentation task \citep{angelopoulos2022conformal}.

\subsection{Task, Baselines, and Implementation}
Following \citep{bates2021distribution}, we aggregate data from five open-source polyp segmentation benchmarks: Kvasir \citep{pogorelov2017kvasir}, Hyper-Kvasir \citep{borgli2020hyperkvasir}, CVC-ClinicDB and CVC-ColonDB \citep{bernal2012towards}, and ETIS-Larib \citep{silva2014toward}, yielding a combined dataset with $1781$ examples of segmented polyps. We adopt a pre-trained PraNet \citep{fan2020pranet} as the base segmentation model $p(y|x)$.

For all schemes and for each trial, we randomly select a reference dataset $\mathcal{D}$ of size $|\mathcal{D}| = 1000$ and a test dataset $\mathcal{D}^{\text{te}}$ of size $|\mathcal{D}^{\text{te}}| = 781$. The reference dataset $\mathcal{D}$ is randomly split into two disjoint datasets, namely the optimization dataset $\mathcal{D}^{\text{opt}}$ used to estimate the parameter $t^*$, and the calibration dataset $\mathcal{D}^{\text{cal}}$ used to identify the parameter $\hat{\lambda}$, with cardinalities $|\mathcal{D}^{\text{opt}}| = 200$ and $|\mathcal{D}^{\text{cal}}| = 800$, respectively. All results are averaged over $1000$ independent trials\footnotemark[2]. \footnotetext[2]{Code can be found at \url{https://github.com/kclip/OCE-RCPS}.}

\begin{figure} [tb] 
    \centering
    \centerline{\includegraphics[scale=0.25]{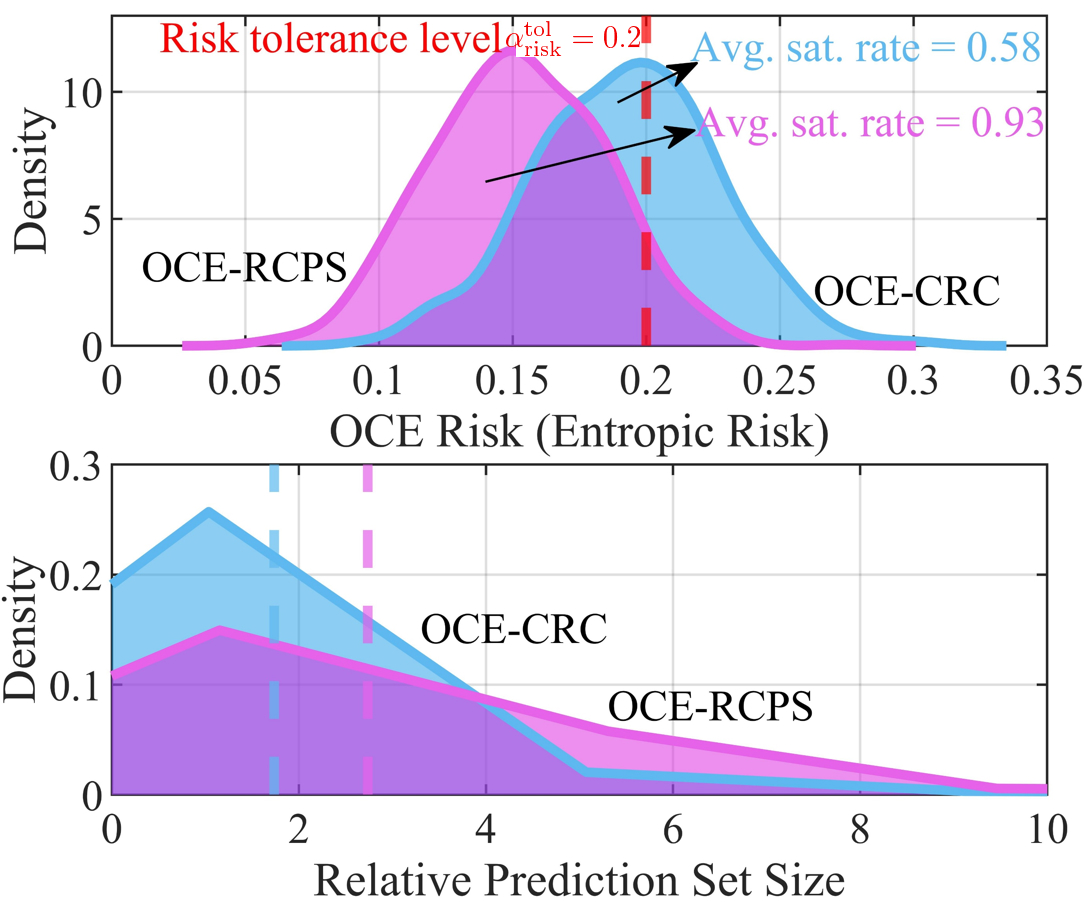}}
     \caption{Distribution of entropic risk (top) and relative prediction set size (bottom) for \gls{oce-crc} and \gls{oce-rcps}, with tolerated risk level $\alpha_{\text{risk}}^{\text{tol}}=0.2$, target satisfaction rate $1-\delta = 0.8$, and entropic risk sensitivity level $\zeta=3$. Dashed lines in the bottom panel indicate the median relative prediction set size for each method.}
    \label{fig:c4_dist_entropic} 
\end{figure}

We adopt the \gls{fnr} loss, i.e., $\ell(y, \Gamma_{\lambda}(x)) = 1-|y \cap \Gamma_{\lambda}(x)|/|y|$, where $y$ is the set of pixels containing the object of interest and $|y|$ its cardinality, and consider the following evaluation metrics:
\begin{itemize}
    \item \emph{Average satisfaction rate}: This is the proportion of trials (i.e., calibration datasets) in which the \gls{oce} risk $R(\hat{\lambda})$ of the selected hyperparameter $\hat{\lambda}$ does not exceed the target tolerance level $\alpha_{\text{risk}}^{\text{tol}}$. This quantity estimates the left-hand side of the requirement (\ref{eq:c4_oce_rcps_requirement}).
    \item \emph{Relative prediction set size}: This is the ratio, $|\Gamma(\hat{\lambda})|/|y|$, between the size of the prediction set and the ground-truth size of the polyp region.
    \item \emph{Distributions of the \gls{oce} risk metric and the relative prediction set size}: The distributions of the \gls{oce} risk metric and of the relative prediction set size are obtained as empirical estimates using \gls{kde} across the different trials.
\end{itemize}

\subsection{Results}

\textbf{\gls{oce-crc} versus \gls{oce-rcps}:}
To start, in the right top panel of Fig.~\ref{fig:c4_overview} and top panel of Fig.~\ref{fig:c4_dist_entropic}, we report the performance of \gls{oce-crc} \citep{yeh2025conformal} and of the proposed \gls{oce-rcps} in terms of risk and of relative prediction set size, for the \gls{cvar} with parameter $\zeta=0.9$ in the right panel of Fig.~\ref{fig:c4_overview} and for the entropic risk with parameter $\zeta = 3$ in Fig.~\ref{fig:c4_dist_entropic}. As shown in the right top panel of Fig.~\ref{fig:c4_overview} and top panel of Fig.~\ref{fig:c4_dist_entropic}, the distribution of the \gls{cvar} and of the entropic risk obtained by \gls{oce-crc} exhibits significant probability mass beyond the target risk tolerance level $\alpha_{\text{risk}}^{\text{tol}}=0.2$, resulting in an average satisfaction rate of only $0.65$ and of only $0.58$, respectively, well below the target of $1-\delta = 0.8$. This confirms that controlling only the expected risk, as in \gls{oce-crc}, provides no guarantee on the tail behavior of the risk distribution. In contrast, \gls{oce-rcps} achieves an average satisfaction rate of $0.83$ and of $0.93$ for the \gls{cvar} and entropic risk, respectively, meeting the target reliability requirement as in (\ref{eq:c4_oce_rcps_requirement}). To achieve these results, as seen in the bottom panel of the figures, \gls{oce-rcps} increases the median of relative prediction set size from $7.98$ to $8.45$ for the \gls{cvar}, and from $1.74$ to $2.74$ for the entropic risk, respectively.

\begin{figure} [tb] 
    \centering
    \centerline{\includegraphics[scale=0.2]{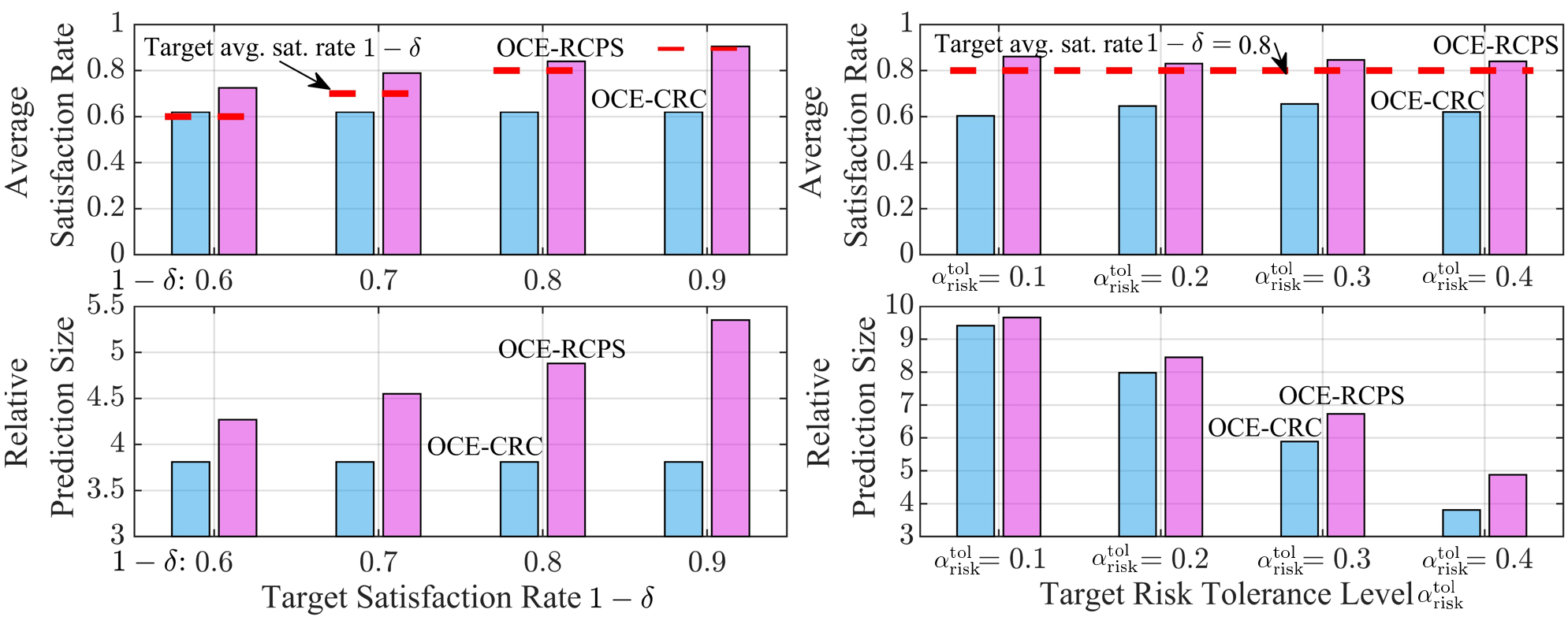}}
     \caption{Average satisfaction rate (top) and relative prediction set size (bottom) for \gls{oce-crc} and \gls{oce-rcps} across (\emph{i}) varying target satisfaction rates $1-\delta \in \{ 0.6, 0.7, 0.8,0.9 \}$, with tolerated risk level $\alpha_{\text{risk}}^{\text{tol}}=0.4$ (left), and (\emph{ii}) various target risk levels $\alpha_{\text{risk}}^{\text{tol}} \in \{ 0.1, 0.2, 0.3, 0.4 \}$, with target average satisfaction rate $1-\delta=0.8$ (right). Both with \gls{cvar} concentration level $\zeta=0.9$. Red dashed lines in the top panel indicate the corresponding target levels $1-\delta$.}
    \label{fig:c4_changing_delta} 
\end{figure}

\textbf{Robustness of \gls{oce-rcps} across reliability configurations:}
We now turn to analyzing the performance of \gls{oce-rcps} across a range of configurations. The left top panel of Fig.~\ref{fig:c4_changing_delta} shows the average satisfaction rate as a function of the target quantile level $1-\delta$. It is observed that, as the target satisfaction rate $1-\delta$ increases from $0.6$ to $0.9$, \gls{oce-rcps} consistently meets the target reliability requirement as in (\ref{eq:c4_oce_rcps_requirement}), while the average satisfaction rate of \gls{oce-crc} remains around $0.6$, irrespective of the choice of the target of $1-\delta$. This confirms that \gls{oce-crc} lacks a mechanism to control the probabilistic requirement on the \gls{oce} risk as in (\ref{eq:c4_oce_rcps_requirement}). By contrast, as shown in the bottom left panel of Fig.~\ref{fig:c4_changing_delta}, \gls{oce-rcps} provides a controllable reliability knob, adaptively inflating the prediction set size to meet the prescribed target.

Right panel of Fig.~\ref{fig:c4_changing_delta} plots the average satisfaction rate and the relative prediction set size as a function of the target risk tolerance level $\alpha_{\text{risk}}^{\text{tol}}$. \gls{oce-rcps} consistently satisfies the target satisfaction rate $1-\delta=0.8$ across various tolerated risk levels $\alpha_{\text{risk}}^{\text{tol}} \in \{ 0.1, 0.2, 0.3, 0.4 \}$, while \gls{oce-crc} does not. The bottom left panel of Fig.~\ref{fig:c4_changing_delta} shows that the prediction set size of \gls{oce-rcps} inflates as the reliability requirement becomes stricter, i.e., larger target average satisfaction rate $1-\delta$ or smaller risk tolerated level $\alpha_{\text{risk}}^{\text{tol}}$.



\begin{figure} [tb] 
    \centering
    \centerline{\includegraphics[scale=0.05]{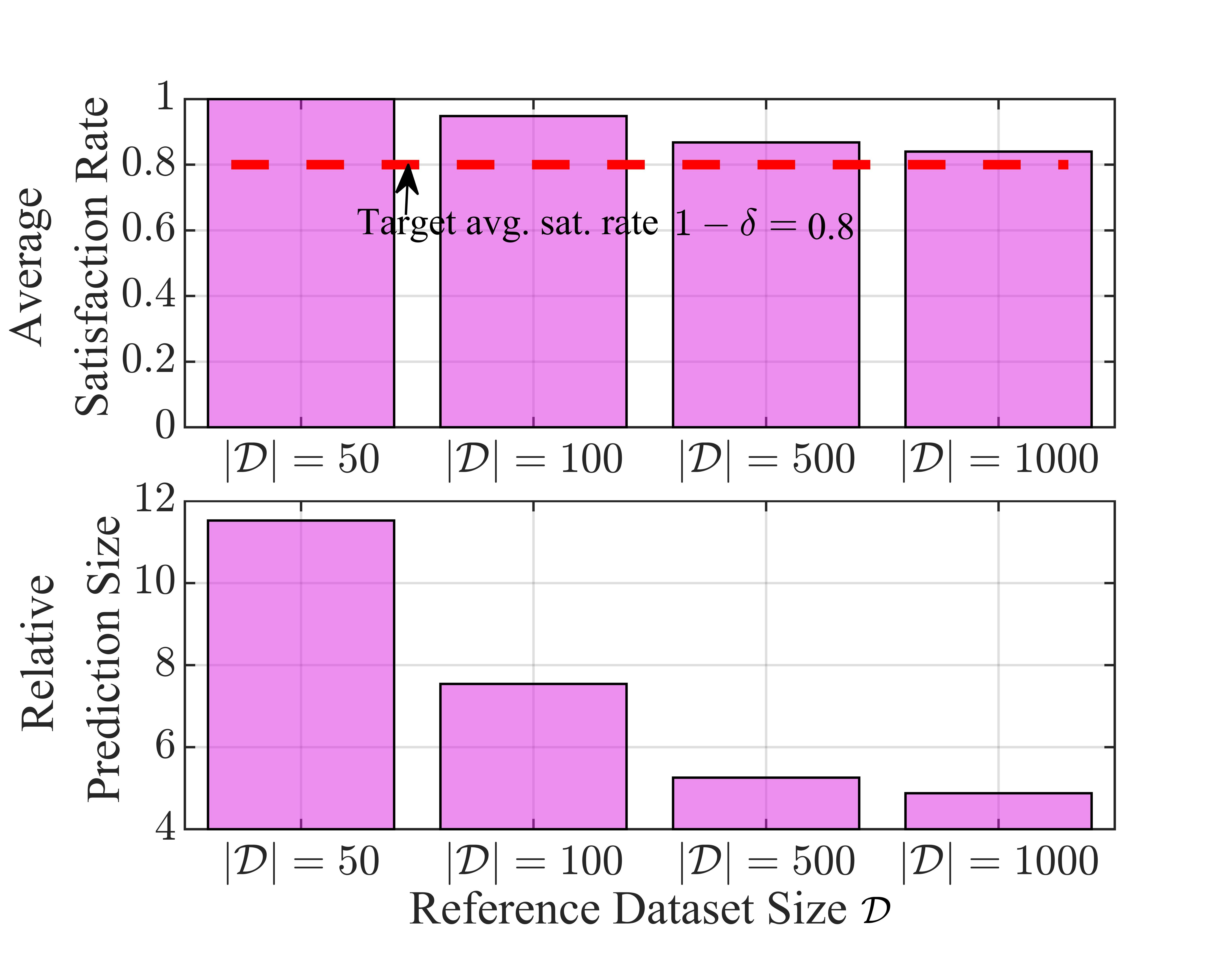}}
     \caption{Performance of \gls{oce-crc} and \gls{oce-rcps} on the tumor segmentation task across various reference dataset sizes $|\mathcal{D}| \in \{ 50, 100, 500, 1000 \}$, with target average satisfaction rate $1-\delta=0.8$, tolerated risk level $\alpha_{\text{risk}}^{\text{tol}}=0.4$ and \gls{cvar} concentration level $\zeta=0.9$. The top panel shows the average satisfaction rate, with a red dashed line indicating a target level $1-\delta$. The bottom panel shows the average set size as a fraction of polyp size. 
     }
    \label{fig:c4_changing_reference_size} 
\end{figure}

\textbf{How much reference data is sufficient?}
In Fig.~\ref{fig:c4_changing_reference_size}, we report the performance of \gls{oce-rcps} by plotting the average satisfaction rate and the relative prediction set size as a function of the reference dataset size $|\mathcal{D}|$, across various reference dataset sizes $|\mathcal{D}| \in \{ 50, 100, 500, 1000 \}$. As shown in the top panel of Fig.~\ref{fig:c4_changing_reference_size}, even with as few as $50$ reference samples, \gls{oce-rcps} achieves a satisfaction rate close to $1$, demonstrating that the reliability guarantee holds uniformly regardless of the amount of the reference data. Moreover, the efficiency of \gls{oce-rcps} improves substantially with more available data, as shown in the bottom panel of Fig.~\ref{fig:c4_changing_reference_size}.

\section{Conclusion} \label{sec:c4_conclusion}
In this chapter, we proposed \gls{oce-rcps}, a framework that can control the tail behavior on the \gls{oce} risk with respect to the calibration data, not just merely on average. Future directions include extending \gls{oce-rcps} to distribution-shift settings and applying the framework to other safety-critical domains such as autonomous driving.
\chapter{Distilling Calibration via Conformalized Credal Inference} \label{chapter:5}

\ifpdf
    \graphicspath{{Chapter5/Chapter5/Figs/}{Chapter5/Chapter5/Figs/PDF/}{Chapter5/Chapter5/Figs/}}
\else
    \graphicspath{{Chapter5/Chapter5/Figs/}{Chapter5/Chapter5/Figs/}}
\fi

\section{Overview}

While the post-hoc calibration developed in Chapter~\ref{chapter:4} eliminates the computational burden of trainable calibration methods described in Chapter~\ref{chapter:3}, the informativeness of the resulting prediction sets fundamentally depends on the quality of the underlying pre-trained model. Applying post-hoc methods to a weak small-scale model yields formally valid but uninformatively large prediction sets, while applying them to a powerful large-scale model restores the computational cost at inference time. More broadly, deploying \gls{ai} models on edge devices involves a delicate balance between meeting stringent complexity constraints, such as limited memory and energy resources, and ensuring reliable performance in sensitive decision-making tasks. One way to enhance reliability is through uncertainty quantification via Bayesian inference as described in Chapter~\ref{chapter:3}. This approach, however, typically necessitates maintaining and running multiple models in an ensemble, which may exceed the computational limits of edge devices.

To resolve this tension, this chapter introduces a low-complexity methodology by distilling calibration information from a large-scale cloud model to a small-scale edge model. In an offline phase, predictive probabilities generated by a high-complexity cloud-based model are leveraged to determine a threshold based on the typical divergence between the cloud and edge models. At run time, this threshold is used to construct \emph{credal sets} -- ranges of predictive probabilities that are guaranteed, with a user-selected confidence level, to include the predictions of the cloud model. The credal sets are obtained through thresholding of a divergence measure in the simplex of predictive probabilities. Experiments on visual and language tasks demonstrate that the proposed approach, termed \gls{cd-ci}, significantly improves calibration performance compared to low-complexity Bayesian methods, such as Laplace approximation, making it a practical and efficient solution for edge \gls{ai} deployments.

\section{Introduction} \label{sec:c5_intro}
\begin{figure}
    \centering
    \centerline{\includegraphics[width=\textwidth]{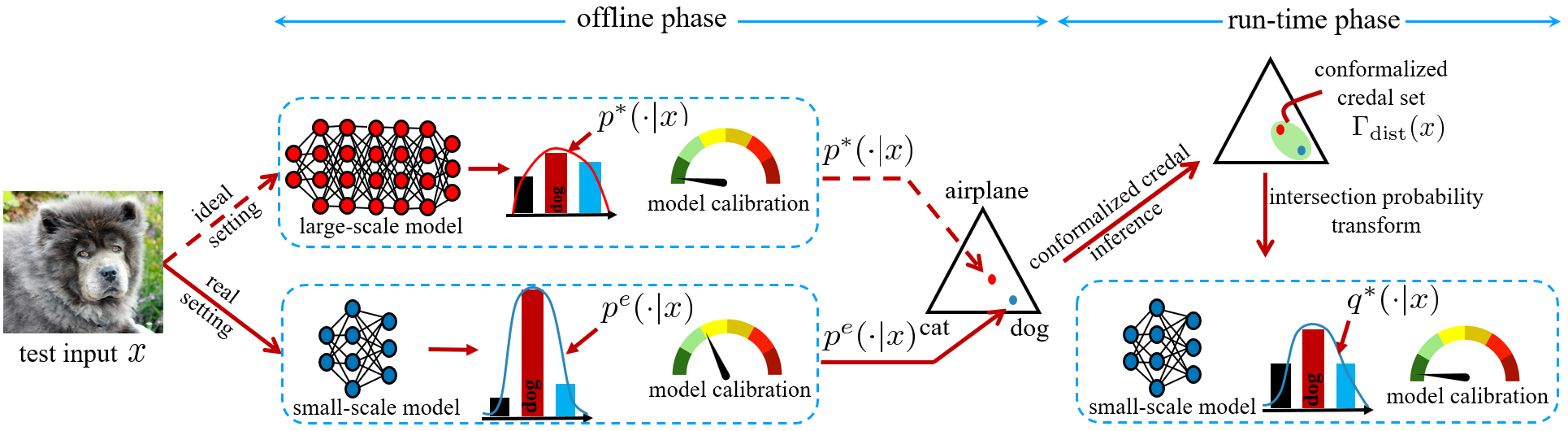}}
    \caption{Given an input $x$, the predictive distribution ideally coincides with that of a large-scale cloud-based model $p^*(\cdot|x)$. In the setting studied in this work, a small-scale edge-based model produces a probabilistic distribution $p^e(\cdot|x)$ that deviates from the reference distribution $p^*(\cdot|x)$, and is thus uncalibrated. The proposed conformalized credal inference-based scheme post-processes the small-scale edge model output $p^e(\cdot|x)$ via a simple thresholding mechanism to produce a subset $\Gamma_{\text{dist}}(x)$ in the simplex of predictive distributions, with the guarantee of containing the reference distribution $p^*(\cdot|x)$ with probability $1-\alpha_{\text{dist}}^{\text{mis}}$. A final calibrated predictive distribution can be obtained via ensembling or via other combining mechanisms.}
    \label{fig:c5_overview}
\end{figure}

\subsection{Context and Motivation}

Modern \gls{ai} models, including \gls{nn} and \gls{llm}, have achieved remarkable success in decision-making and generative tasks. However, two significant challenges persist in their deployment: (\emph{i}) achieving \emph{reliability} in safety-critical applications \citep{guo2017calibration}, and (\emph{ii}) enabling \emph{efficient} deployment on edge devices with constrained resources \citep{singh2023edge}. 

Traditional Bayesian methods, which model epistemic uncertainty by treating model parameters as random variables, are a popular approach to improving reliability by enhancing \emph{calibration} \citep{mackay2003information, li2024arithmetic, khan2021bayesian}. A well-calibrated model is one whose confidence levels match the true accuracy levels, thus providing trustworthy information about the reliability of the model's output.  However, Bayesian methods have several drawbacks. For one, they typically require maintaining and running multiple models in order to carry out ensembling, which is computationally challenging for edge devices. Furthermore, they depend on the choice of prior distributions, whose misspecification may lead to suboptimal calibration.

Against this background, this work introduces a low-complexity, practical approach for calibrating edge models. Our approach distills calibration knowledge from complex cloud models, ensuring reliable performance without the computational overhead of Bayesian methods.

\subsection{Conformalized Distillation for Credal Inference}

As illustrated in Fig.~\ref{fig:c5_overview}, this work introduces a low-complexity methodology to enhance the calibration of a small-scale edge model by distilling calibration information from a more complex cloud model. In the setting under study, prior to deployment at the edge, a small-scale model is calibrated with the use of data collected from a large-scale model. 

The approach, termed \emph{\gls{cd-ci}},  builds on \emph{\gls{cp}}  \citep{angelopoulos2024theoretical, shafer2008tutorial} and \emph{\gls{ip}} \citep{walley1991statistical, beer2013imprecise}. In an \emph{offline calibration phase}, \gls{cd-ci} uses the predictive probabilities generated by a high-complexity cloud-based model to determine a threshold based on the typical divergence between the predictions of the cloud and edge models. 

At \emph{run time}, \gls{cd-ci} uses this threshold to construct \emph{credal sets} \citep{caprio2024Bayesian} -- ranges of predictive probabilities that are guaranteed, with a user-selected confidence level, to include the predictions of the cloud model. The credal sets are obtained through simple thresholding of a divergence measure in the simplex of predictive probabilities \citep{javanmardi2024conformalized}. Credal sets are finally converted into predictive distributions by using methods such as entropy maximization \citep{lukasiewicz2013credal} or intersection probabilities \citep{wang2024credal}. 
  
\subsection{Related Work}

\emph{\gls{ip}} offers a mathematical framework for describing random events that cannot be captured by standard probability theory \citep{dempster2008upper, walley1991statistical, gillies2012philosophical, manski2003partial, walley1982towards, cattaneo2012likelihood, cattaneo2017empirical}. \gls{ip} handles scenarios where multiple probabilities are assigned to a single event, such as: (\emph{i}) when the underlying probability distribution varies over time (e.g., the probability of rain tomorrow differs from yesterday) \citep{manski2003partial, walley1982towards}; (\emph{ii}) when estimating a fixed distribution requires considering an uncertainty set (e.g., the rain probability is estimated between $0.2$ and $0.3$) \citep{cattaneo2012likelihood}; and (\emph{iii}) when subjective beliefs, like Bayesian priors, are not uniquely defined (e.g., expecting rain $2-3$ days per week in London) \citep{walley1991statistical}. For an empirical interpretation of \gls{ip}, see \citep{cattaneo2017empirical}.  

\emph{\gls{ipml}} \citep{zaffalon2002naive, denoeux2000neural, chen2023imprecise, caprio2024Bayesian} applies \gls{ip} in machine learning, addressing issues such as imprecise labelling (e.g., crowdsourced labels \citep{chen2023imprecise}), domain generalization \citep{caprio2024Bayesian}, and Bayesian model misspecification \citep{caprio2024Bayesian}. While \gls{ipml} can enhance robustness and generalization \citep{chen2023imprecise, caprio2024Bayesian}, most approaches require constructing a \emph{credal set} -- a convex set of distributions containing the target distribution. This construction often relies on assumptions about the target distribution \citep{caprio2024Bayesian}, which may not hold in practice.

To address this, \emph{distribution-free \gls{ipml}} \citep{cella2021valid, cella2022validity, javanmardi2024conformalized, caprio2024conformalized} uses \emph{\gls{cp}} \citep{vovk2005algorithmic} to construct credal sets. Early methods \citep{cella2021valid, cella2022validity} focused on confidence intervals for labels using conformal $p$-values \citep{bates2023testing}. Recent work \citep{javanmardi2024conformalized, caprio2024conformalized} simplifies this approach by leveraging ground-truth label distributions for calibration data, justified using the concept of \emph{ambiguous} labels \citep{stutz2023conformal}. In the system proposed in this work, ground-truth label distributions are naturally provided by the cloud model.

\emph{\gls{ai} calibration} aims to adjust \gls{ai} models producing probabilistic outputs, avoiding overconfidence or underconfidence \citep{niculescu2005predicting, guo2017calibration, gupta2020calibration, mukhoti2020calibrating, bohdal2021meta, marx2024calibration} (e.g., mitigating LLM hallucinations \citep{achiam2023gpt, huang2023survey}). While Bayesian learning offers theoretical calibration \citep{simeone2022machine, zuk2012number}, it is limited in practice by model misspecification \citep{masegosa2020learning,zecchin2023robust} and computational complexity \citep{tierney1986accurate, jordan1999introduction, daxberger2021laplace}, particularly for edge devices \citep{katti2023bayesian}. Alternatives include (\emph{i}) {calibration-aware training} \citep{kumar2018trainable, lakshminarayanan2017simple, huang2025calibrating} and (\emph{ii}) {post-hoc calibration} \citep{platt1999probabilistic, vovk2012venn, marx2022modular}. Calibration-aware training modifies model training but increases complexity, especially with ensembling. Post-hoc calibration adjusts pre-trained models using a held-out dataset, though performance can degrade with limited calibration data \citep{shen2024thermometer}. \gls{cp}-based post-hoc methods \citep{vovk2012venn, marx2022modular, vovk2017nonparametric} address this by producing calibrated distributions, referred to as conformal calibration or Venn predictors. This work aims at producing calibrated credal sets, offering a more nuanced and comprehensive characterization of uncertainty compared to calibrated distributions.

Finally, this work aligns with \emph{knowledge distillation} \citep{hinton2015distilling}, where a larger model aids a smaller model. Unlike traditional approaches focused on accuracy \citep{zhu2023rethinking}, we enhance the calibration of the smaller model without compromising its accuracy.

\subsection{Main Contributions}

This chapter proposes a novel, low-complexity calibration method specifically designed for edge \gls{ai} deployment. The primary contributions are as follows:
\begin{itemize}
    \item \emph{Credal set construction via distilled calibration}: We introduce a post-processing approach, \gls{cd-ci}, that distills calibration knowledge from large-scale cloud models. By treating the predictive distributions from cloud models as a reference, we enable edge models to make calibrated predictions with statistical reliability guarantees.
    \item \emph{Robust predictive distribution}: When a single predictive distribution is required, \gls{cd-ci} extracts a distribution from the credal set by using the intersection probability approach \citep{wang2024credal}, achieving a more robust performance compared to low-complexity Bayesian methods. 
    \item \emph{Experimental validation}: We present experiments on visual and language modelling tasks, including the CIFAR-10 dataset \citep{krizhevsky2010cifar} and the SNLI dataset \citep{bowman2015large}, in which small-scale models are obtained via smaller architectures or quantized weights \citep{li2024evaluating, leviathan2023fast}.  We demonstrate significant improvements in calibration performance, as measured by the \emph{\gls{ece}} \citep{guo2017calibration}, over the original small-scale model, with negligible drops in accuracy. Comparisons are also provided with a low-complexity Bayesian learning method, the Laplace approximation \citep{daxberger2021laplace}. These results highlight the effectiveness of our approach in real-world edge deployments. 

\end{itemize}

\subsection{Organization}
The remainder of this chapter is organized as follows. Sec.~\ref{sec:c5_problem_definition} defines the problem, and the proposed methodology for distilling calibration via conformalized credal inference is presented in Sec.~\ref{sec:c5_our_scheme}. Sec.~\ref{sec:c5_hard_decision} describes how to extract a predictive distribution from a credal set. Finally, Sec.~\ref{sec:c5_experiments} illustrates the experimental setting and results and Sec.~\ref{sec:c5_conclusion} concludes the work.

\section{Problem Definition} \label{sec:c5_problem_definition}
\subsection{Distilling Calibration}
As illustrated in Fig.~\ref{fig:c5_overview}, we aim at calibrating a pre-trained small-scale classifier, $p^e(y|x)$, with $d$-dimensional input $x \in \mathbb{R}^d$ and label $y \in \mathcal{Y} = \{1, 2, \cdots, |\mathcal{Y}|\}$, by leveraging a pre-trained large-scale model, $p^*(y|x)$, and an unlabeled dataset $\mathcal{D}^\text{unl} = \{{x_{i}}\}^{|\mathcal{D}^\text{unl}|}_{i = 1}$. As an example, the large-scale model may be cloud-based, while the small-scale model may be intended for edge deployment.

Given a test input $x$, the calibration procedure \emph{post-processes} the output of the small-scale model, given by the distribution $p^e(\cdot|x) = \{p^e(y|x)\}_{y\in \mathcal{Y}}$, to produce a \emph{subset} of predictive probability distributions. This subset aims at capturing the uncertainty of the small-scale model about the predictive distribution of the large-scale model, $p^*(\cdot|x)$ for the given input $x$. This uncertainty arises due to the limited computational power of the small-scale model as compared to the large-scale model. We aim to derive a low-complexity calibration procedure in which the subset is defined by a simple thresholding mechanism.

Let $\mathcal{P}$ denote the simplex of $|\mathcal{Y}|$-dimensional probabilistic distributions.  At test time, as mentioned, the calibration procedure maps the small-scale model output probability $p^e(\cdot|x)$  into a subset $\Gamma_{\text{dist}}(x) \subseteq \mathcal{P}$. The mapping between predictive probabilities and a subset of $\mathcal{P}$ is designed during an offline \emph{distillation} phase in which the designer has access to the unlabeled dataset $\mathcal{D}^{\text{unl}}$ and to the large-scale model. The large-scale model is no longer accessible at test time. 

The design goal is to ensure that the set includes the reference distribution $p^*(\cdot|x)$ that would have been produced by the large-scale model with probability no smaller than a user-defined level $1-\alpha_{\text{dist}}^{\text{mis}}$, i.e.,
\begin{align} \label{eq:c5_coverage}
    \Pr \left[p^*(\cdot|x) \in  \Gamma_{\text{dist}}(x) \right] \geq 1 - \alpha_{\text{dist}}^{\text{mis}},
\end{align}
where $\alpha_{\text{dist}}^{\text{mis}} \in [0,1]$ is the desired \emph{miscoverage} rate. The probability in (\ref{eq:c5_coverage}) is evaluated with respect to the distribution of the unlabeled dataset $\mathcal{D}^{\text{unl}}$ used to design the post-processing mechanism, as well as over the test input $x$. 

The condition (\ref{eq:c5_coverage}) can be satisfied for any miscoverage rate $\alpha_{\text{dist}}^{\text{mis}}$ by setting $\Gamma_{\text{dist}}(x) = \mathcal{P}$, i.e., by producing the set of all possible distributions on the label set $\mathcal{Y}$ in response to any test input $x$. However, this output would be clearly uninformative. Therefore, to gauge the informativeness of the set predictor, we evaluate the normalized average size of the set, also known as \emph{inefficiency}, i.e.,
\begin{align} \label{eq:c5_inefficiency}
     \mathbb{E} \left[ \frac{|\Gamma_{\text{dist}}(x)|}{|\mathcal{P}|} \right].
\end{align}
The expectation in (\ref{eq:c5_inefficiency}) is taken with respect to the same distribution as in (\ref{eq:c5_coverage}). Furthermore, the size $|\Gamma_{\text{dist}}(x)|$ corresponds to the standard volume covered by set $\Gamma_{\text{dist}}(x)$ within the simplex $\mathcal{P}$.

\section{Conformalized Credal Inference} \label{sec:c5_our_scheme}
In this section, we first introduce credal sets, and then we propose a way to conformalize the credal sets so as to satisfy the coverage requirement (\ref{eq:c5_coverage}).

\subsection{Credal Sets} 
Given a test input $x$ and an output $p^e(\cdot|x)$ of the small-scale model, the set predictor $\Gamma_{\text{dist}}(x)$ is constructed by including all distributions $q = q(\cdot|x)$ in a neighbourhood of $p^e(\cdot|x)$. The radius defining the size of the neighbourhood is determined during the offline calibration phase by using the unlabeled data and the large-scale model.

To elaborate, consider the class of divergences between two distributions $q_1$ and $q_2$ in simplex $\mathcal{P}$, i.e.,
\begin{align} \label{eq:c5_f-diveregnce}
    \mathrm{D}_f (q_1 \| q_2) = \mathbb{E}_{z \sim q_2(z)} \left[f \left(\frac{q_1(z)}{q_2(z)}\right) \right],
\end{align}
where $f(\cdot)$ is a convex function satisfying the properties (\emph{i}) $f(1) =0$, and (\emph{ii}) $0 \cdot f(0/0) = 0$ \citep{Polyanskiy_Wu_2024, simeone_cqit}. The class of $f$-divergences encompasses a variety of divergence measures, including the \gls{kl} divergence and the Tsallis divergence.

Given a $f$-divergence $\mathrm{D}_f (\cdot \| \cdot)$, we obtain the credal set as 
\begin{align} \label{eq:c5_credal_set}
    \Gamma (x) = \left \{ q \in \mathcal{P} : \hspace{0.5em} \mathrm{D}_f (q \| p^e(\cdot|x)) \leq \tau_{\text{div}} \right \},
\end{align}
where $\tau_{\text{div}}$ is a threshold to be determined during the offline calibration phase.

\subsection{Distilling Calibration via Conformalized Credal Inference}
Conformalized credal inference determines the threshold $\tau_{\text{div}}$ in (\ref{eq:c5_credal_set}) to guarantee the coverage condition (\ref{eq:c5_coverage}). This is done offline by leveraging the unlabeled dataset $\mathcal{D}^{\text{unl}}$. To this end, we first construct the calibration dataset
\begin{align} \label{eq:c5_calibration_dataset}
    \mathcal{D}^{\text{cal}}_{\text{CD}} = \{{(x_{i}, p^*(\cdot|x_i))}\}^{|\mathcal{D}^\text{unl}|}_{i = 1},
\end{align}
in which the large-scale model is used to assign \emph{soft} label $p^*(\cdot|x_i)$ to the unlabeled input $x_i$. Then, we leverage the calibration dataset  $\mathcal{D}^{\text{cal}}_{\text{CD}}$ to construct the score set
\begin{align} \label{eq:c5_nc-score-set}
    \mathcal{V} = \{ V_i = \mathrm{D}_f(p^*(\cdot|x_i) \| p^e(\cdot|x_i))) \}_{i=1}^{|\mathcal{D}^{\text{cal}}_{\text{CD}}|}.
\end{align} 
The set $\mathcal{V}$ collects the divergence values $\mathrm{D}_f(p^*(\cdot|x_i) \| p^e(\cdot|x_i))$ between the reference and the predictive distributions in the calibration dataset $\mathcal{D}^{\text{cal}}_{\text{CD}}$. Thus, this set supports inferences about the distributions of the divergence $\mathrm{D}_f(p^*(\cdot|x_i) \| p^e(\cdot|x_i))$ as the input $x$ varies according to the underlying population distribution.

Following the standard split \gls{cp} methodology, we compute the $ \lceil (1 + |\mathcal{D}^{\text{cal}}_{\text{CD}}|)(1 - \alpha_{\text{dist}}^{\text{mis}}) \rceil$ smallest element in set $\mathcal{V}$, i.e., the $ \lceil (1 + |\mathcal{D}^{\text{cal}}_{\text{CD}}|)(1 - \alpha_{\text{dist}}^{\text{mis}}) \rceil/|\mathcal{D}^{\text{cal}}_{\text{CD}}|$-th empirical quantile of the divergences in the set $\mathcal{V}$. Finally, the radius threshold in (\ref{eq:c5_credal_set}) is evaluated as 
\begin{align} \label{eq:c5_threshold}
    \tau_{\text{div}} = \lceil (1 + |\mathcal{D}^{\text{cal}}_{\text{CD}}|)(1 - \alpha_{\text{dist}}^{\text{mis}}) \rceil \text{ smallest element of } \mathcal{V}.
\end{align}

The proposed method, referred to as \gls{cd-ci}, is summarized in Algorithm~\ref{alg:c5_CD_CI_offline} and Algorithm~\ref{alg:c5_CD_CI_run time}. Specifically, Algorithm~\ref{alg:c5_CD_CI_offline} defines the offline phase producing the threshold $\tau_{\text{div}}$, while Algorithm~\ref{alg:c5_CD_CI_run time} describes the run-time operation.

\begin{algorithm}[tb] 
\renewcommand{\algorithmicrequire}{\textbf{Input:}}
\renewcommand{\algorithmicensure}{\textbf{Output:}}
\caption{\gls{cd-ci} -- Offline Phase}
\begin{algorithmic}[1] 
  \REQUIRE Unlabeled dataset $\mathcal{D}^{\text{unl}}$, small-scale classifier $p^e(\cdot|x)$, large-scale classifier \\ $p^*(\cdot|x)$, divergence measure $\mathrm{D}_f (\cdot \| \cdot)$, target coverage level $1-\alpha_{\text{dist}}^{\text{mis}}$
  \ENSURE Threshold $\tau_{\text{div}}$
  \STATE Construct the calibration dataset $\mathcal{D}^{\text{cal}}_{\text{CD}}$ in (\ref{eq:c5_calibration_dataset}) by using the unlabeled dataset $\mathcal{D}^{\text{unl}}$ \\
  and the large-scale classifier $p^*(\cdot|x)$
  \STATE Evaluate the dataset $\mathcal{V}$ in (\ref{eq:c5_nc-score-set}) 
  \STATE Evaluate the radius threshold $\tau_{\text{div}}$ in (\ref{eq:c5_threshold}), for the given target coverage level $1-\alpha_{\text{dist}}^{\text{mis}}$ 
  \STATE {\textbf{return } Threshold $\tau_{\text{div}}$}
\end{algorithmic} \label{alg:c5_CD_CI_offline}
\end{algorithm}

\begin{algorithm}[tb] 
\renewcommand{\algorithmicrequire}{\textbf{Input:}}
\renewcommand{\algorithmicensure}{\textbf{Output:}}
\caption{\gls{cd-ci} -- Run-Time Operation}
\begin{algorithmic}[1] 
  \REQUIRE Test input $x$, small-scale classifier $p^e(\cdot|x)$, divergence measure $\mathrm{D}_f (\cdot \| \cdot)$, \\ simplex $\mathcal{P}$, target coverage level $1-\alpha_{\text{dist}}^{\text{mis}}$
  \ENSURE Prediction set $\Gamma_{\text{dist}}(x)$ or predictive distribution $q^*(\cdot|x)$
  \STATE Construct prediction set $\Gamma_{\text{dist}}(x)$ using (\ref{eq:c5_credal_set})
  \IF{predictive distribution is required}
    \STATE Derive bounds $ q_{L_y}$ and $ q_{U_y}$ as in (\ref{eq:c5_credal_bounds})
    \STATE Compute the predictive distribution $q^*(\cdot|x)$ as in (\ref{eq:c5_intersection_non_normalized})
  \ENDIF
  \STATE {\textbf{return } Prediction set $\Gamma_{\text{dist}}(x)$ or predictive distribution $q^*(\cdot|x)$}
\end{algorithmic} \label{alg:c5_CD_CI_run time}
\end{algorithm}

In the offline phase, \gls{cd-ci} only requires the evaluation of the threshold $\tau_{\text{div}}$ in (\ref{eq:c5_credal_set}). This operation requires ordering the nonconformity scores $V_i$ in the score set $\mathcal{V}$. This is a low-complexity operation with order $\mathcal{O}(|\mathcal{D}^{\text{cal}}_{\text{CD}}|\log(|\mathcal{D}^{\text{cal}}_{\text{CD}}|))$. In the run-time phase, as defined in Algorithm~\ref{alg:c5_CD_CI_run time}, \gls{cd-ci} evaluates the set (\ref{eq:c5_credal_set}), from which a predictive distribution can be evaluated as detailed in the next section.

\subsection{Theoretical Reliability Guarantees}

\gls{cd-ci} satisfies the following reliability guarantees.

\textbf{Theorem 5.1:} \textit{If the data samples in the unlabeled dataset $\mathcal{D}^{\text{unl}}$ and the test input $x$ are exchangeable, e.g., i.i.d.,  then \gls{cd-ci} (Algorithm~\ref{alg:c5_CD_CI_offline} and Algorithm~\ref{alg:c5_CD_CI_run time}) produces credal sets $\Gamma_{\text{dist}}(x)$ in (\ref{eq:c5_credal_set}) that satisfy the coverage condition
\begin{align} \label{eq:c5_theorem_1}
    \Pr \left[p^*(\cdot|x) \in  \Gamma_{\text{dist}}(x) \right] \geq 1 - \alpha_{\text{dist}}^{\text{mis}}.
\end{align}
}

The proof of this theorem follows directly from the marginal coverage guarantees of \gls{cp} \citep[Eq.~(1)]{angelopoulos2021gentle} (see also \citep[Thm.~4.1]{javanmardi2024conformalized}).

\section{Predictive Distributions from Credal Sets} \label{sec:c5_hard_decision}

The proposed \gls{cd-ci} method provides a low-complexity procedure to identify, for any given input $x$, a subset $\Gamma_{\text{dist}}(x)$ of predictive distributions $q(\cdot|x)$ that are likely to contain the golden-standard distribution $p^*(\cdot|x)$ of the large-scale model. The set can directly provide actionable information. For example, if the set is deemed to be too large, an edge device may conclude that the local model is insufficiently accurate for the given input $x$, refraining from making a decision. 

In practice, the edge device may wish to produce a single predictive probability $q^*(\cdot|x)$ for decision-making. In this case, it is desirable that the resulting predictive probability $q^*(\cdot|x)$, evaluated from the set $\Gamma_{\text{dist}}(x)$, be better calibrated than the initial distribution $p^e(\cdot|x)$ produced by the small-scale model. 

The distribution $q^*(\cdot|x)$ can be computed from the set $\Gamma_{\text{dist}}(x)$ in different ways, which are explored in the literature on imprecise probabilities \citep{stutz2023conformal}. For example, one can choose the distribution within the credal region $\Gamma_{\text{dist}}(x)$ that achieves the maximum Shannon entropy, thus finding the most conservative decisions \citep{caprio2024Bayesian}.

In this work, we adopt the \emph{intersection probability} \citep{wang2024credal} as the mechanism to obtain the final predictive distribution  $q^*(\cdot|x)$. This method is presented in Sec.~\ref{sec:c5_intersection_prob}. Then, for reference, Sec.~\ref{sec:c2_laplace} discusses a standard low-complexity Bayesian approach that can produce a re-calibrated predictive distribution, namely the Laplace approximation \citep{daxberger2021laplace, simeone2022machine}. Finally, we evaluate the calibration performance of predictive distributions in terms of \gls{ece} as reviewed in Sec.~\ref{sec:c2_empirical}.

\subsection{Intersection Probability} \label{sec:c5_intersection_prob}
Given a credal set $\Gamma_{\text{dist}}(x)$, the intersection probability method obtains a single predictive distribution $q^*(\cdot|x)$ as follows.

First, for every class $y \in \mathcal{Y}$, one obtains lower bound $q_{L_y}$ and upper bound $q_{U_y}$ on the probability $q(y|x)$ assigned by distributions in subset $\Gamma_{\text{dist}}(x)$ as
\begin{align} \label{eq:c5_credal_bounds}
    q_{L_y} = \min_{q(\cdot|x) \in \Gamma_{\text{dist}}(x)} q(y|x), \quad q_{U_y} = \max_{q(\cdot|x) \in \Gamma_{\text{dist}}(x)} q(y|x).
\end{align}
In practice, the set can be represented by a discrete subset of distributions $q(\cdot|x) \in \Gamma_{\text{dist}}(x)$. They can be obtained via grid search, with complexity linear in the grid size, or via importance sampling.

Finally, the predictive distribution $q^*(\cdot|x)$ is evaluated as
\begin{align} \label{eq:c5_intersection_non_normalized}
    q^*(y|x) = q_{L_y} + b \cdot (q_{U_y} - q_{L_y}),
\end{align}
where $b \in [ 0,1 ]$ is a constant chosen to guarantee that the function $q^*(\cdot|x)$ is a valid probabilistic distribution. Accordingly, the constant value $b$ is calculated as 
\begin{align} \label{eq:c5_intersection_normalized_constant}
    b = \frac{1 - \sum_{y=1}^{|\mathcal{Y}|}q_{L_y}}{\sum_{y=1}^{|\mathcal{Y}|} (q_{U_y} - q_{L_y})}.
\end{align}
The distribution in (\ref{eq:c5_intersection_non_normalized}) ensures that each class $y$ is treated equally. This is done by choosing the probability $q^*(y|x)$ to be within the interval $[q_{L_y}, q_{U_y}]$ at the same relative position determined by the fraction $b$ for all $y \in \mathcal{Y}$.

Sec.~\ref{sec:c5_experiments} will compare the performance of the predictive distribution (\ref{eq:c5_intersection_non_normalized}) with other methods used in the literature on \gls{ip}, namely max entropy \citep{lukasiewicz2013credal}. A hard decision $\hat{y}$ can be obtained from the predictive distribution $q^*(\cdot|x)$ as
\begin{align} \label{eq:c5_hard_decision}
    \hat{y}(x) = \arg \max_{y\in \mathcal{Y}} q^*(y|x).
\end{align}

\begin{figure} [tb] 
    \centering
    \centerline{\includegraphics[width=\textwidth]{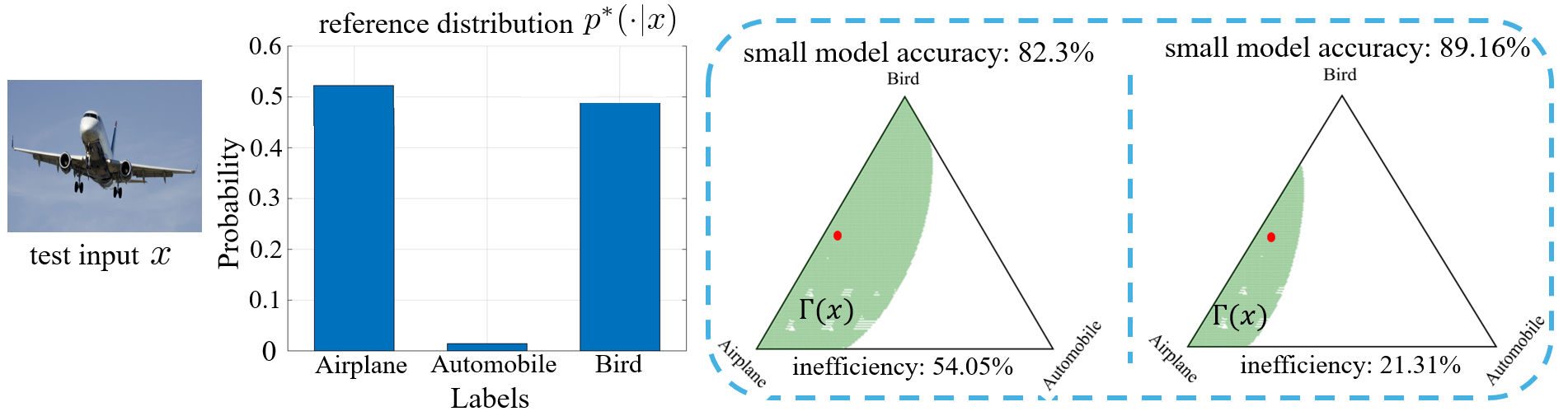}}
    \caption{Test input $x$, reference distribution $p^*(\cdot | x)$ from the large-scale model, and credal sets produced by \gls{cd-ci} for small-scale models with different accuracy on the CIFAR-10 dataset with classes $\{\text{airplane, automobile, bird}\}$ using the \gls{kl} divergence in (\ref{eq:c5_credal_set}) with target coverage rate $1-\alpha_{\text{dist}}^{\text{mis}} = 0.9$. Note that the large-model distribution  $p^*(\cdot | x)$ is marked as red point in the simplex.} 
    \label{fig:c5_cifar_visualization} 
\end{figure}

\section{Experimental Results} \label{sec:c5_experiments}
In this section, we report empirical results for visual and natural language tasks. 

\subsection{Performance Metrics}
For both tasks, we consider the following evaluation metrics: (\emph{i}) \emph{inefficiency}, which evaluates the average size of the credal set $\Gamma_{\text{dist}}(x)$ as in (\ref{eq:c5_inefficiency}); (\emph{ii}) \emph{coverage}, which is the percentage of samples for which the large-scale model predictive distribution $p^*(\cdot|x)$ falls inside the credal set $\Gamma_{\text{dist}}(x)$ as in (\ref{eq:c5_coverage}); (\emph{iii}) the \emph{\gls{ece}} (\ref{eq:c2_ece}); and (\emph{iv}) \emph{accuracy}, measured by the probability that the hard decision obtained as in (\ref{eq:c5_hard_decision}) is correct.

\subsection{Implementations}
Throughout, we use the $\alpha$-divergence to evaluate the conformalized credal set in (\ref{eq:c5_credal_set}). The $\alpha$-divergence is obtained from the general definition of the $f$-divergence in (\ref{eq:c5_f-diveregnce}) with $f(n) = (n^\alpha -1) / (\alpha(\alpha-1))$. Note that, for $\alpha = 1$, the $\alpha$-divergence reduces to the \gls{kl} divergence \citep{cichocki2010families}. Increasing the parameter $\alpha$ yields more constrained sets (\ref{eq:c5_credal_set}), for which the support of the distributions $q(\cdot|x)$ is increasingly forced not to exceed the support of the small-scale distribution $p^e(\cdot|x)$ \citep{minka2005divergence}.

All the experiments reported in this work are implemented via PyTorch \citep{paszke2019pytorch} and run over a GPU server with a single NVIDIA A100 card\footnotemark[3].
\footnotetext[3]{Code can be found at \url{https://github.com/kclip/Distilling-Calibration}.}

\begin{figure} [tb] 
    \centering
    \centerline{\includegraphics[scale=0.35]{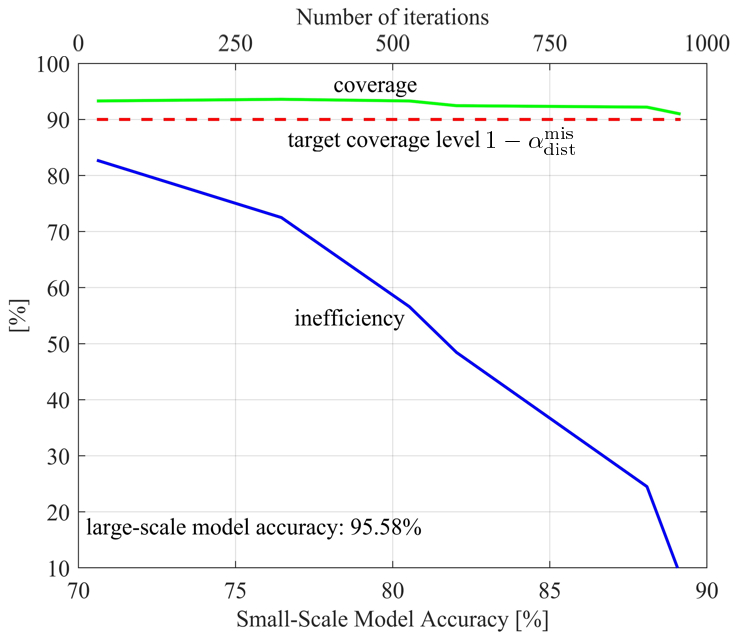}}
    \caption{Coverage and inefficiency versus the small-scale models accuracy on the CIFAR-10 dataset with classes $\{\text{airplane, automobile, bird}\}$ using the \gls{kl} divergence in (\ref{eq:c5_credal_set}) with target coverage rate $1-\alpha_{\text{dist}}^{\text{mis}} = 0.9$. The accuracy of the large-scale model, ResNet-18 network, is $95.58\%$, and the accuracy of the small-scale model, Mini-VGG-8, is controlled by training over different numbers of iterations. }
    \label{fig:c5_cifar_changing_accuracy} 
\end{figure}

\subsection{Image Classification}

\begin{figure} [tb] 
    \centering
    \centerline{\includegraphics[scale=0.18]{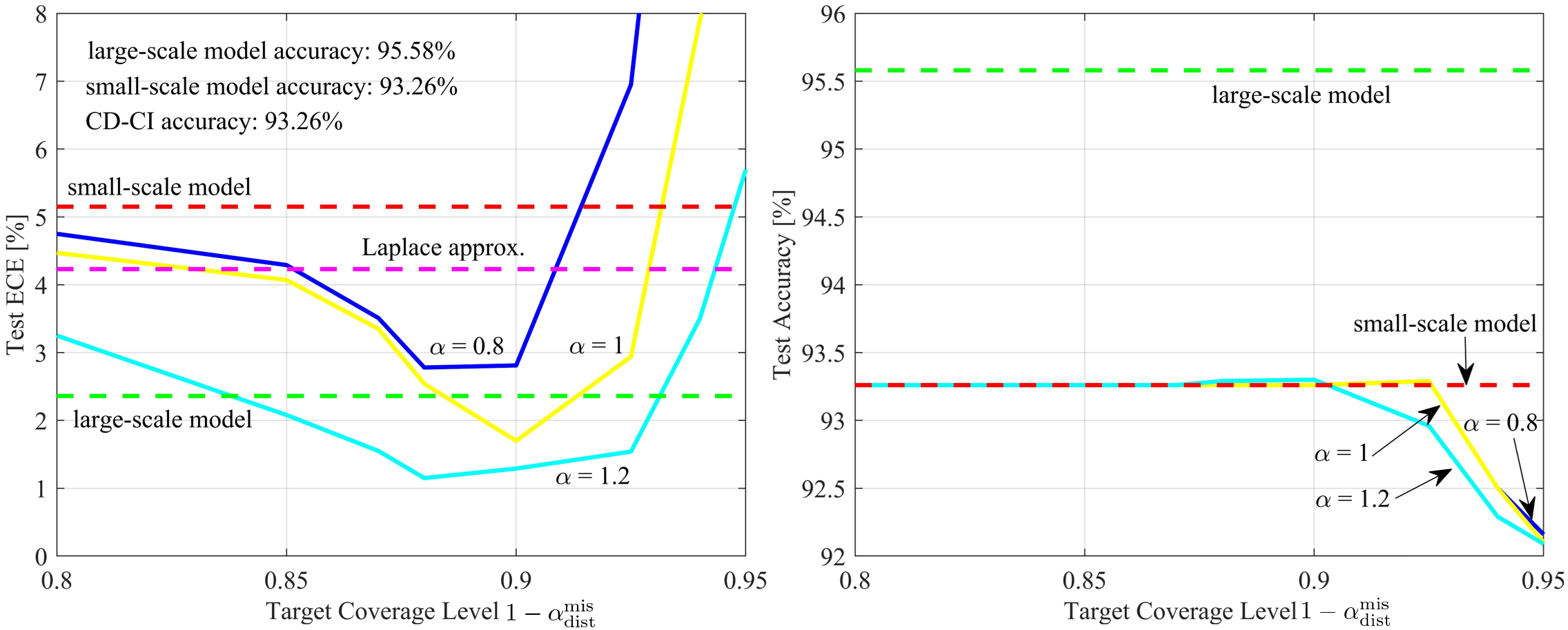}}
    \caption{\gls{ece} (left) and accuracy (right) versus target coverage rate $1-\alpha_{\text{dist}}^{\text{mis}}$ for different values of $\alpha$ for the $\alpha$-divergence used in (\ref{eq:c5_credal_set}) on the CIFAR-10 dataset with classes $\{\text{airplane, automobile, bird}\}$. The dashed lines in the left panel report the \gls{ece} performance of the large-scale model predictive distribution $p^*(\cdot|x)$, of the small-scale model $p^e(\cdot|x)$, and of the Laplace approximation method $q^{\text{La}}(\cdot|x)$ in (\ref{eq:c2_laplace_prob}). The dashed lines in the right panel report the accuracy performance of the large-scale model predictive distribution $p^*(\cdot|x)$, and of the small-scale model $p^e(\cdot|x)$. Note that there is no change to accuracy when applying the Laplace approximation in a post-processing way.}
    \label{fig:c5_cifar_ECE_versus_coverage} 
\end{figure}

For the image classification task, as in \citep{caprio2024conformalized}, we extract the first three classes from the CIFAR-10 \citep{krizhevsky2010cifar} dataset. Furthermore, we adopt the ResNet-18 model \citep{he2016deep} and the Mini-VGG-8 model \citep{simonyan2014very} as the large-scale and small-scale models, respectively. 

To start, in Fig.~\ref{fig:c5_cifar_visualization}, we visualize the impact of the small-scale model accuracy on the inefficiency of the credal set produced by \gls{cd-ci} via Algorithm~\ref{alg:c5_CD_CI_offline}. We control the accuracy of the small-scale model by training over different numbers of iterations, and we set $\alpha=1$, thus relying on the \gls{kl} divergence. The figure illustrates a test example $x$, the large-scale model's predictive distribution $p^*(\cdot|x)$, and credal sets produced by \gls{cd-ci} for small-scale models with different accuracy on the test dataset. As seen, a more accurate small-scale model yields smaller credal sets.

The inefficiency and coverage of \gls{cd-ci} as a function of the test accuracy of the small-scale model are shown in Fig.~\ref{fig:c5_cifar_changing_accuracy}. Note that the accuracy of the small-scale model ranges from $70\%$ to $90\%$, while the large-scale model obtains accuracy $95\%$. We set the target miscoverage rate as $\alpha_{\text{dist}}^{\text{mis}} = 0.1$. Improving the accuracy of the small-scale model is seen to be instrumental in enhancing the efficiency of the conformalized credal set. Furthermore, \gls{cd-ci} maintains a coverage rate close to the target level $1-\alpha_{\text{dist}}^{\text{mis}} = 0.9$, for all small-scale models, validating Theorem 5.1.

\begin{figure} [tb] 
    \centering
    \centerline{\includegraphics[scale=0.21]{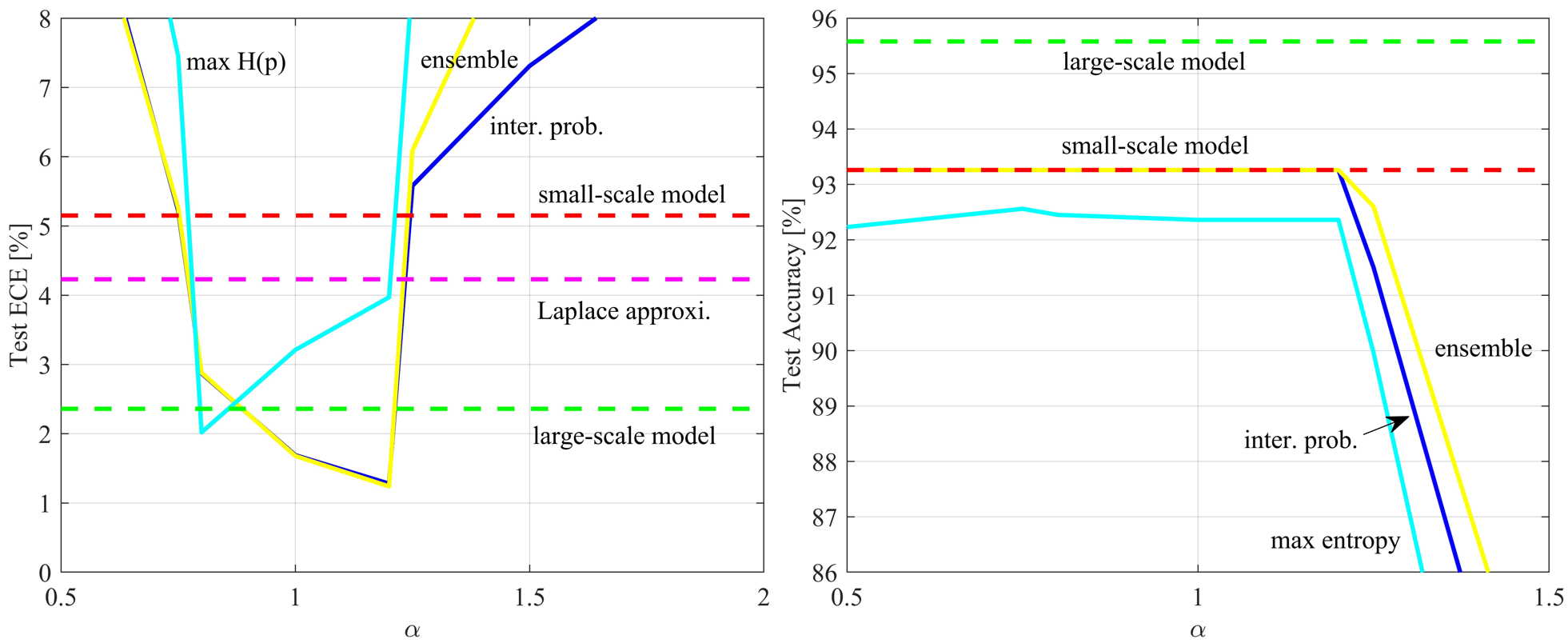}}
    \caption{\gls{ece} (left) and accuracy (right) versus the values of $\alpha$ for the $\alpha$-divergence used in (\ref{eq:c5_credal_set}) for different target coverage rate $1-\alpha_{\text{dist}}^{\text{mis}}$ on the CIFAR-10 dataset with classes $\{\text{airplane, automobile, bird}\}$. The blue line represents the performance of the predictive distribution $ q^*(y|x)$ (\ref{eq:c5_intersection_non_normalized}), in yellow the performance of the ensemble distribution $\mathbb{E}_{\Gamma_{\text{dist}}(x)}[q(\cdot|x)]$, and in cyan the performance of the maximum Shannon entropy distribution $\max_{q(\cdot|x) \in \Gamma_{\text{dist}}(x)}[H(q(\cdot|x)]$.}
    \label{fig:c5_cifar_ablation_fix_epsilon} 
\end{figure}

While Fig.~\ref{fig:c5_cifar_changing_accuracy} focuses on the performance of the credal set $\Gamma_{\text{dist}}(x)$, we now turn to analyzing the performance of the predictive distribution (\ref{eq:c5_intersection_non_normalized}) extracted from the subset $\Gamma_{\text{dist}}(x)$ as described in Sec.~\ref{sec:c5_intersection_prob}. To this end, Fig.~\ref{fig:c5_cifar_ECE_versus_coverage} plot the \gls{ece} and accuracy of the small-scale model versus the target coverage level $1-\alpha_{\text{dist}}^{\text{mis}}$ in the range from $0.8$ to $0.95$, respectively. In Fig.~\ref{fig:c5_cifar_ablation_fix_epsilon}, we also vary the $\alpha$ values used in evaluating the subset (\ref{eq:c5_credal_set}). These figures also report for reference the \gls{ece} and accuracy performance of the large-scale model $p^*(\cdot|x)$, and of the Laplace approximation method $q^{\text{La}}(\cdot|x)$ in (\ref{eq:c2_laplace_prob}).

The key observation from Fig.~\ref{fig:c5_cifar_ECE_versus_coverage} and Fig.~\ref{fig:c5_cifar_ablation_fix_epsilon} is that, through a suitable choice of the hyperparameters $\alpha_{\text{dist}}^{\text{mis}}$ and $\alpha$, \gls{cd-ci} can improve the \gls{ece} of the original, uncalibrated, small-scale model, as well as of the Laplace approximation, without any accuracy loss. Moreover, the performance is robust to the choice of the hyperparameter $\alpha_{\text{dist}}^{\text{mis}}$. In fact, with all values within the range $0.85 \leq 1-\alpha_{\text{dist}}^{\text{mis}} \leq 0.92$, \gls{cd-ci} outperforms both the original small-scale model and the Laplace approximation in terms of \gls{ece}, achieving \gls{ece} reductions of approximately $4\%$ and $3\%$, respectively. For smaller values of $1-\alpha_{\text{dist}}^{\text{mis}}$, \gls{cd-ci} tends to produce credal set $\Gamma_{\text{dist}}(x)$ concentrated around the small-scale model predictive distribution. Conversely, for larger values of $1-\alpha_{\text{dist}}^{\text{mis}}$, the credal set $\Gamma_{\text{dist}}(x)$ becomes larger, eventually leading to a deterioration in the \gls{ece} and of the accuracy of \gls{cd-ci}.

Fig.~\ref{fig:c5_cifar_ECE_versus_coverage} also illustrates the advantages of constraining the credal set through the choice of a value of $\alpha$ larger than $\alpha=1$, thus moving beyond the \gls{kl} divergence. To investigate the impact of the divergence parameter $\alpha$, the \gls{ece} and the accuracy are shown in the left and right panels of Fig.~\ref{fig:c5_cifar_ablation_fix_epsilon}, respectively, as a function of $\alpha$ for $1-\alpha_{\text{dist}}^{\text{mis}}=0.9$. In this figure, we evaluate and compare three approaches for deriving a single predictive distribution, namely (\emph{i}) the intersection probability (\ref{eq:c5_intersection_non_normalized}); (\emph{ii}) the ensemble distribution $\mathbb{E}_{\Gamma_{\text{dist}}(x)}[q(\cdot|x)]$, where the average is over a uniform distribution in the credal set $\Gamma_{\text{dist}}(x)$; and (\emph{iii}) the maximum Shannon entropy distribution $\max_{q(\cdot|x) \in \Gamma_{\text{dist}}(x)}[H(q(\cdot|x)]$ \citep{caprio2024conformalized}.

The results confirm that the intersection probability approach in (\ref{eq:c5_intersection_non_normalized}) consistently achieves the best performance in terms of \gls{ece}, outperforming both the ensemble and maximum Shannon entropy methods. Moreover, with the intersection probability method, the best results are obtained with $\alpha$ larger than $1$, although excessively large values of $\alpha$ cause a performance degradation.

\begin{figure} [tb] 
    \centering
    \centerline{\includegraphics[scale=0.23]{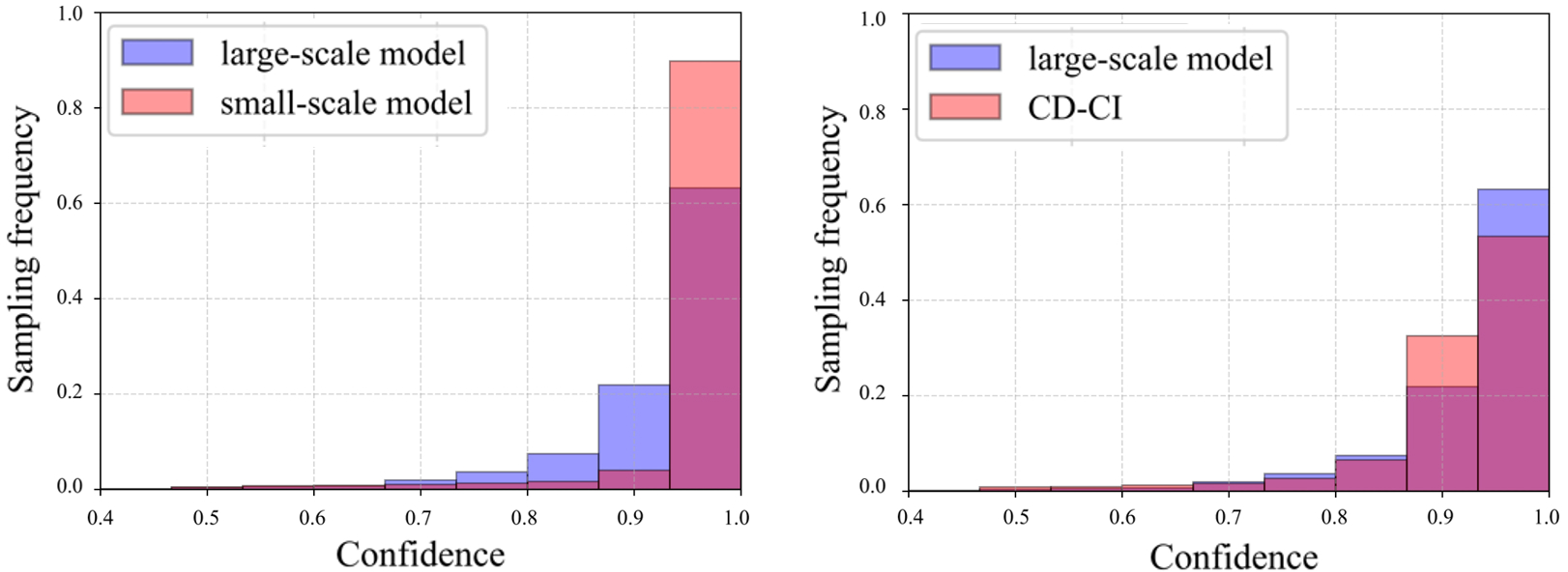}}
    \caption{Confidence histograms for the large-scale model predictive distribution $p^*(y|x)$, for the small-scale model $p^e(y|x)$, and for \gls{cd-ci} $q^*(y|x)$ evaluated on SNLI dataset, using $\alpha = 1$ with target coverage rate $1-\alpha_{\text{dist}}^{\text{mis}} = 0.9$.}
    \label{fig:c5_snli_visualization} 
\end{figure}

\begin{figure} [H] 
    \centering
    \centerline{\includegraphics[scale=0.27]{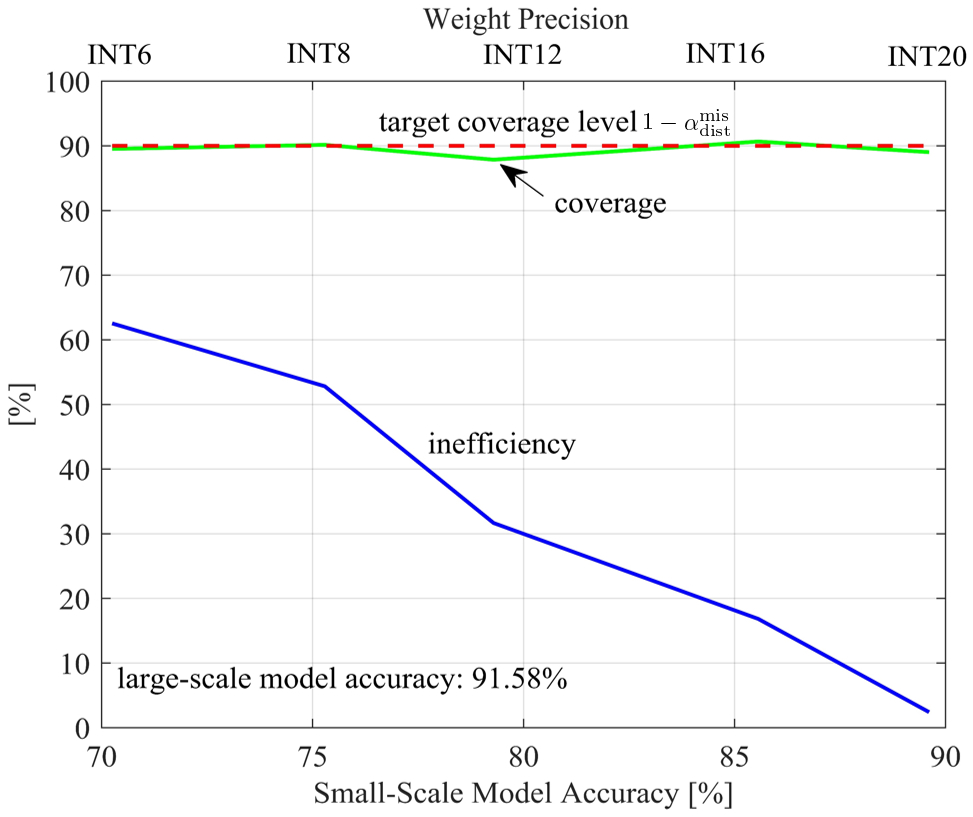}}
    \caption{Coverage and inefficiency versus the small-scale models accuracy on the SNLI dataset using the \gls{kl} divergence in (\ref{eq:c5_credal_set}) with target coverage rate $1-\alpha_{\text{dist}}^{\text{mis}} = 0.9$. The accuracy of the large-scale model is $91.58\%$. Note that INT$X$ denotes the block-floating point format, in which each element of the block is expressed via INT$X$ precision (see e.g., \citep{li2024evaluating, drumond2018training}).}
    \label{fig:c5_snli_changing_accuracy} 
\end{figure}
\subsection{Natural Language Classification}
For the natural language task, we adopt the SNLI dataset \citep{bowman2015large}, an English language sentence pairs classification dataset with three labels: entailment, contradiction, and neutral. We use the NLI-deberta-v3-large adopted temperature scaling \citep{shen2024thermometer} and original NLI-deberta-v3-small \citep{he2020deberta} as the large-scale and small-scale models, respectively. We further vary the quality of the small-scale model via uniform quantization.

Firstly, to visualize the calibration benefits of \gls{cd-ci}, Fig.~\ref{fig:c5_snli_visualization} plots the confidence histograms evaluated over test data $(x,y)$ for the large-scale model predictive distribution $p^*(y|x)$, the small-scale model $p^e(y|x)$, and \gls{cd-ci} $q^*(y|x)$ on the SNLI dataset, using $\alpha = 1$ with target coverage rate $1-\alpha_{\text{dist}}^{\text{mis}} = 0.9$. As illustrated, while the small-scale model is overconfident, \gls{cd-ci} successfully recalibrate the model, ensuring decisions that are calibrated to a level similar to the reference large-scale model.

Then, with the same settings as in Fig.~\ref{fig:c5_cifar_changing_accuracy}, we further investigate the relationship between the accuracy of the small-scale model and the performance of the \gls{cd-ci} in Fig.~\ref{fig:c5_snli_changing_accuracy}. To achieve this, we control the accuracy of the small-scale model through uniform quantization with varying weight precision \citep{li2024evaluating, leviathan2023fast}, a strategy that facilitates the efficient edge deployment of Transformer network architectures. The results in Fig.~\ref{fig:c5_snli_changing_accuracy} reaffirm the conclusions drawn in Fig.~\ref{fig:c5_cifar_changing_accuracy}.

\begin{figure} [tb] 
    \centering
    \centerline{\includegraphics[scale=0.34]{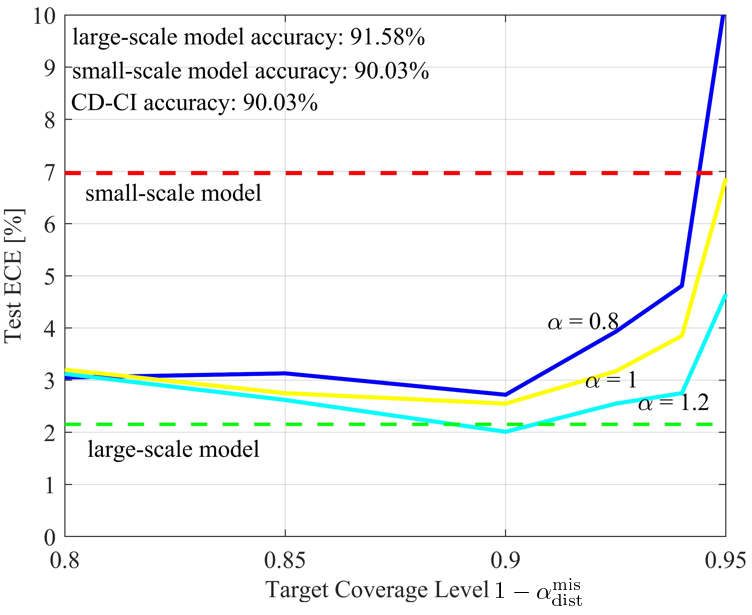}}
    \caption{\gls{ece} versus target coverage rate $1-\alpha_{\text{dist}}^{\text{mis}}$ for different values of the $\alpha$ for $\alpha$-divergence used in (\ref{eq:c5_credal_set}) on the SNLI dataset. The dashed lines report the \gls{ece} performance of the large-scale model predictive distribution $p^*(\cdot|x)$, of the small-scale model $p^e(\cdot|x)$, and of the Laplace approximation method $q^{\text{La}}(\cdot|x)$ in (\ref{eq:c2_laplace_prob}).}
    \label{fig:c5_snli_ECE_versus_coverage} 
\end{figure}

\begin{figure} [htb] 
    \centering
    \centerline{\includegraphics[scale=0.24]{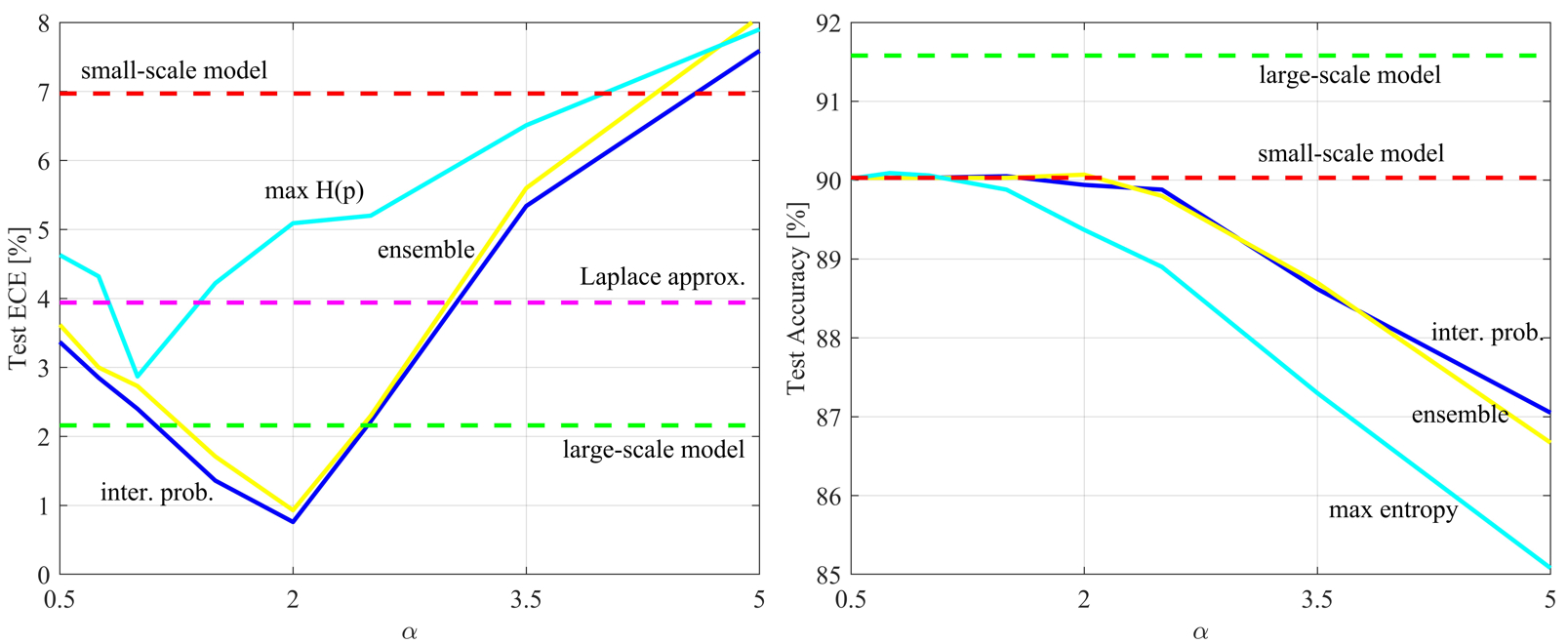}}
    \caption{\gls{ece} (left) and accuracy (right) versus the values of $\alpha$ for the $\alpha$-divergence used in (\ref{eq:c5_credal_set}) for different target coverage rate $1-\alpha_{\text{dist}}^{\text{mis}}$ on the SNLI dataset. The solid lines in the left panel represent the \gls{ece} performance of the predictive distribution $ q^*(y|x)$ (\ref{eq:c5_intersection_non_normalized}), of the ensemble distribution $\mathbb{E}_{\Gamma_{\text{dist}}(x)}[q(\cdot|x)]$, and of the maximum Shannon entropy distribution $\max_{q(\cdot|x) \in \Gamma_{\text{dist}}(x)}[H(q(\cdot|x)]$. The solid lines in the right panel represent the accuracy of the predictive distribution $ q^*(y|x)$ (\ref{eq:c5_intersection_non_normalized}), of the ensemble distribution $\mathbb{E}_{\Gamma_{\text{dist}}(x)}[q(\cdot|x)]$, and of the maximum Shannon entropy distribution $\max_{q(\cdot|x) \in \Gamma_{\text{dist}}(x)}[H(q(\cdot|x)]$.}
    \label{fig:c5_snli_ablation_fix_epsilon} 
\end{figure}


Fig.~\ref{fig:c5_snli_ECE_versus_coverage} shows the \gls{ece} of \gls{cd-ci} as a function of the target coverage level $1-\alpha_{\text{dist}}^{\text{mis}}$, ranging from $80\%$ to $95\%$, for different values of $\alpha$. These results confirm the general conclusions from Fig.~\ref{fig:c5_cifar_ECE_versus_coverage}. In fact, in this experiment, the benefits of \gls{cd-ci} are seen to be even more robust to the choice of hyperparameters $1-\alpha_{\text{dist}}^{\text{mis}}$ and $\alpha$ in terms of the \gls{ece}. Overall, \gls{cd-ci} achieves an improvement of approximately $5\%$ and $2\%$ in \gls{ece} compared to the original small-scale model and the Laplace approximation.

With the same settings as in Fig.~\ref{fig:c5_cifar_ablation_fix_epsilon}, we evaluate the \gls{ece} and accuracy against the divergence hyperparameter $\alpha$ in Fig.~\ref{fig:c5_snli_ablation_fix_epsilon}. This figure confirms again that (\emph{i}) the intersection approach (\ref{eq:c5_intersection_non_normalized}) achieves superior performance in terms of \gls{ece} and accuracy, and (\emph{ii}) relatively large values of $\alpha$ improve performance, whereas excessively large values of $\alpha$ result in performance degradation.

\section{Conclusion} \label{sec:c5_conclusion}
In this work, we have proposed a low-complexity methodology to calibrate a small-scale edge model prior to deployment by leveraging data generated by a large-scale cloud model. The method, called \gls{cd-ci}, ensures that the edge model can produce, at runtime, a set of predictive distributions guaranteed to include the large-scale model's predictive distribution with a pre-specified probability. Unlike standard Bayesian learning techniques, the ensemble of predictive distributions is obtained through a simple thresholding operation applied directly to the small model's output.  

Future research directions may include evaluating the performance of conformalized credal inference under covariate shift  \citep{tibshirani2019conformal}, integrating the approach with prior-data fitted networks \citep{muller2021transformers}, exploring online calibration strategies that enable interactive communication between edge and cloud models \citep{leviathan2023fast}, and studying the interplay between credal sets and imprecise highest density regions \citep{caprio2024conformalized}.
\chapter{Reliable Inference in Edge-Cloud Model Cascades via Conformal Alignment} \label{chapter:6}

\ifpdf
    \graphicspath{{Chapter6/Chapter6/Figs/}{Chapter6/Chapter6/Figs/PDF/}{Chapter6/Chapter6/Figs/}}
\else
    \graphicspath{{Chapter6/Chapter6/Figs/}{Chapter6/Chapter6/Figs/}}
\fi

\section{Overview}

The offline calibration distillation method developed in Chapter~\ref{chapter:5} enables efficient edge inference with formal reliability guarantees, but the calibrated edge model cannot be equally reliable for all inputs, i.e., complex inputs that exceed the capacity of the small-scale model may remain poorly calibrated regardless of the distillation procedure. That said, edge intelligence enables low-latency inference via compact on-device distilled models, but assuring reliability remains challenging.

To address this challenge, this chapter studies edge-cloud cascades that allow the edge and cloud models to work collaboratively at inference time and must preserve conditional coverage: whenever the edge returns a prediction set, it should contain the true label with a user-specified probability, as if produced by the cloud model. We formalize conditional coverage with respect to the cloud predictive distribution, and introduce a \gls{cab} cascading mechanism that certifies this property with user control over the risk level. Our method casts escalation from edge to cloud models as a \gls{mht} problem, tailoring \gls{ca} to select which inputs can be safely handled at the edge. The proposed \gls{cab} model cascading method yields statistical guarantees on the average fraction of edge decisions that satisfy cloud-level conditional coverage. The procedure applies to arbitrary edge prediction sets, including variants of \gls{cp}, and exposes a tunable trade-off among coverage, deferral rate, and set size. Experiments on CIFAR-100 image classification and the TeleQnA \gls{qa} benchmark show that the proposed \gls{cab} cascade maintains the target conditional coverage for edge predictions while substantially reducing offloading to the cloud and incurring modest increases in prediction-set size.

\begin{figure}
    \centering
    \centerline{\includegraphics[scale=0.26]{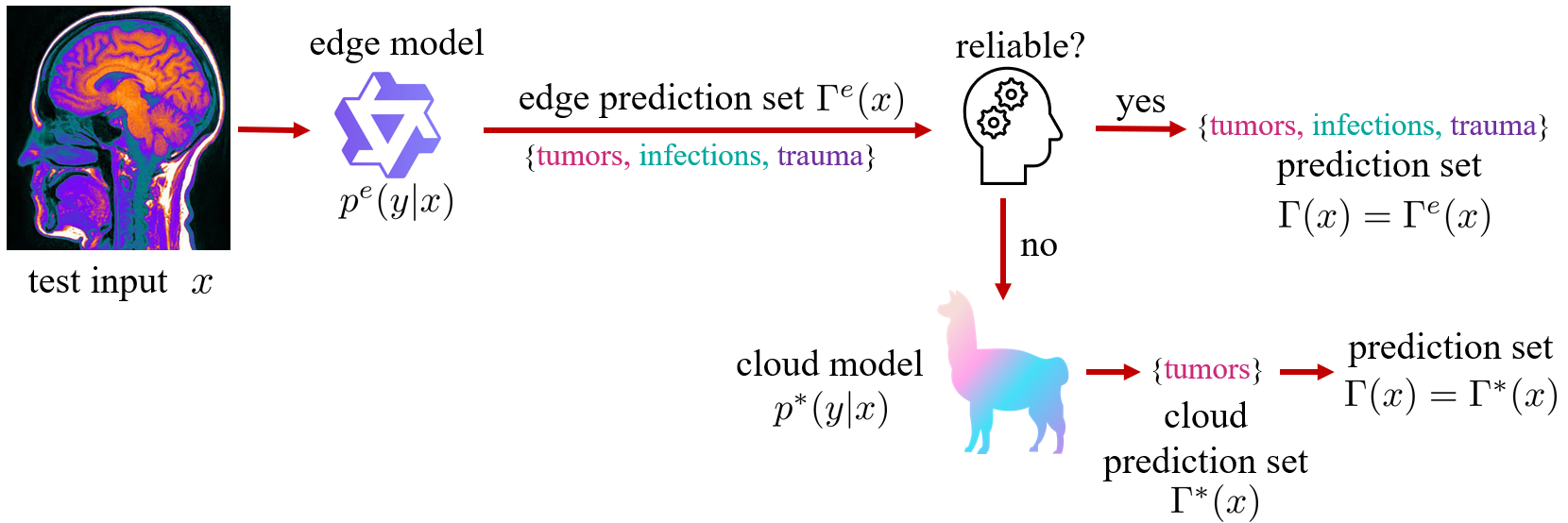}}
    \caption{In the edge-cloud cascade model under study, the goal is to produce a prediction set that is as reliable as the one produced by the cloud model, while leveraging the edge model for as many inputs as possible.}
    \label{fig:c6_overview_1}
\end{figure}

\section{Introduction}

\subsection{Context and Motivation}
Edge computing enables on-device inference with reduced latency and limited bandwidth usage, but replacing a powerful cloud model with a compact edge model raises concerns about reliability \citep{kaur2022trustworthy}. Beyond average accuracy, many real-world applications require coverage guarantees on prediction sets, i.e., sets that contain the ground-truth label with high probability. For instance, in Fig.~\ref{fig:c6_overview_1}, prediction sets are employed in medical imaging to identify the most likely diagnosis, which may require follow-up interventions by a doctor. Generalizing this example, by explicitly quantifying predictive uncertainty, set-valued predictions can strengthen reliability across diverse domains in healthcare \citep{sreenivasan2025conformal, shashikumar2021artificial}, such as radiology triage, diagnostic support, surgical robotics, and personalized dosing, and in engineering \citep{cohen2023calibrating, lekeufack2024conformal}, including autonomous driving, predictive maintenance, power-grid state estimation, and telecommunications. 

A particularly strong and practically relevant notion of reliability for set predictors is that of conditional coverage, which requires that the probability of the ground-truth label lying within the prediction set exceed a user-defined confidence level for any given input. Ensuring conditional coverage at the edge, however, is challenging. Simple knowledge distillation typically fails to transfer calibrated uncertainty from the cloud to the edge model \citep{huang2025distilling}; heuristic confidence thresholds used for deferral or selective prediction lack formal statistical guarantees \citep{chen2023adaptation, kadavath2022language}; and standard \gls{cp} methods \citep{minka2005divergence,shafer2008tutorial} provide only marginal coverage. Marginal coverage only guarantees reliability on average across the population of inputs, rather than conditionally for each input. As such, marginal coverage does not offer any performance guarantee on any given input. Addressing this gap calls for new cascading mechanisms capable of preserving cloud-level conditional coverage properties when inference is performed locally at the edge.
\begin{figure} [tb]
    \centering
    \centerline{\includegraphics[width=\textwidth]{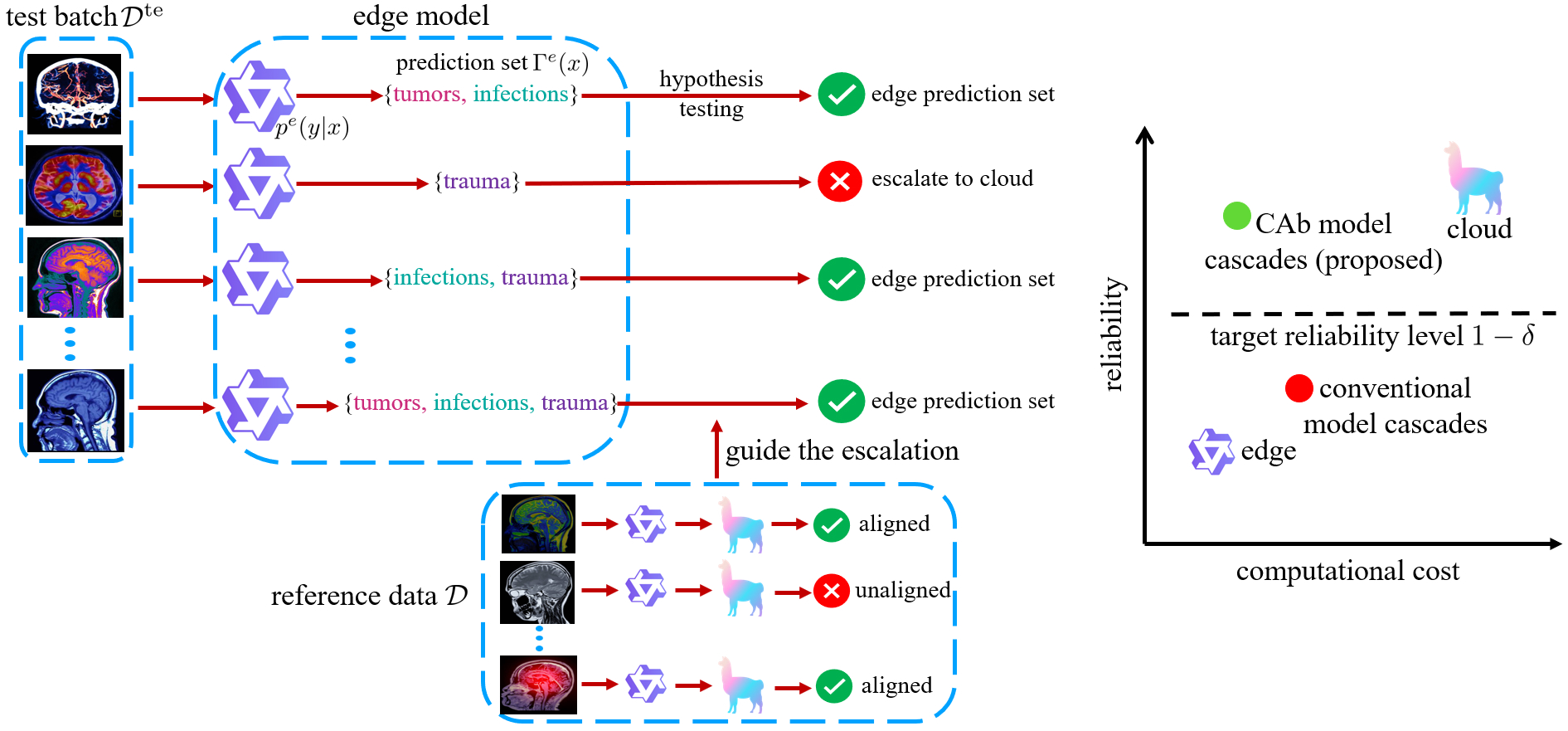}}
    \caption{Given a batch of test input $\mathcal{D}^{\text{te}}$, the small-scale edge model generates prediction sets that may deviate from the prediction sets that would have been produced by a large-scale cloud model, failing to meet a target reliability requirement. The proposed method, \gls{cab} model cascading, casts the edge-cloud escalation as a \gls{mht} problem, determining when to trust the edge prediction set based on reference data $\mathcal{D}$. \gls{cab} controls the fraction of edge-generated prediction sets that satisfy conditional coverage conditions, while minimizing the deferral rate to the cloud model.}
    \label{fig:c6_overview_2}
\end{figure}

In this context, the goal of this work is to ensure that, whenever the edge model outputs a prediction set, the probability that this set contains the ground-truth label meets a user-specified confidence level with respect to the cloud model’s predictive distribution. In essence, edge decisions are required to inherit the reliability guarantees of the cloud, ensuring that the statistical coverage achieved at the edge mirrors that of the cloud model, while allowing users to control the desired level of confidence.

\subsection{Related Work}

\textbf{Conformal prediction and coverage.} \gls{cp} provides finite-sample, distribution-free guarantees for marginal coverage under the assumption of exchangeability \citep{angelopoulos2021gentle}. However, achieving exact conditional coverage is, in general, infeasible without imposing strong distributional assumptions \citep{lei2014distribution, foygel2021limits}. To mitigate conditional under-coverage, several methodological variants of \gls{cp} have been proposed, including group-conditional \gls{cp} \citep{bostrom2021mondrian, vovk2012conditional}, and \gls{lcp} \citep{tibshirani2019conformal, barber2023conformal, qkae103}.

\textbf{Selection with guarantees and alignment.} \gls{ca} identifies outputs that satisfy a desired alignment criterion, e.g., human preference agreement, while providing finite-sample, distribution-free guarantees. This is done by learning an alignment score predictor and calibrating the selection threshold \citep{gui2024conformal, jung2024trust}. Importantly, to the best of our knowledge, this work is the first to design the selection criterion based on the conditional coverage probability, thereby ensuring dual reliability in terms of both statistical coverage and alignment consistency.

\textbf{Cascades and selective deferral.} Model cascades route easy inputs to a lightweight model, while deferring difficult inputs to a powerful model, with the aim of reducing cost and latency \citep{marquez2018deep}. Conventional model cascades rely on a fixed heuristic confidence measure, such as predictive entropy or maximum probability, to decide when to defer \citep{fithian2014optimal, rabanser2025gatekeeper}. These heuristics lack formal reliability guarantees, generally leading to unreliable and unnecessary deferral under distribution shift. Recent advances integrate \gls{cp} into model cascades design. For instance, reference \citep{yadkori2024mitigating} leverages calibration data to tune the selection threshold, thereby ensuring a marginal guarantee on the error rate.

\subsection{Main Contributions}

As illustrated in Fig.~\ref{fig:c6_overview_2}, in this work, we introduce a novel routing methodology for edge-cloud deferral systems that provides set predictions with statistical guarantees in terms of conditional coverage. The proposed approach ensures that edge predictions retain the same probabilistic reliability, in terms of conditional coverage,  as those produced by the cloud, while minimizing unnecessary offloading. The main contributions are summarized as follows:
\begin{enumerate}
    \item \textbf{Cloud-referenced conditional coverage for edge decisions.} We formalize a conditional coverage requirement that evaluates the reliability of edge predictions with respect to the cloud model’s predictive distribution. Specifically, we require that, whenever the edge model produces a prediction set, the probability that it contains the ground-truth label meets a user-specified coverage level relative to the cloud reference.
    
    \item \textbf{\gls{cab} cascading with statistical reliability guarantees.} We cast the edge-cloud routing problem as a \gls{mht} task and develop a \gls{cab} cascade that rigorously controls the \gls{fdr} of violations of the desired conditional coverage among edge-handled inputs. The proposed \gls{cab} mechanism applies to arbitrary edge prediction sets, including those derived from \gls{cp} or other calibration procedures, without requiring any modification of their construction.
    
    \item \textbf{Empirical validation on vision and language tasks.} Experiments on the CIFAR-100 image classification and TeleQnA multiple-choice \gls{qa} benchmarks demonstrate that the proposed \gls{cab} cascade maintains the desired conditional coverage for edge predictions, while substantially reducing cloud offloading and incurring only modest increases in prediction-set size. These results highlight explicit trade-offs among conditional coverage, deferral rate (cloud offloading), and prediction-set inefficiency, confirming the practicality and robustness of the proposed framework.
\end{enumerate}

\subsection{Organization}
The remainder of this work is organized as follows. Sec.~\ref{sec:c6_problem_definition} defines the problem formulation, and the \gls{sota} benchmarks are presented in Sec.~\ref{sec:c6_baselines}. Sec.~\ref{sec:c6_CA} formulates model cascading via \gls{mht}, and proposes the \gls{cab} model cascading mechanism. Finally, Sec.~\ref{sec:c6_results} illustrates the experimental setting and results, and Sec.~\ref{sec:c6_conclusion} concludes the work.

\section{Problem Definition} \label{sec:c6_problem_definition}

\subsection{Setting} \label{subsec:c6_setting}
In the edge-cloud system shown in Fig.~\ref{fig:c6_overview_1}, the cloud implements a reference predictive model $p^*(y|x)$, while the edge has access to a low-quality model $p^e(y|x)$, where $x \in \mathcal{X}$ is an input and $y \in \mathcal{Y}$ is a discrete output. The edge-cloud system is tasked to implement a predictive mapping from any input $x$ to a subset $\Gamma(x)$ of the label space $\mathcal{Y}$. Depending on the input $x$, the prediction set $\Gamma(x)$ may be produced at the edge, based on the edge model $p^e(y|x)$, or at the cloud, using the reference cloud model $p^*(y|x)$. The goal is to ensure that the prediction set $\Gamma(x)$ contains the ground-truth label $y$ with probability no smaller than a predetermined coverage level $1-\alpha_{\text{label}}^{\text{mis}} \in  [0,1]$, while using the edge model for the largest possible fraction of inputs.

Formally, for a given input $x$, we wish to ensure the conditional coverage guarantee
\begin{align} \label{eq:c6_ideal_goal}
    \Pr[y \in \Gamma(x)|x] \geq 1-\alpha_{\text{label}}^{\text{mis}}
\end{align}
for some user-defined miscoverage rate $0 \leq \alpha_{\text{label}}^{\text{mis}} \leq 1$, where $\Pr[\cdot|x]$ represents the conditional distribution of label $y$ given input $x$. We take the distribution $p^*(y|x)$ produced by the cloud model as the reference to evaluate the probability (\ref{eq:c6_ideal_goal}). Specifically, we evaluate the conditional probability in (\ref{eq:c6_ideal_goal}) using the cloud distribution $p^*(y|x)$ as the distribution of the label $y$ given input $x$ as
\begin{align} \label{eq:c6_idea_goal_reformula}
    \Pr[y \in \Gamma(x)|x] = p^*(\Gamma(x)|x) = \sum_{y \in \Gamma(x)} p^*(y|x).
\end{align}

The definition (\ref{eq:c6_idea_goal_reformula}) of conditional coverage is tailored to the given edge-cloud setting in which the cloud model is considered to be reliable but resource-intensive, calling for a targeted use of edge computing where possible.
In particular, in analogy with the notion of self-consistency \citep{certifiedllm}, the requirement (\ref{eq:c6_ideal_goal}) with (\ref{eq:c6_idea_goal_reformula}) can be viewed as a form of cloud-consistency for the decisions made across the edge-cloud system.
Incorporating also the ground-truth distribution $p^*(x)$ over input $x$, the reference data distribution is denoted as
\begin{align} \label{eq:c6_ground-truth_dist}
    p^*(x,y)=p^*(x)p^*(y|x).
\end{align}

The $(1-\alpha_{\text{label}}^{\text{mis}})$-conditional coverage guarantee (\ref{eq:c6_ideal_goal}) is achievable by deferring the input $x$ to the cloud. In fact, using the cloud model $p^*(y|x)$, the $(1-\alpha_{\text{label}}^{\text{mis}})$-\gls{hms}
\begin{align} \label{eq:c6_cloud_prediction_set}
    \Gamma^*(x) = \arg \min_{\Gamma^*(x) \subseteq \mathcal{Y}} |\Gamma^*(x)| \text{  s.t. } p^*(\Gamma^*(x)|x) \geq 1-\alpha_{\text{label}}^{\text{mis}}
\end{align}
satisfies the condition (\ref{eq:c6_ideal_goal}). In fact, by definition, the cloud-generated \gls{hms} (\ref{eq:c6_cloud_prediction_set}) is the smallest set $\Gamma(x) \subseteq \mathcal{Y}$ that satisfies the requirement (\ref{eq:c6_ideal_goal}). The main challenge addressed in this work is how to approximately attain the conditional coverage condition (\ref{eq:c6_ideal_goal}), while processing as many test inputs $x$ as possible at the edge.

\subsection{Design Criteria} \label{subsec:c6_design_criterion}
As explained, our goal is to approximate condition (\ref{eq:c6_ideal_goal}) while allowing for some decisions to be produced at the edge. In the considered edge-cloud system, the prediction set $\Gamma(x)$ is thus given by 
\begin{align} \label{eq:c6_original_prediction}
       \Gamma(x) =
        \begin{cases}
       \Gamma^*(x), & \text{if input $x$ is deferred to the cloud }  \\
       \Gamma^e(x) , & \text{if input $x$ is processed at the edge},
        \end{cases}
\end{align}
where $\Gamma^e(x)$ is any prediction set constructed using only the edge predictive distribution $p^e(y|x)$. 

In general, it is not possible to guarantee the conditional coverage condition (\ref{eq:c6_ideal_goal}) when the prediction set $\Gamma(x)$ differs from the cloud prediction set $\Gamma^*(x)$, unless one choose the trivial prediction set $\Gamma(x)=\mathcal{Y}$ \citep[Sec. 5]{vovk2012conditional}. Therefore, we target a probabilistic version of the guarantee (\ref{eq:c6_ideal_goal}) that can be potentially met while allowing for non-trivial prediction sets at the edge. 

Specifically, considering a batch $\mathcal{D}^\text{te}=\{x_i\}_{i=1}^{|\mathcal{D}^\text{te}|}$ of unlabeled test inputs, instead of imposing that the condition (\ref{eq:c6_ideal_goal}) holds deterministically for all test inputs in $\mathcal{D}^\text{te}$, we target a constraint on the average satisfaction rate over edge-processed inputs. In particular, we wish to ensure a lower bound on the average fraction of edge-processed inputs for which condition (\ref{eq:c6_ideal_goal}) is satisfied. 

Denote as $\mathcal{S} \subseteq \mathcal{D}^\text{te}$ the subset of test examples processed at the edge. Given a tolerated violation level $0 \leq \delta \leq 1$, the requirement on the average satisfaction rate is expressed mathematically as the inequality
\begin{align} \label{eq:c6_batch_goal}
\mathbb{E}\bigg[\frac{ | \{x_i \in \mathcal{S}: \Pr[y_i \in \Gamma(x_i)|x_i] \geq 1-\alpha_{\text{label}}^{\text{mis}} \} | }{|\mathcal{S}|}\bigg] \geq 1 - \delta,
\end{align}
where we follow the convention that $0/0=0$ throughout the work. The inner probability in (\ref{eq:c6_batch_goal}) is taken with respect to the reference distribution of the label $y_i \sim p^*(y_i|x_i)$ given test input $x_i$ as in (\ref{eq:c6_ideal_goal}), while the outer expectation in  (\ref{eq:c6_batch_goal}) is evaluated with respect to the covariates $\{x_i\}_{i=1}^{|\mathcal{D}^\text{te}|}$ in the test input dataset $\mathcal{D}^\text{te}$ and over any reference data used to produce the prediction set $\Gamma(x)$ (see Sec.~\ref{subsec:c6_edge_only} for details). The inequality (\ref{eq:c6_batch_goal}) imposes that the fraction of edge-processed inputs for which the conditional coverage condition (\ref{eq:c6_ideal_goal}) is met is no smaller than $1-\delta$.

Since the requirement (\ref{eq:c6_batch_goal}) can be always guaranteed by a cascading procedure that defers all inputs to the cloud or that returns the trivial prediction set $\Gamma(x) = \mathcal{Y}$, it is important to evaluate the performance of the edge-cloud systems also in terms of the deferral rate and of the informativeness of the prediction set.

The deferral rate evaluates the expected fraction of the test samples deferred to the cloud, i.e.,
\begin{align} \label{eq:c6_deferral_ratio}
   \text{DR} = \mathbb{E}\left[ 1 - \frac{|\mathcal{S}|}{|\mathcal{D}^{\text{te}}|} \right],
\end{align}
where the expectation is taken with respect to the distribution of the selected subset $\mathcal{S}$ and over any reference data used to generate the prediction set $\Gamma(x)$. The deferral rate (\ref{eq:c6_deferral_ratio}) ranges from $0$, indicating that all test samples are processed at the edge, to $1$, indicating that all test inputs are deferred to the cloud.

The informativeness of the prediction set is evaluated by comparing the set size $|\Gamma(x)|$ with the cloud model's set size $|\Gamma^*(x)|$. Accordingly, the expected size of the prediction set $\Gamma(x)$ normalized by the size of the cloud prediction set $\Gamma^*(x)$, referred to normalized inefficiency, is defined as
\begin{align} \label{eq:c6_inefficiency}
     \text{NI} = \frac{1}{|\mathcal{D}^{\text{te}}|} \mathbb{E}\left[ \sum_{x_i \in \mathcal{D}^{\text{te}}} \frac{|\Gamma(x_i)|}{|\Gamma^*(x_i)|}  \right],
\end{align}
where the expectation is taken over the randomness of the covariates $\{x_i\}_{i=1}^{|\mathcal{D}^\text{te}|}$ in the test input dataset $\mathcal{D}^\text{te}$ and over any reference data used to generate the prediction set $\Gamma(x)$.

The normalized inefficiency (\ref{eq:c6_inefficiency}) measures the relative increase in the prediction set size caused by the use of the edge model for some of the test inputs. Accordingly, a normalized inefficiency equal to $1$ indicates an edge-cloud system that is as efficient as the cloud prediction, while a larger normalized inefficiency quantifies the loss of information about the label that is entailed by the use of the edge model.

All in all, a well-designed edge-cloud prediction mechanism should seek to minimize the deferral rate (\ref{eq:c6_deferral_ratio}) and the normalized inefficiency (\ref{eq:c6_inefficiency}), while satisfying the average satisfaction rate guarantee in (\ref{eq:c6_batch_goal}).

\section{Baselines} \label{sec:c6_baselines}

In this section, we introduce baseline prediction strategies based only on the cloud or edge models, as well as a conventional heuristic cascading strategy based on the edge model's confidence \citep{gui2024conformal}.

\subsection{Cloud-Only Inference}
As discussed in Sec.~\ref{subsec:c6_setting}, the cloud-only \gls{hms} $\Gamma^*(x)$ in (\ref{eq:c6_cloud_prediction_set}) is the smallest-cardinality prediction set satisfying the conditional coverage requirement (\ref{eq:c6_ideal_goal}). Since it satisfies (\ref{eq:c6_ideal_goal}), it also directly meets the relaxed requirement (\ref{eq:c6_batch_goal}) for any tolerated violation level $\delta$. Furthermore, the normalized inefficiency (\ref{eq:c6_inefficiency}) equals $1$. However, this scheme has the highest deferral rate, i.e., deferral rate is $1$, since all inputs are escalated to the cloud.

\subsection{Edge-Only Inference} \label{subsec:c6_edge_only}

At the other side of the spectrum with respect to cloud-only schemes are methods that leverage only the edge model $p^e(y|x)$, without requiring access to the cloud. We review three such methods, a baseline edge-only \gls{hms} scheme, \gls{cp}, and \gls{lcp}. By definition, all these schemes exhibit the minimum deferral rate of $0$.

\emph{1) Edge Highest Mass Set: }When we replace cloud predictive distribution $p^*(y|x)$ with the edge predictive distribution $p^e(y|x)$ in the \gls{hms} (\ref{eq:c6_cloud_prediction_set}), we obtain the prediction set
\begin{align} \label{eq:c6_edge_HDS}
    \Gamma^e(x) = \arg \min_{\Gamma^e(x) \subseteq \mathcal{Y}} |\Gamma^e(x)| \text{  s.t. } p^e(\Gamma^e(x)|x) \geq 1-\alpha_{\text{label}}^{\text{mis}}.
\end{align}
The performance of this prediction set is highly sensitive to the edge model's calibration performance. Over-confident edge models tend to produce excessively small edge \gls{hms} (\ref{eq:c6_edge_HDS}), possibly with normalized inefficiency smaller than $1$, violating the target coverage constraint (\ref{eq:c6_batch_goal}). In contrast, under-confident edge models produce excessively large, and thus very inefficient, prediction set (\ref{eq:c6_edge_HDS}), with normalized inefficiency greater than $1$. In general, this approach does not satisfy the target coverage requirement (\ref{eq:c6_batch_goal}). 

\emph{2) Conformal Prediction: }To mitigate edge model miscalibration, \gls{cp} leverages a held-out labeled calibration dataset $\mathcal{D}^{\text{cal}} = \{(x_i, y_i)\}_{i=1}^{|\mathcal{D}^{\text{cal}}|}$ generated from the reference data distribution $p^*(x,y)$ (\ref{eq:c6_ground-truth_dist}) to obtain a prediction set with marginal validity guarantees, as described in Sec.~\ref{sec:c2_certified}. 

Fix a function $J(x,y)$ measuring the discrepancy between the prediction produced by the edge model $p^e(y|x)$ and the true label $y$, such as the negative log-loss $J(x,y) = -\log p^e(y|x)$. This function is applied to all data points in the calibration dataset, producing the set of scores 
\begin{align} \label{eq:c6_edge_CP_error_set}
    \mathcal{J} = \{ J(x_i, y_i) \}_{i=1}^{|\mathcal{D}^{\text{cal}}|}.
\end{align}

Given an input $x$, \gls{cp} constructs the edge prediction set by including all the labels $y\in \mathcal{Y}$ for which the score $J(x,y)$ does not exceed a threshold $q$, i.e., 
\begin{align} \label{eq:c6_edge_CP}
    \Gamma^e(x) = \{ y \in \mathcal{Y}: J(x,y) \leq q \},
\end{align}
where the threshold $q$ is selected as (\ref{eq:c2_cp_quantile}), i.e., the $(1-\alpha_{\text{label}}^{\text{mis}})$-th lower quantile of the empirical distribution of the scores in set $\mathcal{J}$ (\ref{eq:c6_edge_CP_error_set}).


\gls{cp} provides only marginal validity guarantees \citep[Eq.~(1)]{angelopoulos2021gentle}, that is, the prediction set $\Gamma^e(x)$ (\ref{eq:c6_edge_CP}) satisfies the inequality
\begin{align} \label{eq:c6_marginal_validity}
    \Pr \left[ y \in \Gamma^e(x) \right] \geq 1-\alpha_{\text{label}}^{\text{mis}},
\end{align}
where the probability is evaluated with respect to the joint distribution $p^*(x,y)$ of the test pair $(x, y)$ and to the calibration dataset used to generate the edge prediction set $\Gamma^e(x)$. The condition (\ref{eq:c6_marginal_validity}) is weaker than the conditional coverage requirement (\ref{eq:c6_ideal_goal}), and thus \gls{cp} does not guarantee the required inequality (\ref{eq:c6_batch_goal}).

\emph{3) Localized Conformal Prediction: }While \gls{cp}-based methods can only guarantee the marginal coverage condition (\ref{eq:c6_marginal_validity}), a modified version of \gls{cp}, known as \gls{lcp}, attempts to improve conditional coverage by selecting the threshold $q$ in (\ref{eq:c2_cp_quantile}) as a function of the test input $x$ \citep{qkae103}.

To elaborate, fix any localization kernel, such as the Gaussian kernel
\begin{align} \label{eq:c6_gaussian_kernel}
    \kappa(x_1, x_2) = \exp \left( -\frac{\lVert x_1 - x_2 \rVert_2^2}{2h^2} \right),
\end{align}
with kernel bandwidth $h > 0$. Then, given a test input $x$, \gls{lcp} draws a random perturbation $\tilde{x}$ of the test input $x$ by sampling from a distribution with density proportional to the kernel $\kappa(x, \cdot)$. Then, \gls{lcp} evaluates the threshold
\begin{align} \label{eq:c6_localized_quantile}
    \hat{q}(x) = 
    \text{Quantile}_{1-\alpha_{\text{label}}^{\text{mis}}} \left( 
        \sum_{i=1}^{|\mathcal{D}^{\text{cal}}|} w_{x_i}  \delta_{J(x_i, y_i)} 
        + w_{x} \delta_{\infty} 
    \right),
\end{align}
where the normalized weights are
\begin{align} \label{eq:c6_normalized_weight}
    w_{x_i} = \frac{\kappa(x_i, \tilde{x})}{\kappa(x,\tilde{x}) + \sum_{i=1}^{|\mathcal{D}^{\text{cal}}|} \kappa(x_i, \tilde{x})}, 
    w_{x} = \frac{\kappa(x, \tilde{x})}{\kappa(x,\tilde{x}) + \sum_{i=1}^{|\mathcal{D}^{\text{cal}}|} \kappa(x_i, \tilde{x})}.
\end{align}
This approach localizes the threshold (\ref{eq:c6_localized_quantile}) around the test input $x$ by assigning higher weights to calibration points closer to $x$.

Finally, the \gls{lcp} set is 
\begin{align} \label{eq:c6_edge_LCP}
    \Gamma^e(x) = \{ y \in \mathcal{Y}: J(x,y) \leq \hat{q}(x) \}.
\end{align}
By the definition of the localized threshold $\hat{q}(x)$ (\ref{eq:c6_localized_quantile}), a small kernel bandwidth $h$ yields more localized prediction sets, while a large kernel bandwidth $h$ reduces \gls{lcp} to \gls{cp} (\ref{eq:c6_edge_CP}).

Although there is numerical evidence that \gls{lcp} can enhance conditional coverage over \gls{cp} \citep[Thm. 2]{qkae103}, it still guarantees only the marginal validity condition (\ref{eq:c6_marginal_validity}) \citep[Thm. 1]{qkae103}, not meeting the target requirement (\ref{eq:c6_batch_goal}).

\subsection{Confidence-Based Model Cascading} \label{sec:c6_CbD}

As seen, edge-only schemes can not offer the target conditional coverage guarantees (\ref{eq:c6_batch_goal}). In this subsection, we review conventional edge-cloud systems in which the deferral option is implemented by following a heuristic confidence-based rule \citep{gui2024conformal}. 

Given an input $x$, the edge system evaluates a measure of confidence on its output, and decides to defer the decision to the cloud when the confidence level is below a pre-determined threshold. In this work, we adopt the common top-$1$ confidence measure, i.e., $\max_{y \in \mathcal{Y}} p^e(y|x)$ \citep{gui2024conformal}, as briefly described in Sec.~\ref{sec:c2_cascading}. Accordingly, the edge-cloud system produces the prediction sets based on the rule
\begin{align} \label{eq:c6_conventional_cascade}
       \Gamma(x) =
        \begin{cases}
       \Gamma^*(x), & \text{if $ \max_{y \in \mathcal{Y}} p^e(y|x) < \tau_{\text{def}}$ }  \\
       \Gamma^e(x), & \text{if $\max_{y \in \mathcal{Y}} p^e(y|x) \geq \tau_{\text{def}}$ },
        \end{cases}
\end{align}
with a pre-determined threshold $\tau_{\text{def}} \in [0,1]$, where $\Gamma^e(x)$ is an edge-only prediction set, such as \gls{hms} (\ref{eq:c6_edge_HDS}), \gls{cp} (\ref{eq:c6_edge_CP}), or \gls{lcp} (\ref{eq:c6_edge_LCP}).

The threshold $\tau_{\text{def}}$ is typically selected as $\tau_{\text{def}}=1-\delta$ \citep{gui2024conformal}. This way, the edge-only prediction sets for which the edge confidence exceeds the target average satisfaction level $1-\delta$ in (\ref{eq:c6_batch_goal}) are processed by the edge model, while others are outsourced to the cloud.

\section{Conformal Alignment-based Cascading} \label{sec:c6_CA}

In this section, we introduce a \gls{cab} model cascading mechanism that provably meets the target coverage requirement (\ref{eq:c6_batch_goal}). To this end, we formulate the escalation procedure as a \gls{mht} problem by tailoring the \gls{ca} method \citep{gui2024conformal} to adopt the conditional coverage probability (\ref{eq:c6_ideal_goal}) as the alignment score.

\subsection{Model Cascading via Multiple Hypothesis Testing}
The proposed \gls{cab} methodology is based on the observation that the requirement (\ref{eq:c6_batch_goal}) can be interpreted as a \gls{fdr} constraint in an \gls{mht} procedure \citep{sedgwick2014understanding}. To elaborate, given any edge-only prediction set $\Gamma^e(x)$, such as \gls{hms} (\ref{eq:c6_edge_HDS}), \gls{cp} (\ref{eq:c6_edge_CP}), or \gls{lcp} (\ref{eq:c6_edge_LCP}), we write the conditional coverage probability as
\begin{align} \label{eq:c6_alignment_score}
      C^*(x) = p^*(\Gamma^e(x)|x).
\end{align}
In the following, we interpret the probability $C^*(x)$ as an alignment score, measuring how well the edge-only prediction set $\Gamma^e(x)$ aligns with the oracle prediction set in $\Gamma^*(x)$ (\ref{eq:c6_cloud_prediction_set}). In particular, if the edge-only prediction set $\Gamma^e(x)$ aligns well with the cloud-only \gls{hms} $\Gamma^*(x)$, the alignment score must be no smaller than the target conditional coverage probability $1-\alpha_{\text{label}}^{\text{mis}}$.

For any test input $x_i \in \mathcal{D}^{\text{te}}$, we wish to decide whether the edge model prediction set meets the conditional coverage requirement (\ref{eq:c6_ideal_goal}). To formalize this problem, we assign each test input $x_i \in \mathcal{D}^{\text{te}}$ to a null hypothesis $\mathcal{H}_i$ that the edge-only prediction set $\Gamma^e(x)$ fails to satisfy the conditional coverage requirement (\ref{eq:c6_ideal_goal}). This can be expressed mathematically via the inequality
\begin{align} \label{eq:c6_null_hypothesis}
    \mathcal{H}_i: C^*(x_i) < 1-\alpha_{\text{label}}^{\text{mis}}.
\end{align} 
While the hypothesis (\ref{eq:c6_null_hypothesis}) pertains to an individual test input $x_i \in \mathcal{D}^{\text{te}}$, the average satisfaction rate guarantee (\ref{eq:c6_batch_goal}) requires the simultaneous consideration of the hypothesis for all test inputs $x_i \in \mathcal{D}^{\text{te}}$, inducing an \gls{mht} problem.
\begin{figure} [tb] 
    \centering
    \centerline{\includegraphics[width=\textwidth]{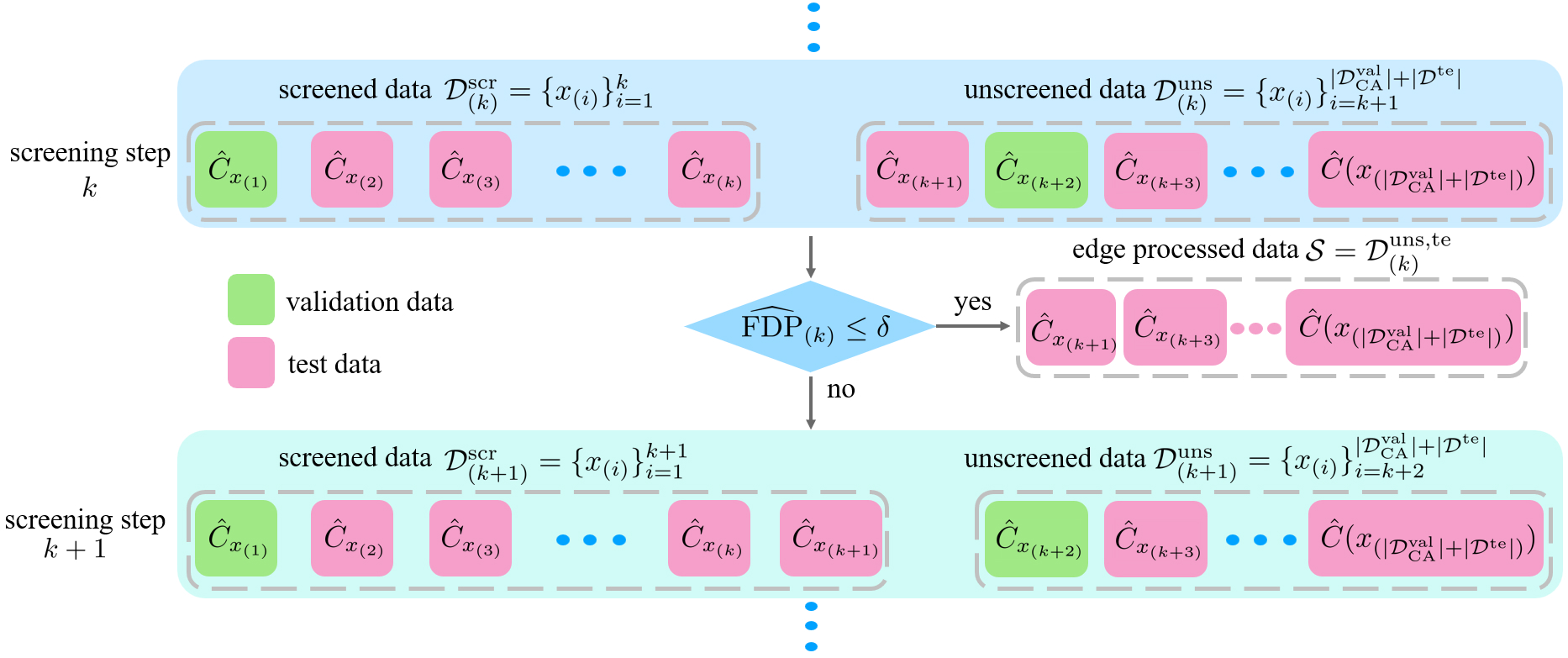}}
     \caption{The proposed \gls{cab} model cascading adopts a sequential screening methodology to ensure that the edge-processed subset $\mathcal{S}$ satisfies the constraint (\ref{eq:c6_batch_goal}). This procedure evaluates inputs in the joint test and validation dataset $\mathcal{D}^{\text{te}} \cup \mathcal{D}^{\text{val}}_{\text{CA}}$ in the order (\ref{eq:c6_screen_order}) of increasing estimate alignment score $\hat{C}(x)$. At each step, the \gls{cab} method estimates the \gls{fdp} of the unscreened test inputs based on the unscreened validation data using (\ref{eq:c6_FDP_estimator}). The procedure terminates at the earliest step $k_{\text{CA}}$ in (\ref{eq:c6_stopping_rule}) when the estimated \gls{fdp} of unscreened test inputs falls below the pre-determined tolerated level $\delta$.}
    \label{fig:c6_ACS} 
\end{figure}
In this \gls{mht} problem, the subset $\mathcal{S} \subseteq \mathcal{D}^{\text{te}}$ of test inputs that are processed at the edge corresponds to the subset of null hypotheses $\{ \mathcal{H}_i\}_{i=1}^{|\mathcal{D}^{\text{te}}|}$ in (\ref{eq:c6_null_hypothesis}) that are rejected. Accordingly, we can reformulate the edge-cloud prediction set $\Gamma(x_i)$ in (\ref{eq:c6_original_prediction}) as
\begin{align} \label{eq:c6_final_prediction}
       \Gamma(x_i) =
        \begin{cases}
       \Gamma^*(x_i), & \text{if input $x_i \notin \mathcal{S}$, i.e., $\mathcal{H}_i$ is accepted}  \\
       \Gamma^e(x_i) , & \text{if input $x_i \in \mathcal{S}$, i.e., $\mathcal{H}_i$ is rejected}.
        \end{cases}
\end{align}

Furthermore, the average satisfaction rate guarantee (\ref{eq:c6_batch_goal}) can be expressed in terms of the \gls{fdp}, i.e., the fraction of test samples in the edge-processed subset $\mathcal{S}$ for which the null hypothesis (\ref{eq:c6_null_hypothesis}) is incorrectly rejected \citep{jin2023selection}. By the definition of the null hypothesis in (\ref{eq:c6_null_hypothesis}), the \gls{fdp} is defined as 
\begin{align} \label{eq:c6_FDP}
   \text{FDP}(\mathcal{S}) = 
\frac{\left| \{ x_i \in \mathcal{S} : C^*(x_i) < 1 - \alpha_{\text{label}}^{\text{mis}} \} \right|}
{|\mathcal{S}|}.
\end{align}
Then, the average satisfaction rate guarantee (\ref{eq:c6_batch_goal}) can be equivalently written as the inequality
\begin{align} \label{eq:c6_batch_goal_reformulated}
    \text{FDR} = \mathbb{E}[\text{FDP}(\mathcal{S})] \leq \delta,
\end{align}
where the expectation is evaluated with respect to both the distribution of the subset $\mathcal{S}$ and the reference data used to generate the prediction set $\Gamma(x)$. The expected value of the \gls{fdp} in (\ref{eq:c6_batch_goal_reformulated}) is known as the \gls{fdr}. 

\subsection{Conformal Alignment-Based Model Cascading}

In this subsection, we describe the proposed \gls{cab} model cascading procedure that enforces the constraint (\ref{eq:c6_batch_goal_reformulated}). As illustrated in Fig.~\ref{fig:c6_ACS}, we adopt a sequential screening approach \citep{mary2022semi, gui2025acs}, whereby the inputs that are likely to violate the coverage condition (\ref{eq:c6_ideal_goal}) are progressively eliminated until the remaining unscreened test inputs satisfy the requirement (\ref{eq:c6_batch_goal_reformulated}). 

The \gls{cab} method assumes the availability of a reference dataset $\mathcal{D}$ consisting of pairs $(x, C^*(x))$, where $C^*(x)$ is the true alignment score (\ref{eq:c6_alignment_score}). Note that the label $C^*(x)$ is obtained by querying the cloud model during an offline phase. The reference dataset $\mathcal{D}$ is partitioned into two disjoint datasets, i.e., a training dataset $\mathcal{D}^{\text{tr}}_{\text{CA}}$ and a validation dataset $\mathcal{D}^{\text{val}}_{\text{CA}}$, which are used as detailed next.

Since the true alignment score $C^*(x)$ is not available for test inputs, we introduce an alignment score predictor $\hat{C}(x)$ \citep{jin2023selection, gui2024conformal}. This predictor is trained on the training dataset $\mathcal{D}^{\text{tr}}_{\text{CA}}= \{ (x_i, C^*(x_i)) \}_{i=1}^{|\mathcal{D}^{\text{tr}}_{\text{CA}}|}$ in an offline phase using any supervised learning method. No specific assumption is imposed on the quality of this predictor.

\emph{1) Sequential Screening: }Given a pre-trained alignment score predictor $\hat{C}(x)$, and given an input batch $\mathcal{D}^{\text{te}}$, the proposed \gls{cab} methodology uses the validation dataset $\mathcal{D}^{\text{val}}_{\text{CA}} = \{ (x_i, C^*(x_i)) \}_{i=1}^{|\mathcal{D}^{\text{val}}_{\text{CA}}|}$ to guide the selection of the subset $\mathcal{S} \subseteq \mathcal{D}^{\text{te}}$ of test inputs to process at the edge. To start, the test dataset $\mathcal{D}^{\text{te}}$ is combined with the validation dataset $\mathcal{D}^{\text{val}}_{\text{CA}}$ to form a joint set
\begin{align}
    \mathcal{D}^{\text{te}} \cup \mathcal{D}^{\text{val}}_{\text{CA}} = \{x_1, \cdots, x_{|\mathcal{D}^{\text{val}}_{\text{CA}}|}, \cdots, x_{|\mathcal{D}^{\text{val}}_{\text{CA}}|+|\mathcal{D}^{\text{te}}|} \},
\end{align} 
where $\{x_1, \cdots, x_{|\mathcal{D}^{\text{val}}_{\text{CA}}|} \}$ represents the inputs for the labeled validation samples in the validation dataset $\mathcal{D}^{\text{val}}_{\text{CA}}$. Then, the edge device ranks all samples in the union $\mathcal{D}^{\text{te}} \cup \mathcal{D}^{\text{val}}_{\text{CA}}$ as $x_{(1)}, \cdots, x_{(|\mathcal{D}^{\text{val}}_{\text{CA}}|+|\mathcal{D}^{\text{te}}|)}$ in ascending order of their predicted alignment score $\hat{C}(x)$, i.e.,
\begin{align} \label{eq:c6_screen_order}
    \hat{C}(x_{(1)}) \leq \cdots \leq \hat{C}(x_{(|\mathcal{D}^{\text{val}}_{\text{CA}}|)}) \leq \cdots \leq \hat{C}(x_{(|\mathcal{D}^{\text{val}}_{\text{CA}}|+|\mathcal{D}^{\text{te}}|)}).
\end{align}
Intuitively, this step lists the inputs in order from least to most promising for edge processing.

As illustrated in Fig.~\ref{fig:c6_ACS}, the edge proceeds to screen the data points in the joint dataset $\mathcal{D}^{\text{te}} \cup \mathcal{D}^{\text{val}}_{\text{CA}}$ following the order in (\ref{eq:c6_screen_order}), with screening steps indexed by an integer $k =1,2,\cdots, |\mathcal{D}^{\text{val}}_{\text{CA}}|+|\mathcal{D}^{\text{te}}|$. Accordingly, at each screening step $k$, we screen the new input $x_{(k)}$, and we define the screened inputs and the unscreened inputs as 
\begin{align} \label{eq:c6_unscreend_test_inputs}
     \mathcal{D}^{\text{scr}}_{(k)} = \{x_{(i)}\}_{i=1}^{k} \quad \text{and}  \quad \mathcal{D}^{\text{uns}}_{(k)} = \{x_{(i)}\}_{i=k + 1}^{|\mathcal{D}^{\text{val}}_{\text{CA}}|+|\mathcal{D}^{\text{te}}|},
\end{align}
respectively. Since the unscreened input subset $\mathcal{D}^{\text{uns}}_{(k)}$ generally includes both validation and test data, we also partition this set into unscreened validation and test subsets as
\begin{align} \label{eq:c6_selected_set}
    \mathcal{D}^{\text{uns,val}}_{(k)}  =\mathcal{D}^{\text{uns}}_{(k)} \cap \mathcal{D}^{\text{val}}_{\text{CA}} \quad \text{and}\quad
    \mathcal{D}^{\text{uns,te}}_{(k)}  =\mathcal{D}^{\text{uns}}_{(k)} \cap \mathcal{D}^{\text{te}},
\end{align}
respectively. Initially, at step $k = 0$, the screened input subset is an empty set, $\mathcal{D}^{\text{scr}}_{(0)} = \emptyset$.

The screening procedure proceeds along steps $k =1,2,\cdots, |\mathcal{D}^{\text{val}}_{\text{CA}}|+|\mathcal{D}^{\text{te}}|$, until a certain condition is met at some step $k_{\text{CA}} \leq |\mathcal{D}^{\text{val}}_{\text{CA}}|+|\mathcal{D}^{\text{te}}|$. Once this occurs, the \gls{cab} procedure returns the set 
\begin{align} \label{eq:c6_edge_process_set}
    \mathcal{S} = \mathcal{D}^{\text{uns,te}}_{(k_{\text{CA}})}
\end{align}
of unscreened test inputs. By the ordering (\ref{eq:c6_screen_order}), this set contains all test inputs $x_i \in \mathcal{D}^{\text{te}}$ with an estimated alignment score $\hat{C}(x)$ no smaller than $\hat{C}(x_{(k_{\text{CA}})})$, i.e.,
\begin{align}
    \mathcal{S} = \{ x_i \in  \mathcal{D}^{\text{te}}: \hat{C}(x_i) \geq \hat{C}(x_{(k_{\text{CA}})})\}.
\end{align}

To determine the stopping time $k_{\text{CA}}$, as illustrated in Fig.~\ref{fig:c6_ACS}, at each step $k$, the \gls{cab} method estimates the \gls{fdp} (\ref{eq:c6_FDP}) of the subset of unscreened test inputs $\mathcal{D}^{\text{uns,te}}_{(k)}$ by using the corresponding \gls{fdp} of the subset of unscreened validation inputs $\mathcal{D}^{\text{uns,val}}_{(k)}$ as
\begin{align} \label{eq:c6_FDP_estimator}
    \widehat{\text{FDP}}_{( k)} = \frac{|\mathcal{D}^{\text{te}}|}{1 + |\mathcal{D}^{\text{val}}_{\text{CA}}|} \frac{1+| \{x_i \in \mathcal{D}^{\text{uns,val}}_{( k)}: C^*(x_i) < 1-\alpha_{\text{label}}^{\text{mis}}      \}|}{|\mathcal{D}^{\text{uns,te}}_{(k)}|}.
\end{align}
Intuitively, the multiplicative term $|\mathcal{D}^{\text{te}}| / (1 + |\mathcal{D}^{\text{val}}_{\text{CA}}|)$ in (\ref{eq:c6_FDP_estimator}) compensates for the discrepancy in the sizes of the validation and the test dataset \citep[Eq.~(2)]{gui2025acs}. Note that the \gls{fdp} for the unscreened validation inputs, which is obtained as the ratio $| \{ x_i \in \mathcal{D}^{\text{uns,val}}_{(k)}: C^*(x_i) < 1 - \alpha_{\text{label}}^{\text{mis}} \} | / |\mathcal{D}^{\text{uns,val}}_{(k)}|$ by the definition (\ref{eq:c6_FDP}), can be evaluated since the ground-truth alignment scores $C^*(x_i)$ are available for the validation samples $x_i \in \mathcal{D}^{\text{val}}_{\text{CA}}$. 

With this estimate, the \gls{cab} method terminates the sequential screening procedure at the first step that meets the condition (\ref{eq:c6_batch_goal_reformulated}), with the estimate (\ref{eq:c6_FDP_estimator}) used in lieu of the true \gls{fdp}, i.e.,
\begin{align} \label{eq:c6_stopping_rule}
    k_{\text{CA}} = \inf \{ k \geq 0: \widehat{\text{FDP}}_{( k)} \leq \delta \}.
\end{align}
As discussed next, this procedure satisfies the requirement (\ref{eq:c6_batch_goal}).

\emph{2) Theoretical Guarantees: }The output $\mathcal{S} = \mathcal{D}_{(k_\text{CA})}^\text{uns,te}$ of the \gls{cab} model cascading methodology satisfies the target \gls{fdr} constraint (\ref{eq:c6_batch_goal_reformulated}), which coincides with the target average satisfaction rate guarantee (\ref{eq:c6_batch_goal}).

\textbf{Proposition 6.1:} \textit{If the examples in the reference dataset $\mathcal{D}$, and the test dataset $\mathcal{D}^{\text{te}}$ are exchangeable, then, for any pre-determined average satisfaction level $1-\delta \in [0,1]$, the output subset $\mathcal{S}$ in (\ref{eq:c6_edge_process_set}) satisfies the average satisfaction rate requirement (\ref{eq:c6_batch_goal}).
}

\emph{Proof}: The proof of this proposition follows directly from the \gls{fdr} control of \gls{ca} (see \citep[Thm.~1]{gui2025acs} for details). For completeness, a proof tailored to the sequential screening procedure presented in this subsection, which was introduced in \citep{gui2025acs}, can be found in Appendix B.1. \hfill $\blacksquare$

\section{Experiments} \label{sec:c6_results}

In this section, to validate the proposed approaches, we report empirical results for vision and \gls{qa} tasks.

\subsection{Performance Metrics}
For both tasks, we consider the following evaluation metrics:
\begin{itemize}
    \item {Average satisfaction rate}, the average proportion of edge-processed test samples whose conditional coverage probability (\ref{eq:c6_idea_goal_reformula}) is no smaller than the desired requirement $1-\alpha_{\text{label}}^{\text{mis}}$, estimating the left-hand side of (\ref{eq:c6_batch_goal}).
    \item {Deferral rate}, the averaged fraction of test samples deferred to the cloud, estimating (\ref{eq:c6_deferral_ratio}).
    \item {Normalized inefficiency}, the expected size of the prediction set $\Gamma(x)$ normalized by the size of the oracle cloud prediction set $\Gamma^*(x)$, estimating (\ref{eq:c6_inefficiency}).
\end{itemize}

\subsection{Implementation}
The calibration dataset $\mathcal{D}^{\text{cal}} = \{(x_i, y_i)\}_{i=1}^{|\mathcal{D}^{\text{cal}}|}$ is used by the edge model $p^e(y|x)$ to construct the edge prediction set $\Gamma^e(x)$ as in (\ref{eq:c6_edge_CP}) or in (\ref{eq:c6_edge_LCP}). We randomly partition the reference dataset $\mathcal{D}$ into two disjoint datasets, namely the training dataset $\mathcal{D}^{\text{tr}}_{\text{CA}} = \{ (x_i, C^*(x_i)) \}_{i=1}^{|\mathcal{D}^{\text{tr}}_{\text{CA}}|}$ used to train the alignment score predictor $\hat{C}(x)$ in (\ref{eq:c6_screen_order}), and the validation dataset $\mathcal{D}^{\text{val}}_{\text{CA}} = \{ (x_i, C^*(x_i)) \}_{i=1}^{|\mathcal{D}^{\text{val}}_{\text{CA}}|}$ used for the \gls{cab} deferral decision in (\ref{eq:c6_stopping_rule}). We fix the sizes for each dataset as $|\mathcal{D}^{\text{cal}}| = 500$, $|\mathcal{D}^{\text{tr}}_{\text{CA}}|=200$, $|\mathcal{D}^{\text{val}}_{\text{CA}}|=500$, and $|\mathcal{D}^{\text{te}}|=100$. For the alignment score predictor $\hat{C}(x)$ in (\ref{eq:c6_screen_order}), we adopt a regression model that takes as input the probability $p^e(\Gamma^e(x)|x)$, which represents the conditional coverage probability for the edge prediction set $\Gamma^e(x)$ as estimated by the edge model.

All results are averaged over $200$ independent runs, with each run corresponding to an independent split of the datasets. All the experiments are implemented via PyTorch \citep{paszke2019pytorch}, and run over a GPU server with a single NVIDIA A100 card\footnotemark[4].
\footnotetext[4]{Code can be found at \url{https://github.com/kclip/Edge-Cloud-Conformal-Alignment}.}

\subsection{Image Classification}

For the image classification task, we use the CIFAR-100 dataset \citep{krizhevsky2010cifar}, with a Bayesian WideResNet-40-2 network \citep{huang2025calibrating} and a standard WideResNet-40-2 model \citep{zagoruyko2016wide} as the cloud and edge models, respectively. The alignment score predictor $\hat{C}(x)$ is trained via XGBoost \citep{gui2024conformal}.

\begin{figure} [tb] 
    \centering
    \centerline{\includegraphics[width=\textwidth]{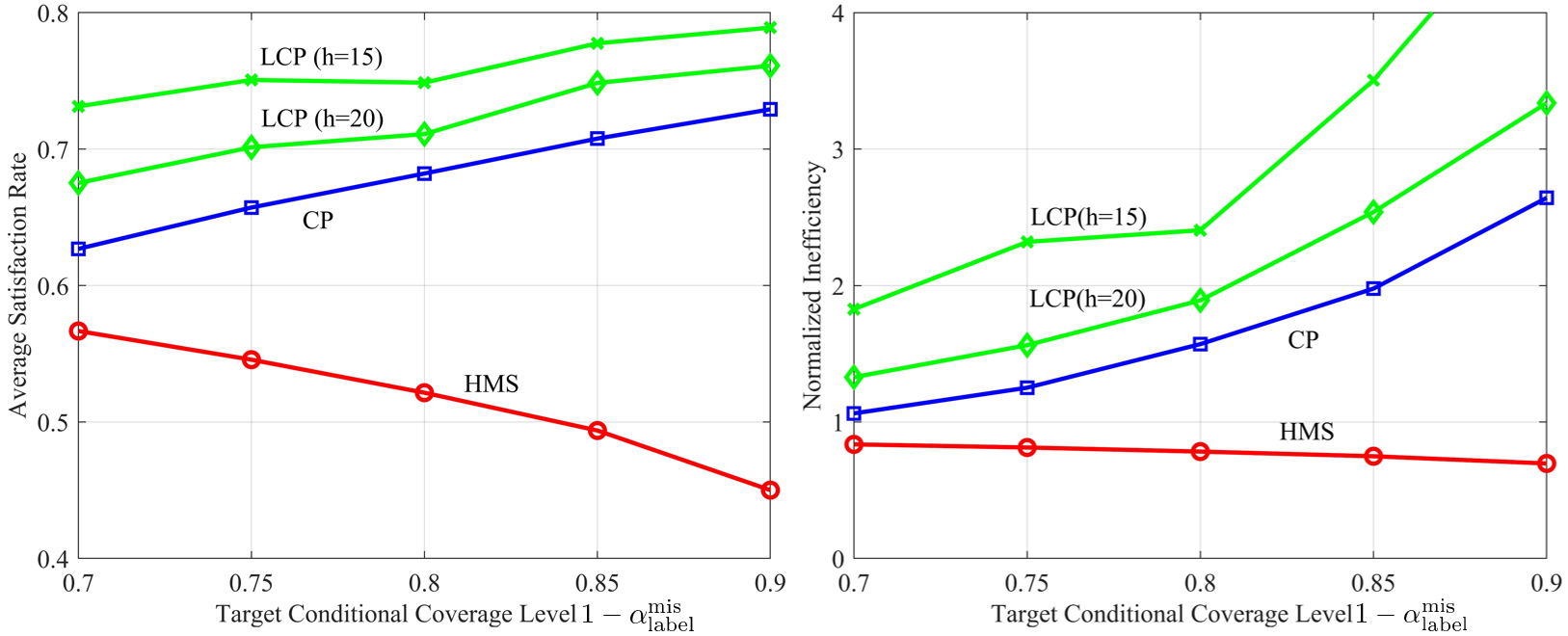}}
     \caption{Average satisfaction rate (left) and normalized inefficiency (right) versus target conditional coverage levels $1-\alpha_{\text{label}}^{\text{mis}} \in \{0.7, 0.75, 0.8, 0.85, 0.9\}$ on the CIFAR-100 dataset, considering edge-only schemes: \gls{hms} in (\ref{eq:c6_edge_HDS}), \gls{cp} in (\ref{eq:c6_edge_CP}), and \gls{lcp} in (\ref{eq:c6_edge_LCP}) with Gaussian kernel bandwidth $h=15$ and $h=20$, respectively.}
    \label{fig:c6_image_CD} 
\end{figure}

\textbf{Empirical average satisfaction rates for edge-only schemes.} To start, we present reference results for edge-only schemes in Fig.~\ref{fig:c6_image_CD}, where we report the average satisfaction rate and the normalized inefficiency at target conditional coverage levels $1-\alpha_{\text{label}}^{\text{mis}} \in \{0.7, 0.75,0.8,0.85,0.9 \}$ for \gls{hms} (\ref{eq:c6_edge_HDS}), \gls{cp} (\ref{eq:c6_edge_CP}), and \gls{lcp} (\ref{eq:c6_edge_LCP}). For \gls{lcp}, we fix the Gaussian kernel bandwidth to $h=15$ and to $h=20$, respectively. It is emphasized that the performance of edge-only schemes in terms of conditional coverage can only be evaluated using empirical means, as edge-only schemes do not offer any formal mechanism to control the average satisfaction rate as in (\ref{eq:c6_batch_goal}). That said, the results in Fig.~\ref{fig:c6_image_CD} provide useful benchmarks for the cloud-aided cascading techniques studied in this work.

The empirical results in Fig.~\ref{fig:c6_image_CD} show that edge-only schemes achieve low values of the average satisfaction rate, e.g., \gls{lcp} with bandwidth $h=15$ obtaining average satisfaction rate $0.71$ and $0.79$ at target level $1-\alpha_{\text{label}}^{\text{mis}}=0.7$ and $1-\alpha_{\text{label}}^{\text{mis}}=0.9$, respectively. \gls{lcp} with bandwidth $h=15$ attains a higher satisfaction rate than any other edge-only scheme, including \gls{lcp} with bandwidth $h=20$, but at the cost of much larger prediction sets. This indicates that a more localized kernel (\ref{eq:c6_gaussian_kernel}) helps enhance the conditional coverage by increasing the size of the prediction sets. Based on this observation, in the following, we set the bandwidth of \gls{lcp} to $h=20$ to balance conditional coverage and inefficiency.

\begin{figure} [tb] 
    \centering
    \centerline{\includegraphics[scale=0.28]{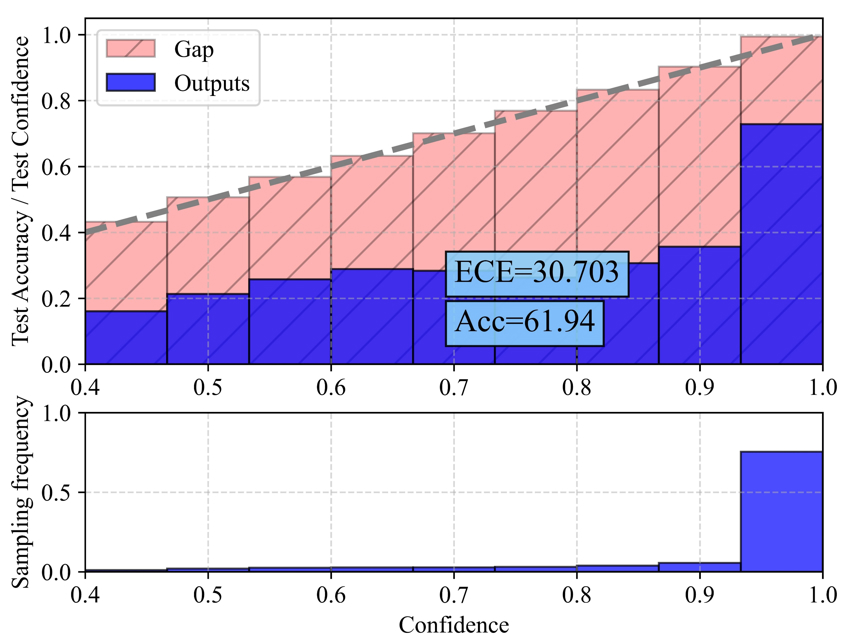}}
    \caption{Reliability diagram for the edge model, namely WideResNet-40-2 model, on the CIFAR-100 dataset.}
    \label{fig:c6_image_rd} 
\end{figure}

To understand why the edge model tends to undercover the true conditional distribution of the output label in this setting, Fig.~\ref{fig:c6_image_rd} shows the reliability diagram of the edge model. The reliability diagram plots the test accuracy as a function of the model confidence \citep{guo2017calibration}. The diagram highlights how the edge model, namely the WideResNet-40-2 model, is highly over-confident, having large positive gaps between accuracy and confidence. The model's over-confidence is reflected in predictive distributions $p^e(y|x)$ that are highly peaked around the top-$1$ label. This, in turn, leads to excessively small \gls{hms} (\ref{eq:c6_edge_HDS}). Consequently, the average satisfaction rate for \gls{hms} declines as the target coverage requirement $1-\alpha_{\text{label}}^{\text{mis}}$ increases, i.e., becomes stricter.

While \gls{cp} and \gls{lcp} guarantee only marginal validity \citep{qkae103}, Fig.~\ref{fig:c6_image_CD} demonstrates that they generally improve the achievable satisfaction rate for conditional coverage as compared to \gls{hms}. This is done by suitably increasing the size of the prediction set (see right panel of Fig.~\ref{fig:c6_image_CD}). In particular, \gls{lcp} achieves a satisfaction rate higher than \gls{cp} at the expense of further inflating the prediction sets. However, as mentioned, even with \gls{lcp} with a small bandwidth $h=15$, the average satisfaction rate remains quite low as a result of the poor calibration of the edge model.

\begin{figure} [tb] 
    \centering
    \centerline{\includegraphics[width=\textwidth]{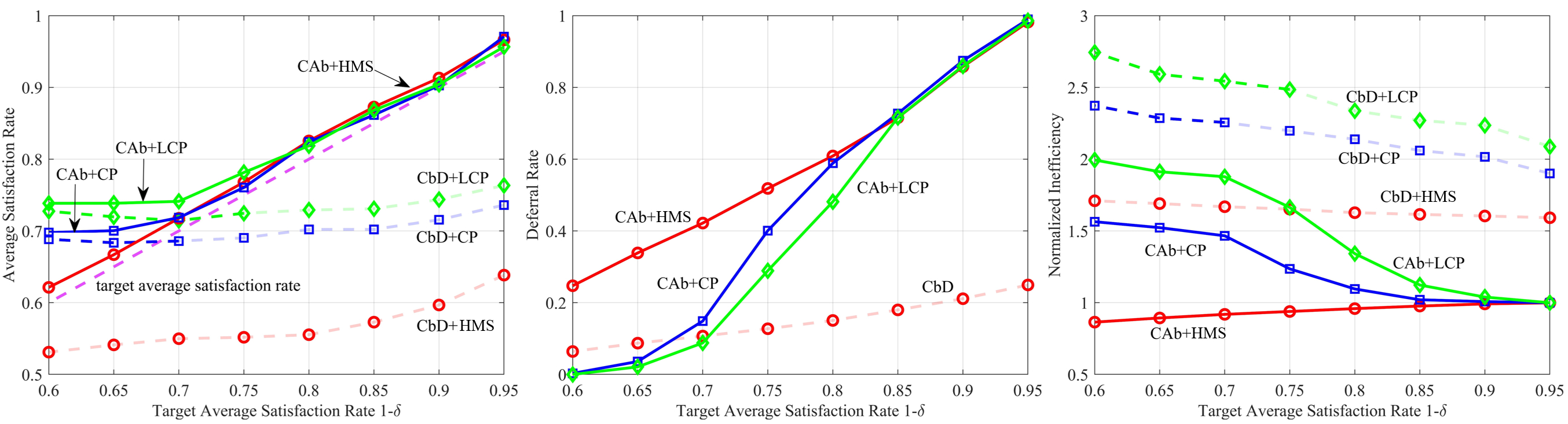}}
    \caption{Average satisfaction rate (left), deferral rate (middle), and normalized inefficiency (right) for conventional \gls{cbd} schemes and the proposed \gls{cab} schemes versus different target average satisfaction rates $1-\delta \in \{ 0.6, 0.65, \dots, 0.95\}$ on the CIFAR-100 dataset for required conditional coverage level $1-\alpha_{\text{label}}^{\text{mis}}=0.8$. The dashed straight line in the left figure indicates the target average satisfaction rate $1-\delta$. The other dashed lines in the figure represent the \gls{cbd} schemes, with transparent segments indicating regimes in which \gls{cbd} does not meet the target average satisfaction requirements (\ref{eq:c6_batch_goal}).}
    \label{fig:c6_image_CA} 
\end{figure}

\textbf{Confidence-based versus conformal alignment-based model cascading.} As discussed, edge-only schemes, such as \gls{hms}, \gls{cp}, and \gls{lcp}, do not offer any formal mechanism to enforce a constraint on the average satisfaction rate as in (\ref{eq:c6_batch_goal}). This requirement can only be met by implementing a deferral option to outsource inference to the cloud. To elaborate on the relative merits of different cascading techniques, we now compare the performance of \gls{cbd} schemes, which operate according to the heuristic rule (\ref{eq:c6_conventional_cascade}), to the proposed \gls{cab} schemes, which operate as detailed in Sec.~\ref{sec:c6_CA}. We emphasize that only \gls{cab} schemes can formally guarantee the average satisfaction rate constraint (\ref{eq:c6_batch_goal}). In this analysis, we vary the target average satisfaction level in the set $1-\delta \in \{ 0.6, 0.65, \dots, 0.95\}$, with a fixed target conditional coverage requirement $1-\alpha_{\text{label}}^{\text{mis}} = 0.8$. Following a conventional thresholding strategy, we set the confidence threshold for the deferral rule (\ref{eq:c6_conventional_cascade}) of \gls{cbd} as $\tau_{\text{def}} = 1-\delta$ (see Sec.~\ref{sec:c6_CbD}).

Fig.~\ref{fig:c6_image_CA} reports the average satisfaction rate, deferral rate, and the normalized inefficiency as a function of the required average satisfaction rate $1-\delta$. \gls{cbd} schemes, due to the heuristic nature of the deferral rule (\ref{eq:c6_conventional_cascade}), do not generally meet the average satisfaction requirement (\ref{eq:c6_batch_goal}). In contrast, as formalized by Proposition 6.1, \gls{cab} schemes can always guarantee the condition (\ref{eq:c6_batch_goal}), regardless of the choice of the edge prediction set, namely \gls{hms}, \gls{cp}, or \gls{lcp}.

For a fixed target average satisfaction rate $1-\delta$, the choice of the edge prediction set strategy, namely \gls{hms}, \gls{cp}, or \gls{lcp}, determines different trade-offs between deferral rate and normalized inefficiency. In particular, as seen in Fig.~\ref{fig:c6_image_CD}, \gls{hms} yields smaller prediction sets than \gls{cp}, which in turn produces smaller prediction sets than \gls{lcp}. This ensures that the deferral rate decreases when switching from \gls{hms} to \gls{cp} and from \gls{cp} to \gls{lcp}. Overall, \gls{hms} yields the smallest prediction sets with the largest deferral rate, while \gls{lcp} produces the largest prediction sets with the smallest deferral rate. \gls{cp} offers an intermediate solution in terms of the trade-off between deferral rate and prediction set size.

\begin{figure} [tb] 
    \centering
    \centerline{\includegraphics[scale=0.24]{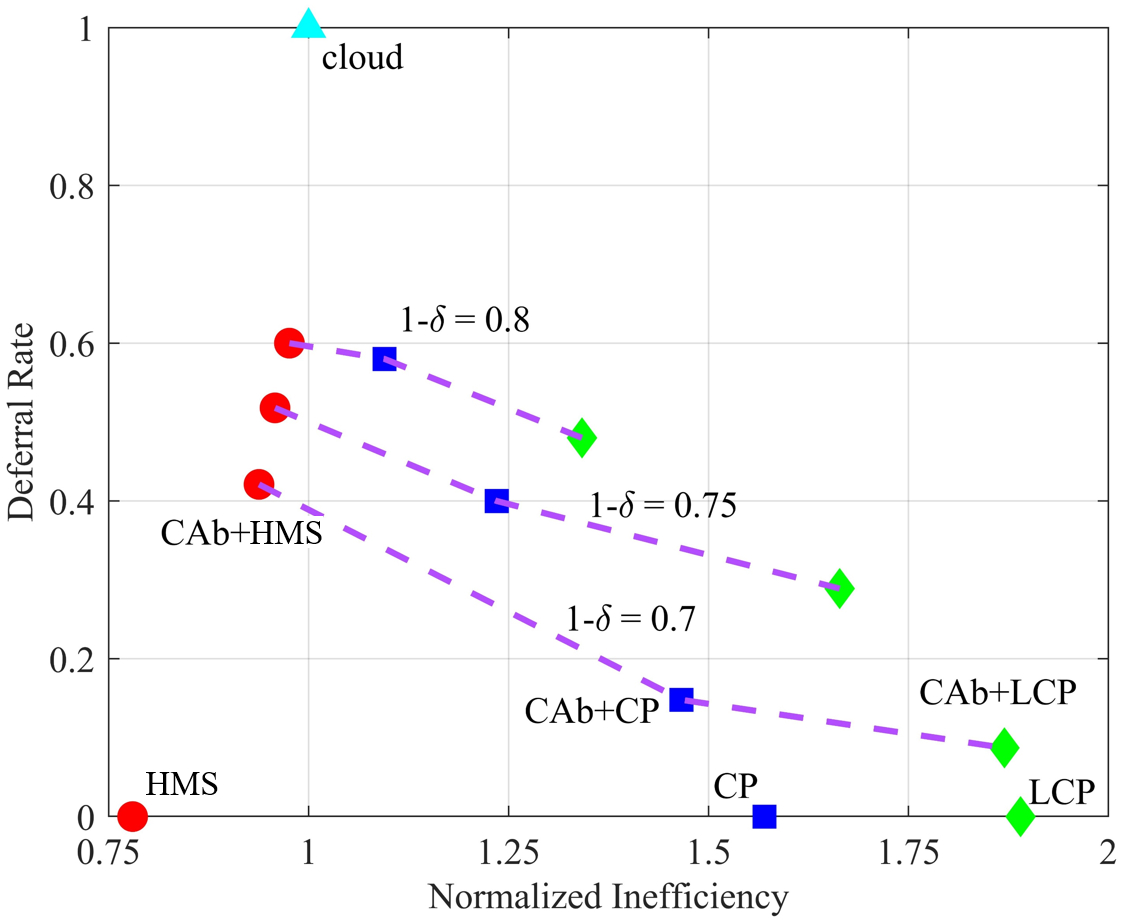}}
    \caption{Deferral rate versus normalized inefficiency obtained by changing the target average satisfaction level $1-\delta$ on CIFAR-100 dataset for \gls{cab} schemes, namely \gls{cab}+\gls{hms}, \gls{cab}+\gls{cp}, and \gls{cab}+\gls{lcp}. Points on the same dashed line share the same target average satisfaction rate $1-\delta$.}
    \label{fig:c6_image_pareto} 
\end{figure}

\textbf{Trade-off between deferral rate and normalized inefficiency.} The trade-offs between the deferral rate and prediction set size observed in Fig.~\ref{fig:c6_image_CA} are further analyzed in Fig.~\ref{fig:c6_image_pareto}, which plots the deferral rate versus the normalized inefficiency for different target average satisfaction rate $1-\delta$ in (\ref{eq:c6_batch_goal}). We focus on \gls{cab} schemes given their capacity to guarantee the average satisfaction rate constraint (\ref{eq:c6_batch_goal}). First, we note that increasing the value of the requirement $1-\delta$ consistently raises the deferral rate for all schemes, while driving normalized inefficiency toward $1$. Furthermore, for a fixed value of the requirement $1-\delta$, larger prediction sets achieve lower deferral rates, with \gls{hms}, \gls{cp}, and \gls{lcp} yielding increasingly large prediction sets.

\subsection{Question Answering}
We now consider TeleQnA \citep{maatouk2023teleqna}, a real-world multiple-choice \gls{qa} dataset, which is used for assessing the knowledge of \gls{llm} in the field of telecommunications. The TeleQnA dataset contains $10,000$ multiple-choice questions, including $6441$ five questions-options pairs, $3456$ four questions-options pairs, and a small number with two or three questions-options pairs, spanning five distinct categories: lexicon, research overview, research publications, standards overview, and standards specifications. We focus here on the four questions-options pairs.

We adopt language models Qwen2-7B-Instruct and Qwen2-1.5B-Instruct \citep{qwen2} as the cloud and edge models, respectively, without fine-tuning. Treating the \gls{llm} as black boxes, we approximate the cloud conditional distribution $p^*(y|x)$ and edge conditional distribution $p^e(y|x)$ by randomly sampling $10$ answers per question as in \citep{wang2022probabilistic}. The alignment score predictor $\hat{C}(x)$ is trained via XGBoost \citep{gui2024conformal}.
\begin{figure} [tb] 
    \centering
    \centerline{\includegraphics[scale=0.28]{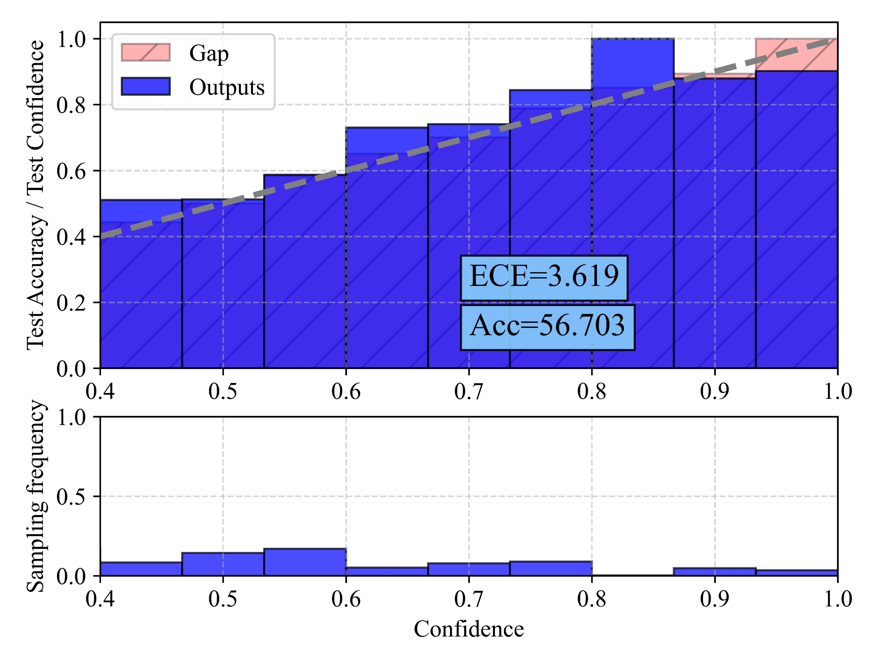}}
    \caption{Reliability diagram for the edge language models, namely Qwen2-7B-Instruct, on the TeleQnA dataset.}
    \label{fig:c6_qa_rd} 
\end{figure}

Based on the insights obtained from the previous experiment, we start by analyzing the calibration properties of the edge model. To this end, Fig.~\ref{fig:c6_qa_rd} shows the reliability diagram \citep{guo2017calibration} for the edge model, namely Qwen2-1.5B-Instruct. It is observed that, in stark contrast to the previous setting, here the edge model is generally under-confident but well-calibrated, exhibiting small negative gaps between accuracy and confidence. As we will see, this modifies the relative performance of cascading schemes based on \gls{hms}, \gls{cp}, and \gls{lcp} prediction sets, as compared to the previous experiment.
\begin{figure} [tb] 
    \centering
    \centerline{\includegraphics[width=\textwidth]{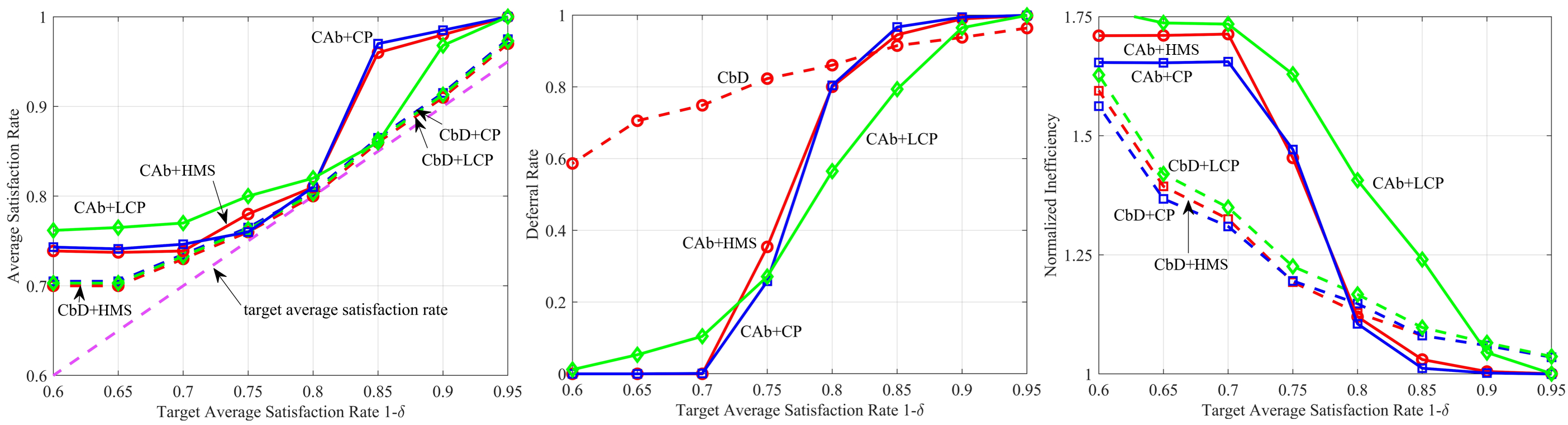}}
    \caption{Average satisfaction rate (left), deferral rate (middle), and normalized inefficiency (right) for \gls{cbd} schemes and \gls{cab} schemes versus different target average satisfaction levels $1-\delta \in \{ 0.6, 0.65, \dots, 0.95\}$ on the TeleQnA dataset for required conditional coverage level $1-\alpha_{\text{label}}^{\text{mis}}=0.8$. The dashed straight line in the figure indicates the target average satisfaction rate $1-\delta$. The other dashed lines in the figure represent the \gls{cbd} schemes. }
    \label{fig:c6_qa_CA} 
\end{figure}

\begin{figure} [tb] 
    \centering
    \centerline{\includegraphics[scale=0.24]{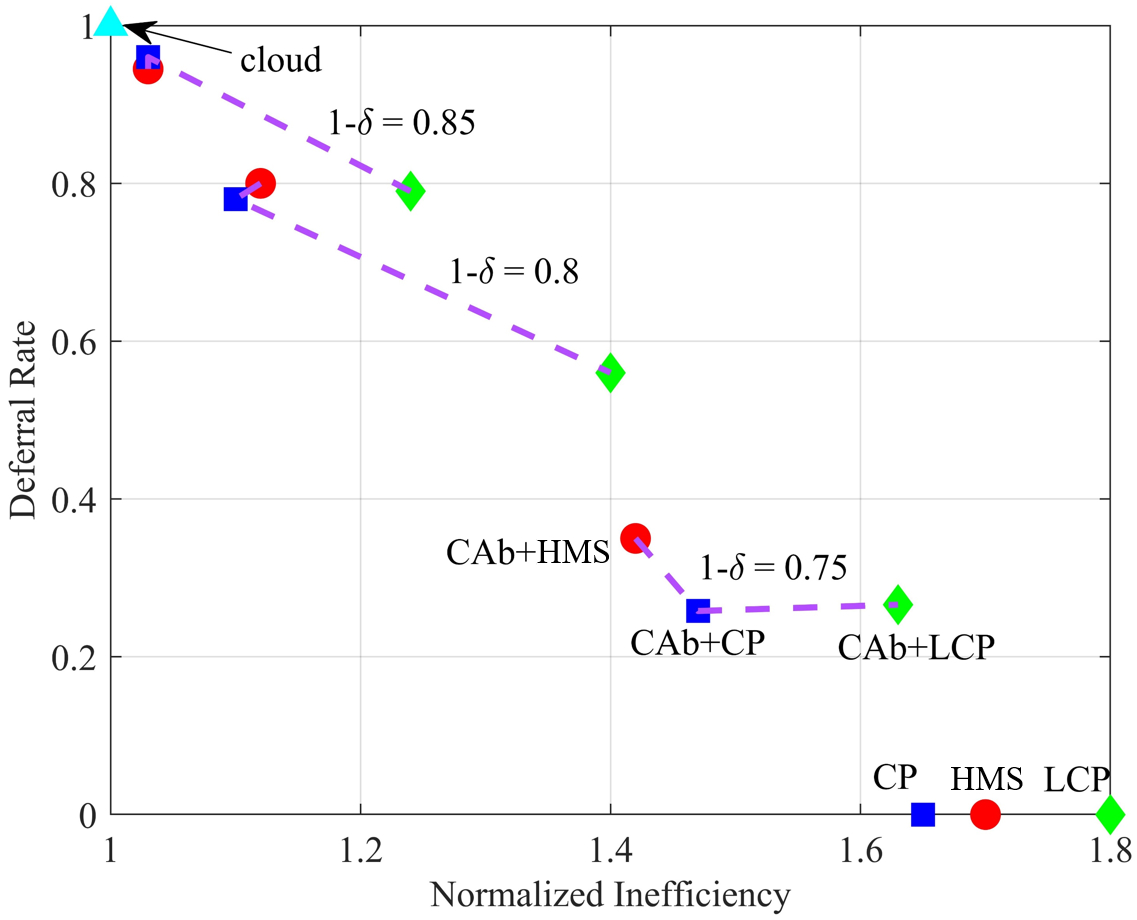}}
    \caption{Deferral rate versus normalized inefficiency obtained by changing the target average satisfaction level $1-\delta$ on TeleQnA dataset for \gls{cab} schemes, namely \gls{cab}+\gls{hms}, \gls{cab}+\gls{cp}, and \gls{cab}+\gls{lcp}. Points on the same dashed line share the same target average satisfaction rate $1-\delta$.}
    \label{fig:c6_qa_pareto} 
\end{figure}

\textbf{Confidence-based versus conformal alignment-based model cascading.} To elaborate, in a manner similar to Fig.~\ref{fig:c6_image_CA}, Fig.~\ref{fig:c6_qa_CA} evaluates average satisfaction rate, deferral rate, and normalized inefficiency for \gls{cbd} and \gls{cab} schemes against the target average satisfaction level in the set $1-\delta \in \{ 0.6, 0.65, \dots, 0.95\}$ with a fixed conditional coverage requirement $1-\alpha_{\text{label}}^{\text{mis}}=0.8$.

Since the edge model is better calibrated, even \gls{cbd} schemes can meet the target average satisfaction requirement (\ref{eq:c6_batch_goal}) in this example. It is emphasized, however, that this is a purely empirical observation, and there is a priori no guarantee that \gls{cbd} schemes would satisfy the condition (\ref{eq:c6_batch_goal}). In contrast, \gls{cab} schemes adapt the overly conservative prediction regions generated by the under-confident edge model into less conservative regions that still provably satisfy the average satisfaction rate requirement (\ref{eq:c6_batch_goal}). Furthermore, \gls{cab} schemes are seen in the figure to obtain far lower deferral rates and only modest increases in normalized inefficiency, as compared to \gls{cbd} methods. For instance, at the fixed target average satisfaction level $1-\delta=0.75$, \gls{cab} methods reduce deferral rate by approximately $60\%$, while incurring a $20\%$ increase in normalized inefficiency.

In terms of the relative performance of different prediction sets, while \gls{lcp} continues to produce the largest prediction sets with the smallest deferral rate, \gls{hms} and \gls{cp} exhibit similar deferral rates and normalized inefficiency levels, especially for a higher target average satisfaction requirement, e.g., $1-\delta \geq 0.8$. This result is expected given that a well-calibrated model generally yields HMS with good marginal coverage guarantees.

\textbf{Trade-off between deferral rate and normalized inefficiency.} To further elaborate on this point, Fig.~\ref{fig:c6_qa_pareto} demonstrates the trade-off between the deferral rate and normalized inefficiency for \gls{cab} schemes by varying the target average satisfaction levels $1-\delta$. The figure confirms that with a better calibrated model, \gls{hms} and \gls{cp} tend to yield similar results in both deferral rate and normalized inefficiency, while \gls{lcp} remains the most conservative solution, producing the largest prediction sets with the lowest deferral rate.

\section{Conclusion} \label{sec:c6_conclusion} 
In this chapter, we have proposed a novel edge-cloud model cascading mechanism producing prediction sets that have the same conditional coverage properties of sets produced at the cloud model only. The proposed method, namely \gls{cab} model cascading, provides statistical guarantees on the average fraction of edge-processed decisions that satisfy cloud-level conditional coverage, while minimizing reliance on cloud resources. This guarantee is achieved by casting the escalation from edge to cloud models as a \gls{mht} problem, where the conditional coverage probability serves as the tailored alignment score. Empirical results demonstrate that \gls{cab} methods exhibit a tunable trade-off among conditional coverage, deferral rate, and set size. For instance, compared to \gls{cbd} schemes, at the fixed target average satisfaction rate $1-\delta=0.75$, \gls{cab} schemes reduce deferral rate by approximately $60\%$, at the cost of $20\%$ increase in set size, while maintaining provable reliability guarantees.

Future research directions may include evaluating the robustness of \gls{cab} schemes under covariate shift \citep{tibshirani2019conformal}, extending the proposed framework to localized conformal alignment \citep{wu2024conditional}, and integrating conformal e-values into the alignment process to offer anytime-valid guarantees for sequential inputs in edge-cloud systems \citep{ramdas2024hypothesis, xu2021unified}. 

\chapter{Conclusions} \label{chapter:7}

\ifpdf
    \graphicspath{{Chapter7/Figs/}{Chapter7/Figs/PDF/}{Chapter7/Figs/}}
\else
    \graphicspath{{Chapter7/Figs/}{Chapter7/Figs/}}
\fi

\section{Summary of Thesis Achievements}
\label{sec:c7_summary}

The deployment of \gls{ai} models in safety-critical applications demands both (\emph{i}) \emph{reliability}, the ability to faithfully report uncertainty, and (\emph{ii}) \emph{efficiency}, the ability to reduce computational overhead and to support deployment on resource-constrained devices. These two requirements are two sides of the same coin: the most reliable methods tend to demand the most computation, while the most efficient models typically sacrifice trustworthy uncertainty quantification. In this thesis, we investigated a series of methods to push the frontier of reliable inference via reliability-efficiency co-design.

In Chapter~\ref{chapter:3}, we studied the problem of enabling \gls{ai} models to know what they know and what they do not know, within the framework of Bayesian learning. Standard Bayesian learning captures epistemic uncertainty via ensembling over the posterior, but model misspecification and approximate inference limit its calibration benefits, and improving \gls{id} calibration often degrades \gls{ood} detection performance. We proposed \gls{scbnn-ocm}, a general framework that enhances variational inference-based Bayesian learning to target both \gls{id} and \gls{ood} calibration. The key insight is that calibration regularization and \gls{ocm} are complementary when combined with selective inference, which filters inputs on which the two objectives conflict. Experiments showed that the proposed \gls{scbnn-ocm} achieves the best \gls{id} and \gls{ood} performance as compared to existing benchmarks.

In Chapter~\ref{chapter:4}, we turned to the question of whether reliability can be formally guaranteed in an efficient way. The training-time cost involved in Chapter~\ref{chapter:3} motivates post-hoc methods that operate on any pre-trained model without retraining. While existing frameworks such as \gls{rcps} control only the expected loss, safety-critical applications often require control over tail behavior. We proposed \gls{oce-rcps}, a framework that can control the tail behavior on the \gls{oce} risk with respect to the calibration data, not merely on average.

In Chapter~\ref{chapter:5}, we addressed the question of how to preserve both certified reliability and informativeness when efficiency forces the use of a small-scale model. Directly applying the post-hoc methods developed in Chapter~\ref{chapter:4} to a weak edge model yields uninformative predictions, while applying them to a powerful cloud model preserves informativeness but reintroduces the computational burden. We proposed \gls{cd-ci}, a low-complexity methodology to calibrate a small-scale edge model prior to deployment by leveraging calibration data generated by a large-scale cloud model. Furthermore, unlike standard Bayesian learning techniques, the ensemble of predictive distributions is obtained through a conformalized credal set. Experiments demonstrated that the proposed \gls{cd-ci} significantly improves calibration performance at low computational cost, making it a reliable and efficient solution for edge \gls{ai} deployment.

Finally, in Chapter~\ref{chapter:6}, we investigated whether the reliability-efficiency frontier can be pushed further beyond what a single calibration-distilled model achieves. Rather than forcing the edge model to handle all inputs, we proposed a novel edge-cloud model cascading mechanism that produces prediction sets with the same conditional coverage properties as those produced by the cloud model. The proposed \gls{cab} method casts the escalation from edge to cloud models as a \gls{mht} problem, providing formal conditional reliability guarantees ensuring that edge-processed decisions satisfy cloud-level conditional coverage, while minimizing reliance on cloud resources. Experiments demonstrated that the proposed \gls{cab} method performs competitively with \gls{sota} methods in terms of the average satisfaction rate and deferral ratio, with only a modest reduction in informativeness.

\section{Open Research Directions}
\label{sec:c7_future}

Many aspects of the reliability-efficiency co-design developed in this thesis could be further explored. In this section, we detail some of these open questions.

\begin{itemize}[leftmargin=*]

\item \textbf{Robustness under distribution shift.} The certified reliability guarantees in Chapters~\ref{chapter:4}, Chapter \ref{chapter:5}, and Chapter \ref{chapter:6} rely on the assumption that calibration and test data are exchangeable. In practice, the data distribution may shift over time due to changing environments, such as medical imaging with varying patient populations or wireless communications with changing channel conditions. Extending the proposed frameworks to handle covariate shift is an important direction. Several lines of work provide promising starting points. For instance, adaptive conformal inference~\citep{gibbs2021adaptive, zaffran2022adaptive} dynamically adjusts the miscoverage level over time to maintain long-run coverage under temporal distribution drift. Integrating these techniques into the post-processing methods developed in this thesis could enable robust \gls{ai} model deployment under distributional shift.

\item \textbf{Anytime-valid guarantees for sequential inputs.} The post-processing calibration methods developed in this thesis assume a batch setting in which the calibration and test datasets are fixed. In many real-world edge-cloud systems, however, test inputs arrive sequentially, and it is desirable to maintain valid reliability guarantees at \emph{any stopping time}, not just after processing the entire batch. \emph{Conformal e-values}~\citep{ramdas2024hypothesis, bashari2023derandomized} and the associated framework of \emph{e-processes}~\citep{xu2021unified} provide a natural tool for this purpose. Unlike p-value-based methods, e-values can be multiplied across time steps and yield anytime-valid guarantees by construction. Integrating e-values into the proposed \gls{cab} cascading mechanism of Chapter~\ref{chapter:6} is a concrete avenue: the sequential screening procedure could be reformulated using e-processes, allowing the edge-cloud system to make deferral decisions one input at a time while maintaining reliability control at every point in time, rather than only after the entire test batch is processed. This would make the cascading framework substantially more practical for streaming deployments involving \gls{llm} and real-time inference services.

\item \textbf{Latent-space information for more efficient inference.} The methods developed in this thesis rely on \emph{output-level} information, such as predictive probabilities and prediction sets, to make reliable and efficient inference. However, the output-level information captures only a compressed summary of the model's internal representation. The \emph{latent space}, i.e., the activations of intermediate layers, contains richer information about the model's confidence, input difficulty, and proximity to training data \citep{gelada2019deepmdp}. Recent work on \emph{latent reasoning} has demonstrated that operating directly in continuous latent space, rather than decoding into discrete tokens, can yield substantial gains in both reasoning quality and inference efficiency \citep{saunshi2025reasoning}. Rethinking from the latent space perspective may potentially boost the escalation power with lower computational overhead, as latent features are available before the full forward pass is completed.

\end{itemize}


\begin{spacing}{0.9}


\bibliographystyle{apalike}
\cleardoublepage
\bibliography{References/references, References/references3, References/references4, References/references5, References/references6}



\end{spacing}


\begin{appendices} 

\chapter{Chapter 3 Supplementary Material}  \label{app:a1}

This appendix provides supplementary material for Chapter \ref{chapter:3}, consisting of five sections. Sec.~\ref{sec:a1_differentiable} demonstrates the generality of the proposed framework by instantiating it with a differentiable calibration-aware regularizer. Ablation studies on \gls{id} calibration, \gls{ood} detection, and selective calibration are conducted to illustrate the impact of hyperparameters, e.g., weight of calibration-aware regularizer, in Sec.~\ref{sec:a1_ablation}. In Sec.~\ref{sec:a1_backbone}, to showcase the general performance gains of the proposed framework, we present the additional evaluations on a different backbone. Sec.~\ref{sec:a1_outlier_score} details the non-parametric outlier score vectors used in the selective calibration module. We refer to Sec.~\ref{sec:a1_experiments} for all the experimental details.

\section{Discussion on Differentiable Calibration Regularizer} \label{sec:a1_differentiable}

As noted in Chapter \ref{chapter:3}, the proposed framework is compatible with any calibration-aware regularizer. The main experiments adopt the \gls{wmmce} \citep{kumar2018trainable}, which uses non-differentiable confidence and correctness scores. In contrast, in this section, we instantiate the framework with a fully differentiable alternative based on a smoothed maximum operator and a differentiable rank function. As shown in Sec.~\ref{sec:a1_differentiable_results}, this variant yields modest \gls{ece} improvements without affecting accuracy, confirming that the framework readily benefits from improved regularizers. The main text in Chapter \ref{chapter:3} retains the \gls{wmmce} for direct comparability with prior work. All qualitative conclusions of Chapter \ref{chapter:3} hold regardless of the choice of regularizer.

\subsection{Differentiable Calibration Measures} \label{sec:a1_diff_cal_measures}

In this section, we review and extend the differentiable approximate \gls{ece} measure, $\mathcal{E}(\theta|\mathcal{D}^\text{tr}_{\text{id}})$, introduced in \citep{kumar2018trainable} with the name \emph{\gls{wmmce}}. As discussed in the Chapter \ref{chapter:3}, this measure can be used in the learning objectives (\ref{eq:c3_ca_fnn}) and (\ref{eq:c3_CA-BNN_gen_recipe}) for CFNN and \gls{cbnn}, respectively. After reviewing the \gls{wmmce} score in \citep{kumar2018trainable}, we introduce an extension of \gls{wmmce} that may potentially improve calibration performance by leveraging an alternative differentiable metric introduced in \citep{bohdal2021meta}.

\emph{1) Weighted \gls{mmce}: }In \citep{kumar2018trainable}, the \gls{wmmce} metric was defined as an estimate of the \gls{ece} (\ref{eq:c2_ece}). To introduce it, define as $\kappa(\cdot,\cdot)$ a kernel function operating on scalar inputs, such as $\kappa(x_1,x_2) = \text{exp}(-|x_1-x_2|/h)$ for some $h>0$. Using the training dataset $\mathcal{D}^\text{tr}_{\text{id}}$, the \gls{wmmce} first computes confidence scores $\{r_i\}_{i=1}^{|\mathcal{D}^\text{tr}_{\text{id}}|}$ and correctness scores $\{c_i\}_{i=1}^{|\mathcal{D}^\text{tr}_{\text{id}}|}$ as defined in (\ref{eq:c2_confidence_score}) and (\ref{eq:c2_correctness_score}), respectively, with the parametric classifier $p(y|x,\theta)$ in lieu of the trained classifier $p(y|x, \mathcal{D}^{\text{tr}})$ as discussed in Sec.~\ref{sec:c3_ Calibration-Regularized}. Then, the \gls{wmmce} evaluates the metric
\begin{align} \label{eq:a1_wmmce}
     \mathcal{E}(\theta|\mathcal{D}^\text{tr}_{\text{id}}) = \Bigg(\sum_{i,j:{c}_i = {c}_j = 0} \frac{{r}_i {r}_j \kappa({r}_i , {r}_j)}{(|\mathcal{D}^\text{tr}_{\text{id}}|-n_c)(|\mathcal{D}^\text{tr}_{\text{id}}|-n_c)}  &+ \sum_{i,j:{c}_i = {c}_j = 1} \frac{(1-{r}_i) (1-{r}_j) \kappa({r}_i , {r}_j)}{n_c ^2}
      \nonumber \\ &-2 \sum_{i,j:{c}_i = 1,  {c}_j = 0} \frac{(1-{r}_i) {r}_j \kappa({r}_i , {r}_j)}{(|\mathcal{D}^\text{tr}_{\text{id}}|-n_c)n_c} \Bigg)^{\frac{1}{2}},
\end{align}
where $n_c = \sum_{i = 1}^n {{c}}_{i}$ is the number of correct examples. The sums in (\ref{eq:a1_wmmce}) are extended over all examples in training dataset $\mathcal{D}^\text{tr}_{\text{id}}$.

\emph{2) Gradient of the Weighted \gls{mmce}: }In order to evaluate the gradient $\nabla_{\theta}\mathcal{E}(\theta|\mathcal{D}^\text{tr}_{\text{id}})$ required for both CFNN and \gls{cbnn}, one needs to calculate the gradients $\nabla_{\theta} r_i$ and $\nabla_{\theta} c_i$ of the confidence and correctness scores, respectively. To simplify this calculation, reference \citep{kumar2018trainable} implicitly ignored the dependence of the point classification decision $\hat{y}(x)$ in (\ref{eq:c2_hard_decision_fnn}) on the model parameter $\theta$. Accordingly, the gradient of the correctness score was set to zero; and the gradient $\nabla_{\theta} r_i$ was evaluated as $\nabla_{\theta} p (\hat{y}(x) | x, \theta)$, where $\hat{y}(x)$ is treated as a constant. This approximation of the gradient is motivated by the non-differentiable nature of the decision $\hat{y}(x) = \arg \max_{y \in \mathcal{Y}} p(y | x, \theta)$ with respect to $\theta$. 

As a discussion, we propose potentially more accurate estimators of the gradient of the confidence and correctness scores, replacing the non-differentiable maximum operator in (\ref{eq:c2_hard_decision_fnn}) with a differentiable \emph{smoothed maximum} operator \citep{cuturi2019differentiable}.

Specifically, the \emph{differentiable confidence score} is defined by taking the smoothed maximum among all the available candidate confidence scores $\{p (y |x_i,\theta)\}_{y \in \mathcal{Y}}$ as
\begin{align} \label{eq:a1_rhat}
    \hat{r_i} = \sum_{y \in \mathcal{Y}} p (y |x_i,\theta) \frac{e^{p (y |x_i,\theta) / \tau_r}}{\sum_{y' \in \mathcal{Y}} e^{p (y' |x_i,\theta) / \tau_r}},
\end{align}
with temperature parameter $\tau_r > 0$ controlling the smoothness of the approximation (\ref{eq:a1_rhat}). In the limit $\tau_r \rightarrow 0$, the differentiable confidence score $\hat{r_i}$ recovers the true score $r_i$. Furthermore, we adopt the \emph{differentiable correctness score} \citep{bohdal2021meta}
\begin{align}
    \label{eq:a1_chat}
    \hat{c_i} = \text{ReLU} (2 - [\hat{R}(x_i)]_{y_i})
\end{align}
with $\text{ReLU}(a)=\max(0,a)$ and  differentiable rank function
\begin{align} \label{eq:a1_Rhat}
    [\hat{R}(x)]_{y} = 1 + \sum_{y^{\prime} \in \mathcal{Y}, y \neq y^{\prime}} \frac{1}{1 + e^{S_{y, y^{\prime}} / \tau_c}},
\end{align}
with $S_{y, y^{\prime}} = p(y |x, \theta) - p(y^{\prime} |x, \theta)$. With small enough temperature parameter $\tau_c > 0$, the differentiable correctness score $\hat{c_i}$ recovers the true correctness score $c_i$ in (\ref{eq:c2_correctness_score}). 

By replacing $r_i, r_j$ and $c_i, c_j$ in (\ref{eq:a1_wmmce}) with $\hat{r}_i, \hat{r}_j$ and $\hat{c}_i$, $\hat{c}_j$ via (\ref{eq:a1_rhat}) and (\ref{eq:a1_chat}), respectively for $i,j=1,...,|\mathcal{D}^\text{tr}_{\text{id}}|$, one obtains a \emph{fully differentiable} version of $\mathcal{E}(\theta|\mathcal{D}^\text{tr}_{\text{id}})$, which can be differentiated with respect to the model parameter $\theta$.

\subsection{Experiments}
\label{sec:a1_differentiable_results}


\begin{figure}[tb]
    \centering
    \centerline{\includegraphics[scale=0.35]{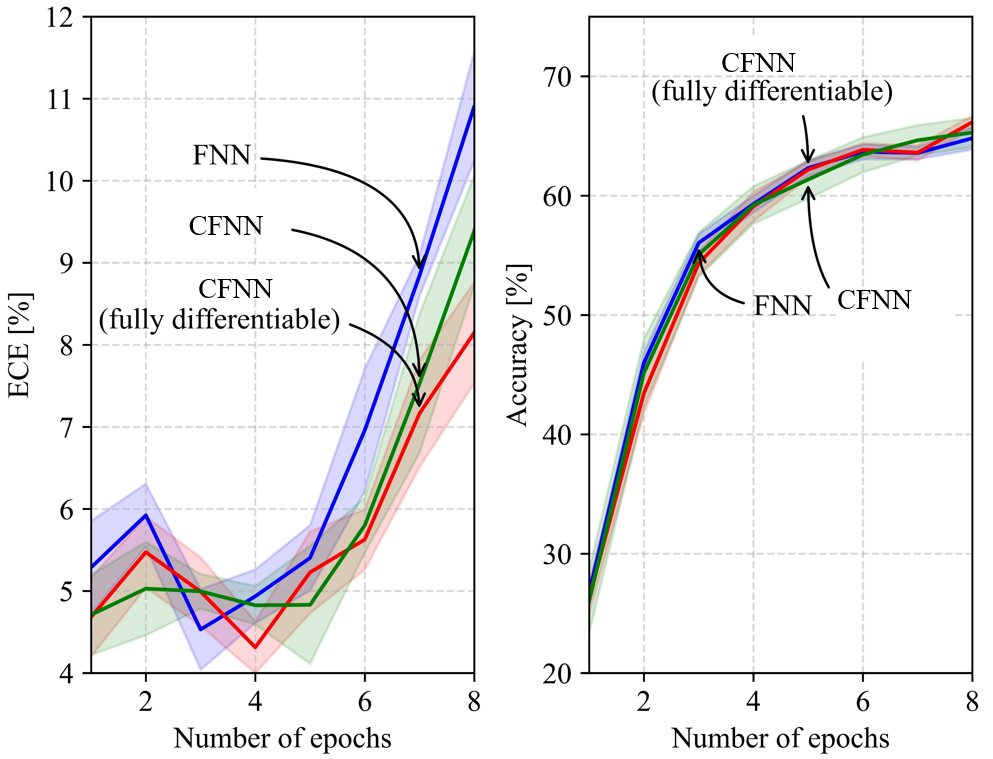}}
    \caption{\gls{ece} and accuracy as a function of number of epochs for 20 Newsgroups classification task for FNN, CFNN \citep{kumar2018trainable}, and the modified CFNN with fully differentiable batch regularizer as discussed in Sec.~\ref{sec:a1_diff_cal_measures}. The shaded areas correspond to  $75\%$  intervals of the realized values.}
    \label{fig:a1_differentiable}
\end{figure}

In this section, we compare the calibration performance and the accuracy of conventional FNN and BNN, as well as CFNN \citep{kumar2018trainable} and \gls{cbnn} (ours). We validate the proposed differentiable \gls{wmmce} on the 20 Newsgroups dataset \citep{lang1995newsweeder}.

For the 20 Newsgroups classification task, as in \citep{lin2013network}, we adopt a convolutional NN with global pooling for all schemes. For Bayesian learning, i.e., for BNN and \gls{cbnn}, we choose $\beta=0.1$ in (\ref{eq:c3_CA-BNN_gen_recipe}), with zero-mean Gaussian prior $p(\theta)$ with standard deviation $0.05$. The temperature parameters in (\ref{eq:a1_rhat}) and (\ref{eq:a1_Rhat}) are set to $\tau_r = 0.001$ and $\tau_c = 0.01$; and we set $h=0.4$ for the \gls{wmmce} kernel. We use the RMSprop optimizer with learning rate $0.002$.

We examine the impact of the differentiable confidence and correctness scores introduced in (\ref{eq:a1_rhat}) and (\ref{eq:a1_chat}). Fig.~\ref{fig:a1_differentiable} shows \gls{ece} and accuracy as a function of number of training epochs for (\emph{i}) FNN; (\emph{ii}) CFNN in (\ref{eq:c3_ca_fnn}); and (\emph{iii}) CFNN with the fully differentiable batch regularizer discussed in Sec.~\ref{sec:a1_diff_cal_measures}. We fix the weight of calibration-aware regularizer $\gamma_{\text{id}}$ to $10$. It is observed that the calibration performance in terms of \gls{ece} is enhanced by the proposed fully differentiable calibration-aware regularizer, while not affecting the accuracy.


\section{Ablation Study} \label{sec:a1_ablation}

\subsection{\gls{id} Calibration}
\begin{figure} [htb] 
    \centering
    \centerline{\includegraphics[scale=0.25]{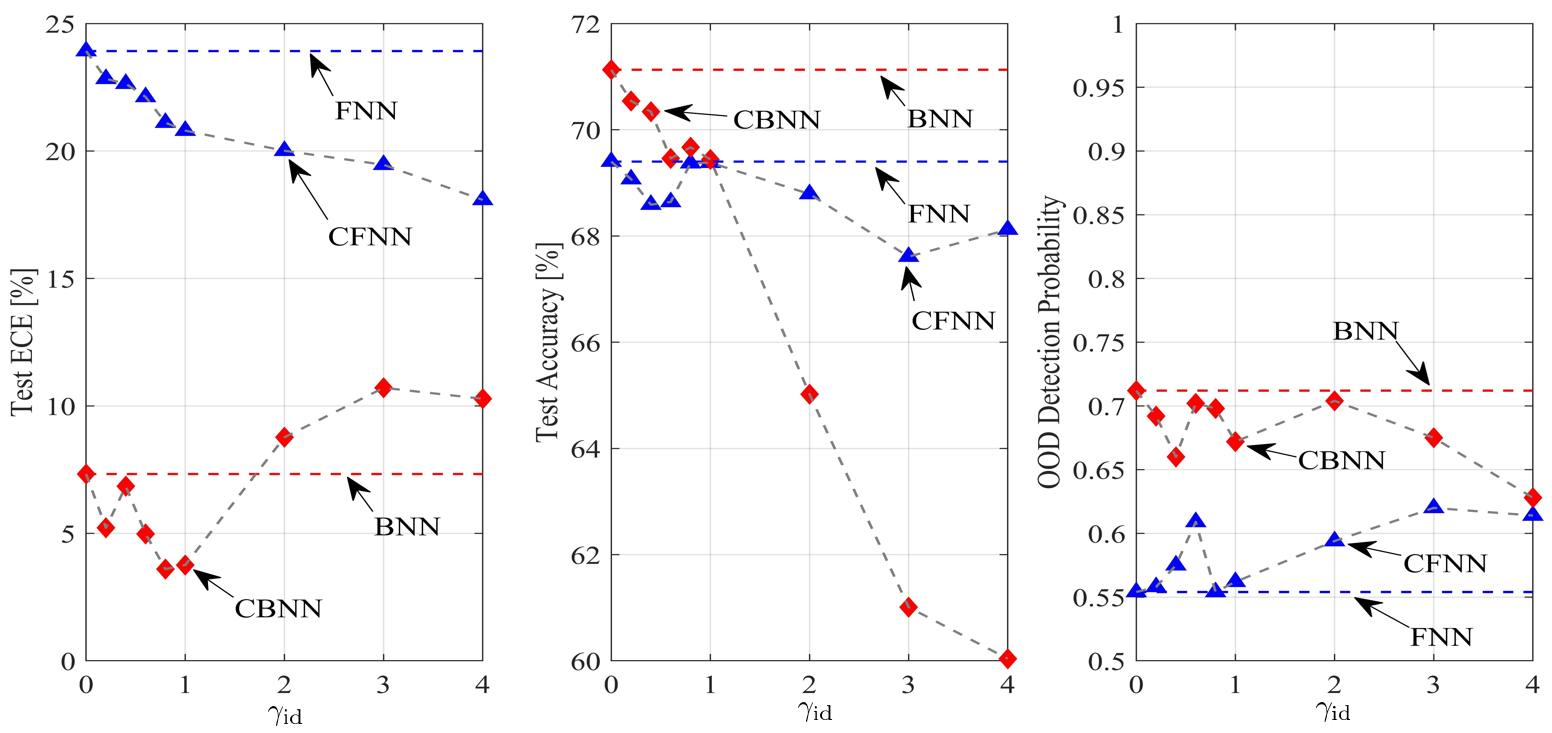}}
    \caption{\gls{ece}, accuracy and \gls{ood} detection probability as a function of hyperparameter $\gamma_{\text{id}}$ on CIFAR-100 (\gls{id}) and TinyImageNet (\gls{ood}) dataset for FNN, BNN, CFNN, and \gls{cbnn}.}
    \label{fig:a1_CA_1} 
\end{figure}

The impact of the calibration-based regularizer hyperparameter, $\gamma_{\text{id}}$ in (\ref{eq:c3_ca_fnn}) and (\ref{eq:c3_CA-BNN_gen_recipe}), for \gls{id} calibration is analyzed in Fig.~\ref{fig:a1_CA_1} under the same conditions as for Fig.~\ref{fig:c3_CA-pareto}. The figure confirms that the best performance for CFNN and \gls{cbnn} on CIFAR-100 (\gls{id}) and TinyImageNet (\gls{ood}) datasets is achieved for $\gamma_{\text{id}} = 4$ and $\gamma_{\text{id}} = 0.8$, respectively. \gls{cbnn} scheme seems to be more sensitive to the selection of hyperparameters $\gamma_{\text{id}}$, possibly because of the interplay with the choice of the prior distribution in the Bayesian learning objective (\ref{eq:c3_CA-BNN_gen_recipe}). That said, \gls{cbnn} outperforms BNN in terms of \gls{ece} for a wide range of values of hyperparameter $\gamma_{\text{id}}$.

\subsection{\gls{ood} Detection}
\begin{figure} [htb] 
    \centering
    \centerline{\includegraphics[scale=0.25]{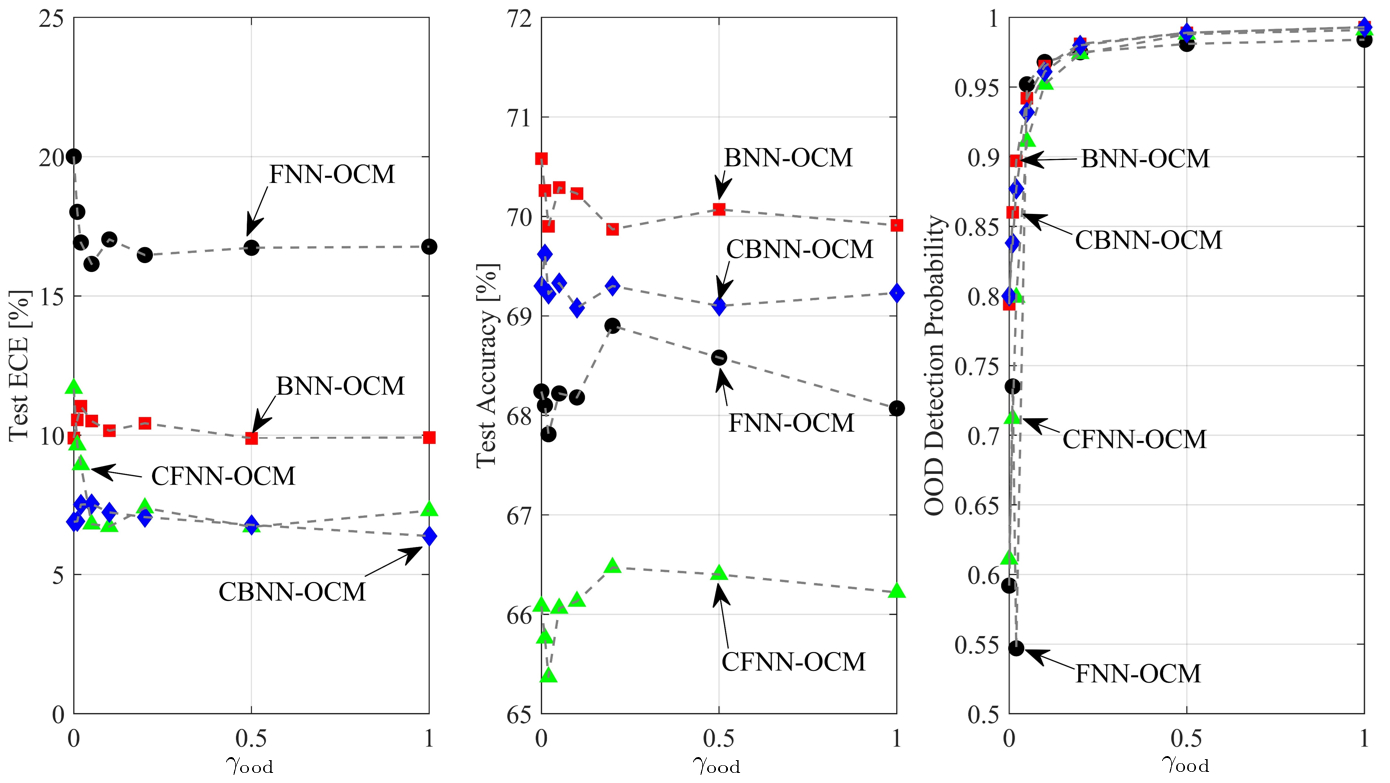}}
    \caption{\gls{ece}, accuracy, and \gls{ood} detection probability as a function of hyperparameter $\gamma_{\text{ood}}$ on CIFAR-100 (\gls{id}) and TinyImageNet (\gls{ood}) dataset for FNN-OCM, BNN-OCM, CFNN-OCM, and \gls{cbnn}-OCM.}
    \label{fig:a1_CM_1} 
\end{figure}

Fig. \ref{fig:a1_CM_1} presents an ablation study for the \gls{ocm} regularizer weight,  $\gamma_{\text{ood}}$ in (\ref{eq:c3_CM}) and (\ref{eq:c3_general_CM_gen_recipe}), for the same setting as in the previous subsection. The \gls{ece}, test accuracy, and \gls{ood} detection probability are relatively stable when $\gamma_{\text{ood}} > 0.1$, which implies that our default choice of $\gamma_{\text{ood}} = 0.5$, also reported in \citep{choi2023conservative}, is well suited for all schemes.

\subsection{Selective Calibration}
\begin{figure} [htb] 
    \centering
    \centerline{\includegraphics[scale=0.25]{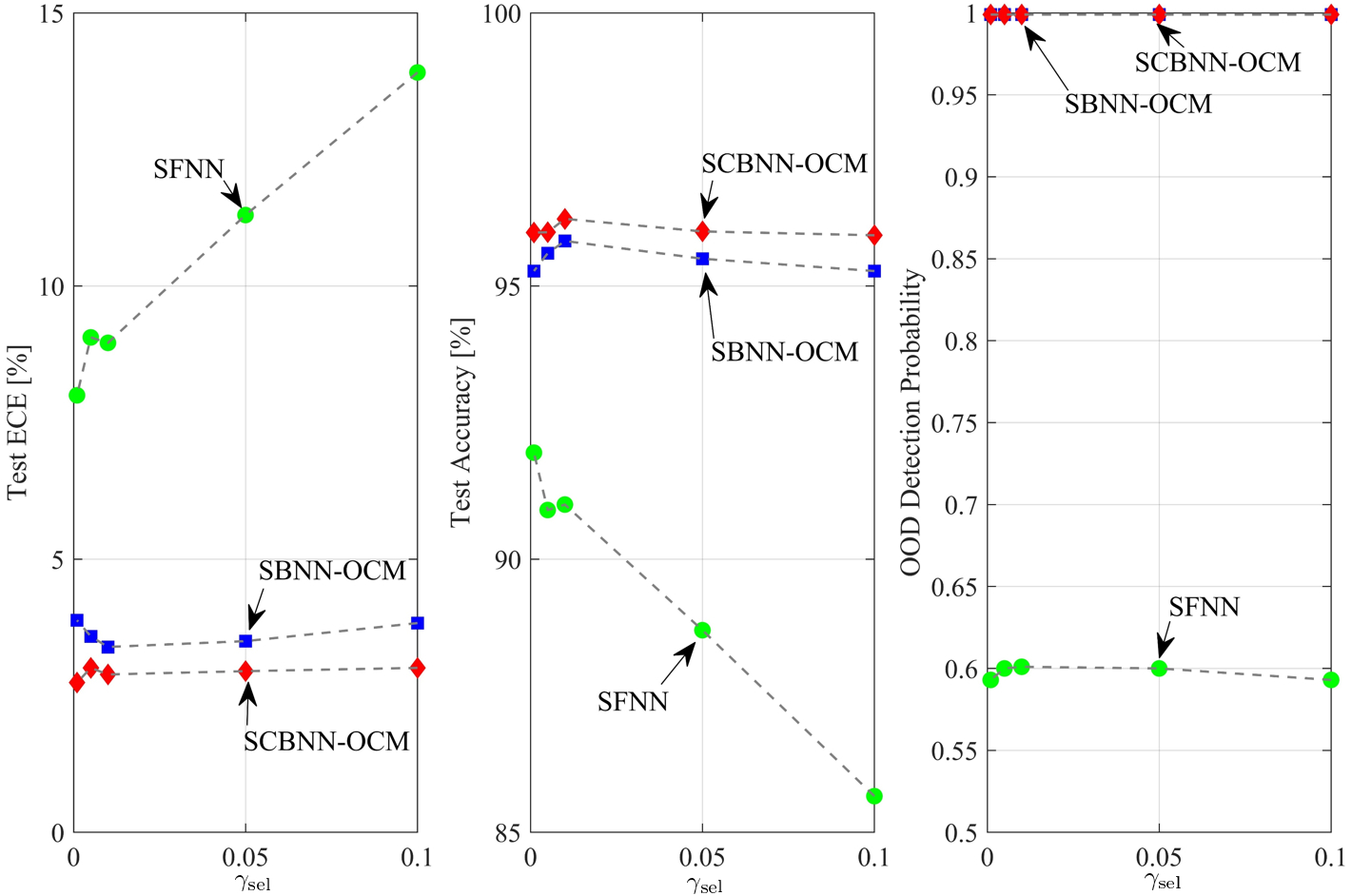}}
    \caption{\gls{ece}, accuracy and \gls{ood} detection probability as a function of hyperparameter $\gamma_{\text{sel}}$ on CIFAR-100 (\gls{id}) and TinyImageNet (\gls{ood}) dataset for SFNN, \gls{sbnn-ocm}, and \gls{scbnn-ocm} with \gls{id} coverage rate as $40\%$.}
    \label{fig:a1_SEL_1} 
\end{figure}

By varying the selector regularizer weight, $\gamma_{\text{sel}}$ in (\ref{eq:c3_calibration_selector}) and (\ref{eq:c3_selector_gen_gen_recipe}), Fig.~\ref{fig:a1_SEL_1} shows that the proposed Bayesian methods have a better performance in terms of all metrics as compared to SFNN, and are more robust to the choice of hyperparameter $\gamma_{\text{sel}}$. To maintain consistency with \citep{fisch2022calibrated}, we set $\gamma_{\text{sel}} = 0.01$ as the default choice for $\gamma_{\text{sel}}$.

\section{Backbone Study} \label{sec:a1_backbone}
In order to showcase that the gain of \gls{scbnn-ocm} is not restricted to a particular NN architecture or to a specific \gls{id}/\gls{ood} dataset, we present additional evaluations on a different backbone, namely ResNet-18, and on different datasets by considering CIFAR-10 as the \gls{id} and LSUN resized as the \gls{ood} dataset. For all the experiments in this section, we use the same settings as the experimental results in Chapter \ref{chapter:3}.
    
\subsection{\gls{id} Calibration}
\begin{figure} [htb] 
    \centering
    \centerline{\includegraphics[scale=0.25]{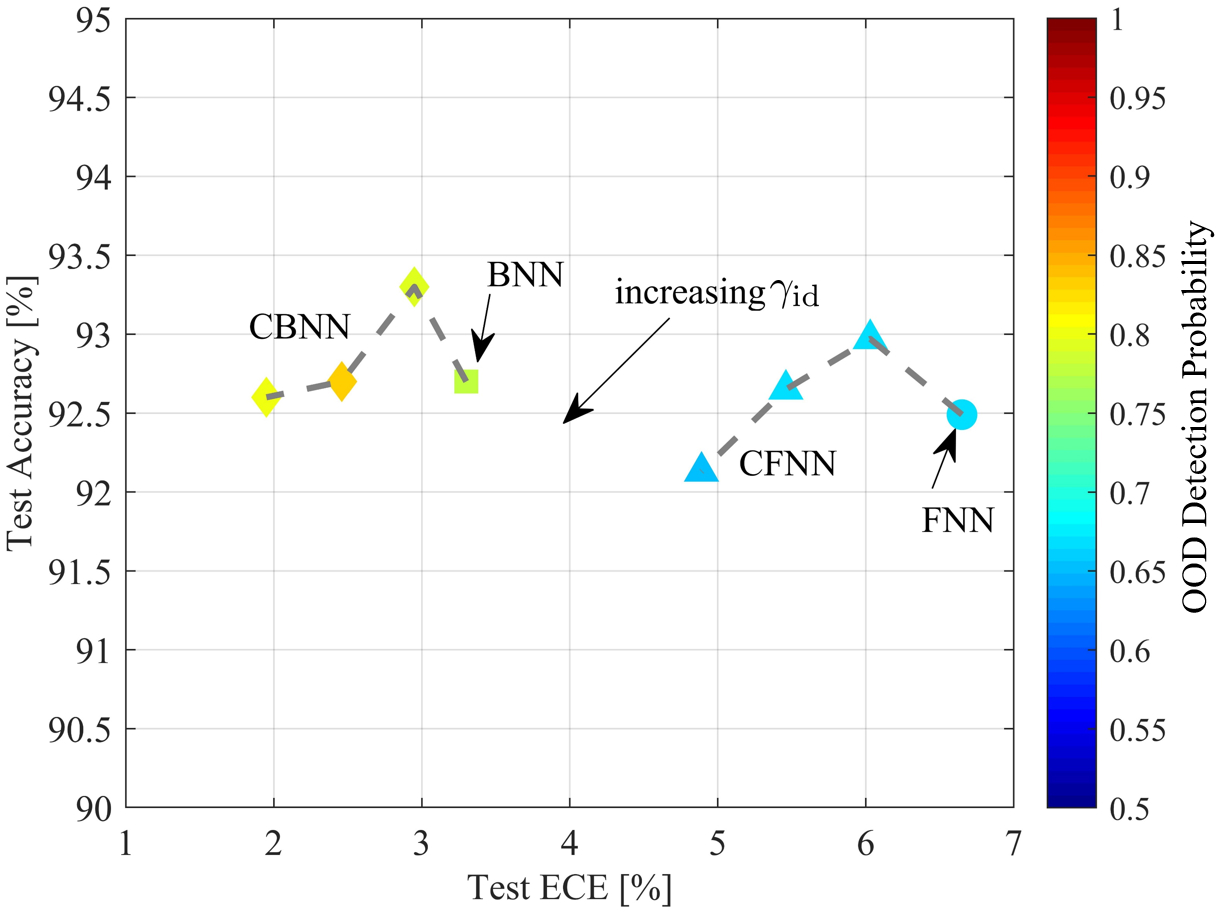} }
    \caption{Accuracy versus \gls{ece} obtained by changing the hyperparameter $\lambda$ on CIFAR-10 dataset for FNN, CFNN (benchmark), BNN, and \gls{cbnn} (ours), with \gls{ood} detection probability indicated by the marker's color.} 
    \label{fig:a1_backbone_1}  
\end{figure}

The impact of the calibration-based regularization hyperparameter for \gls{id} calibration, $\gamma_{\text{id}}$ in (\ref{eq:c3_ca_fnn}) and (\ref{eq:c3_CA-BNN_gen_recipe}), is analyzed in Fig.~\ref{fig:a1_backbone_1}. The figure confirms that the best performance for CFNN and \gls{cbnn} on CIFAR-10 (\gls{id}) and LSUN (\gls{ood}) datasets is achieved for $\gamma_{\text{id}} = 10$ and $\gamma_{\text{id}} = 4$, respectively. Also, increasing the value of $\gamma_{\text{id}}$ is seen to decrease the \gls{ece} for both CFNN and \gls{cbnn}, with similar test accuracy.
    
\subsection{\gls{ood} Detection}
\begin{figure} [htb] 
    \centering
    \centerline{\includegraphics[scale=0.25]{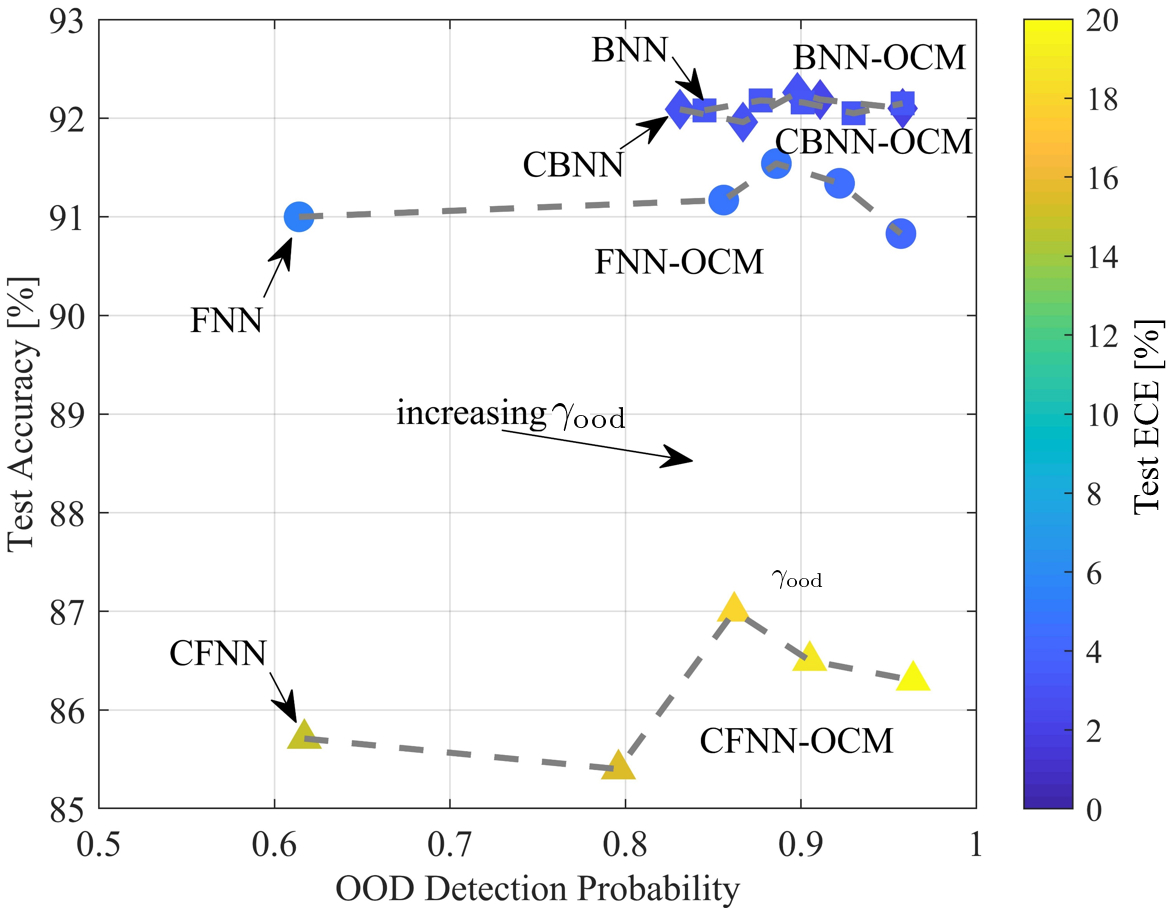} }
    \caption{Test accuracy versus \gls{ood} detection probability on CIFAR-10 (\gls{id}) and LSUN (\gls{ood}) for FNN-OCM (benchmark), CFNN-OCM, BNN-OCM, and CBNN-OCM (ours), with test \gls{ece} as the marker's color. Note that hyperparameter $\gamma = 0$ recovers the original schemes without \gls{ocm} regularizer.} 
    \label{fig:a1_backbone_2}  
\end{figure}

To investigate the impact of \gls{ocm} regularizer on \gls{id} and \gls{ood} performance, Fig.~\ref{fig:a1_backbone_2} plots the \gls{id} and \gls{ood} performance by varying the \gls{ocm} hyperparameter $\gamma_{\text{ood}}$ in (\ref{eq:c3_CM}) and (\ref{eq:c3_general_CM_gen_recipe}). Note that the deterioration of the \gls{id} performance, especially for CFNN-OCM, shows the side effects of introducing \gls{ocm}, although it entails \gls{ood} detection probability improvement. Additionally, the proposed CBNN-OCM better balances the \gls{id} and \gls{ood} performance levels as compared to CFNN-OCM.

\subsection{Selective Calibration}

\begin{figure} [htb] 
    \centering
    \centerline{\includegraphics[width=\textwidth]{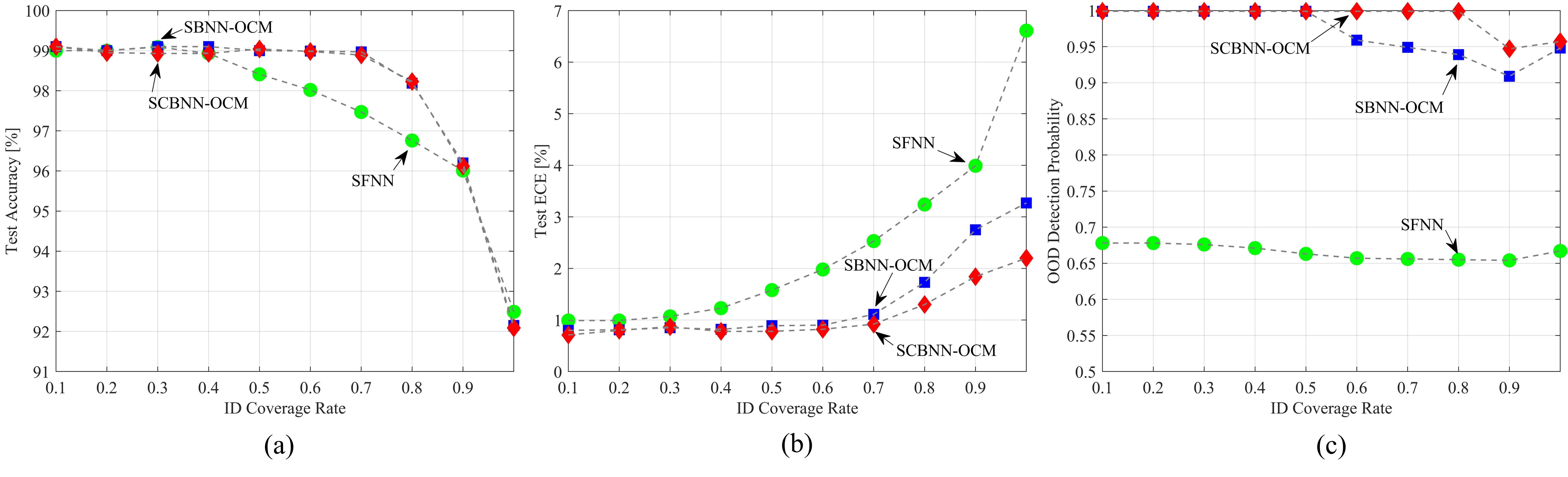}}
    \caption{Test accuracy, \gls{ece}, and \gls{ood} detection probability versus \gls{id} coverage rate on CIFAR-10 (\gls{id}) and LSUN (\gls{ood}) for SFNN (benchmark), \gls{sbnn-ocm}, and \gls{scbnn-ocm} (ours).}
    \label{fig:a1_SEL-cal} 
\end{figure}

To investigate the impact of selective calibration, Fig.~\ref{fig:a1_SEL-cal} plots the \gls{id} and the \gls{ood} performance for different \gls{id} coverage rates ranging from $0.1$ to $1$. The figure shows that \gls{scbnn-ocm} outperforms \gls{sbnn-ocm} in terms of \gls{ece} and \gls{ood} detection probability in different \gls{id} coverage rate regimes, while attaining similar test accuracy.
    
As compared to Fig.~\ref{fig:c3_SEL-cal}, \gls{scbnn-ocm} is seen to achieve better \gls{id} and \gls{ood} performance with a relatively small reduction in terms of \gls{id} coverage rate. For instance, with an \gls{id} coverage rate of $0.7$, the \gls{id} accuracy is around $99\%$, the \gls{ece} is lower than $1$, and the \gls{ood} detection probability is nearly $1$.

\section{Non-Parametric Outlier Score Vectors} \label{sec:a1_outlier_score}
In this section, we specify the details for non-parametric outlier score vector $ s (x|\theta^\text{FNN})$ described in Sec.~\ref{sec:c3_background_sel_ca}, which can be implemented by following the sklearn package in Python. As mentioned in Sec.~\ref{sec:c3_background_sel_ca}, the first entry is given by (\ref{eq:c3_KDE}) with the Gaussian kernel $\kappa \left( || z-z^{\text{tr}}_{i} || \right) = \exp \left( - || z - z_i^{\text{tr}} ||^2/h \right)$ given the bandwidth parameter $h>0$.
   
The isolation forest score for $x$, $s_2 (x|\theta^\text{FNN})$, is obtained by evaluating the \emph{depth} of $z$ in the \emph{isolation forest}, which is a collection of $T$ binary trees in which each binary tree is constructed so that every element in $\{ z_i^\text{tr} \}_{i=1}^{|\mathcal{X}^\text{tr}|} \cup \{z\}$ is assigned to a leaf node. Specifically, the isolation forest score is evaluated as \citep{liu2008isolation}
\begin{align}
   s_2 (x|\theta^\text{FNN}) = 2^{-\frac{\sum_{i=1}^T h_i(z)}{T\cdot c}}, \label{eq:isolation_forest}
\end{align}
where $h_i(z)$ is the depth of $z$ in the $i$-th binary tree given a normalization constant $c$. 

The one-class support vector machine score for $x$, $s_3 (x|\theta^\text{FNN})$, can be expressed as \citep{scholkopf2001estimating}
\begin{align}
   s_3 (x|\theta^\text{FNN}) = \ \sum_{i=1}^{|\mathcal{X}^{\text{tr}}|} \alpha_i \kappa \left( || z-z^{\text{tr}}_{i} || \right) - \rho ,
\end{align}
in which the coefficients $\alpha_i \geq 0$ for $i=1,...,|\mathcal{X}^\text{tr}|$, satisfying $\sum_{i=1}^{|\mathcal{X}^\text{tr}|} \alpha_i=1$, are optimized along with the offset parameter $\rho \in \mathbb{R}$ in order to separate the training data $\{ z_i^\text{tr} \}_{i=1}^{|\mathcal{X}^\text{tr}|}$ from the origin in the feature space associated with the kernel $\kappa(\cdot)$ \citep[Sec. 5]{scholkopf2001estimating}.

Finally, $k$-nearest neighbor distance for $x$, $s_4 (x|\theta^\text{FNN})$, can be written as \citep{loftsgaarden1965nonparametric}
\begin{align}
    s_4 (x|\theta^\text{FNN}) = || z - z_i^{\text{tr}} ||_{(k)},
\end{align}
where $|| z - z_i^{\text{tr}} ||_{(k)}$ is the $k$-th smallest value in the set of distances $\{|| z - z_i^{\text{tr}} ||\}_{i=1}^{|\mathcal{X}^{\text{tr}}|}$.

\section{Experiment Details} \label{sec:a1_experiments}

In this section, we specify the implementation details for the experimental results in Sec.~\ref{sec:c3_results}. All the experiments are implemented by PyTorch \citep{paszke2019pytorch} and run over a GPU server with a single NVIDIA A100 card.

\subsection{Architecture and Training Details for \gls{id} Calibration}\label{subsec:cal}

\emph{1) Architecture: } For all the experiments related to calibration-regularized learning, we adopt the WideResNet-40-2 architecture \citep{zagoruyko2016wide}.

\emph{2) Hyperparameters:} For fair comparison, we use the same training policy for both frequentist and Bayesian learning. Specifically, we use the \gls{sgd} optimizer with momentum factor $0.9$ and train the model during $100$ epochs. We also decrease the learning rate by dividing its value by a factor of $5$ for every $30$ epochs with an initial learning rate $0.1$ in a manner similar to \citep{choi2023conservative}.

The minibatch size used for single \gls{sgd} update is set to $128$.  For Bayesian learning, we set as prior $p(\theta)$ the Gaussian distribution that has zero-mean vector with diagonal covariance matrix having each element as $0.001$. The hyperparamter $\beta$ for the free energy (\ref{eq:c2_free_energy}) is set to $0.00035$. The ensemble size for Bayesian learning during training is set to $1$, while the ensemble size during testing is set to $20$.

In a manner similar to \citep{yoon2023esd}, the weight of calibration-based regularizer $\gamma_{\text{id}}$ in (\ref{eq:c3_ca_fnn}) and (\ref{eq:c3_CA-BNN_gen_recipe}) is chosen as the value in set $\{0.2, 0.4, 0.6, 0.8, 1.0, 2.0, 3.0, ..., 10.0\}$ that achieves the lowest \gls{ece}, while preserving the accuracy drop no larger than $1.5\%$ as compared to a setting with $\gamma_{\text{id}}=0$. The corresponding accuracy and \gls{ece} are evaluated based on the validation dataset $\mathcal{D}^\text{val}$, and the weights $\gamma_{\text{id}}$ for CFNN and for \gls{cbnn} are chosen independently, which result in the values of $4$ and $0.8$ respectively in our experiments. With these choices, CFNN and \gls{cbnn} can achieve lower \gls{ece} with acceptable accuracy drops. We use \gls{wmmce} regularizer for calibration-aware regularization term $\mathcal{E}(\theta|\mathcal{D}^\text{tr}_{\text{id}})$, which is defined in (\ref{eq:a1_wmmce}).


\emph{3) Dataset Split and Augmentations: } As mentioned in Sec.~\ref{subsec:c3_setting_and_metrics}, we choose CIFAR-100 dataset \citep{krizhevsky2010cifar} for the \gls{id} samples, which is a dataset composed of $60,000$ images each with label information chosen among $100$ different classes. In particular, CIFAR-100 splits the dataset into two parts: $50,000$ for training and $10,000$ for testing; and we further split the training dataset into $45,000$ examples and $5,000$ examples to define the training dataset $\mathcal{D}^\text{tr}_{\text{id}}$ and the validation dataset $\mathcal{D}^\text{val}$. We adopt the standard random flip and random crop augmentations provided by Pytorch \citep{paszke2019pytorch} during the training process.

\subsection{Architecture and Training Details for \gls{ood} Detection}  \label{sec:a1_OOD-detection-settings} 
\emph{1) Architecture: } Since we use the same predictor for \gls{ood} detection, the corresponding architecture remains the same, i.e., WideResNet-40-2, as described above.

\emph{2) Hyperparameters: } Following the original \gls{ocm} paper \citep{choi2023conservative}, \gls{ood} confidence minimization (\ref{eq:c3_CM}) and (\ref{eq:c3_general_CM_gen_recipe}) is carried out by fine-tuning based on the corresponding pre-trained models, e.g., CBNN-OCM is obtained via fine-tuning with the \gls{ocm}-regularized training loss given the pre-trained \gls{cbnn}. Uncertainty dataset $\mathcal{D}^\text{unl}_{\text{ood}}$ is constructed by randomly choosing $6,000$ input data from TinyImageNet. During fine-tuning, \gls{sgd} optimizer with momentum factor of $0.9$ is adopted during $10$ epochs, each consisting $282$ iterations. The initial learning rate is set to $0.001$, and we update the learning rate for each iteration by following \citep{hendrycks2019oe}. Specifically, during \gls{sgd} training, for each epoch, we first sample $9,000$ examples from $\mathcal{D}^\text{tr}_{\text{id}}$ and evaluate the \gls{id}-related training measures for each \gls{sgd} update via sampling $32$ examples (minibatch size being $32$) without replacement among the $9,000$ examples; for the \gls{ood}-related training measures, we sample $64$ examples (minibatch size being $64$) among $\mathcal{D}^\text{unl}_{\text{ood}}$ with replacement. The hyperparameter of \gls{ocm} regularizer $\gamma_{\text{ood}}$ in (\ref{eq:c3_CM}) and (\ref{eq:c3_general_CM_gen_recipe}) in the main text is chosen as in reference \citep{choi2023conservative}, i.e, $\gamma_{\text{ood}} = 0.5$, which increases \gls{ood} detection probability nearly to $1$ for all \gls{ocm} schemes. Other hyperparameters are the same as calibration-regularized learning described in the previous subsection.

\emph{3) Dataset Split and Augmentations: } TinyImageNet dataset \citep{liang2017principled} contains $10,000$ images that corresponds to different $200$ classes. We split the input data of TinyImageNet dataset into two parts, $6,000$ and $4,000$, and use them for the uncertainty dataset $\mathcal{D}^\text{unl}_{\text{ood}}$ and for the \gls{ood} test dataset, respectively. During fine-tuning, we apply the same standard random flip and random crop augmentations to both $\mathcal{D}^\text{tr}_{\text{id}}$ and $\mathcal{D}^\text{unl}_{\text{ood}}$.

\subsection{Architecture and Training Details for Selective Calibration}  \label{appendix:selector-settings} 

\emph{1) Architecture: } For the selector implementation, we use a $3$-layer feed-forward neural network with $64$ neurons in each hidden layer, activated by ReLU. 

\emph{2) Hyperparameters: } For selector training (\ref{eq:c3_calibration_selector}) and (\ref{eq:c3_selector_gen_gen_recipe}), we use the Adam optimizer, with learning rate $0.001$ and weight decay coefficient $10^{-5}$. We train the model for $5$ epochs, each epoch consisting $50,000$ iterations, and each iteration samples $32$ examples (minibatch size being $32$) among $\mathcal{D}^\text{val}$. Since the proposed \gls{scbnn-ocm} is more robust to the choices of $\gamma_{\text{sel}}$, the hyperparameter $\gamma_{\text{sel}}$ in (\ref{eq:c3_calibration_selector}) and (\ref{eq:c3_selector_gen_gen_recipe}) is set to $0.01$ as reported in \citep{fisch2022calibrated}, and we set the kernel function in (\ref{eq:c3_sel-cal-loss-bayes}) as $ \kappa(r_i, r_j) = \exp (-||r_i - r_j||/0.2)$. We use the same ensemble size for training and testing as in calibration-regularized learning. 

\emph{3) Dataset Split and Augmentations: } As described above, $\mathcal{D}^\text{val}$ has $5,000$ examples obtained from CIFAR-100 dataset. We utilize the standard random flip and random crop augmentations during selector training. 

\chapter{Chapter 6 Supplementary Material}  \label{app:a2}
This appendix provides supplementary material for Chapter \ref{chapter:6}, consisting of the proof of Proposition 6.1. 

\section{Proof of Proposition 6.1}

Given the output $\mathcal{S}= \mathcal{D}_{(k_{\text{CA}})}^{\text{uns,te}}$ of the \gls{cab} model cascading methodology, by the definition of the \gls{fdr} in (\ref{eq:c6_batch_goal_reformulated}), we have

\begin{align}  \label{eq:appendix_proof}
   & \text{FDR}  =  \mathbb{E} \left[ 
\frac{\left| \{x_i \in \mathcal{D}^{\text{uns,te}}_{( k_{\text{CA}})}: C^*(x_i) < 1-\alpha^{\text{mis}}_{\text{label}}      \} \right|}
{\left|\mathcal{D}^{\text{uns,te}}_{( k_{\text{CA}})}\right|} \right] \nonumber  \\ & \overset{(a)}{=} \mathbb{E} \left[ \widehat{\text{FDP}}_{( k_{\text{CA}})} \cdot \frac{1 + |\mathcal{D}^{\text{val}}|}{|\mathcal{D}^{\text{te}}|}  \frac{\left| \{x_i \in \mathcal{D}^{\text{uns,te}}_{( k_{\text{CA}})}: C^*(x_i) < 1-\alpha^{\text{mis}}_{\text{label}}      \}\right|}{1+\left| \{x_i \in \mathcal{D}^{\text{uns,val}}_{( k_{\text{CA}})}: C^*(x_i) < 1-\alpha^{\text{mis}}_{\text{label}}      \}\right|}   \right] \nonumber \\
& \overset{(b)}{\leq} \delta \cdot \frac{1 + |\mathcal{D}^{\text{val}}|}{|\mathcal{D}^{\text{te}}|} \mathbb{E} \left[ \frac{\left| \{x_i \in \mathcal{D}^{\text{uns,te}}_{( k_{\text{CA}})}: C^*(x_i) < 1-\alpha^{\text{mis}}_{\text{label}}      \}\right|}{1+\left| \{x_i \in \mathcal{D}^{\text{uns,val}}_{( k_{\text{CA}})}: C^*(x_i) < 1-\alpha^{\text{mis}}_{\text{label}}\}\right|}  \right], 
\end{align}
where $(a)$ follows from the definition of the \gls{fdp} estimator in (\ref{eq:c6_FDP_estimator}), and $(b)$ is obtained by adopting the stopping rule in (\ref{eq:c6_stopping_rule}), which implies the inequality $\mathbb{E} \left[\widehat{\text{FDP}}_{(k_{\text{CA}})}\right] \leq \delta$.

For notational convenience, we define the data-dependent statistic
\begin{align} \label{eq:martingale_RV}
    \mathcal{M}_{(k)} = \frac{\left| \{x_i \in \mathcal{D}^{\text{uns,te}}_{( k)}: C^*(x_i) < 1-\alpha^{\text{mis}}_{\text{label}}      \}\right|}{1+\left| \{x_i \in \mathcal{D}^{\text{uns,val}}_{( k)}: C^*(x_i) < 1-\alpha^{\text{mis}}_{\text{label}}\}\right|}. 
\end{align}
The sequence $\{ \mathcal{M}_{(k)}\}_{k \geq 0}$ can be shown to be a super-martingale with respect to the filtration $\{ \mathcal{G}_{(k)}\}_{k \geq 0}$, where
\begin{align}
    \mathcal{G}_{(k)}  = \sigma ( \{x_i, C^*(x_i), \hat{C}(x_i)\}_{i=1}^{k} ) 
\end{align}
represents the information set observed up to screening step $k$ \citep[Lem.~1]{gui2025acs}. Therefore, by the defining condition of a super-martingale, we have the inequality
\begin{align} \label{eq:martingale_definition}
    \mathbb{E} \left[\mathcal{M}_{(k+1)} |\mathcal{G}_{(k)}\right] \leq \mathcal{M}_{(k)}.
\end{align}
Then, taking expectations on both sides in (\ref{eq:martingale_definition}) and applying the law of iterated expectations, we have the inequality
\begin{align} \label{eq:martingale_property_1}
    \mathbb{E} \left[\mathcal{M}_{(k+1)}\right] \leq \mathbb{E} \left[ \mathcal{M}_{(k)} \right].
\end{align}
Furthermore, since the sequential screening process terminates at step $k_{\text{CA}}$, we can leverage the optional stopping theorem for super-martingale \citep{billingsley2017probability} to obtain the inequality 
\begin{align} \label{eq:martingale_property_2}
    \mathbb{E} \left[\mathcal{M}_{(k_{\text{CA}})}\right] \leq \mathbb{E} \left[ \mathcal{M}_{(0)} \right], 
\end{align}
where, by the definition of $\mathcal{M}_{(k)}$ in (\ref{eq:martingale_RV}), we have 
\begin{align} \label{eq:martingale_bounds}
    \mathbb{E} \left[ \mathcal{M}_{(0)} \right] = \mathbb{E} \left[ \frac{\left| \{x_i \in \mathcal{D}^{\text{te}}: C^*(x_i) < 1-\alpha^{\text{mis}}_{\text{label}}      \}\right|}{1+\left| \{x_i \in \mathcal{D}^{\text{val}}: C^*(x_i) < 1-\alpha^{\text{mis}}_{\text{label}}\}\right|} \right] =  \frac{|\mathcal{D}^{\text{te}}|}{1 + |\mathcal{D}^{\text{val}}|}.
\end{align}
Finally, using (\ref{eq:appendix_proof}), (\ref{eq:martingale_property_2}), and (\ref{eq:martingale_bounds}), we obtain the desired result
\begin{align}
    \mathbb{E}\bigg[\frac{ | \{x_i \in \mathcal{S}: \Pr[y_i \in \Gamma(x_i)|x_i] \geq 1-\alpha^{\text{mis}}_{\text{label}} \} | }{|\mathcal{S}|}\bigg] = 1- \text{FDR} \geq 1 - \delta, 
\end{align}
which ensures that the output $\mathcal{S}= \mathcal{D}_{(k_{\text{CA}})}^{\text{uns,te}}$ of the \gls{cab} model cascading methodology satisfies the guarantee (\ref{eq:c6_batch_goal}).

\end{appendices}

\printthesisindex 

\end{document}